%% file: main.tex
\documentclass{article}

\PassOptionsToPackage{numbers,compress,sort}{natbib}
\usepackage{natbib}

\usepackage{wrapfig}
\usepackage[utf8]{inputenc} %
\usepackage[T1]{fontenc}    %
\usepackage{url}            %
\usepackage{booktabs}       %
\usepackage{longtable}
\usepackage{amsfonts}       %
\usepackage{nicefrac}       %
\usepackage{xspace}         %
\usepackage[
  protrusion=true,
  expansion=true,
  final,
  factor=1100,
  stretch=20,
  shrink=20
]{microtype}
\usepackage{xcolor}         %
\usepackage{amsmath} 
\usepackage{graphicx}
\usepackage[rightcaption,raggedright]{sidecap}
\usepackage{subcaption}
\usepackage{multirow}
\usepackage{todonotes}
\usepackage{float}
\usepackage{adjustbox}
\usepackage{placeins}
\usepackage{pdflscape}
\usepackage{titletoc}      %

\usepackage[a4paper,margin=1in]{geometry}

\usepackage[table]{xcolor}  %
\definecolor{medalGold}{HTML}{FFE89C}    %
\definecolor{medalSilver}{HTML}{FFD2A6}  %
\definecolor{medalBronze}{HTML}{E8C49C}  %

\definecolor{gold}{RGB}{218,165,32}
\definecolor{silver}{RGB}{160,160,160}
\definecolor{bronze}{RGB}{205,127,50}

\definecolor{priorlight}{RGB}{247,248,250} %
\usepackage[colorlinks=true,linkcolor=black,urlcolor=blue]{hyperref}
\usepackage{cleveref}
\usepackage[most]{tcolorbox}
\definecolor{priorAccent}{HTML}{4A3FB7}  %

\tcbset{
colback=priorlight,
colframe=black!15,
boxrule=0.5pt,
arc=6pt,
left=16pt,right=16pt,top=14pt,bottom=14pt,
enhanced,
}

\newcommand{\field}[2]{\textbf{#1:}\enspace #2\par}
\renewcommand{\field}[2]{%
  \par\noindent
  {\small\color{black!55}\textsc{#1}}\quad #2\par\addvspace{2pt}}

\usepackage{tikz}
\usepackage{circuitikz}
\usepackage{amsmath, amssymb}
\usepackage{xcolor}
\usetikzlibrary{
  arrows.meta,
  positioning,
  fit,
  shapes.geometric,
  shapes.symbols,
  shapes.misc,
  shapes.multipart,
  matrix,
  backgrounds,
  calc,
  decorations.pathreplacing,
  decorations.markings
}

\usepackage{caption}
\usepackage{amsmath,amssymb}
\usepackage{xcolor}

\pgfdeclarelayer{midground}
\pgfsetlayers{background,midground,main}

\colorlet{groupCyan}{cyan!15}
\colorlet{groupGreen}{green!15}
\colorlet{groupOrange}{orange!15}
\colorlet{groupPurple}{violet!12}
\definecolor{darkgreen}{RGB}{0,100,0}

\definecolor{groupCyanEdge}{RGB}{ 30, 110, 135}
\definecolor{groupGreenEdge}{RGB}{ 60, 120,  55}
\definecolor{groupOrangeEdge}{RGB}{180,  95,  20}
\definecolor{groupPurpleEdge}{RGB}{ 95,  60, 140}
\definecolor{groupGray}{RGB}{238, 238, 240}
\definecolor{groupGrayEdge}{RGB}{ 90,  90,  95}

\newcommand{\ourmodeltwofive}{\mbox{TabPFN-2.5}\xspace}
\newcommand{\ourmodel}{\mbox{TabPFN-3}\xspace}
\newcommand{\ourmodelenhanced}{\mbox{TabPFN-3-Plus} (Thinking)\xspace}
\newcommand{\ourmodelplus}{\mbox{TabPFN-3-Plus}\xspace}
\newcommand{\rtzero}{$\text{RT}_{\text{zero}}$}

\newcommand{\err}{\operatorname{err}}
\newcommand{\besterr}{\operatorname{best\_err}}

\begin{document}

\vspace*{-1cm}
\begin{tcolorbox}

\vspace{-0.35em}\hspace*{-0.2cm}\includegraphics[width=0.2\linewidth]{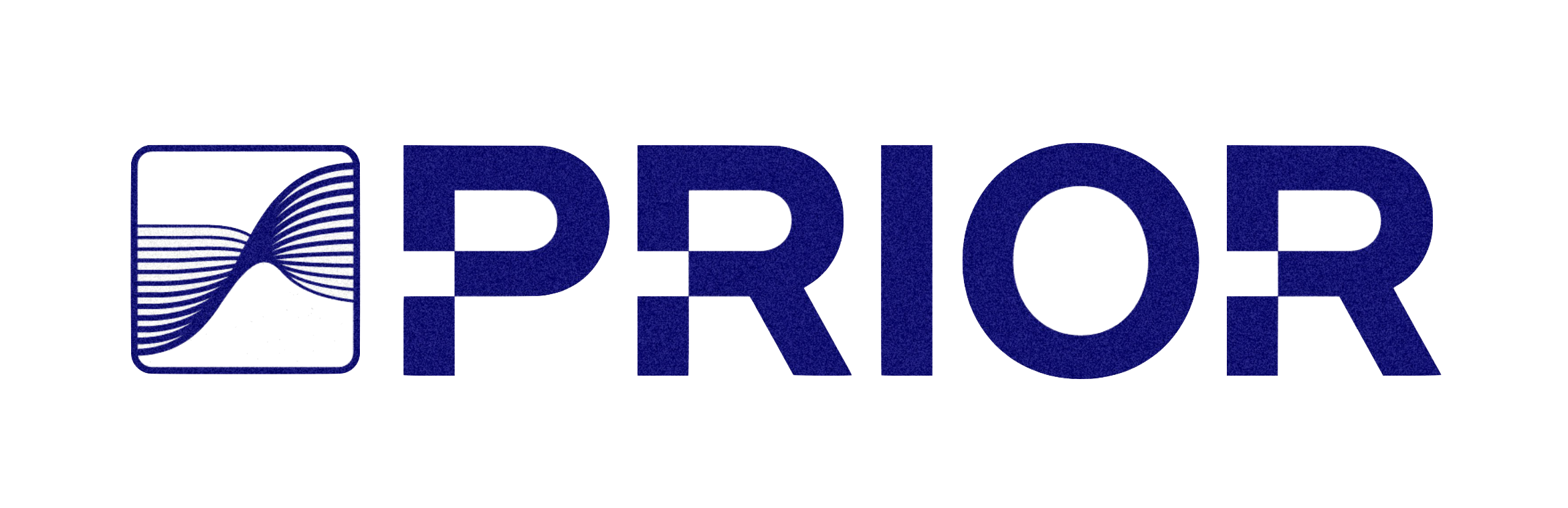}

{\LARGE\bfseries TabPFN-3: Technical Report\par}

\vspace{0.35em}
{\large \hyperref[app:contributors]{Prior Labs Team}}  (see Appendix \ref{app:contributors} for the list of contributors)

\vspace{1.5em}

Tabular data underpins most high-value prediction problems in science and industry, and TabPFN has driven the foundation model revolution for this modality. Designed with feedback from our users, \ourmodel builds on this foundation to scale state-of-the-art performance to datasets with 1M training rows and substantially reduce training and inference time. 
Pretrained exclusively on synthetic data from our prior, \ourmodel dramatically pushes the frontier of tabular prediction and brings substantial gains on time series, relational, and tabular-text data.

\vspace{-1.0em}
\paragraph{A new performance standard.}
On the standard tabular benchmark TabArena, a forward pass of \ourmodel outperforms all other models, including tuned and ensembled baselines, by a significant margin, and pareto-dominates the speed/performance frontier. \ourmodel also scales to more diverse datasets: it ranks first on datasets with many classes, and beats 8-hour-tuned gradient-boosted-tree baselines on datasets up to 1M training rows and 200 features.

\vspace{-1.0em}
\paragraph{Thinking mode.}
\ourmodel introduces test-time compute scaling to tabular foundation models. Our API offering \ourmodelenhanced exploits this to beat all non-TabPFN models by over 200 Elo on the standard TabArena benchmark, rising to 420 Elo on the largest data subset, and outperforming AutoGluon 1.5 extreme in less than a tenth of its runtime, without using LLMs, real data, internet search or any other model besides TabPFN.

\vspace{-1.0em}
\paragraph{Broader capabilities.} 
TabPFN-3 extends the capabilities of our models, enabling SOTA prediction on many-class datasets, relational data (new SOTA foundation model on RelBenchV1) and tabular-text datasets (SOTA on TabSTAR via \ourmodelplus). 
It also directly improves existing integrations of TabPFN: 
a specialized \ourmodel checkpoint, 
TabPFN-TS-3, ranks 2$^{\text{nd}}$ on the time-series benchmark fev-bench,
and SHAP-value computation through \texttt{shapiq} is up to $120\times$ faster with KV caching.

\vspace{-1.0em}
\paragraph{An enterprise-ready model.} \ourmodel achieves this performance while being up to 20x faster than \ourmodeltwofive. In addition, a reduced KV cache and row-chunking scale to 1M rows on a single H100 with fast inference speed.

\vspace{.5em}

We release TabPFN-3 under the \texttt{TABPFN-3.0 License v1.0}, permissive for research and internal evaluation. \ourmodelenhanced is available via API and enterprise licensing including on-prem and VPC environments (AWS SageMaker, Azure AI Foundry).

\vspace{0.5em}

  \field{Date}{May 12, 2026}
   \field{License}{\texttt{TABPFN-3.0 License v1.0} (see Section \ref{sec:license} for details)}
  \field{Docs}{\url{https://docs.priorlabs.ai}}

\end{tcolorbox}

\begin{figure}[h]
    \centering
    \includegraphics[width=\linewidth,trim={0 0 0 0.69cm}, %
    clip]{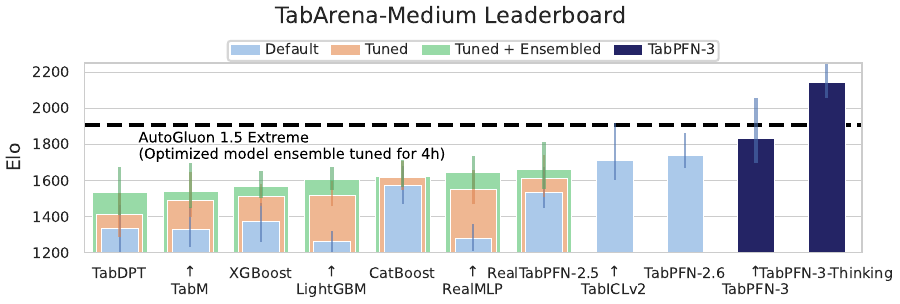}
    \vspace{-0.5cm}
    \caption{\textbf{Performance on the TabArena benchmark \citep{erickson2025tabarena}, largest data subset (10k-100k samples)}.   TabPFN-3 outperforms any other model in a forward pass. \ourmodelenhanced is dramatically better yet, outperforming AutoGluon 1.5 extreme \citep{autogluon_tabular}, 
    a complex ensemble of models tuned for 4 hours, while being 10x faster.
    }
    \label{fig:tabarena-hero-plot}
\end{figure}

\newpage
{
  \setcounter{tocdepth}{2}
  \tableofcontents
}
\newpage

\input{sections/01_introduction}
\input{sections/02_tabpfn3}
\input{sections/03_experimental_results}
\input{sections/04_adoption}
\input{sections/05_license}

\input{sections/A_contributors}

\input{sections/B_acknowledgements}

\input{sections/C_architectural_hyperparams}

\input{sections/D_prior_visualizations}

\input{sections/E_experimental_results_details}

\input{sections/F_additional_internal_benchmarks}

\input{sections/G_supplementary_inference_time}

\input{sections/H_time_series_results}

\input{sections/I_use_cases}

\end{document}

%% file: sections/01_introduction.tex
\begin{figure}[t]
  \centering
  \begin{minipage}[b]{0.48\linewidth}
  \vspace*{-6.5cm}
    \centering
    \includegraphics[width=\linewidth]{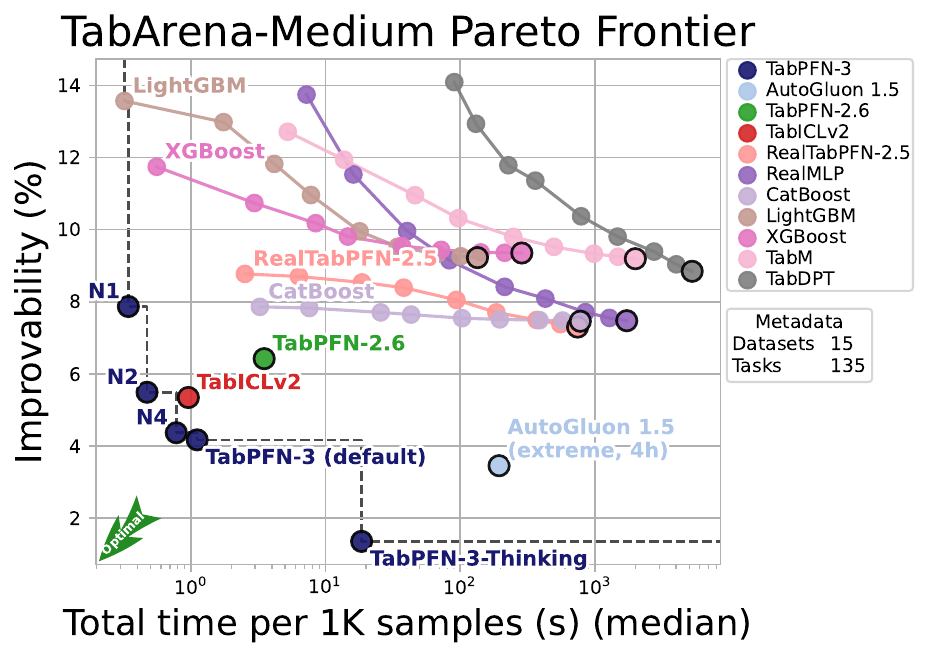}
    \caption{\textbf{TabPFN-3 dominates the Pareto frontier on the largest datasets in TabArena}
    (10k--100k rows). %
    N1, N2, and N4 are model versions with 1, 2, and 4 estimators. Improvability measures how much worse a model is than the best per-dataset model. See Appendix \ref{app:tabarena-metrics} and \ref{app:tabarena_pareto_frontier_explanation} for details.}
    \label{fig:tabarena_pareto_medium}
  \end{minipage}\hfill
  \begin{minipage}[b]{0.48\linewidth}
    \centering

    \includegraphics[width=\linewidth]{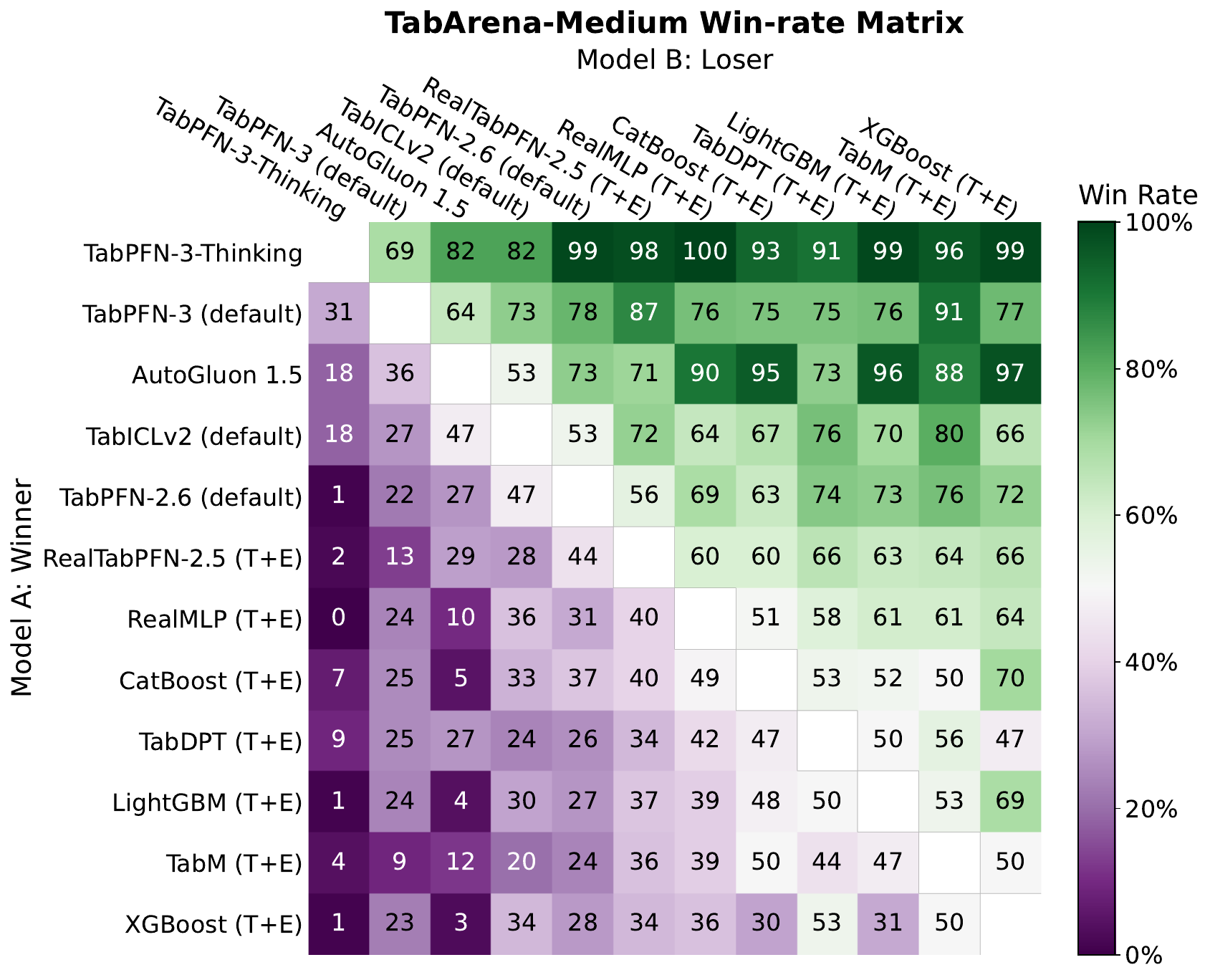}
    \caption{\textbf{Pairwise win rates on TabArena-medium} (10k--100k rows) for a curated set of the strongest models on TabArena. See Appendix \ref{sec:tabarena_leaderboard_tables} for the full results.}
    \label{fig:tabarena_winrate_medium}
  \end{minipage}
\end{figure}

\section{Introduction}

Tabular data sits at the core of operational decision-making across science and industry, including clinical risk prediction \cite{henry2015targeted, johnson2016mimic, ophir2020deep}, credit scoring \cite{abdou2011credit, khandani2010consumer, lessmann2015benchmarking}, predictive maintenance \cite{carvalho2019systematic, dalzochio2020machine}, and scientific measurement \cite{baldi2014searching, dunn2020benchmarking}. While gradient-boosted trees were the reliable default for decades \cite{shwartz2022tabular, grinsztajn2022tree,salinas2024tabrepo}, tabular foundation models have displaced them as the strongest predictors on standard small-to-medium-sized benchmarks over the last year~\citep{erickson2025tabarena}.

Earlier TabPFN releases established and extended this paradigm. TabPFN~v1~\citep{hollmann2022tabpfnv1} showed that a transformer pretrained on synthetic tasks could approximate Bayesian inference in a single forward pass, though only on a thousand rows of clean numerical data. TabPFN~v2~\citep{Hollmann2025tabpfnv2} scaled this to 10,000 rows datasets with categorical features, missing values, and outliers, becoming the first tabular foundation model to outperform tuned gradient-boosted trees on standard benchmarks. TabPFN-2.5~\citep{TabPFN-2.5} extended the strong performance to 100{,}000 rows and 2{,}000 features and matched four-hour-tuned ensembles in a single forward pass. 
Across these releases, an active research ecosystem of extensions grew on top of the core model -- domains include time-series forecasting~\citep{hoo2024tabpfn_ts}, causal inference~\citep{robertson_dopfn, balazadeh_causalpfn, feuerriegel_causalfm}, Bayesian optimization~\citep{Yu2025GITBO}, graph learning~\citep{Hayler2025GraphsTablesZeroShot, eremeev2025turningtabularfoundationmodels}, interpretability \cite{rundel2024interpretable, ye2026closer}, reinforcement learning \cite{Schiff2025TabPFNRL}
-- with over 200 published applications (see Appendix \ref{app:use_cases}) and more than three million PyPI downloads.

TabPFN-3 is shaped by the feedback from users and the entire ecosystem. To remove common bottlenecks, we scaled beyond a hundred thousand rows to one million rows, cut the memory and latency of inference at scale, added support for many-class classification, and honed our calibrated predictive distributions in a single forward pass.
Furthermore, we carefully designed the TabPFN-3 model
and training process to lift performance on both core tabular prediction as well as the
many downstream extensions built on top of the open-source model, in particular time-series forecasting, multi-table relational data, and interpretability.

The remainder of this report describes the architecture, prior, and inference-time optimizations of TabPFN-3 (Section~\ref{sec:methods}); evaluates its performance on public and internal benchmarks across classification, regression, many-class, time-series, and relational data (Section~\ref{sec:results}); surveys the adoption and ecosystem the model is built for (Section~\ref{sec:usecases_extensions}); and details licensing and availability (Section~\ref{sec:license}). Appendices provide architectural hyperparameters, prior visualizations, additional internal benchmarks, more detailed benchmark results and an extensive list of published TabPFN use cases. For installation and usage, see \url{https://docs.priorlabs.ai/}.

\begin{figure}[!t]
  \centering

  \begin{subtable}[b]{0.47\textwidth}
    \centering
    \small
    \begin{tabular}{lrrrr}
      \toprule
      \multirow{2}{*}{Model} & \multirow{2}{*}{Rows} & \multirow{2}{*}{Features}
        & \multicolumn{2}{c}{Parameters} \\
      \cmidrule(lr){4-5}
      & & & Clf. & Reg. \\
      \midrule
      TabPFN-v1  & $1{,}000$       & $100$      & $26$\,M & ---       \\
      TabPFN-v2  & $10{,}000$      & $500$      & $7$\,M  & $11$\,M \\
      TabPFN-2.5 & $100{,}000$     & $2{,}000$  & $11$\,M & $10$\,M \\
      TabPFN-2.6 & $100{,}000$     & $2{,}000$  & $11$\,M & $13$\,M \\
      \midrule
      \multirow{3}{*}{TabPFN-3}
        & $1{,}000{,}000$ & $200$      & \multirow{3}{*}{$53$\,M} & \multirow{3}{*}{$58$\,M} \\
        & $100{,}000$     & $2{,}000$  & & \\
        & $1{,}000$       & $20{,}000$ & & \\
      \bottomrule
    \end{tabular}
    \vspace{1.8cm}
    \caption{\textbf{Overview of previous TabPFN releases}, the maximal numbers of rows and features that they yielded state-of-the-art performance in, and their parameter counts. 
    TabPFN-v1 supports classification datasets only.}
    \label{tab:tabpfn-variants}
  \end{subtable}
  \hfill
  \begin{subfigure}[b]{0.47\textwidth}
    \centering
    \includegraphics[width=0.8\linewidth]{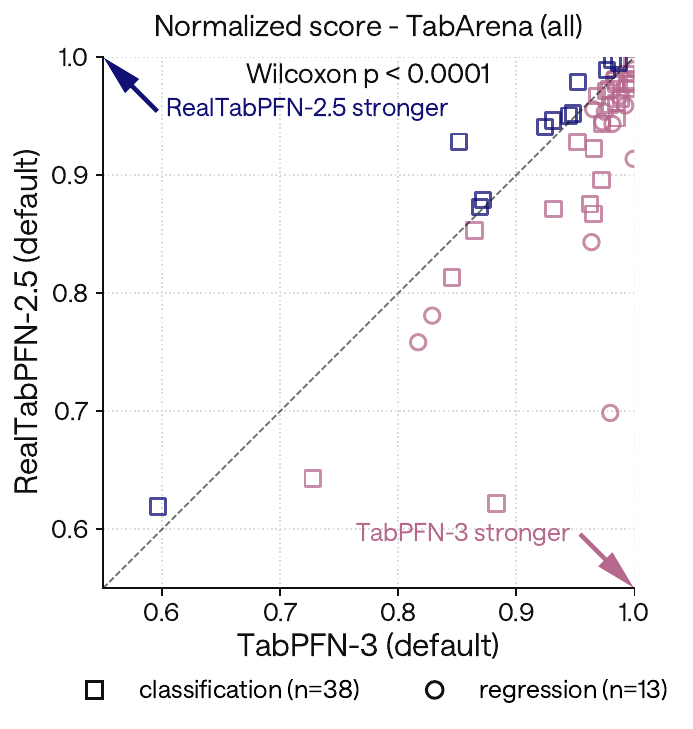}
    \caption{\textbf{TabPFN-3 presents a significant improvement against TabPFN-2.5}. We report the per-datasets scores on TabArena. The normalization procedure is described in Section \ref{app:methodology}.}
    \label{fig:tabpfn3-vs-v2p5}
  \end{subfigure}

  \caption{\textbf{Evolution and performance of the TabPFN model family.}
    The row and feature counts in the table denote benchmark-validated regimes
    where public and internal evaluations demonstrate state-of-the-art (SOTA) performance;
    larger or different row--feature configurations may be feasible, but are outside
    the validated SOTA envelope summarized here. Earlier releases are evaluated
    within a single row--feature regime, whereas TabPFN-3 is benchmarked along a
    cell-budget frontier: up to 1M rows at 200 features, 100k rows at 2{,}000
    features, or 1k rows at 20{,}000 features. The right panel shows per-dataset
    scores across TabArena; points below the diagonal indicate stronger TabPFN-3
    performance, with a Wilcoxon test confirming that the improvement is significant
    ($p < 0.0001$).}
  \label{fig:table-and-figure}
\end{figure}

%% file: sections/02_tabpfn3.tex
\section{TabPFN-3}\label{sec:methods}

TabPFN-3 comes with a new architecture (Section~\ref{sec:arch_overview}), including an attention-based many-class decoder (Section~\ref{sec:many-class-decoder}), an improved preprocessing pipeline (Section~\ref{sec:preprocessing}), inference-time optimizations that enable scaling to one million rows on a single GPU (Section~\ref{sec:inference-optimization}), and an improved synthetic SCM prior used for pre-training (Section~\ref{sec:synthetic-prior}). We also introduce the API and enterprise features \ourmodelplus which handles text in tables natively and \ourmodelenhanced which applies test-time-compute for dramatically improved performance (Section \ref{sec:tabpfn3plus}).

\subsection{Architecture}
\label{sec:arch_overview}

\begin{figure}[!t]
    \centering
    \adjustbox{width=0.75\linewidth}{
    \input{figures/tikz/arch_diagram}}
    \caption{\textbf{Architecture of TabPFN-3, adapted from the TabICLv2 architecture.} Changes include adding novel orthogonal embeddings, the many class decoder, NaN/Inf indicator variables, and the option to use low-memory chunked inference (dotted paths; dashed paths signal fully parallel path). C refers to number of colums, N to the number of rows, K is the number of inducing points, and $\mathcal{Y}$ is the set of labels. Shown is TabPFN-3 for classification; the regression variant does not use the many-class decoder.
    }
    \label{fig:arch_diagram}
\end{figure}
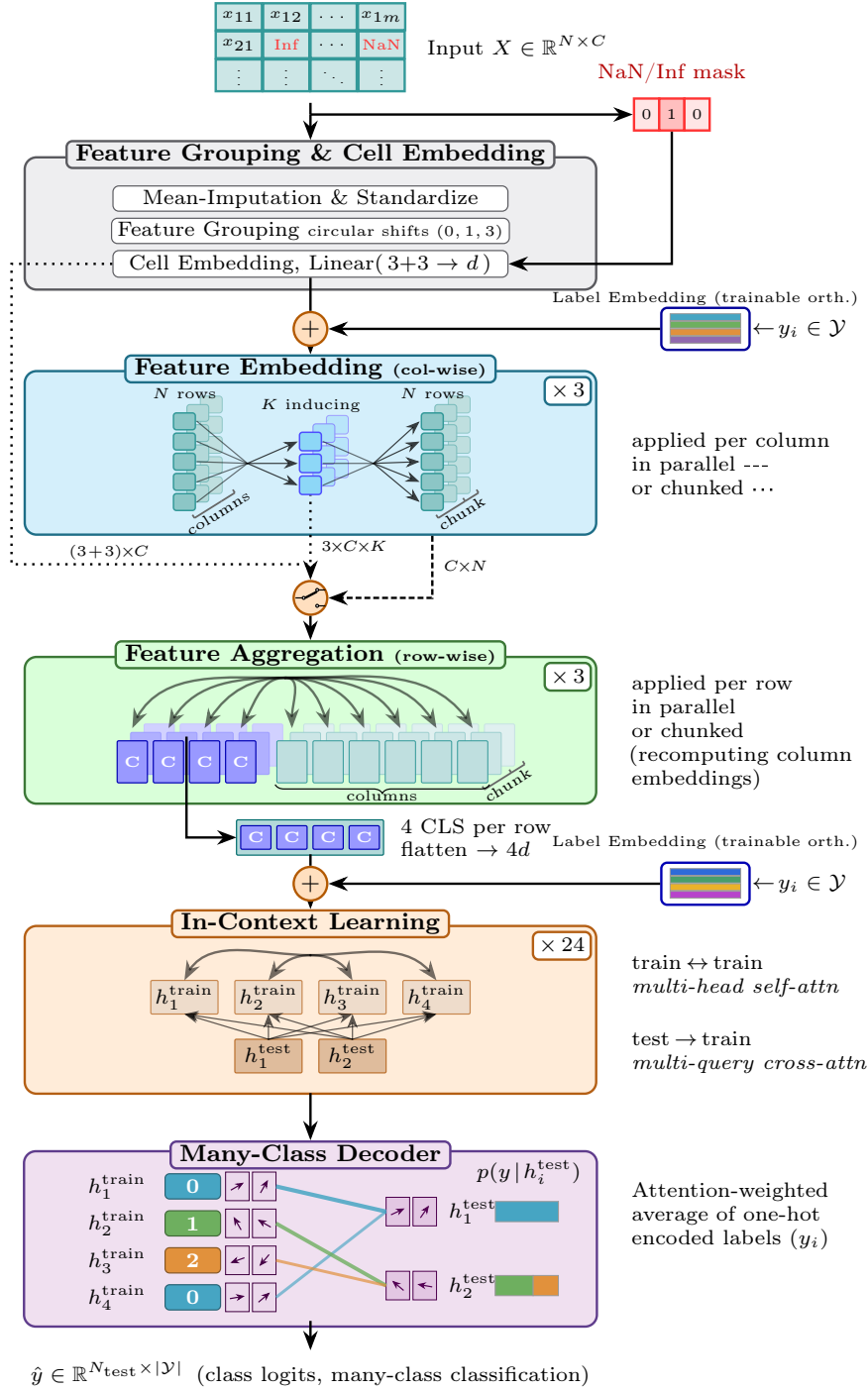

An overview of TabPFN-3's full architecture is shown in~\Cref{fig:arch_diagram}. 
TabPFN-3 introduces a substantially redesigned architecture that scales in-context learning to datasets with one million rows.
\\

TabPFN~v1~\citep{hollmann2022tabpfnv1} used a transformer architecture to perform in-context learning (ICL) on embeddings of entire rows.
TabPFN-2.x (v2, v2.5, v2.6) \citep{Hollmann2025tabpfnv2, TabPFN-2.5} used a transformer architecture that alternates row-wise and feature-wise attention layers; this improves performance, but becomes prohibitively expensive as the dataset size grows.
TabPFN-3 returns to TabPFN~v1's ICL for embeddings of entire rows. 
It builds on the two-stage row-compression design introduced by \citet{qu2025tabicl, qu2026tabiclv2} in the TabICL architecture, which uses a column-wise feature embedding layer followed by row-wise feature aggregation to obtain the row representation that is used in a TabPFN~v1-like ICL layer.

Before entering the two compression stages, we group features, similar to TabPFN-2.x, while adopting TabICLv2's\citep{qu2026tabiclv2} group assignment, which creates triplets by grouping each feature with two cyclically shifted neighbors. Each triplet is mapped to the hidden dimension of the model by a learned linear projection (cell embedding), and target-aware embeddings are added to the cell embeddings of training rows \citep{qu2026tabiclv2}.

The resulting grouped feature embeddings are processed by the following three stages:

\begin{itemize}
    \item \textbf{Stage~1: Feature distribution embedding (column-wise).}
    Each feature column is embedded independently using a transformer with an efficient inducing-point attention mechanism.
    This avoids the quadratic cost of full cross-row attention while still capturing column-level statistics at arbitrary dataset scales.

    \item \textbf{Stage~2: Feature aggregation (row-wise).}
    For each data point, a set of learned \textsc{cls} tokens and the feature embeddings of that row attend to one another via non-causal attention, allowing cross-feature information to be distilled into a fixed number of vectors.
    Concatenating the \textsc{cls} tokens' hidden states yields a single, fixed-dimensional embedding per row, decoupling the subsequent in-context learning stage from the number of input features.

    \item \textbf{Stage~3: In-context learning.}
    The row embeddings for the training and test sets are jointly passed to a transformer that performs in-context learning: training-row embeddings attend to one another to capture relationships within the training set, while test-row embeddings attend to training-row embeddings to produce predictions.
    Because each data point is now a single vector, this stage operates on a sequence proportional only to the number of rows, enabling efficient scaling to large datasets.
\end{itemize}

In Stages~1 and~3, and in the many-class decoder (introduced below), every attention layer applies the query-aware scalable softmax (QASSMax) \citep{qu2026tabiclv2}, itself inspired from SSMAX \citep{ssmax_vanilla}, which rescales attention queries as a function of input length, improving  length generalization of in-context learning to large training sets.
Detailed architectural hyperparameters are provided in Appendix~\ref{app:architecture-hyperparams}.

TabPFN-3 introduces several architectural innovations on top of the three-stage architecture: 

\begin{itemize}
    \item \textbf{Attention-based many-class decoder.}
    For classification, the fixed-width MLP output head of previous TabPFN versions is replaced with an attention-based retrieval decoder that treats class prediction as soft nearest-neighbor retrieval over the in-context training set, inspired by \citet{arbel2026equitabpfn} and \citet{koshil-etal-2025-context}.
    The decoder is non-parametric in the class count, enabling native support for an arbitrary number of classes. 
    A detailed description is given in Section \ref{sec:many-class-decoder}.

    \item \textbf{Row-chunking.} A two-phase inference scheme that decouples peak GPU activation memory from dataset size (rows $\times$ columns), while producing outputs equivalent to the unchunked computation: we precompute the distribution embedder's inducing-vector summary once over the full training set, then stream rows through feature embedding and column aggregation in fixed-size chunks that reuse this cached summary as their attention key/value set. See Section \ref{sec:row_chunking} for more details.
    
    \item \textbf{Reduced KV cache via multi-query attention.}
    In the ICL transformer, test-row queries attend to train-row keys and values using a single KV head (multi-query attention), while train rows retain full multi-head attention.
    This allows reducing the per-estimator KV cache to approximately 7\,GB for datasets of one million rows, enabling ultra-fast inference on common GPUs. This is described in detail in Section~\ref{sec:kv_cache}.

    \item \textbf{Orthogonal target embeddings.}
    Training labels are encoded with learned embeddings initialized via orthogonal decomposition, providing near-maximally separated class representations at the start of training and improving gradient flow in the many-class regime.

    \item \textbf{RMSNorm.}
    All normalization layers use RMSNorm in place of the layer normalization used in \ourmodeltwofive. RMSNorm omits the mean-centering term, reducing compute while preserving training stability.

    \item \textbf{Native missing-value handling.}
    For each cell that is \texttt{NaN}, TabPFN-3 computes a binary indicator and concatenates it with the cell value before embedding. The model therefore receives an explicit signal about missing data and can condition its predictions accordingly, rather than relying on upstream imputation.
\end{itemize}

\subsection{Many-class Decoder}
\label{sec:many-class-decoder}
For multiclass classification, \ourmodel replaces the fixed-width MLP classification head used in TabPFN-2.6 (and earlier versions) with an \emph{attention-based retrieval decoder} over the in-context training set,
which treats class prediction as a soft nearest-neighbor retrieval, inspired by \citet{arbel2026equitabpfn} and \citet{koshil-etal-2025-context}: the final-layer train embeddings $\{h^\mathrm{train}_n\}_{n=1}^{N_\mathrm{train}}$ act as keys, the corresponding one-hot label vectors $\mathbf{y}_n \in \{0,1\}^{C}$ as values, and test embeddings $h^\mathrm{test}_m$ as queries. After the usual learned linear projections $W_Q, W_K$ and a multi-head split, the decoder computes
\[
p_m \;=\; \frac{1}{H}\sum_{h=1}^{H}\sum_{n=1}^{N}
            \alpha^{(h)}_{m,n}\,\mathbf{y}_n,
\qquad
\alpha^{(h)}_{m,n}=\mathrm{softmax}_n\!\left(
            \tfrac{q^{(h)}_m \cdot k^{(h)}_n}{\sqrt{D_h}}\right),
\]
that is: a (head-averaged) attention-weighted average of the in-context one-hot labels, which is then converted to logits via $\log\!\big(\mathrm{clip}(p_m)\big)$. This formulation has two consequences. First, classes are no longer tied to fixed output positions of a parametric head
so the decoder is naturally permutation-equivariant in the class indices.
Second, decoding is non-parametric in $C$: the decoder's parameters depend only on the embedding dimension and the number of attention heads, not on some $C_{\max}$, decoupling the head's capacity from the supported label cardinality.

\textbf{Class-count limit from pre-training.}
Although the decoder is non-parametric in $C$, the trained \ourmodel still fixes a hard ceiling $C_{\max} = 160$ at pre-training time via three checkpoint-bound tensors: the trainable orthogonal label embeddings $E_{\mathrm{col}}, E_{\mathrm{icl}} \in \mathbb{R}^{C_{\max} \times D}$ used by the column encoder and the ICL transformer, and the one-hot value tensor consumed by the decoder. Enlarging $C_{\max}$ at pre-training therefore costs only $\mathcal{O}(C_{\max}\,D)$ extra parameters and no extra decode-time memory.

\subsection{Preprocessing}
\label{sec:preprocessing}

As in previous versions, TabPFN-3 aggregates predictions across multiple estimators, each operating on a distinct combination of dataset permutations and feature transformations, forming an effective ensemble that enhances robustness and generalization. Individual estimators apply complementary feature transformations—combining robust scaling and soft clipping (following \citep{holzmuller2024realmlp}) with quantile transformations and standard scaling—to balance stability and sensitivity across varying feature distributions. As in TabPFN-2.5, a subset of estimators augments the feature matrix with singular value decomposition (SVD) components, capturing high-energy directions of global variance.

TabPFN-3 introduces two further improvements to this pipeline. First, features are subsampled in a round-robin fashion, ensuring that each feature appears in at least one estimator and is never systematically excluded from the ensemble. For datasets exceeding 100{,}000 rows, random feature subsampling is replaced by an informed selection based on Gini importance derived from a lightweight tree model fitted on a subsample, focusing each estimator on the most discriminative features rather than an arbitrary subset. Second, feature transformations such as quantile normalization are now executed on GPU, substantially reducing preprocessing latency and making the pipeline practical at the larger dataset scales supported by TabPFN-3. As in TabPFN-2.5~\citep{TabPFN-2.5}, post-processing capabilities are available, including decision threshold tuning for metric-specific optimization (e.g., F1-score) and temperature scaling for probability calibration.

\subsection{Inference Optimization}\label{sec:inference-optimization}
\ourmodel introduces several inference-time optimizations that together reduce its compute and memory footprint enough to scale to one-million-rows on a single GPU with sub-second inference latency.

\subsubsection{Row-Chunking} \label{sec:row_chunking}

\ourmodel's pre-ICL stages---cell embedding, feature distribution embedding, and feature aggregation---materialize an $(n_\mathrm{train}+n_\mathrm{test}){\times}n_\mathrm{features}{\times}d$ activation, so peak memory can saturate the GPU well before any operation becomes compute-bound. One solution is to offload activations to CPU memory or disk, as in TabICLv2~\citep{qu2026tabiclv2}. This however requires a large amount of CPU memory (250GB for a $1\text{M} \times 500$ table in \citet{qu2026tabiclv2}), or otherwise incurs substantial I/O overhead (\citet{qu2026tabiclv2} report a 4x slowdown).
We instead stream the row dimension in fixed-size slices and keep all activations on the GPU.

\begin{figure}[!t]
    \centering
    \includegraphics[width=\linewidth]{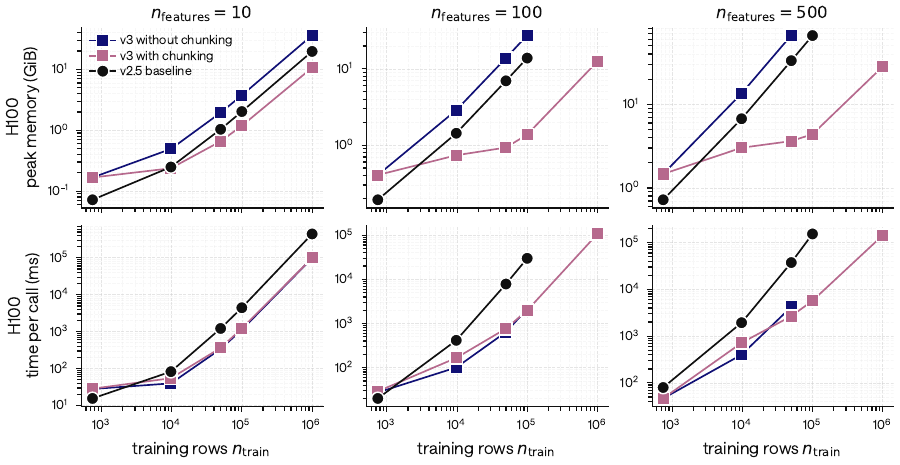}
    \caption{\textbf{Chunking flattens the peak-memory without impacting the time-per-call.}
    Model forward pass without preprocessing, measured on a H100, for $n_\mathrm{features} \in \{10, 100, 500\}$. Top row: peak GPU memory (GiB) versus number of training rows; bottom row: time per call (ms). Three series per panel: \ourmodel without chunking (blue), \ourmodel with chunking (pink), and the \ourmodeltwofive baseline (black). Both axes are log-scaled.
    Note that \emph{TabPFN-3 is much faster than TabPFN-2.5, especially at large feature counts.}}
    \label{fig:chunked_vs_nonchunked}
\end{figure}

A naive row-wise stream is not directly applicable: the distribution embedder summarizes the training set into a fixed-size\footnote{We use 128 inducing points, much smaller than the dataset sizes of interest, which often exceed 100,000 rows.} set of inducing points via cross-attention over all training rows, and splitting that call across chunks would change its semantics. \ourmodel resolves this with a two-phase scheme exactly equivalent to the unchunked computation: (i) the inducing states are computed once over the full training set, chunked along the (independent) column dimension to bound its own memory cost; (ii) rows are then streamed through feature distribution embedding and the feature aggregator in fixed-size chunks, each reusing the precomputed inducing states as its attention key/value set, and the per-chunk row embeddings are concatenated along the row axis. The scheme adds a small overhead from recomputing cell embeddings in phase~(ii) but avoids the disk-bandwidth bottleneck. We enable chunking when $n_\mathrm{train}+n_\mathrm{test} > 2048$.

Figure~\ref{fig:chunked_vs_nonchunked} highlights the different memory–compute trade-offs of \ourmodel and \ourmodeltwofive. Without chunking, the peak memory of \ourmodel grows steeply with $n_\mathrm{train}$ and $n_\mathrm{features}$. This is because the model carries a pre-ICL activation $n_\mathrm{features}$-wide through cell embedding, feature distribution embedding, and feature aggregation before collapsing the feature axis into a single row representation for the ICL transformer. By contrast, \ourmodeltwofive alternates row- and column-attention layers over a representation grouped into $n_\mathrm{features}/3$ tokens, and therefore never materialises a tensor wider than this. This explains why \ourmodel's unchunked peak memory exceeds \ourmodeltwofive's. Applying row-chunking to \ourmodel flattens peak memory with respect to $n_\mathrm{features}$ and yields an approximately ${\sim}5{\times}$ reduction at the largest shapes, enabling 1M-row inference, while incurring only a small wall-clock overhead of a few percent near $n_\mathrm{train}\approx 10^4$ that becomes amortised at larger scales once the $n_\mathrm{train}^2$ ICL row-attention dominates. At the same time, the feature-collapsed row representation gives \ourmodel a substantial runtime advantage at large $n_\mathrm{train}$ or $n_\mathrm{features}$ since its ICL row-attention scales as $n_\mathrm{train}^2$ independently of $n_\mathrm{features}$, whereas \ourmodeltwofive's row attention retains linear dependence on $n_\mathrm{features}$ and scales with $n_\mathrm{features} \cdot n_\mathrm{train}^2$.

\begin{figure}[!t]
    \centering
    \begin{subfigure}{\linewidth}
        \centering
        \includegraphics[width=\linewidth]{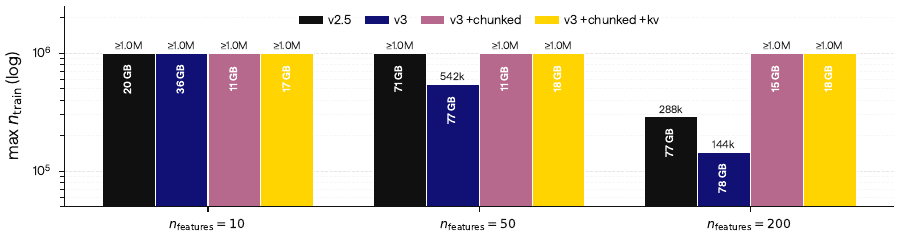}
        \caption{\textbf{Chunking eliminates the OOM frontier; the KV-cache adds memory essentially constant in $n_\mathrm{feature}$.} Maximum $n_\mathrm{train}$ that fits on one 80~GiB H100 for $n_\mathrm{features}\in\{10,50,200\}$. Bars: \ourmodeltwofive, \ourmodel, \ourmodel\,+\,chunking, \ourmodel\,+\,chunking\,+\,KV-cache. White labels: peak memory; ``$\geq 1.0$M'' marks bars that hit the search cap.}
        \label{fig:max_n_train_probe}
    \end{subfigure}
    \begin{subfigure}{\linewidth}
        \centering
        \includegraphics[width=\linewidth]{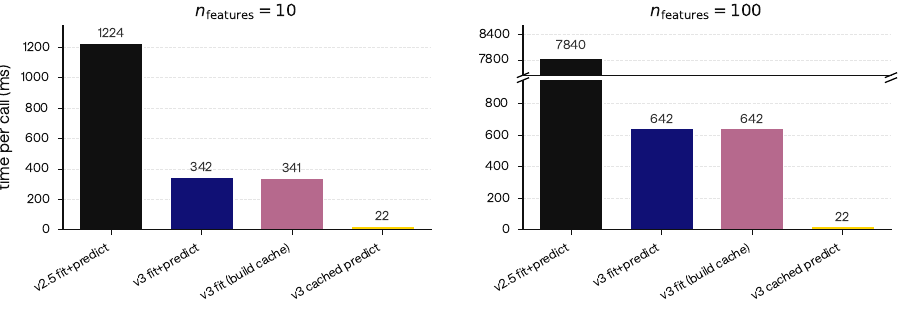}
        \caption{\textbf{Cached predict is 1--2 orders of magnitude faster than the uncached \ourmodel.} Time per model forward pass without preprocessing on H100 at $n_\mathrm{train}=50{,}000$, $n_\mathrm{test}=100$, $n_\mathrm{features}\in\{10,100\}$. Bars: \ourmodeltwofive cold fit+predict, \ourmodel cold fit+predict, \ourmodel fit-with-cache, \ourmodel cached predict.}
        \label{fig:kv_cache_h100}
    \end{subfigure}\\[2pt]
    \caption{KV-cache on H100 for a single estimator without preprocessing: OOM frontier with chunking and KV-cache (\subref*{fig:max_n_train_probe}) and cached-predict latency vs.\ uncached paths (\subref*{fig:kv_cache_h100}).}
    \label{fig:kv_cache}
\end{figure}

\subsubsection{Fast Inference with a Small KV-cache} \label{sec:kv_cache}

\begin{figure}[!t]
      \centering
      \hspace{-3em}
      \includegraphics[width=0.85\linewidth]{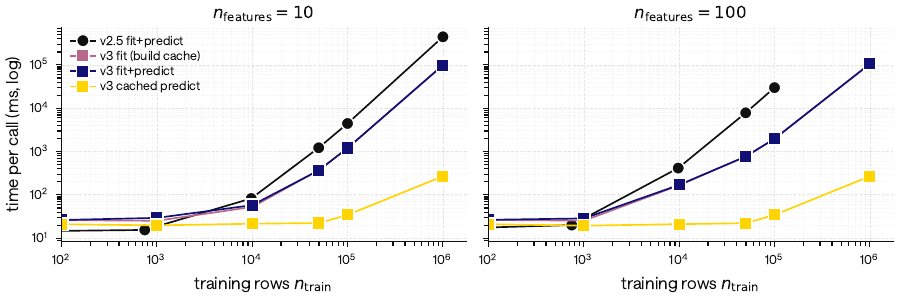}
      \caption{\textbf{\ourmodel's KV-cached predict allows for one to three orders of magnitude speedup}. We report results for a single estimator without preprocessing on an H100, for $n_\mathrm{features} \in \{10, 100\}$ and $n_\mathrm{test} = 100$. Four series per panel: \ourmodeltwofive \texttt{fit+predict} (black, baseline), \ourmodel cold \texttt{fit+predict} (blue, no cache reuse), \ourmodel \texttt{fit (build cache)} that builds the cache (magenta -- overlaps the cold curve since the train-side work is identical, the cache is just retained), and \ourmodel cached \texttt{predict} (yellow). The KV-Cache is built under the deployed multi-query test-side configuration ($n_\mathrm{kv,test}=1$).}
      \label{fig:kv_cache_scaling_h100}
  \end{figure}

Being an in-context-learning model, \ourmodel combines training (fit) and inference (predict) in one forward pass. While this allows for very fast training, it can make online or batched predictions too slow for production usecases. Caching the keys and values (KVs) from the train set removes this issue. While KV-caching has been available in our previous models, the memory cost of the cache was prohibitive for larger datasets. \ourmodel solves this in two ways:
\begin{itemize}
    \item Compared to \ourmodeltwofive, which needs to store an embedding for each cell of the table, \ourmodel only needs to store three components: the per-block inducing states produced by the feature distribution embedder, the train-side keys and values of the ICL self-attention at every transformer block in the ICL stage; as well as the train embeddings of the final ICL layer, which are consumed by the many-class decoder. The inducing states are small and the two other components only scale with the number of rows rather than rows × features.
    \item We use multi-query with only a single head for cross attention between test and train samples, reducing KV-cache size by a factor of eight.
\end{itemize}
This achieves a KV-cache size of 7GiB per estimator for 1M rows datasets, making \ourmodel's default 8 estimators usable on common GPUs even for the largest datasets we support. As can be seen in Figure~\ref{fig:max_n_train_probe}, peak memory of (chunked) cache-predict is basically flat across feature sizes.
On an H100, cached-predict is one to three orders of magnitude faster than either the \ourmodeltwofive baseline or \ourmodel's own cold ``fit+predict'' path (Figure~\ref{fig:kv_cache_h100}),  achieving between 0.1 and 3 ms/test point for batches of 100 test points. The fit-with-cache call costs essentially the same as the cold fit+predict at every measured shape, including $n_\mathrm{train}=10^6$ where both complete in ${\sim}107$~s (Figure~\ref{fig:kv_cache_scaling_h100}).

\subsubsection{Model Distillation} \label{sec:distillation}
In production environments constrained by latency or memory budgets, hardware availability, or regulatory requirements that mandate familiar model classes, \ourmodel also supports distillation into dataset-specific MLPs or tree ensembles via the engine introduced with \ourmodeltwofive \citep{TabPFN-2.5}. The distilled artifact runs on CPU at the sub-millisecond latency of a standard MLP or tree ensemble while retaining most of \ourmodel's predictive performance on the dataset it was distilled for.  

\subsubsection{Compilation and FlashAttention-3} \label{sec:compile_and_fa3}

\ourmodel ships with two opt-in performance features that target different bottlenecks: \texttt{torch.compile}, which fuses dispatch on the non-attention hot paths, and FlashAttention-3 (FA3) \cite{flash_attention_3}, a Hopper-specific kernel for the in-context-learning attention. On MI-250x, \texttt{torch.compile} reaches up to $1.58{\times}$ speedup on the non-chunked forward pass; on H100, FA3 reaches $1.5\text{--}1.7{\times}$ at $n_\mathrm{train}=10^6$ over the SDPA fallback. Both compose cleanly with row chunking and are auto-detected at runtime; see Appendix~\ref{app:compile-fa3} for the full measurements and per-shape breakdowns.

\subsubsection{Improved interpretability for TabPFN}

TabPFN-3's reduced KV-cache (Section~\ref{sec:kv_cache}) and fast inference make interpretability extensions significantly more practical.

Through the \texttt{tabpfn-extensions} package, TabPFN is directly integrated with the popular \texttt{shapiq} library~\citep{shapiq}, enabling efficient approximation of any-order Shapley interactions. \autoref{fig:shap-kv-speedup} in the Appendix shows both the absolute runtime and the relative speed-ups achieved by KV caching. For large datasets, KV caching provides more than $120\times$ efficiency gains, reducing the runtime per test row to 1.08 seconds even for a training table with 200k rows and 500 features.

\subsection{Synthetic Prior}\label{sec:synthetic-prior}
Following previous TabPFN model variants \cite{TabPFN-2.5, hollmann2022tabpfnv1, Hollmann2025tabpfnv2}, TabPFN-3 is trained on synthetically generated data based on our Structural Causal Model (SCM) prior. A schematic flow chart demonstrating how our SCM prior works is shown in Figure \ref{fig:prior}.

Our philosophy in designing our prior is to maximize breadth of possible datasets while capturing the structure models will encounter in real-world data. The result is an updated, more sophisticated prior that allows us to scale up training and continue extracting signal from the wide range of synthetic datasets it generates: our final TabPFN-3 model was trained on more than 8 trillion tokens.

\input{figures/tikz/scm_prior}

\begin{enumerate}
    \item \textbf{Graph generation.} We expand the distribution of graphs
    underlying the SCM by introducing new sampling algorithms, enabling
    richer structural diversity. Sample graphs are shown in
    Figure~\ref{fig:graph_sampling}.
    
    \item \textbf{Combiner mechanisms.} We introduce a host of new combiner mechanisms that combine values of parent nodes to propagate values to child nodes, some examples of which are visualized for a simple two-dimensional case in Figure \ref{fig:mechanisms}. Increasing the variety of functional forms by which child nodes depend on the respective parent nodes allows for richer node relationships in the SCM.

    \item \textbf{Categorical variables.} Compared to TabPFN-2.5, we reworked the treatment of categorical variables in our SCM, moving from a comparatively simple categorical data model to more expressive variants.

    \item \textbf{High-frequency oscillators.} TabPFN-2.5 struggled with
    high-frequency oscillations despite performing well on sinusoidal data
    generally. Improved sinusoidal activations give TabPFN-3 strong
    performance across the full frequency spectrum.
    
    \item \textbf{Spatial prior.} Many tabular datasets have underlying spatial structure (e.g. datasets containing longitude and latitude as covariates, grids of sensors, etc.). We add spatial activations that allow our prior to encode spatial relationships between variables. %
    
    \item \textbf{Many-class prior.} The flexible many-class decoder in TabPFN-3 enables native classification support for an arbitrary number of classes. We match this architectural design in the prior, ensuring high quality datasets that enable state-of-the-art downstream performance from binary datasets to datasets with hundreds of classes.

    \item \textbf{Temporal prior.} Many tabular datasets have temporal structure: rows are collected over intervals of time, train and test splits are often ordered by time rather than drawn i.i.d., and temporal dependencies between variables are common. We extend the SCM into a discrete-time Dynamic Structural Causal Model \citep{boeken2024dynamicstructuralcausalmodels}.

    \item \textbf{Out-of-distribution prior.} We add out-of-distribution prediction tasks, allowing models trained on our prior data to remain performant under distribution shifts, as well as moving from pure interpolation to extrapolation. A simple example highlighting how our o.o.d. prior allows \ourmodel to perform extrapolation is shown in Figure \ref{fig:ood} - a capability that is notably absent from most tree-based algorithms as well as most other tabular foundation models.
    
\end{enumerate}

\subsection{TabPFN-3-Plus and Thinking mode}
\label{sec:tabpfn3plus}

On top of \ourmodel, which we release open-source, our API and enterprise deployments provide access to \ourmodelplus and its thinking mode "TabPFN-3-Plus (Thinking)" (named TabPFN-3-Thinking in our plots)
These variants are fully compatible with the open-source \ourmodel interface and can be used as a drop-in replacement, while offering additional capabilities:

\paragraph{Native text-feature support.} \ourmodelplus accepts string-valued columns directly, without requiring upstream featurization. Free-text fields -- such as product names, insurance claim descriptions, or customer reviews -- are encoded jointly with numeric and categorical features inside the model, so cross-feature interactions between text and structured columns are learned end-to-end rather than imposed by a fixed encoder.

\paragraph{Thinking mode.} \ourmodelenhanced applies additional inference-time computation on top of \ourmodelplus to push prediction quality further. Thinking mode composes with native text-feature support, so a single call can handle mixed numerical, categorical, and text columns under the same inference-time-compute regime. We emphasize that our Thinking mode achieves this strong performance while only relying on TabPFN, without using LLMs, real data, internet search, or any other model.

\ourmodelplus, including Thinking mode, is available through our API and through enterprise deployments including on-prem and VPC deployment on AWS SageMaker and Azure AI Foundry; see Section~\ref{sec:license} for licensing and access. Benchmark results are reported in Sections~\ref{sec:tabarena-result} (TabArena), \ref{sec:tabstar-result} (TabSTAR), and \ref{sec:large-data} (Large data).
\FloatBarrier

%% file: figures/tikz/arch_diagram.tex
\begin{tikzpicture}[
    >=Stealth,
    font=\small,
    cell/.style={
      rectangle, draw=teal!80, thick, fill=teal!20,
      minimum width=0.55cm, minimum height=0.38cm,
      anchor=center, inner sep=1pt
    },
    cellEmpty/.style={
      rectangle, draw=teal!80, thick, fill=white,
      minimum width=0.55cm, minimum height=0.38cm,
      anchor=center, inner sep=1pt
    },
    cellRed/.style={
      rectangle, draw=red!80, thick, fill=red!20,
      minimum width=0.55cm, minimum height=0.38cm,
      anchor=center, inner sep=1pt
    },
    nanCell/.style={
      rectangle, draw=red!80, thick, fill=red!15,
      minimum width=0.30cm, minimum height=0.38cm,
      anchor=center, inner sep=1pt, font=\tiny
    },
    block/.style={
      rectangle, draw=black, thick, fill=white,
      minimum width=6.0cm, minimum height=0.95cm,
      align=center, inner sep=4pt
    },
    smallblock/.style={
      rectangle, draw=black, thick, fill=white,
      minimum width=2.2cm, minimum height=0.50cm,
      align=center, font=\scriptsize, inner sep=2pt
    },
    decblock/.style={
      rectangle, draw=violet!60!black, thick, fill=violet!8,
      minimum width=2.6cm, minimum height=0.55cm,
      align=center, font=\footnotesize, inner sep=3pt
    },
    sidelabel/.style={
      align=left, font=\scriptsize\itshape, text=black!60,
      anchor=west, inner sep=2pt
    },
    novelty/.style={
      rectangle, draw=red!80, thick, fill=red!15,
      rounded corners=2pt, align=left,
      inner sep=4pt, font=\scriptsize, anchor=west
    },
    arrow/.style={-Stealth, thick, black},
    bluearrow/.style={-Stealth, thick, black},
    purplearrow/.style={-Stealth, thick, black},
    thinarrow/.style={-Stealth, thin, black},
    plus/.style={
      circle, draw=black, thick, inner sep=0.5pt,
      minimum size=0.40cm, font=\footnotesize, fill=white
    },
    embedbar/.style={
      rectangle, draw=teal!80, thick, fill=teal!20,
      minimum width=0.26cm, minimum height=0.55cm,
      anchor=center, inner sep=0pt
    },
    embedbarY/.style={
      rectangle, draw=orange!70!black, thick, fill=orange!25,
      minimum width=0.26cm, minimum height=0.55cm,
      anchor=center, inner sep=0pt
    },
    indvec/.style={
      rectangle, draw=blue!70, thick, fill=cyan!25,
      minimum width=0.26cm, minimum height=0.50cm,
      anchor=center, inner sep=0pt
    },
    tensorbox/.style={
      rectangle, draw=teal!80, very thick, fill=teal!8,
      minimum width=2.6cm, minimum height=0.60cm,
      align=center, font=\footnotesize
    },
    tensorboxV/.style={
      rectangle, draw=violet!60!black, very thick, fill=violet!6,
      minimum width=2.6cm, minimum height=0.60cm,
      align=center, font=\footnotesize
    }
  ]

  \definecolor{cA}{RGB}{ 70, 165, 195}
  \definecolor{cB}{RGB}{110, 175,  95}
  \definecolor{cX}{RGB}{225, 145,  60}
  \definecolor{cD}{RGB}{145, 105, 175}
  \colorlet{cA2}{cA!65!blue}
  \colorlet{cB2}{cB!65!teal}
  \colorlet{cX2}{cX!70!yellow}
  \colorlet{cD2}{cD!65!magenta}

  \pgfmathsetmacro{\orthXshift}{4.7}

  \def\stageW{6.8cm}

  \matrix (rawtbl) [
    matrix of nodes,
    nodes={cell, minimum height=0.32cm},
    row 3/.style={nodes={minimum height=0.15cm, inner sep=0pt}},
    column sep=-\pgflinewidth, row sep=-\pgflinewidth,
    nodes in empty cells,
  ] at (0, 7.1)
  {
    {\tiny $x_{11}$} & {\tiny $x_{12}$} & {\tiny $\cdots$} & {\tiny $x_{1m}$} \\
    {\tiny $x_{21}$} & {\tiny \textcolor{red!80}{Inf}} & {\tiny $\cdots$} & {\tiny \textcolor{red!80}{NaN}} \\
    {\tiny \raisebox{2pt}{\scalebox{1}[0.6]{$\vdots$}}} & {\tiny \raisebox{2pt}{\scalebox{1}[0.6]{$\vdots$}}} & {\tiny \raisebox{2pt}{\scalebox{1}[0.6]{$\ddots$}}} & {\tiny \raisebox{2pt}{\scalebox{1}[0.6]{$\vdots$}}} \\
  };
  \node[anchor=west, font=\scriptsize] at (rawtbl.east) {Input $X \in \mathbb{R}^{N \times C}$};

  \coordinate (mainSplit) at (0, 6.30);
  \draw[thick, solid] (rawtbl.south) -- (mainSplit);

  \tikzset{
    preprocstep/.style={
      align=center, font=\scriptsize, inner ysep=1pt, inner xsep=2pt,
      draw=groupGrayEdge, fill=white, rounded corners=2pt,
      minimum width=4.7cm, minimum height=0.30cm
    }
  }

  \coordinate (grayPanelTop) at (0, 5.65);

  \node[preprocstep] (stdblk) at (0, 5.30)
  {Mean-Imputation~\&~Standardize};

  \node[preprocstep, below=2pt of stdblk] (group)
  {Feature Grouping \tiny circular shifts $(0,1,3)$};

  \node[preprocstep, below=2pt of group] (linear)
  {Cell Embedding, Linear$(\,3{+}3 \to d\,)$};

  \node[anchor=south, font=\footnotesize\bfseries, text=black,
    draw=groupGrayEdge, thick, rounded corners=2pt,
    fill=groupGray, inner xsep=3pt, inner ysep=1pt] (grayTitle)
    at (grayPanelTop) {Feature Grouping \& Cell Embedding};

  \begin{scope}[on background layer]
    \node[fit=(grayPanelTop)(stdblk)(group)(linear), minimum width=\stageW,
      fill=groupGray, draw=groupGrayEdge, thick,
      rounded corners=4pt, inner xsep=10pt, inner ysep=4pt] (grayGroup) {};
  \end{scope}

  \draw[arrow] (mainSplit) -- (grayTitle.north);

  \node[nanCell, fill=red!10] (nm1) at (4.0, 6.30) {0};
  \node[nanCell, fill=red!25] (nm2) at (4.3, 6.30) {1};
  \node[nanCell, fill=red!10] (nm3) at (4.6, 6.30) {0};
  \node[anchor=south, font=\scriptsize, text=red!70!black]
    at (4.3, 6.55) {NaN/Inf mask};

  \draw[arrow] (mainSplit) -- (nm1.west);
  \draw[arrow] (nm2.south) |- (linear.east);

  \coordinate (addPos) at (0, 3.75);
  \node[plus, fill=orange!25, draw=orange!70!black] (taeplus) at (addPos) {$+$};

  \begin{scope}[shift={(\orthXshift, 3.75)}]
    \foreach \i/\col in {0/cA, 1/cB, 2/cX, 3/cD} {
      \filldraw[draw=black!40, line width=0.15pt, fill=\col]
      (-0.42, {0.18 - \i*0.09}) rectangle (0.42, {0.10 - \i*0.09});
    }
    \node[draw=blue!60!black, thick, rounded corners=2pt,
    minimum width=1.05cm, minimum height=0.50cm, inner sep=0pt]
    (taeTable) at (0, 0) {};
    \node[anchor=east, font=\scriptsize, text=black] at (1.8, 0) {$\leftarrow\!y_i \in \mathcal{Y}$};
    \node[anchor=south, font=\tiny, text=black, align=center,
    inner sep=0.5pt]
    at (0, 0.25) {Label Embedding (trainable orth.)};
  \end{scope}

  \draw[bluearrow] (taeTable.west) -- (taeplus.east);
  \draw[thick, black] (linear.south) -- (taeplus.north);

  \begin{scope}[yshift=0.6cm]
    \begin{scope}[yshift=1.1cm]
      \coordinate (colCenter) at (0, 0.45);

      \node[anchor=south, font=\footnotesize\bfseries, text=black,
        draw=groupCyanEdge, thick, rounded corners=2pt,
      fill=groupCyan, inner xsep=3pt, inner ysep=1pt] (cyanTitle)
      at ($(colCenter) + (0, 0.95)$)
      {Feature Embedding \tiny{(col-wise)}};
      \node[draw=groupCyanEdge, thick, fill=white, rounded corners=2pt,
        font=\scriptsize\bfseries, text=black,
        inner sep=2pt, anchor=north east]
      at (3.35, 1.49) {$\times\,3$};

      \foreach \g in {2, 1} {
        \pgfmathsetmacro{\dx}{\g * 0.16}
        \pgfmathsetmacro{\dy}{\g * 0.10}
        \pgfmathsetmacro{\sc}{0.90 - \g * 0.13}
        \pgfmathsetmacro{\tealpct}{int(40 * \sc)}
        \pgfmathsetmacro{\cyanpct}{int(40 * \sc)}
        \pgfmathsetmacro{\tealdraw}{int(70 * \sc)}
        \pgfmathsetmacro{\bluedraw}{int(60 * \sc)}
        \foreach \r in {0, 1, 2, 3, 4} {
          \pgfmathsetmacro{\yc}{0.95 - \r * 0.24 + \dy}
          \pgfmathsetmacro{\xc}{-1.50 + \dx}
          \filldraw[draw=teal!\tealdraw, line width=0.3pt, fill=teal!\tealpct,
          rounded corners=1pt]
          (\xc - 0.13, \yc - 0.10) rectangle (\xc + 0.13, \yc + 0.10);
        }
        \foreach \i in {0, 1, 2} {
          \pgfmathsetmacro{\yi}{0.71 - \i * 0.26 + \dy}
          \pgfmathsetmacro{\xi}{0.00 + \dx}
          \filldraw[draw=blue!\bluedraw, line width=0.3pt, fill=cyan!\cyanpct,
          rounded corners=1pt]
          (\xi - 0.13, \yi - 0.11) rectangle (\xi + 0.13, \yi + 0.11);
        }
        \foreach \r in {0, 1, 2, 3, 4} {
          \pgfmathsetmacro{\yc}{0.95 - \r * 0.24 + \dy}
          \pgfmathsetmacro{\xc}{1.45 + \dx}
          \filldraw[draw=teal!\tealdraw, line width=0.3pt, fill=teal!\tealpct,
          rounded corners=1pt]
          (\xc - 0.13, \yc - 0.10) rectangle (\xc + 0.13, \yc + 0.10);
        }
      }

      \foreach \r in {0, 1, 2, 3, 4} {
        \pgfmathsetmacro{\yc}{0.95 - \r * 0.24}
        \filldraw[draw=teal!80, line width=0.5pt, fill=teal!45, rounded corners=1pt]
        (-1.63, \yc - 0.10) rectangle (-1.37, \yc + 0.10);
      }
      \node[anchor=south, font=\tiny, text=black]
      at (-1.50, 1.10) {$N$ rows};

      \foreach \i in {0, 1, 2} {
        \pgfmathsetmacro{\yi}{0.71 - \i * 0.26}
        \filldraw[draw=blue!70, line width=0.5pt, fill=cyan!40, rounded corners=1pt]
        (-0.13, \yi - 0.11) rectangle (0.13, \yi + 0.11);
      }
      \coordinate(inducingSouth) at (0, 0.19);

      \node[anchor=south, font=\tiny, text=black, align=center]
      at (0.00, 0.94) {$K$ inducing};

      \foreach \r in {0, 1, 2, 3, 4} {
        \pgfmathsetmacro{\yc}{0.95 - \r * 0.24}
        \filldraw[draw=teal!80, line width=0.5pt, fill=teal!45, rounded corners=1pt]
        (1.32, \yc - 0.10) rectangle (1.58, \yc + 0.10);
      }
      \node[anchor=south, font=\tiny, text=black]
      at (1.45, 1.10) (colEmbOutRows) {$N$ rows};

      \coordinate (bcAnchor) at (1.60, -0.15);
      \begin{scope}[semithick, black!60]
        \draw ($(bcAnchor) + (-0.08, -0.05)$) -- ($(bcAnchor) + (-0.104, -0.012)$);
        \draw ($(bcAnchor) + (-0.08, -0.05)$) -- ($(bcAnchor) + (0.24, 0.15)$);
        \draw ($(bcAnchor) + (0.24, 0.15)$) -- ($(bcAnchor) + (0.216, 0.188)$);
        \draw ($(bcAnchor) + (0.24, 0.15)$) -- ($(bcAnchor) + (0.40, 0.25)$);
        \draw ($(bcAnchor) + (0.40, 0.25)$) -- ($(bcAnchor) + (0.376, 0.288)$);
        \node[anchor=south, font=\tiny, text=black, rotate=32] at ($(bcAnchor) + (0.3, -0.15)$)  {chunk};
      \end{scope}

      \coordinate (bcolAnchor) at (-1.30, -0.15);
      \begin{scope}[semithick, black!60]
        \draw ($(bcolAnchor) + (-0.08, -0.05)$) -- ($(bcolAnchor) + (-0.104, -0.012)$);
        \draw ($(bcolAnchor) + (-0.08, -0.05)$) -- ($(bcolAnchor) + (0.24, 0.15)$);
        \draw ($(bcolAnchor) + (0.24, 0.15)$) -- ($(bcolAnchor) + (0.40, 0.25)$);
        \draw ($(bcolAnchor) + (0.40, 0.25)$) -- ($(bcolAnchor) + (0.376, 0.288)$);
        \node[anchor=south, font=\tiny, text=black, rotate=32] at ($(bcolAnchor) + (0.3, -0.15)$)  {columns};
      \end{scope}

      \coordinate (funnelA) at (-0.75, 0.45);
      \begin{scope}[every path/.style={
        -, thin, draw=black, opacity=0.55, line cap=round}]
        \foreach \rowy in {0.95, 0.71, 0.47, 0.23, -0.01} {
          \draw (-1.35, \rowy) -- (funnelA);
        }
      \end{scope}
      \begin{scope}[every path/.style={
        -Stealth, thin, draw=black, opacity=0.75, line cap=round}]
        \foreach \indy in {0.71, 0.45, 0.19} {
          \draw (funnelA) -- (-0.15, \indy);
        }
      \end{scope}

      \coordinate (funnelB) at (0.75, 0.45);
      \begin{scope}[every path/.style={
        -, thin, draw=black, opacity=0.55, line cap=round}]
        \foreach \indy in {0.71, 0.45, 0.19} {
          \draw (0.15, \indy) -- (funnelB);
        }
      \end{scope}
      \begin{scope}[every path/.style={
        -Stealth, thin, draw=black, opacity=0.75, line cap=round}]
        \foreach \rowy in {0.95, 0.71, 0.47, 0.23, -0.01} {
          \draw (funnelB) -- (1.30, \rowy);
        }
      \end{scope}

      \node[anchor=west, font=\scriptsize, text=black, align=left,
      text width=2.9cm]
      at (3.70, 0.45)
      {applied per column\\in parallel
        \tikz[baseline=-0.5ex]{\draw[dash pattern={on 2pt off 1pt},black,thin](0,0)--(9pt,0);}
      \\ or chunked~\tikz[baseline=-0.5ex]{\draw[dotted,black,thick](0,0)--(8pt,0);}};

      \node[fit={(-2.05, -0.25) (2.05, 1.40)}, inner sep=0pt] (tfcol) {};
    \end{scope}

    \begin{pgfonlayer}{midground}
      \draw[arrow] (taeplus.south) -- (taeplus.south |- cyanTitle.north);
    \end{pgfonlayer}

    \begin{scope}[yshift=-0.4cm]
      \coordinate (featRow) at (0, -1.60);

      \foreach \g in {2, 1} {
        \pgfmathsetmacro{\dx}{\g * 0.16}
        \pgfmathsetmacro{\dy}{\g * 0.10}
        \pgfmathsetmacro{\sc}{0.45 - \g * 0.13}
        \pgfmathsetmacro{\bluepct}{int(100 * \sc)}
        \pgfmathsetmacro{\tealpct}{int(65 * \sc)}
        \pgfmathsetmacro{\bluedraw}{int(120 * \sc)}
        \pgfmathsetmacro{\tealdraw}{int(120 * \sc)}
        \foreach \k in {0, 1, 2, 3} {
          \pgfmathsetmacro{\xc}{-2.30 + \k * 0.43 + \dx}
          \filldraw[draw=blue!\bluedraw, line width=0.3pt, fill=blue!\bluepct,
          rounded corners=0.5pt]
          (\xc, -1.85 + \dy) rectangle (\xc + 0.36, -1.35 + \dy);
        }
        \foreach \k in {0, 1, 2, 3, 4, 5} {
          \pgfmathsetmacro{\xf}{-0.40 + \k * 0.43 + \dx}
          \filldraw[draw=teal!\tealdraw, line width=0.3pt, fill=teal!\tealpct,
          rounded corners=0.5pt]
          (\xf, -1.85 + \dy) rectangle (\xf + 0.36, -1.35 + \dy);
        }
      }

      \foreach \k in {0, 1, 2, 3} {
        \pgfmathsetmacro{\xc}{-2.30 + \k * 0.43}
        \filldraw[draw=blue!70!black, line width=0.4pt, fill=blue!40,
        rounded corners=0.5pt]
        (\xc, -1.85) rectangle (\xc + 0.36, -1.35);
        \node[font=\tiny\bfseries, text=white] at (\xc + 0.18, -1.60) {C};
      }
      \foreach \k in {0, 1, 2, 3, 4, 5} {
        \pgfmathsetmacro{\xf}{-0.40 + \k * 0.43}
        \filldraw[draw=teal!80, line width=0.3pt, fill=teal!25,
        rounded corners=0.5pt]
        (\xf, -1.85) rectangle (\xf + 0.36, -1.35);
      }

      \coordinate (brAnchor) at (2.15, -1.88);
      \begin{scope}[semithick, black!60]
        \draw ($(brAnchor) + (-0.08, -0.05)$) -- ($(brAnchor) + (-0.104, -0.012)$);
        \draw ($(brAnchor) + (-0.08, -0.05)$) -- ($(brAnchor) + (0.24, 0.15)$);
        \draw ($(brAnchor) + (0.24, 0.15)$) -- ($(brAnchor) + (0.216, 0.188)$);
        \draw ($(brAnchor) + (0.24, 0.15)$) -- ($(brAnchor) + (0.40, 0.25)$);
        \draw ($(brAnchor) + (0.40, 0.25)$) -- ($(brAnchor) + (0.376, 0.288)$);
        \node[anchor=south, font=\tiny, text=black, rotate=32] at ($(brAnchor) + (0.3, -0.15)$)  {chunk};
      \end{scope}

      \draw[decorate, decoration={brace, mirror, amplitude=3pt}, black]
      (-0.42, -1.87) -- (2.13, -1.87);
      \node[anchor=north, font=\scriptsize, text=black]
      at (0.85, -1.82) {\tiny{columns}};
    \end{scope}

    \begin{scope}[yshift=0.15cm]
      \node[anchor=south, font=\footnotesize\bfseries, text=black,
        draw=groupGreenEdge, thick, rounded corners=2pt,
      fill=groupGreen, inner xsep=3pt, inner ysep=1pt] (greenTitle)
      at (0, -1.05) {Feature Aggregation \tiny{(row-wise)}};
      \node[draw=groupGreenEdge, thick, fill=white, rounded corners=2pt,
        font=\scriptsize\bfseries, text=black,
        inner sep=2pt, anchor=north east]
      at (3.35, -0.96) {$\times\,3$};

      \coordinate (bowAnchor) at (-0.30, -1.15);
      \begin{scope}[every path/.style={
            -Stealth, thick, draw=black, opacity=0.55,
        line cap=round}]
        \foreach \k in {0, 1, 2, 3} {
          \pgfmathsetmacro{\xc}{-2.12 + \k * 0.43}
          \draw (bowAnchor) to [out=180,in=70 - 4 * \k] (\xc, -1.75);
        }
        \foreach \k in {0, 1, 2, 3, 4, 5} {
          \pgfmathsetmacro{\xf}{-0.22 + \k * 0.43}
          \draw (bowAnchor)to [out=0,in=100 + 4 * \k] (\xf, -1.75);
        }

      \end{scope}

      \node[anchor=west, font=\scriptsize, text=black, align=left,
      text width=2.7cm]
      at (3.70, -1.80)
      {applied per row\\
      in parallel \\ or chunked\\ (recomputing column embeddings)};

      \node[fit={(-2.55, -1.05) (2.55, -2.50)}, inner sep=0pt] (tfrow) {};

      \pgfmathsetmacro{\yc}{-3.05}
      \filldraw[draw=teal!85!black, line width=0.4pt, fill=teal!30, rounded corners=0.5pt] (-0.88, {\yc - 0.21}) rectangle (-0.88 + 0.43 * 3 + 0.36 + 0.1, {\yc + 0.21});
      \foreach \k in {0,1,2,3} {
        \pgfmathsetmacro{\xc}{-0.83 + \k * 0.43}
        \filldraw[draw=blue!70!black, line width=0.4pt, fill=blue!40,
        rounded corners=0.5pt]
        (\xc, {\yc - 0.14}) rectangle (\xc + 0.36, {\yc + 0.14});
        \node[font=\tiny\bfseries, text=white]
        at (\xc + 0.18, \yc) {C};
      }
      \node[anchor=west, font=\scriptsize, text=black, align=left]
      at (0.95, \yc)
      {4 CLS per row\\flatten $\to 4d$};

      \draw[arrow] (-1.475, -1.85) |- (-0.88, \yc);

      \coordinate (addPos) at (0, -3.60);
      \node[plus, fill=orange!25, draw=orange!70!black] (iclplus) at (addPos) {$+$};

      \draw[thick, black] (0, \yc -0.2) -- (iclplus.north);

      \begin{scope}[shift={(\orthXshift, -3.60)}]
        \foreach \i/\col in {0/cA2, 1/cB2, 2/cX2, 3/cD2} {
          \filldraw[draw=black!40, line width=0.15pt, fill=\col]
          (-0.42, {0.18 - \i*0.09}) rectangle (0.42, {0.10 - \i*0.09});
        }
        \node[draw=blue!70!black, thick, rounded corners=2pt,
        minimum width=1.05cm, minimum height=0.50cm, inner sep=0pt]
        (iclTable) at (0, 0) {};
        \node[anchor=east, font=\scriptsize, text=black] at (1.8, 0) {$\leftarrow\!y_i \in \mathcal{Y}$};
        \node[anchor=south, font=\tiny, text=black, align=center]
        at (0, 0.28) {Label Embedding (trainable orth.)};
      \end{scope}

      \draw[bluearrow] (iclTable.west) -- (iclplus.east);
      \begin{scope}[yshift=0.4cm]
        \def\iclTitleY{-4.65}      %
        \def\iclTrainYC{-5.35}     %
        \def\iclTestYC{-6.05}      %
        \def\iclBarHH{0.20}        %
        \def\iclBarHW{0.40}        %
        \def\iclBotY{-6.35}        %

        \foreach \i/\xc in {1/-1.50, 2/-0.50, 3/0.50, 4/1.50} {
          \filldraw[draw=groupOrangeEdge!75, line width=0.4pt,
            fill=groupOrangeEdge!22, rounded corners=0.5pt]
            (\xc - \iclBarHW, \iclTrainYC - \iclBarHH)
            rectangle
            (\xc + \iclBarHW, \iclTrainYC + \iclBarHH);
          \node[font=\scriptsize, text=black, inner sep=0pt]
            at (\xc, \iclTrainYC)
            {$h^{\mathrm{train}}_{\i}$};
          \coordinate (trBar\i N) at (\xc, \iclTrainYC + \iclBarHH);
          \coordinate (trBar\i S) at (\xc, \iclTrainYC - \iclBarHH);
        }
        \foreach \i/\xc in {1/-0.50, 2/0.50} {
          \filldraw[draw=groupOrangeEdge!90, line width=0.4pt,
            fill=groupOrangeEdge!42, rounded corners=0.5pt]
            (\xc - \iclBarHW, \iclTestYC - \iclBarHH)
            rectangle
            (\xc + \iclBarHW, \iclTestYC + \iclBarHH);
          \node[font=\scriptsize, text=black, inner sep=0pt]
            at (\xc, \iclTestYC)
            {$h^{\mathrm{test}}_{\i}$};
          \coordinate (teBar\i N) at (\xc, \iclTestYC + \iclBarHH);
          \coordinate (teBar\i S) at (\xc, \iclTestYC - \iclBarHH);
        }

        \coordinate (iclBowAnchor) at (0, -4.85);
        \begin{scope}[every path/.style={
          -Stealth, thick, draw=black, opacity=0.55, line cap=round}]
          \foreach \k/\xc in {0/-1.50, 1/-0.50} {
            \pgfmathsetmacro{\inAng}{70 - 12 * \k}
            \draw (iclBowAnchor) to[out=180, in=\inAng]
              (\xc, {\iclTrainYC + \iclBarHH});
          }
          \foreach \k/\xc in {0/0.50, 1/1.50} {
            \pgfmathsetmacro{\inAng}{110 + 12 * \k}
            \draw (iclBowAnchor) to[out=0, in=\inAng]
              (\xc, {\iclTrainYC + \iclBarHH});
          }
        \end{scope}
        \node[anchor=west, font=\scriptsize, text=black, align=left]
          at (3.70, -5.10)
          {train$\,\leftrightarrow\,$train\\\textit{multi-head self-attn}};

        \begin{scope}[every path/.style={
          -Stealth, thin, black, line cap=round}]
          \foreach \teX in {-0.50, 0.50} {
            \foreach \trX in {-1.50, -0.50, 0.50, 1.50} {
              \draw[opacity=0.65]
                (\teX, \iclTestYC + \iclBarHH)
                -- (\trX, \iclTrainYC - \iclBarHH);
            }
          }
        \end{scope}
        \node[anchor=west, font=\scriptsize, text=black, align=left]
          at (3.70, -6.00)
          {test$\,\rightarrow\,$train\\\textit{multi-query cross-attn}};

        \node[anchor=south, font=\footnotesize\bfseries, text=black,
          draw=groupOrangeEdge, thick, rounded corners=2pt,
          fill=groupOrange, inner xsep=3pt, inner ysep=1pt] (orangeTitle)
          at (0, \iclTitleY)
          {In-Context Learning};
        \node[draw=groupOrangeEdge, thick, fill=white, rounded corners=2pt,
          font=\scriptsize\bfseries, text=black, inner sep=2pt,
          anchor=north east]
          at (3.35, -4.56) {$\times\,24$};

        \node[fit={(-2.00, \iclBotY) (2.00, \iclTitleY)},
          inner sep=0pt] (tficl) {};

        \begin{pgfonlayer}{midground}
          \draw[arrow] (iclplus.south) -- (orangeTitle.north);
        \end{pgfonlayer}

        \coordinate (decTop) at (0, -7.40);

        \def\trainx{-1.55}
        \def\testx{1.20}
        \def\rowdy{0.44}
        \def\rowytop{-7.60}

        \node[anchor=south, font=\footnotesize\bfseries, text=black,
          draw=groupPurpleEdge, thick, rounded corners=2pt,
        fill=groupPurple, inner xsep=3pt, inner ysep=1pt] (purpleTitle)
        at (0, -7.40) {Many-Class Decoder};

        \begin{pgfonlayer}{midground}
          \draw[arrow] (0, {\iclBotY - 0.14}) -- (purpleTitle.north);
        \end{pgfonlayer}

        \foreach \i/\cls/\letter/\angA/\angB in {
          1/cA/0/30/60,
          2/cB/1/120/150,
          3/cX/2/200/230,
          4/cA/0/10/40
        } {
          \pgfmathsetmacro{\rowy}{\rowytop - (\i - 1) * \rowdy}
          \node[anchor=east, font=\scriptsize] at ($(\trainx - 0.28, \rowy)$)
          {$h^{\mathrm{train}}_{\i}$};
          \filldraw[draw=\cls!60!black, line width=0.4pt, fill=\cls, rounded corners=1.5pt]
          ($(\trainx - 0.18, \rowy - 0.15)$) rectangle
          ($(\trainx + 0.48, \rowy + 0.15)$);
          \node[font=\scriptsize\bfseries, text=white, anchor=center]
          at ($(\trainx + 0.15, \rowy)$) {\letter};
          \foreach \k/\ang in {0/\angA, 1/\angB} {
            \pgfmathsetmacro{\cx}{\trainx + 0.68 + \k*0.32}
            \filldraw[draw=violet!50!black, line width=0.3pt, fill=violet!15]
            ($(\cx - 0.14, \rowy - 0.17)$) rectangle
            ($(\cx + 0.14, \rowy + 0.17)$);
            \draw[-{Stealth[length=3pt, width=2.4pt]},
              line width=0.4pt, violet!55!black]
              ($(\cx, \rowy) + ({0.09*cos(\ang+180)}, {0.09*sin(\ang+180)})$) --
              ($(\cx, \rowy) + ({0.09*cos(\ang)},     {0.09*sin(\ang)})$);
          }
          \coordinate (tr\i) at ($(\trainx + 1.14, \rowy)$);
        }

        \def\testytop{-7.90}
        \def\testdy{0.88}
        \foreach \i/\angA/\angB in {1/25/55, 2/140/170} {
          \pgfmathsetmacro{\rowy}{\testytop - (\i - 1) * \testdy}
          \foreach \k/\ang in {0/\angA, 1/\angB} {
            \pgfmathsetmacro{\cx}{\testx + \k*0.32 - 0.16}
            \filldraw[draw=violet!60!black, line width=0.3pt, fill=violet!22]
            ($(\cx - 0.14, \rowy - 0.17)$) rectangle
            ($(\cx + 0.14, \rowy + 0.17)$);
            \draw[-{Stealth[length=3pt, width=2.4pt]},
              line width=0.4pt, violet!65!black]
              ($(\cx, \rowy) + ({0.09*cos(\ang+180)}, {0.09*sin(\ang+180)})$) --
              ($(\cx, \rowy) + ({0.09*cos(\ang)},     {0.09*sin(\ang)})$);
          }
          \node[anchor=west, font=\scriptsize] at ($(\testx + 0.32, \rowy)$)
          {$h^{\mathrm{test}}_{\i}$};
          \coordinate (te\i) at ($(\testx - 0.30, \rowy)$);
          \coordinate (teR\i) at ($(\testx + 0.85, \rowy)$);
        }

        \draw[cA, line width=1.4pt, opacity=0.85] (te1) -- (tr1);
        \draw[cA, line width=1.0pt, opacity=0.65] (te1) -- (tr4);
        \draw[cB, line width=1.3pt, opacity=0.80] (te2) -- (tr2);
        \draw[cX, line width=1.0pt, opacity=0.70] (te2) -- (tr3);

        \node[anchor=west, font=\scriptsize, text=black, align=left]
          at (3.70, -7.95)
          {Attention-weighted\\ average of one-hot \\ encoded labels $(y_i)$};

        \foreach \i/\wA/\wB/\wC/\wD in {
          1/1.00/0.00/0.00/0.00,
          2/0.00/0.60/0.40/0.00
        } {
          \pgfmathsetmacro{\rowy}{\testytop - (\i - 1) * \testdy}
          \pgfmathsetmacro{\xstart}{\testx + 1.00}
          \pgfmathsetmacro{\barH}{0.12}
          \pgfmathsetmacro{\barW}{0.75}
          \pgfmathsetmacro{\sA}{\wA * \barW}
          \pgfmathsetmacro{\sB}{\wB * \barW}
          \pgfmathsetmacro{\sC}{\wC * \barW}
          \pgfmathsetmacro{\sD}{\wD * \barW}
          \filldraw[fill=cA, draw=black!40, line width=0.2pt]
          (\xstart, \rowy - \barH) rectangle (\xstart + \sA, \rowy + \barH);
          \pgfmathsetmacro{\xa}{\xstart + \sA}
          \filldraw[fill=cB, draw=black!40, line width=0.2pt]
          (\xa, \rowy - \barH) rectangle (\xa + \sB, \rowy + \barH);
          \pgfmathsetmacro{\xb}{\xa + \sB}
          \filldraw[fill=cX, draw=black!40, line width=0.2pt]
          (\xb, \rowy - \barH) rectangle (\xb + \sC, \rowy + \barH);
          \pgfmathsetmacro{\xc}{\xb + \sC}
          \filldraw[fill=cD, draw=black!40, line width=0.2pt]
          (\xc, \rowy - \barH) rectangle (\xc + \sD, \rowy + \barH);
        }
        \node[anchor=south, font=\scriptsize\bfseries, text=black, align=center]
        at ($(\testx + 1.40, \testytop + 0.2)$) {$p(y\!\mid\!h^{\mathrm{test}}_i)$};

        \node[anchor=north, font=\scriptsize, align=center] at (0, -9.55)
        {$\hat{y} \in \mathbb{R}^{N_\text{test} \times |\mathcal{Y}|}$ \,(class logits, many-class classification)};
        \draw[arrow]
        (0, -9.20) -- (0, -9.55);

        \node[fit={(\trainx - 1.50, -7.40) (\trainx - 0.50, -9.05)},
        inner sep=0pt] (purpleAnchorL) {};
        \node[fit={(\testx + 0.85, -7.40) (\testx + 1.85, -9.05)},
        inner sep=0pt] (purpleAnchorR) {};

      \end{scope}

      \def\stageW{6.8cm}
      \begin{scope}[on background layer]
        \node[fit=(tfcol), minimum width=\stageW,
          fill=groupCyan, draw=groupCyanEdge, thick,
        rounded corners=4pt, inner xsep=10pt, inner ysep=4pt] (cyanGroup) {};

        \node[fit=(tfrow), minimum width=\stageW,
          fill=groupGreen, draw=groupGreenEdge, thick,
        rounded corners=4pt, inner xsep=10pt, inner ysep=4pt] (greenGroup) {};

        \node[fit=(tficl), minimum width=\stageW,
          fill=groupOrange, draw=groupOrangeEdge, thick,
        rounded corners=4pt, inner xsep=10pt, inner ysep=4pt] (orangeGroup) {};

        \node[fit=(purpleAnchorL)(purpleAnchorR), minimum width=\stageW,
          fill=groupPurple, draw=groupPurpleEdge, thick,
        rounded corners=4pt, inner xsep=10pt, inner ysep=6pt] (purpleGroup) {};
      \end{scope}
    \end{scope}

    \begin{pgfonlayer}{midground}
      \node[plus, fill=orange!25, draw=orange!70!black] (cyansw) at (0, -0.05) {};
      \draw (0, -0.05) node[spdt, scale=0.3] {};
      \draw[arrow, dotted] (inducingSouth.south -| cyansw) -- node[right, pos=0.65, font=\tiny, text=black] {$3\mkern-6mu\times\mkern-6muC\mkern-6mu\times\mkern-6muK$} (cyansw.north);
      \draw[arrow, dash pattern={on 2pt off 1pt}] let
      \p1 = (colEmbOutRows |- cyanGroup.south),
      \p2 = (cyansw.east)
      in
      (\p1) -- node[right, pos=0.5, font=\tiny, text=black] {$C\mkern-6mu\times\mkern-6muN$} (\x1, \y2) -- (\p2);
      \draw[arrow] (cyansw.south) -- (greenTitle.north);
      \draw[dotted, thick] (linear.west) -- ++(-1.2, 0) coordinate (lhsCorner) -- (lhsCorner |- {$(cyansw.west) + (0.2, 0.4)$}) -- node[above=4pt, left, font=\tiny, text=black] {$(3\!+\!3)\mkern-6mu\times\mkern-6muC$} ($(cyansw.west) + (0.2, 0.4)$);

    \end{pgfonlayer}
  \end{scope}
\end{tikzpicture}

%% file: figures/tikz/scm_prior.tex
\definecolor{PriorInk}{HTML}{0C0C0C}
\definecolor{PriorOff}{HTML}{F8F8F9}
\definecolor{PriorPlum}{HTML}{101075}
\definecolor{PriorPurple}{HTML}{3F2670}
\definecolor{PriorMauve}{HTML}{B6698D}
\definecolor{PriorPlumLight}{HTML}{8585BB}
\definecolor{PriorPurpleSoft}{HTML}{C5BAD2}
\definecolor{PriorMauvePale}{HTML}{E5CFD9}
\definecolor{PriorYellow}{HTML}{FFD400}
\definecolor{HeaderBg}{HTML}{E0E8F5}
\colorlet{PriorGrid}{PriorPurpleSoft}
\colorlet{SourceColor}{PriorPlumLight!55!white}
\colorlet{HiddenColor}{PriorOff!92!PriorInk}
\colorlet{FeatColor}{PriorPlum}
\colorlet{TargetColor}{PriorYellow}
\colorlet{HidColor}{PriorPlumLight}
\colorlet{ChildGreen}{PriorMauvePale}
\colorlet{ArrowBlue}{PriorPurple}
\colorlet{MidBlue}{PriorPurple}
\colorlet{HdrLabel1}{PriorPlum}
\colorlet{HdrLabel2}{PriorPurple}

\begin{center}
\begin{tikzpicture}[
    >=Latex, font=\small,
    scmnode/.style={circle,draw=PriorInk!80,line width=0.5pt,
                    minimum size=7mm,inner sep=0pt},
    src/.style ={scmnode,fill=SourceColor},
    hid/.style ={scmnode,fill=HidColor,text=PriorInk},
    feat/.style={scmnode,fill=FeatColor,text=white},
    tgt/.style ={scmnode,fill=TargetColor,text=PriorInk},
    par/.style ={scmnode,fill=SourceColor},
    chld/.style={scmnode,fill=ChildGreen},
    edge/.style={->,line width=0.5pt,draw=PriorInk!70},
    flow/.style={-{Triangle[length=3mm,width=2.4mm]},
                 line width=1.4pt,draw=ArrowBlue,line cap=round},
    panelttl/.style={font=\bfseries\small,text=PriorInk,anchor=west},
    pcap/.style={font=\itshape\footnotesize,text=PriorInk!90},
    boxout/.style={draw=PriorInk!80,rounded corners=2pt,line width=0.5pt,
                   inner sep=6pt,fill=white},
    minicard/.style={draw=PriorGrid,rounded corners=2pt,
                     line width=0.4pt,fill=white},
    dashbox/.style={draw=PriorInk!70,dashed,rounded corners=2pt,
                    line width=0.5pt,fill=white,inner sep=0pt},
]

\begin{scope}[on background layer]
  \foreach \xa/\ya/\xb/\yb in {%
        0/8.5/7.5/12.5,
        8.0/8.5/15.5/12.5,
        0/4.0/15.5/8.0,
        0/0/7.5/3.5,
        8.0/0/15.5/3.5}{
    \fill[white,rounded corners=3pt] (\xa,\ya) rectangle (\xb,\yb);
    \begin{scope}
      \clip[rounded corners=3pt] (\xa,\ya) rectangle (\xb,\yb);
      \fill[HeaderBg] (\xa,{\yb-0.7}) rectangle (\xb,\yb);
    \end{scope}
    \draw[PriorGrid,line width=0.3pt] (\xa,{\yb-0.7}) -- (\xb,{\yb-0.7});
    \draw[PriorGrid,rounded corners=3pt,line width=0.6pt]
          (\xa,\ya) rectangle (\xb,\yb);
  }
\end{scope}

\node[panelttl] at (0.20,12.15) {Step 1: Sample Hyper-parameters};
\node[panelttl] at (8.20,12.15) {Step 2: Sample DAG};
\node[panelttl] at (0.20, 7.65) {Step 3: Compute SCM};
\node[panelttl] at (0.20, 3.15) {Step 4: Extract Dataset};
\node[panelttl] at (8.20, 3.15) {Step 5: Post-processing};

\begin{scope}[shift={(0,8.5)}]
  \draw[minicard] (0.40,2.10) rectangle (1.95,2.90);
  \begin{scope}
    \clip (0.40,2.10) rectangle (1.95,3.80);
    \fill[SourceColor!70] plot[domain=-1.5:1.5,samples=40]
          ({1.175 + \x*0.42},{2.20 + 0.60*exp(-\x*\x)})
          -- ({1.175+1.5*0.42},2.20) -- ({1.175-1.5*0.42},2.20) -- cycle;
    \draw[MidBlue,line width=0.6pt] plot[domain=-1.5:1.5,samples=40]
          ({1.175 + \x*0.42},{2.20 + 0.60*exp(-\x*\x)});
    \draw[dashed,gray!70] (1.175,2.20) -- (1.175,2.80);
  \end{scope}

  \draw[minicard] (0.40,0.90) rectangle (1.95,1.70);
  \foreach \i/\h/\c in {0/0.55/PriorPlum, 1/0.70/PriorPlum, 2/0.40/PriorYellow,
                        3/0.30/PriorMauve, 4/0.50/PriorPlum}{
    \fill[\c] ({0.55+\i*0.27},1.00) rectangle
              ({0.78+\i*0.27},{1.30+\h*0.45});
  }
  \draw[gray!70] (0.45,1.00) -- (1.90,1.00);

  \foreach \y in {2.7,2.3,1.5,1.1}{
    \draw[gray!55,line width=2pt,line cap=round] (2.20,\y) -- (3.10,\y);
  }
  \foreach \y/\f in {2.7/0.55, 2.3/0.30, 1.5/0.65, 1.1/0.45}{
    \draw[MidBlue,line width=2pt,line cap=round]
          (2.20,\y) -- ({2.20+\f*0.90},\y);
    \fill[white] ({2.20+\f*0.90},\y) circle (0.10);
    \draw[MidBlue,line width=0.6pt] ({2.20+\f*0.90},\y) circle (0.10);
  }

  \draw[flow] (3.35,1.90) -- (4.25,1.9)
       node[midway,above=2pt,font=\bfseries\footnotesize] {Sample};

  \node[boxout,anchor=west,font=\scriptsize,inner sep=5pt, thin]
       at (4.45,1.90) {%
    \begin{tabular}{l@{\hspace{0.5em}}r}
      \texttt{num\_rows:}       & $N$ \\
      \texttt{num\_features:}   & $P$ \\
      \texttt{num\_classes:}    & $C$ \\
      \multicolumn{2}{c}{$\dots$} \\
    \end{tabular}};
\end{scope}

\begin{scope}[shift={(8.0,8.5)}]
  \node[font=\scriptsize\bfseries,anchor=south,text=PriorInk]
        at (1.80,2.82) {DAG Sampler};
  \draw[minicard] (0.30,1.92) rectangle (3.30,2.78);
  \draw[minicard] (0.50,1.95) rectangle (1.55,2.75);
  \foreach \px/\py in {0.70/2.55, 1.35/2.55, 0.80/2.25, 1.30/2.30}{
    \fill[PriorPlum] (\px,\py) circle (0.06);
    \draw[PriorInk!70,line width=0.3pt] (\px,\py) circle (0.06);
  }
  \draw[->,line width=0.55pt] (1.65,2.40) -- (1.80,2.40);
  \draw[minicard] (1.85,1.95) rectangle (3.05,2.75);
  \foreach \sx/\sy/\tx/\ty in {%
      2.05/2.55/2.80/2.55,
      2.05/2.55/2.10/2.20,
      2.80/2.55/2.80/2.20,
      2.10/2.20/2.80/2.20}{
    \draw[line width=0.35pt,draw=PriorInk!70] (\sx,\sy) -- (\tx,\ty);
  }
  \draw[line width=0.65pt,draw=PriorYellow] (2.05,2.55) -- (2.80,2.20);
  \foreach \px/\py in {2.05/2.55, 2.80/2.55, 2.10/2.20, 2.80/2.20}{
    \fill[PriorPlum] (\px,\py) circle (0.06);
    \draw[PriorInk!70,line width=0.3pt] (\px,\py) circle (0.06);
  }

  \node[font=\scriptsize\bfseries,anchor=south west,text=PriorInk]
        at (0.42,1.30) {Noise processes};
  \node[font=\tiny\itshape,anchor=south east,text=PriorInk!60]
        at (3.18,1.30) {$\varepsilon_i$};
  \draw[minicard] (0.30,0.20) rectangle (3.30,1.25);

  \foreach \xs/\ys in {%
      0.55/1.18, 0.78/1.02, 0.98/1.15, 1.20/1.00, 1.42/1.19,
      1.65/1.05, 1.88/1.13, 2.10/1.20, 2.32/1.04, 2.55/1.16,
      2.78/1.08, 3.00/1.12}{
    \fill[PriorPlum] (\xs,\ys) circle (0.04);
  }
  \foreach \xs/\ys in {%
      0.55/0.78, 0.80/0.92, 1.00/0.81, 1.22/0.95, 1.45/0.83,
      1.68/0.89, 1.90/0.76, 2.12/0.93, 2.35/0.85, 2.58/0.79,
      2.80/0.91, 3.02/0.84}{
    \fill[PriorMauve] (\xs,\ys) circle (0.04);
  }
  \foreach \xs/\ys in {%
      0.50/0.68, 0.75/0.52, 0.95/0.65, 1.18/0.55, 1.40/0.70,
      1.62/0.58, 1.85/0.51, 2.08/0.66, 2.30/0.60, 2.52/0.54,
      2.75/0.69, 2.98/0.57}{
    \fill[PriorPurple] (\xs,\ys) circle (0.04);
  }
  \foreach \xs/\ys in {%
      0.55/0.42, 0.78/0.28, 1.00/0.38, 1.23/0.45, 1.45/0.30,
      1.68/0.40, 1.90/0.26, 2.12/0.36, 2.35/0.43, 2.57/0.29,
      2.80/0.41, 3.03/0.33}{
    \fill[PriorPlumLight] (\xs,\ys) circle (0.045);
  }

  \draw[flow] (3.40,2.37) -- (4.10,2.37);
  \draw[flow] (3.40,0.73) -- (4.10,0.73);

  \node[hid] (n1) at (4.55,1.65) {$1$};
  \node[hid] (n4) at (5.65,2.90) {$4$};
  \node[hid] (n2) at (5.65,1.65) {$2$};
  \node[hid] (n3) at (6.95,2.30) {$3$};
  \node[hid] (n5) at (6.20,0.55) {$5$};
  \foreach \u/\v in {n1/n4,n1/n2,n1/n5,n4/n2,n4/n3,n2/n3,n2/n5,n3/n5}
    \draw[edge] (\u) -- (\v);
\end{scope}

\begin{scope}[shift={(0,4.0)}]
  \node[hid,scale=0.85] (m1) at (0.80,1.85) {$1$};
  \node[hid,scale=0.85] (m4) at (1.65,2.75) {$4$};
  \node[hid,scale=0.85] (m2) at (1.65,1.85) {$2$};
  \node[hid,scale=0.85] (m3) at (2.65,2.25) {$3$};
  \node[hid,scale=0.85] (m5) at (2.10,1.05) {$5$};
  \foreach \u/\v in {m1/m4,m1/m2,m1/m5,m4/m2,m4/m3,m2/m3,m2/m5,m3/m5}
    \draw[edge] (\u) -- (\v);

  \draw[flow] (3.20,1.90) -- (3.95,1.90);

  \node[pcap] at (5.40,3.10) {zoom-in};
  \draw[dashbox] (4.15,0.85) rectangle (6.65,2.95);
  \node[par] (zp4) at (4.80,2.40) {$4$};
  \node[par] (zp2) at (6.00,2.40) {$2$};
  \node[chld] (zc3) at (5.40,1.30) {$3$};
  \draw[edge] (zp4) -- (zp2);
  \draw[edge] (zp4) -- (zc3);
  \draw[edge] (zp2) -- (zc3);
  \node[pcap,align=center] at (5.40,0.40)
        {Child node value combines\\\& aggregates parents};

  \node[anchor=west,font=\small\bfseries,text=PriorInk]
        at (7.30,2.75) {Per-node structural equation:};
  \node[anchor=west,font=\small] at (7.50,2.25)
        {General:\quad $X_i = f_i\bigl(\mathrm{pa}(X_i)\bigr) + \varepsilon_i$};
  \node[anchor=west,font=\small] at (7.50,1.65)
        {Specific (child~$3$):};
  \node[anchor=west,font=\small] at (7.85,1.15)
        {$X_{3} = f\!\bigl(X_{4}, X_{2}\bigr) + \varepsilon_{3}$};

  \node[pcap,anchor=west] at (7.50,0.45)
        {$\ast$ Computed in topological order over $G(V,E)$};
\end{scope}

\begin{scope}[shift={(0,0)}]
  \node[feat] (e1) at (0.95,1.30) {$1$};
  \node[feat] (e4) at (2.10,2.40) {$4$};
  \node[hid] (e2) at (2.10,1.30) {$2$};
  \node[tgt]  (e3) at (3.40,1.85) {$3$};
  \node[feat]  (e5) at (2.65,0.40) {$5$};
  \foreach \u/\v in {e1/e4,e1/e2,e1/e5,e4/e2,e4/e3,e2/e3,e2/e5,e3/e5}
    \draw[edge] (\u) -- (\v);

  \node[feat,scale=0.85] at (4.65,2.25) {};
  \node[anchor=west,font=\footnotesize] at (4.90,2.25) {Features ($X$)};
  \node[tgt,scale=0.85]  at (4.65,1.45) {};
  \node[anchor=west,font=\footnotesize] at (4.90,1.45) {Target ($Y$)};
  \node[hid,scale=0.85]  at (4.65,0.65) {};
  \node[anchor=west,font=\footnotesize] at (4.90,0.65) {Hidden};
\end{scope}

\begin{scope}[shift={(8.0,0)}]
  \def\cellW{0.375}
  \def\cellH{0.25}
  \begin{scope}[shift={(0.20,0.50)}]
    \foreach \i in {0,1,2,3}{
      \fill[gray!45] ({\i*\cellW},1.50)
                     rectangle ({(\i+1)*\cellW-0.03},1.75);
      \draw[gray!50,line width=0.3pt]
            ({\i*\cellW},1.50) rectangle ({(\i+1)*\cellW-0.03},1.75);
    }
    \foreach \r in {0,1,2,3}{
      \foreach \i in {0,1,2,3}{
        \fill[white] ({\i*\cellW},{1.20 - \r*\cellH})
                     rectangle ({(\i+1)*\cellW-0.03},{1.45 - \r*\cellH});
        \draw[gray!50,line width=0.3pt]
              ({\i*\cellW},{1.20 - \r*\cellH})
              rectangle ({(\i+1)*\cellW-0.03},{1.45 - \r*\cellH});
      }
    }
    \node[pcap] at (0.74,-0.10) {Raw data};
  \end{scope}

  \draw[flow] (1.85,1.50) -- (2.55,1.50);

  \draw[minicard] (2.65,0.75) rectangle (4.65,2.40);
  \foreach \cx/\cy in {3.10/1.925, 4.20/1.925, 3.10/1.225, 4.20/1.225}{
    \draw[minicard] ({\cx-0.30},{\cy-0.25}) rectangle ({\cx+0.30},{\cy+0.25});
  }
  \foreach \pt in {(2.95,1.75),(3.05,1.85),(3.15,1.95),(3.25,2.05)}
    \fill[PriorInk!85] \pt circle (0.024);
  \draw[->,line width=0.6pt] (4.05,1.75) -- (4.37,2.05);
  \foreach \i/\h in {0/0.16,1/0.26,2/0.14,3/0.28}{
    \fill[PriorInk!85] ({2.85+0.14*\i},1.08)
                       rectangle ({2.95+0.14*\i},{1.08+\h});
  }
  \draw[PriorInk!85,line width=0.6pt]
        (3.95,1.05) -- (4.10,1.05) -- (4.10,1.22) --
        (4.25,1.22) -- (4.25,1.40) -- (4.43,1.40);
  \node[pcap] at (3.65,0.35) {Post-processing};

  \draw[flow] (4.80,1.50) -- (5.50,1.50);

  \begin{scope}[shift={(5.60,0.50)}]
    \foreach \i/\c in {0/PriorPlum, 1/PriorPurple,
                       2/PriorMauve, 3/PriorYellow}{
      \fill[\c] ({\i*\cellW},1.50)
                rectangle ({(\i+1)*\cellW-0.03},1.75);
      \draw[gray!50,line width=0.3pt]
            ({\i*\cellW},1.50) rectangle ({(\i+1)*\cellW-0.03},1.75);
    }
    \foreach \r in {0,1,2,3}{
      \foreach \i in {0,1,2,3}{
        \fill[white] ({\i*\cellW},{1.20 - \r*\cellH})
                     rectangle ({(\i+1)*\cellW-0.03},{1.45 - \r*\cellH});
        \draw[gray!50,line width=0.3pt]
              ({\i*\cellW},{1.20 - \r*\cellH})
              rectangle ({(\i+1)*\cellW-0.03},{1.45 - \r*\cellH});
      }
    }
    \node[pcap] at (0.74,-0.18) {Synthetic Dataset};
  \end{scope}
\end{scope}

\draw[flow] (7.55,10.50) -- (7.95,10.50);   %
\draw[flow] (11.75,8.45) -- (11.75,8.05);   %
\draw[flow] (3.75,3.95)  -- (3.75,3.55);    %
\draw[flow] (7.55,1.75)  -- (7.95,1.75);    %

\end{tikzpicture}

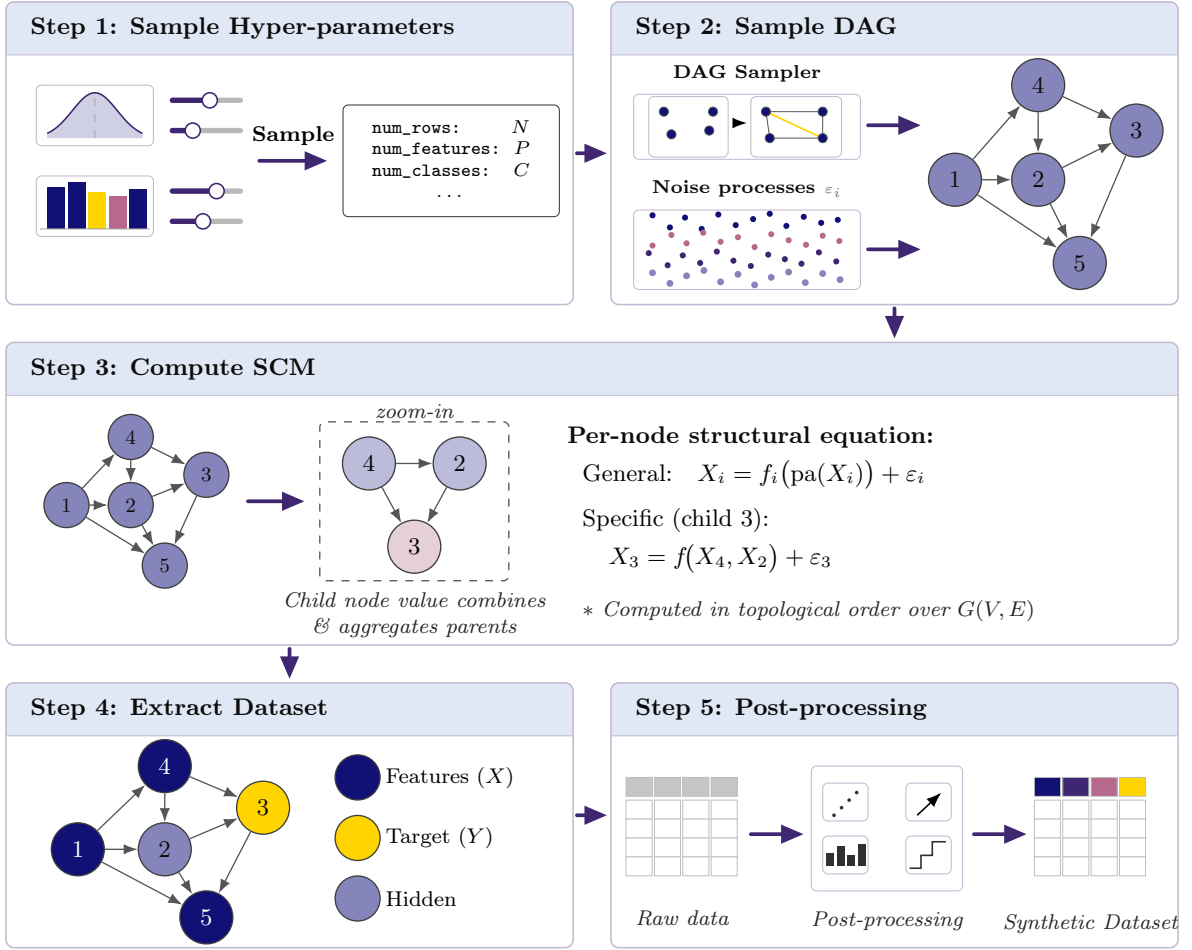
\captionof{figure}{\textbf{Schematic visualization of our SCM prior.} (i) We first sample high-level hyperparameters for the dataset, including number of features and number of rows. (ii) Based on the hyperparameters, we utilize our graph sampling algorithms to generate a directed acyclic graph (DAG) underlying our SCM; in parallel, an i.i.d.\ noise sample $\varepsilon_i$ is drawn per node (each colour shade in the lower-left mini-panel corresponds to a different node). (iii) We compute a topological ordering of the DAG. Based on this, we create a computational graph: First we fill root nodes (i.e.\ exogenous variables) and subsequently traverse the computational graph in topological order, combining parent nodes using our combiner mechanisms and activations to propagate values to the child nodes. (iv) We choose suitable features and target variables from our fully computed SCM. (v) We apply post-processing to the dataset.}  
\label{fig:prior}  \end{center}

%% file: sections/03_experimental_results.tex
\section{Experimental Results}\label{sec:results}

In this section, we report experimental results across a variety of benchmarks. In Section~\ref{subsec:results_tabular}, we focus on public tabular benchmarks: TabArena \citep{erickson2025tabarena}, TALENT \citep{talent_benchmark_jmlr}, and the text-tabular TabSTAR collection \citep{arazi_tabstar_2025}. Section~\ref{subsec:results_internal} describes internal benchmarks spanning various subtypes of tabular learning, including large-scale datasets and features, many-class classification, and quantile regression. The subsequent sections extend beyond classic tabular learning: Section~\ref{subsec:results_ts} addresses time-series data, Section~\ref{subsec:results_rel} covers relational learning, and Section~\ref{subsec:results_embed} focuses on embeddings.

\subsection{Public Tabular Benchmarks}\label{subsec:results_tabular}

\subsubsection{TabArena}
\label{sec:tabarena-result}

\begin{figure}[!t]
  \centering
  \includegraphics[width=\linewidth]{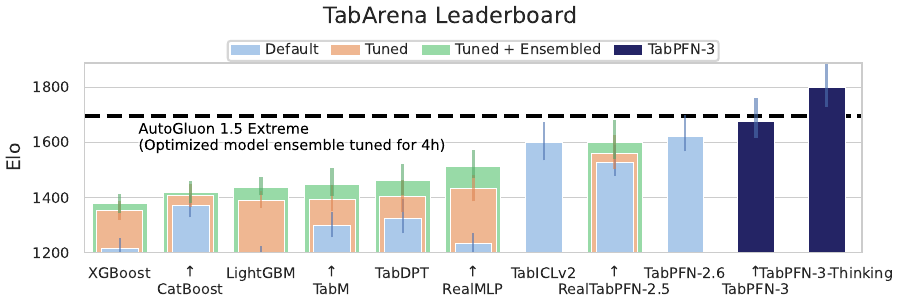}
  \caption{\textbf{TabPFN-3 performance on the standard TabArena benchmark} \citep{erickson2025tabarena}, including all 51 datasets (up to 100K rows). TabPFN-3 outperforms any other model in a forward pass, while \ourmodelenhanced strongly outperforms all existing methods, including AutoGluon 1.5 extreme \citep{autogluon_tabular}, a complex ensemble of models including TabPFN~v2 tuned for 4 hours, in less than a tenth of the runtime. 
  }
  \label{fig:tabarena_enhanced}
\end{figure}

TabArena \citep{erickson2025tabarena} (NeurIPS 2025 Datasets \& Benchmarks) is a recent  and heavily curated tabular benchmark, based on the largest number of candidate datasets considered, and created and maintained by open-source contributors from a wide range of institutions.
In particular, it compares a large and regularly updated list of recent models, including tree-based models like CatBoost \citep{prokhorenkova2018catboost}, LightGBM \citep{lightgbm} or XGBoost \citep{chen2016xgboost}, as well as newer deep-learning models like RealMLP \citep{holzmuller2024realmlp}, TabM \citep{gorishniy2024tabm}, ModernNCA \citep{ye2025revisitingnearestneighbortabular} or xRFM \citep{beaglehole2025xrfmaccuratescalableinterpretable}, the AutoML system AutoGluon \citep{autogluon_tabular}, and other Tabular Foundation Models like TabICL \citep{qu2025tabicl,qu2026tabiclv2}, TabDPT \cite{ma2025tabdptscalingtabularfoundation}, TabSTAR \cite{arazi_tabstar_2025}, LimiX \citep{zhang2025limix}, Mitra \citep{zhang2025mitramixedsyntheticpriors} or TabPFN~v2 \citep{Hollmann2025tabpfnv2}.
The benchmark contains a set of 51 datasets selected from 1053 to be representative of real-world tabular data. See \citet{erickson2025tabarena} for the list of datasets and Section \ref{app:tabarena-metrics} for definitions of TabArena's Elo and Improvability metrics.

\begin{figure}[!t]
  \centering
  \begin{minipage}[t]{0.48\linewidth}
    \centering
    \includegraphics[width=\linewidth]{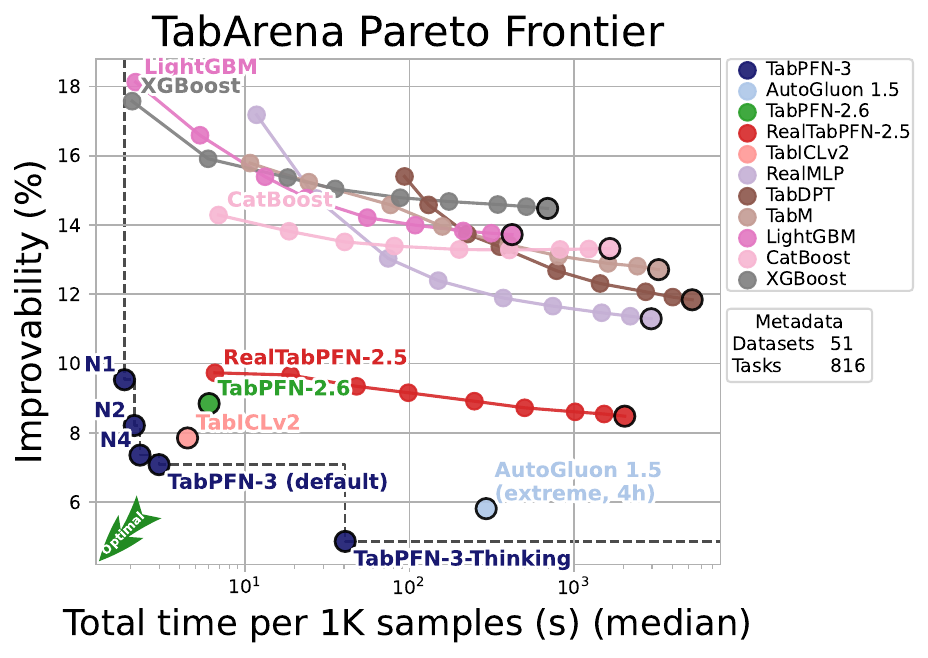}
    \caption{\textbf{Pareto frontier on TabArena}: trade-off between prediction quality and total training + inference cost. N1, N2, and N4 are TabPFN-3 versions with 1, 2, and 4 estimators. Improvability measures how much a model would improve by switching to the best model on each individual dataset, see Appendix \ref{app:tabarena-metrics}.}
    \label{fig:tabarena_pareto}
  \end{minipage}\hfill
  \begin{minipage}[t]{0.48\linewidth}
    \centering
    \includegraphics[width=\linewidth]{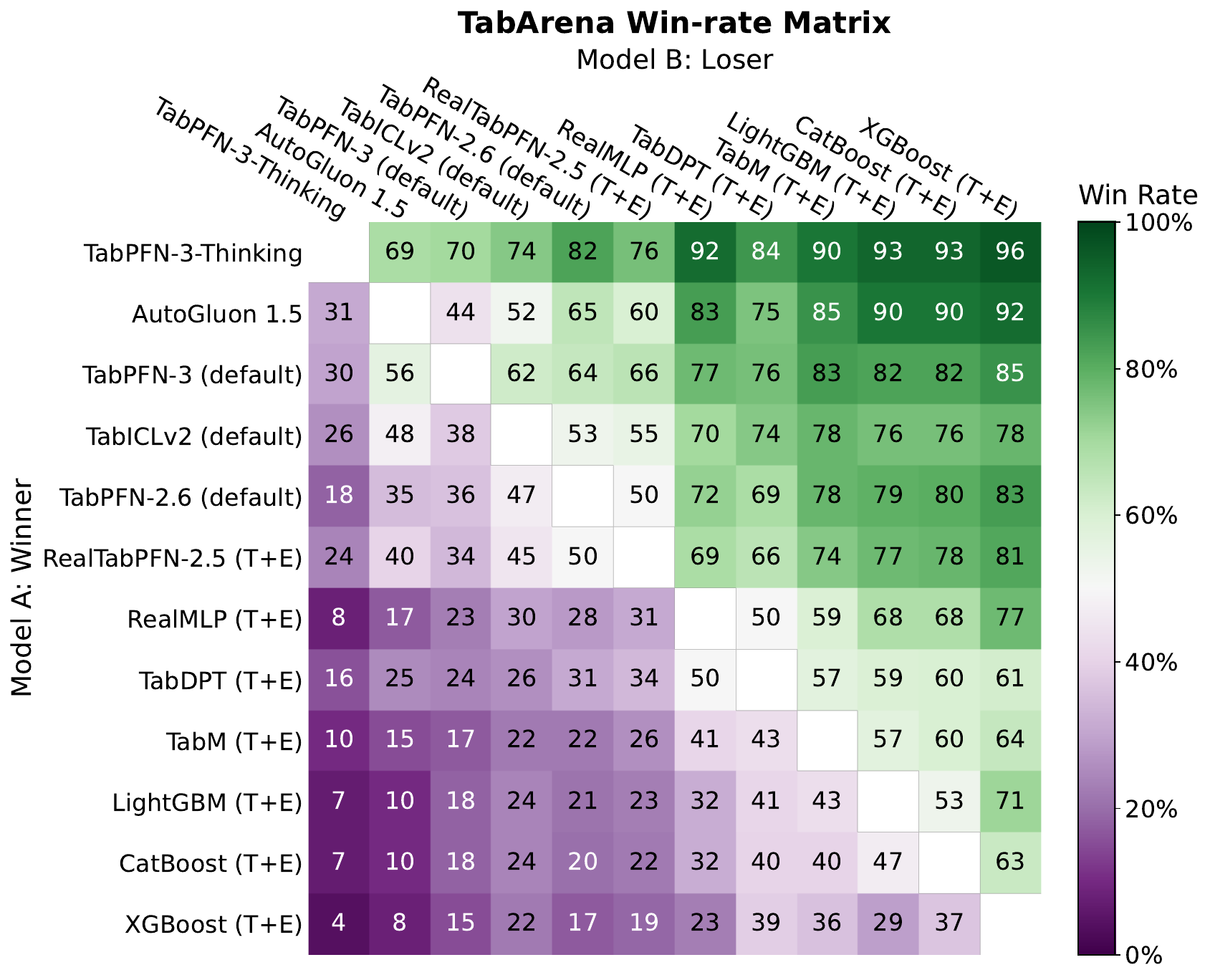}
    \caption{\textbf{Pairwise win rates on TabArena} for a curated set of the strongest models on TabArena. See Appendix \ref{sec:tabarena_leaderboard_tables} for the full results.}
    \label{fig:tabarena_leaderboard}
  \end{minipage}
\end{figure}

\paragraph{Pushing the performance frontier on TabArena.}
Figure \ref{fig:tabarena_enhanced} shows the performance of \ourmodel and \ourmodelenhanced on TabArena.
\ourmodel outperforms in one forward pass all other models, including tuned and ensembled baselines, by a significant margin, gaining 72 Elo points over our previous Real-TabPFN-2.5 tuned and ensembled.
\ourmodelenhanced, leveraging test-time computation, significantly outperforms open-source \ourmodel on TabArena, beating any non-TabPFN model (including tuned and ensembled baselines) by over 200 Elo points,
and outperforming AutoGluon 1.5 extreme, a complex ensemble of models including TabPFN~v2, tuned for 4 hours, by over 100 Elo points while being 10x faster.
Looking at the win rate matrix in Figure \ref{fig:tabarena_leaderboard}, we can see that \ourmodelplus with Thinking mode (respectively \ourmodel) has over 93\% (respectively 80\%) win rate against tuned and ensembled CatBoost, LightGBM and XGBoost, and a 69\% (respectively 56\%) win rate against AutoGluon 1.5 extreme tuned for 4 hours.

\paragraph{Dominating the time / performance Pareto-frontier.}
The strong results of our models are achieved while being much faster to train than the baselines.
On Figure \ref{fig:tabarena_pareto}, we can see that our model family, (\ourmodel with 1, 2, and 4 estimators and \ourmodelplus with Thinking mode) strictly dominates the combined training + inference time/performance pareto-frontier on TabArena by a large margin.

\paragraph{Scaling to larger datasets.}
\ourmodel was built to scale to large datasets, and \ourmodelenhanced benefits from this scalability. While TabArena only contains datasets up to 100k rows, we can still observe very strong performance on the 15 largest datasets in TabArena with between 10k and 100k rows, as shown in Figure \ref{fig:tabarena-hero-plot}.
In particular, on this subset \ourmodel outperforms any other model by 100 Elo, and \ourmodelenhanced dramatically outperforms any other non-TabPFN model (including tuned and ensembled baselines) by over 420 Elo points, and beats AutoGluon 1.5 extreme (4h) by 220 Elo points.
Looking at the win rate matrix in Figure \ref{fig:tabarena_winrate_medium}, \ourmodelenhanced has over 99\% win rate against tuned and ensembled LightGBM and XGBoost, 98\% win rate against CatBoost tuned and ensembled, and 82\% win rate against AutoGluon 1.5 extreme tuned for 4 hours. In Section \ref{sec:large-data}, we study the performance of our model beyond 100K rows, going up to 1M training rows.

\subsubsection{TALENT}
\label{sec:talent-result}

\begin{figure}[!t]
    \centering
    \includegraphics[width=0.9\linewidth]{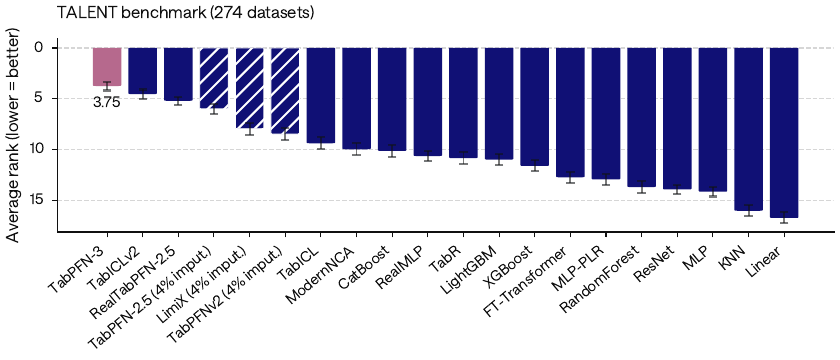}
    \caption{\textbf{Average rank on the TALENT benchmark, using the TabICLv2 evaluation protocol from \citet{qu2026tabiclv2}
      (274 datasets).} The original 300-dataset TALENT
      \citep{talent_benchmark_jmlr} minus the 26 development datasets used for TabPFN-2 / TabICLv2 development removed in the TabICLv2 paper), spanning regression, binary and multiclass classification. Bars show mean rank (lower is better); error bars are
      95\% bootstrap confidence intervals over datasets (see appendix
      \ref{app:TALENT}). Methods tagged \emph{(N imputed, X\%)} failed on some datasets and have that fraction of their score cells filled with K-nearest-neighbour values.}
    \label{fig:TALENT-tabicl-datasets}
  \end{figure}

The TALENT benchmark \citep{talent_benchmark_jmlr} provides a complementary view on the performance of \ourmodel. Instead of a smaller curated list of datasets, this benchmark uses a large number of diverse datasets (300) from a wide range of domains. The strong results of TabPFN-3 on this benchmark confirm the robustness of its performance.
Indeed, TabPFN-3 ranks first on the TALENT benchmark in aggregate, as shown in Figure \ref{fig:TALENT-tabicl-datasets}, as well as for
each task type (regression, binary and multiclass classification) in Figure \ref{fig:per-task-TALENT-rank}.

\subsubsection{TabSTAR}
\label{sec:tabstar-result}

\begin{figure}[!t]
  \centering
  \includegraphics[width=0.8\linewidth]{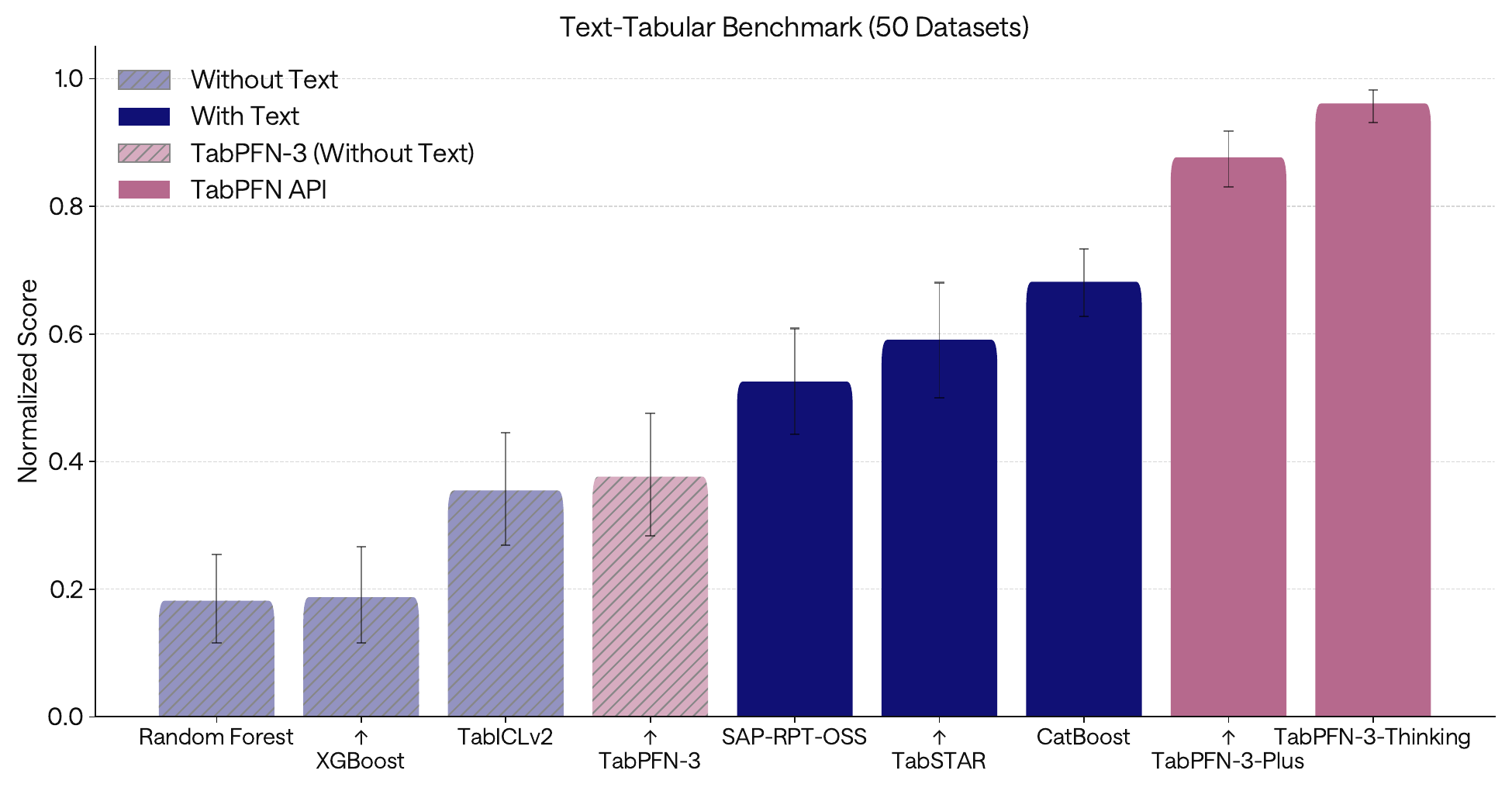}
  \caption{\textbf{Performance over the TabSTAR Text-Tabular Collection}. \ourmodelenhanced and TabPFN-3-Plus significantly outperform text-aware models such as CatBoost, TabSTAR and SAP-RPT-OSS. In turn, these models dominate over numerical-only baselines, for which TabPFN-3 gets the best results.}
  \label{fig:text_leaderboard}
\end{figure}

The TabSTAR study \cite{arazi_tabstar_2025} assembled 50 text-tabular datasets, gathered from previous work \cite{shi2021benchmarking, grinsztajn2023vectorizing, kim2024carte}. These datasets represent real world tasks, where at least one feature is text-based and cannot be faithfully represented without text processing methods. While the open-source version of \ourmodel only supports numerical and categorical variables, \ourmodelplus also offers native support for text features. We compare TabPFN API models with both text-aware models and numerical-only baselines.
Figure \ref{fig:text_leaderboard} shows that \ourmodelplus dominates the leaderboard by a significant margin, and combining our thinking mode with native text support pushes performance further. Furthermore, among models that omit text features due to lack of native support, \ourmodel remains the top performer. Appendix~\ref{app:TABSTAR} provides further details on the benchmark, as well as a performance breakdown by task type.

\subsection{Internal Benchmarks}\label{subsec:results_internal}

To complement the public TabArena~\citep{erickson2025tabarena} and
TALENT~\citep{talent_benchmark_jmlr} benchmarks, we evaluate TabPFN-3 on a
set of internal benchmarks designed to stress capabilities that are only
partially covered by existing public evaluations. These benchmarks test
whether TabPFN-3 pushes the frontier of tabular foundation models beyond the
small- and medium-data regimes emphasized in prior work. In particular, we
evaluate scaling to more than one million samples, high-dimensional feature
spaces, many-class classification, and quantile regression.

Our primary comparisons are against the leading gradient boosted tree frameworks XGBoost \citep{chen2016xgboost}, CatBoost \citep{prokhorenkova2018catboost}, and LightGBM \citep{lightgbm}, as well as TabICLv2 \citep{qu2026tabiclv2}, a recent foundation model for tabular data with strong results on public benchmarks.

\subsubsection{Large Data}
\label{sec:large-data}

\paragraph{Evaluation Protocol.} 
The primary baselines for our large-data evaluation are tree-based methods, which recent large-scale tabular benchmarks have shown to be highly competitive beyond 100,000 samples~\citep{talent_benchmark_jmlr}. Our large-data benchmarking effort focuses on datasets with 100,000 to 1 million training rows and up to 200 features. 

This benchmark targets the large-row regime for which \ourmodel was designed. As described in Section~\ref{sec:arch_overview}, \ourmodel first compresses feature information into fixed-dimensional row representations and subsequently performs in-context learning over these rows.
This architectural decomposition enables inference on datasets with up to one million rows on a single GPU.
At the same time, it induces a scaling trade-off: when both the number of rows and the number of features are very large, the early compression of feature information can become a bottleneck.
We treat the high-dimensional, low-sample regime as a separate evaluation setting, studied in Section~\ref{sec_meany_feats}, rather than conflating it with the large-row setting considered here.

Our benchmark datasets span diverse real-world domains including healthcare,
finance, logistics, and environmental science. For regression, the datasets in our benchmark exhibit temporal structure, where models are trained on past data and must generalize to future data. We found this setting to be the most common and representative of real-world deployment conditions.

\begin{figure}[!t]
    \centering
    \begin{subfigure}[t]{0.48\textwidth}
        \centering
        \includegraphics[width=\textwidth]{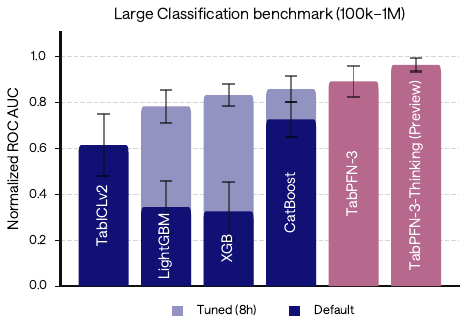}
    \end{subfigure}
    \begin{subfigure}[t]{0.48\textwidth}
    \centering
    \includegraphics[width=\textwidth]{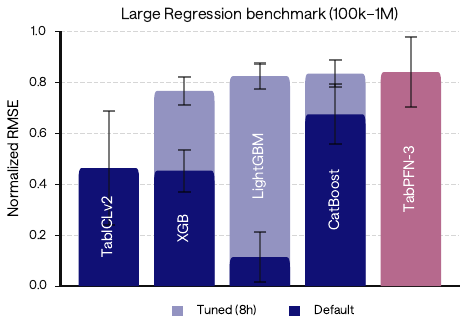}
\end{subfigure}
\caption{%
    \textbf{TabPFN-3 achieves state-of-the-art performance on the large-rows benchmark
    (up to 1M training rows and 200 features, 13 datasets)}, outperforming both default and 8-hour-tuned
    gradient-boosted tree baselines as well as TabICLv2 in a single forward pass.
    \textbf{(a)} Classification (9 datasets).
    \textbf{(b)} Regression (4 datasets) use temporal splits.
    Normalized scores are higher-is-better; see Section~\ref{sec:metric_norm} for the
    normalization procedure and Appendix~\ref{app:large_data_datasets} for critical
    difference diagrams.
}
\label{fig:large_data_all}
\end{figure}

\begin{figure}[!t]
    \centering
    \includegraphics[width=0.8\linewidth]{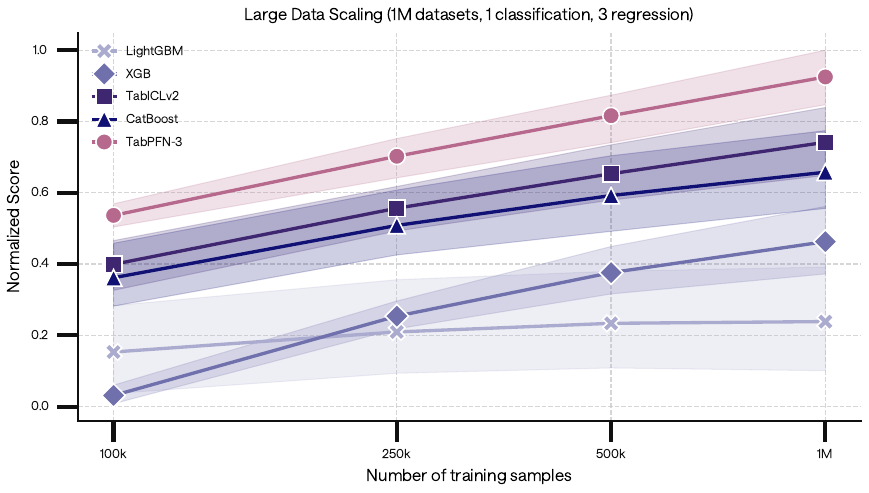}
    \caption{\textbf{\ourmodel{} tops the normalized scaling curves for ROC-AUC OvR classification and RMSE regression across dataset scales.} Results are shown on the four large-data benchmark datasets that reach at least 1M training rows (one classification, three regression). For each dataset we subsample the training set to 100k, 250k, 500k and 1M rows with 3 random repeats. Shaded bands are 95\% bootstrap confidence intervals across the four datasets and 3 repeats. .
    }
    \label{fig:large_data_scaling}
\end{figure}

\paragraph{Results.} \ourmodel achieves state-of-the-art performance on our large-data benchmark, outperforming default and 8-hour-tuned
gradient-boosted tree baselines in a single forward pass, as shown in Figure \ref{fig:large_data_all}. Further, we show a preview version of \ourmodelenhanced on large data, which improves \ourmodel performance further for classification datasets (as \ourmodelplus with Thinking mode does not yet support temporal datasets as of the time of writing, we could not evaluate it on our regression benchmark). To better understand how \ourmodel performance scales with training size, we report performance on subsampled versions of our datasets (keeping test set constant, and only considering datasets with 1M training samples) in Figure ~\ref{fig:large_data_scaling}. Across the 100k–1M range, TabPFN-3 scales smoothly and retains the top normalized score at every training-set size.

\paragraph{Large data results from TALENT benchmark.}

To confirm our internal results, we also extract the 14 available datasets in the TALENT benchmark
with more than 100K and less than 1M training samples (see Appendix \ref{app:large_data_datasets}).
On this subset, TabPFN-3 is again the best ranked model against the baselines provided by the TALENT benchmark,
as shown in Figure \ref{fig:large_rows_rank}.

\subsubsection{Many-Class Classification}
\label{sec:many-class-eval}

\ourmodel introduces a many-class decoder (Section~\ref{sec:many-class-decoder}) that we trained to support up to 160 classes, a regime where most tabular foundation models fail entirely. Creating a benchmark from real-world datasets with naturally many classes is challenging; we therefore evaluate on a synthetic benchmark derived by bucketing regression targets from real regression benchmark datasets. We also confirm the strong performance of \ourmodel on the 4 datasets from the TALENT benchmark that have more than 50 classes in Section \ref{app:TALENT-many-class}.

\paragraph{Synthetic many-class benchmark.}
We construct
a synthetic benchmark by converting the TabArena regression datasets into many-class
classification problems via jittered quantile binning; full construction details
are given in Appendix~\ref{app:many_class_construction}.
Figure~\ref{fig:many_class_synthetic} shows the ROC-AUC (OvR) and accuracy.
TabPFN-3 achieves the highest normalized ROC-AUC of $1.00$, ranking first overall and outperforming all baselines by a large margin.
On ROC-AUC (OvR), the next best model is TabICLv2 at $0.89$ using its many-class wrapper to go beyond its 10 classes limit. TabPFN-2.5 achieves $0.83$, using its own many-class error-correcting-code-based wrapper\footnote{\url{https://github.com/PriorLabs/tabpfn-extensions/tree/main/src/tabpfn_extensions/many_class}}.
Conventional tree-based methods and KNN all perform notably worse, even after $1$ hour of tuning.

\begin{figure}[!t]
    \centering
    \begin{subfigure}[t]{0.48\textwidth}
        \centering
        \includegraphics[width=\textwidth]{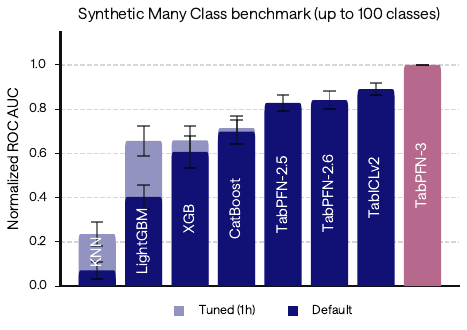}        
    \end{subfigure}
    \begin{subfigure}[t]{0.48\textwidth}
        \centering
        \includegraphics[width=\textwidth]{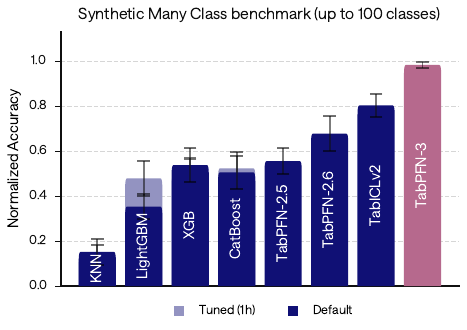}
    \end{subfigure}
    \caption{%
        \textbf{On the synthetic many-class benchmark \ourmodel achieves a normalized ROC-AUC (OvR) of $1.00$, outperforming all GBT baselines by a large margin}. The benchmark contains up to 100 classes, 9 datasets that are derived from TabArena regression tasks via Dirichlet-jittered quantile binning
        with shuffled labels). Normalized scores are higher-is-better; see Section~\ref{sec:metric_norm} for the normalization procedure. The corresponding Critical Difference diagram can be seen in Figure \ref{fig:many_class_roc_auc_cd}.
    }
    \label{fig:many_class_synthetic}
\end{figure}

\subsubsection{Many Features}
\label{sec_meany_feats}

The high-dimensional, low-sample regime poses a qualitatively different challenge from the large-row setting studied in Section~\ref{sec:large-data}. Whereas large-row benchmarks primarily test scalability to many training examples, the many-features setting tests robust generalization and feature-subset selection when the number of candidate features far exceeds the number of samples. 

We evaluate this setting on a dedicated \emph{many-features} slice of six real-world classification datasets with 100--320 samples, 1{,}100--22{,}200 features, and 2--4 classes, mostly from biomedical or gene-expression-style domains. Such large feature-to-sample ratios are challenging for tree-based methods because they increase the risk of selecting spurious feature interactions.

\begin{figure}[!htbp]
    \centering
    \begin{minipage}[t]{0.48\textwidth}
        \centering
        \includegraphics[width=\linewidth]{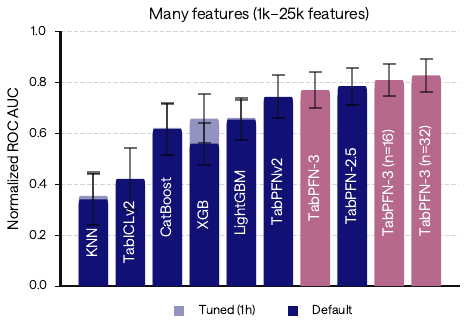}
        \caption{%
            \textbf{TabPFN scales well to high-dimensional, low-sample classification.}
            Normalized ROC-AUC on the many-features benchmark slice, consisting of 6 classification datasets with 102--322 samples and 1{,}117--22{,}215 features.
            This high-dimensional, low-sample regime is particularly challenging for standard tree-based baselines.
            Increasing the number of \ourmodel estimators improves feature-space coverage and substantially boosts performance.
        }
        \label{fig:many_feats_roc_auc}
    \end{minipage}\hfill
    \begin{minipage}[t]{0.48\textwidth}
        \centering
        \includegraphics[width=\linewidth]{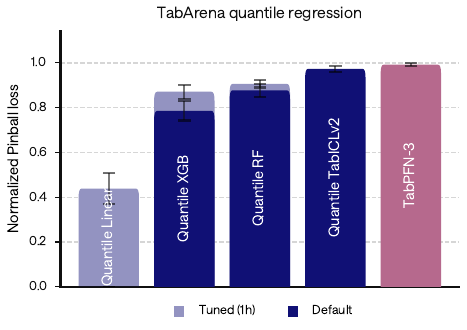}
        \caption{%
             \textbf{TabPFN-3 exhibits strong predictive distribution modeling on quantile regression.}
             Normalized pinball loss on our quantile regression benchmark, constructed from TabArena regression datasets and averaged across 10 quantile levels $q \in \{0.1, 0.2, \ldots, 0.9\}$~\cite{koenker_regression_quantiles}.
            Normalized scores are higher-is-better; see Section~\ref{sec:metric_norm} for
            the normalization procedure.
        }
        \label{fig:quantile_normalized}
    \end{minipage}
\end{figure}

Figure~\ref{fig:many_feats_roc_auc} shows that \ourmodel performs strongly on this challenging slice, reaching the best normalized ROC-AUC with 32 estimators.
Earlier TabPFN variants, in particular Real-TabPFN-2.5 and TabPFN~v2, also perform competitively, suggesting that TabPFN-style pretraining provides a robust inductive bias for high-dimensional, low-sample problems.

As described in Section~\ref{sec:preprocessing}, each \ourmodel estimator is restricted to at most 200 input features per default. Thus, for datasets with tens of thousands of raw features, individual estimators operate on feature subsets rather than compressing the full feature set.
At the same estimator budget, Real-TabPFN-2.5 can slightly outperform \ourmodel; we hypothesize that this reflects two factors: Real-TabPFN-2.5 uses up to 500 features per estimator, providing broader feature-space coverage on some datasets, and its alternating row-wise and feature-wise attention may better exploit the selected feature subset.
For \ourmodel, increasing the number of estimators improves coverage of the raw feature space and raises the probability that informative feature subsets are included.
In our OSS version, this estimator budget is scaled automatically for high-dimensional inputs, making the ensemble substantially more effective in this regime.

Overall, the many-features slice suggests that TabPFN estimators can be ensembled effectively in a high-noise feature-selection regime, where conventional tree-based methods are prone to overfitting to noisy or spurious feature interactions.

\subsubsection{Quantile Regression}
\label{sec:quantile_regression}

Beyond point predictions, TabPFN-3 provides full predictive distributions via a
bar-distribution regression head (Section~\ref{app:architecture-hyperparams}), from
which arbitrary quantiles are decoded at inference by inverting the predicted CDF ---
all from a single forward pass, with no retraining per quantile level. Since TabArena does not natively support quantile regression evaluation, we construct
a dedicated benchmark by downloading the TabArena regression datasets and evaluating
all models on pinball loss~\cite{koenker_regression_quantiles}, averaged across 10
quantile levels $q \in \{0.1, 0.2, \ldots, 0.9\}$. We compare against four baselines spanning the typical strategies for quantile regression: a linear quantile regressor, which fits a separate pinball-loss model per quantile level; XGBoost in quantile mode, which uses a single multi-output booster but adds one tree per quantile per boosting round, scaling training cost roughly linearly in the number of levels; quantile random forests~\cite{meinshausen2006qrf}, which train a single MSE-objective forest and read off all quantiles from leaf-level empirical CDFs at no extra training cost; and TabICL-v2, a tabular foundation model with a quantile head.

TabPFN-3 achieves a normalized pinball loss score very close to $1.00$, ranking first overall
and outperforming all baselines, demonstrating that the bar-distribution head produces
well-calibrated predictive distributions superior to dedicated quantile regression
baselines at no additional training cost per quantile level. The normalized Pinball loss is shown in Figure~\ref{fig:quantile_normalized}, while the corresponding Critical Difference plot can be found in the Appendix in Figure~\ref{fig:quantile_cd}.
\FloatBarrier

\subsection{Time-Series Forecasting}\label{subsec:results_ts}

In addition to the classification and regression checkpoints, we
release a new TabPFN-3 checkpoint for TabPFN-TS \citep{hoo2024tabpfn_ts} fine-tuned on \textbf{synthetic} time-series data for
probabilistic time-series forecasting. This checkpoint can be used in our \href{https://github.com/PriorLabs/tabpfn-time-series}{\texttt{tabpfn-time-series}} library. We evaluate it on fev-bench~\citep{shchur2025fev}, a benchmark containing 100 diverse time-series forecasting tasks.
Following this benchmark, we report win rates and skill scores relative to the Seasonal Naive baseline in Table \ref{tab:fev-bench} (full version in Appendix Table \ref{tab:fev-bench-full}).

\begin{table}[!htbp]
\centering
\caption{Forecasting performance on fev-bench (100 tasks),
sorted by skill score. 
\textbf{TabPFN-TS-3 ranks 2nd among foundation models on both SQL and MASE skill score while being trained only on synthetic data.}
The full 18-baseline leaderboard can be found in Appendix \ref{app:time_series}.}
\label{tab:fev-bench}
\vspace{-3mm}

\begin{minipage}{0.49\textwidth}
\centering
\textbf{(a) SQL (probabilistic)} \\[2pt]
\resizebox{\linewidth}{!}{\input{tables/time_series/leaderboard_SQL_subset}}
\end{minipage}
\hfill
\begin{minipage}{0.49\textwidth}
\centering
\textbf{(b) MASE (point)} \\[2pt]
\resizebox{\linewidth}{!}{\input{tables/time_series/leaderboard_MASE_subset}}
\end{minipage}
\end{table}

\footnotetext{The fev-bench
authors report 28.8 MASE skill for the original TabPFN-TS
\citep{shchur2025fev}; our re-run in Table~\ref{tab:fev-bench-full}
yields 27.6 — we report our own re-run for like-for-like comparison
across the cohort.}

\begin{figure}[!htbp]
\centering
\includegraphics[width=\textwidth]{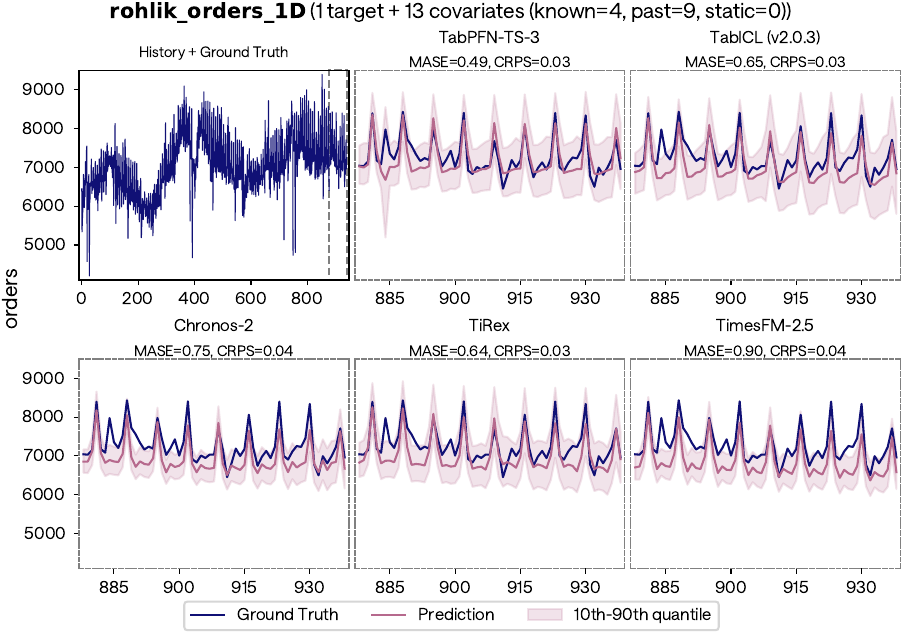}
\caption{\textbf{Qualitative forecast comparison on a fev-bench task (\texttt{rohlik\_order\_1D}).} Each model column shows the forecast horizon (zoomed to time 880-935) against the held-out ground truth, with the shaded band indicating the 10th-90th quantile. The leftmost panel shows the full training history. MASE and CRPS scores are reported per model. Additional examples, including covariate panels, are in~\Cref{app:time_series}.}
\label{fig:fev-bench-qualitative}
\end{figure}

Our checkpoint is evaluated with up to 32k historical time steps of
context, well beyond the budgets typically used by patch- or
window-based time-series foundation models.
Compared to the original TabPFN-TS~\citep{hoo2024tabpfn_ts} as evaluated
by the fev-bench authors (39.6 SQL skill, 28.8 MASE skill;
\citealt{shchur2025fev}), our fine-tuned variant improves to
\textbf{43.1 SQL skill} and \textbf{30.6 MASE skill}. On the full
100-task cohort it ranks 2nd on mean SQL skill scores (ahead of TiRex and TimesFM-2.5)
and 2nd on MASE (ahead of TimesFM-2.5, which has $10\%$ flagged
train/test leakage, and TiRex), in both cases behind only Chronos-2.
Looking at the win-rate results, TabPFN-TS-3's ranking drops to the 4th place, although we found these rates to be very sensitive to tiny differences on a few datasets.

The strong performance of TabPFN-TS-3 is particularly noteworthy seeing that it is trained purely on synthetic data, while most other time-series models, including Chronos-2 \citep{ansari2025chronos2univariateuniversalforecasting}, TiRex \citep{auer:25tirex} and TimesFM-2.5 \citep{pmlr-v235-das24c} are trained on real-world data. This property of our model prevents many issues from real-data pretraining: historical series are leaky and frequently recirculated across forecasting libraries (fev-bench flags 10\% leakage in TimesFM-2.5 and 28\% in Moirai-2.0; see Table~\ref{tab:fev-bench}), forecasting the future from historical pretraining is fundamentally out-of-distribution, and the supply of public real-world time-series data is finite, so any model relying on it inherits both its biases and its ceiling. Our synthetic prior by design has zero contamination from any specific real time series.

We also show qualitative examples in Figure \ref{fig:fev-bench-qualitative} to give a better intuition of our model forecasts. Appendix \ref{app:time_series} complements this section with the full
leaderboards (Table \ref{tab:fev-bench-full}), pairwise comparisons (Figure \ref{fig:fev-bench-pairwise}), additional qualitative forecasts and per-task SQL results.

\subsection{Relational Data}\label{subsec:results_rel}

\begin{figure}[!htbp]
    \centering
    \begin{subfigure}[t]{0.46\textwidth}
        \centering
        \includegraphics[width=\textwidth]{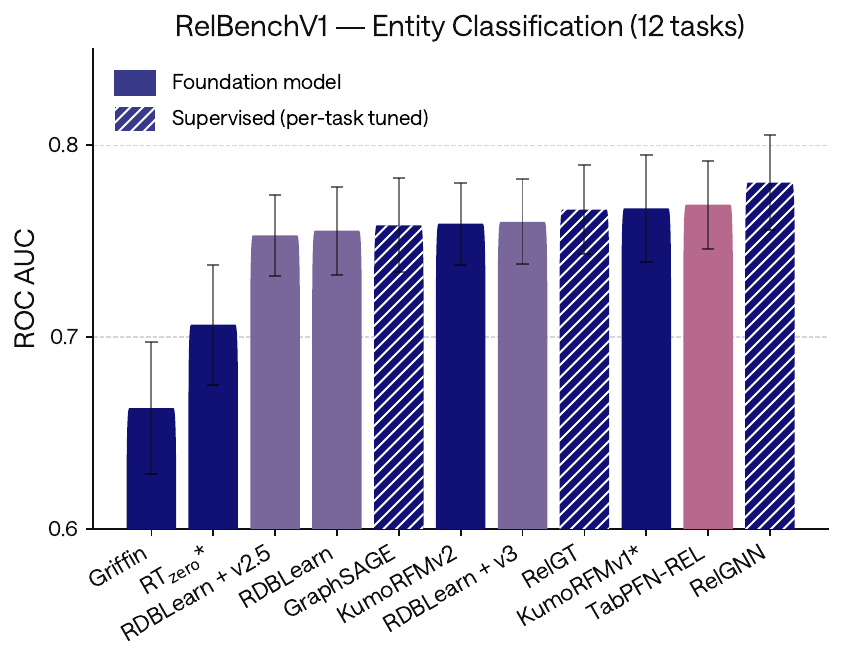}
    \end{subfigure}
    \begin{subfigure}[t]{0.46\textwidth}
        \centering
        \includegraphics[width=\textwidth]{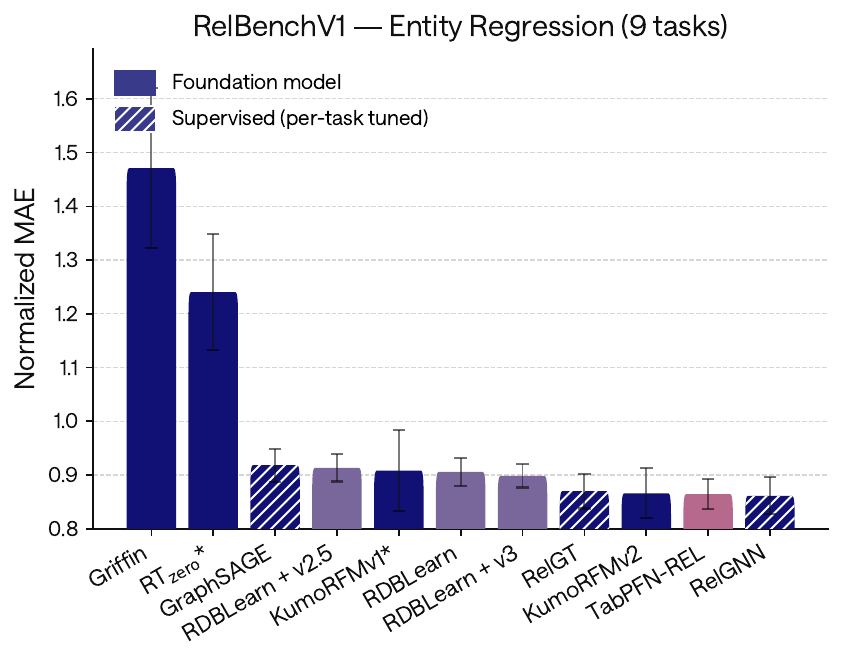}
    \end{subfigure}
    \caption{\textbf{\ourmodel tops performance on RelBenchV1 among foundation models.}
    Following \citet{kumorfmv2}, we report the mean ROC AUC for entity classification and MAE scores for entity regression normalized by LightGBM's MAE. RelGNN~\cite{relgnn} achieves SOTA performance on both tasks, followed by TabPFN-REL, which sets a new SOTA for foundation models.
    Methods marked with $^{*}$ in their name (KumoRFMv1, \rtzero) indicate methods that are likely following a different evaluation protocol than the one outlined in RelBench, which overestimates model performance.
    }
    \label{fig:relbench}
\end{figure}

Real-world data is often relational: commercial enterprises, healthcare systems, and financial institutions routinely store their core operational data across multiple interconnected tables in relational databases. Unlocking predictive insights from such data is therefore of substantial practical importance, and requires to reason jointly over heterogeneous tables linked by complex foreign-key relationships. This has motivated the development of dedicated relational foundation models (RFMs) that aim to provide accurate, up-to-date predictions via In-Context Learning (ICL) without the need for costly per-task model training and hyperparameter tuning.

This has sparked the emergence of dedicated solutions for relational data, e.g., fully supervised solutions particularly tailored for relational data such as GraphSAGE~\cite{graphsage}, RelGT~\cite{relgt} and RelGNN~\cite{relgnn}, closed-source relational foundation models like KumoRFMv1~\cite{kumorfmv1} and KumoRFMv2~\cite{kumorfmv2}, as well as open-source RFMs, Griffin~\cite{griffin} and \rtzero~\cite{rt_zero}. Recently, RDBLearn~\cite{rdblearn} has shown that TFMs including TabPFN can be converted into RFMs by automatically flattening the underlying database into a table.

In this section, we build on this research and show how \emph{TabPFN-REL} using TabPFN-3 achieves state of the art performance on the popular RelBenchV1~\cite{relbenchv1} benchmark for entity classification and regression. 

For RelBench, we follow the general guidelines by truncating each database at the pre-specified test timestamp before constructing the featurization and context for all test entities. Following \citet{kumorfmv2}, we generally report baseline results as provided by the authors of the methods to ensure well-tuned baselines. For methods that likely follow a different evaluation regime, we rerun the evaluation using RelBench's data regime, falling back to author-reported numbers where rerunning is not possible due to model deprecation or missing checkpoints (as is the case for KumoRFMv1 and \rtzero); we note that these may not be directly comparable due to potentially different data setups. For KumoRFMv2 we adapt the original scripts provided by the authors and use four estimators and a context size of $10000$ (the respective maxima for each), which we found to slightly outperform the script defaults of one estimator and a context size of $5000$ samples. We compare three different versions of RDBLearn: Vanilla RDBLearn that tunes over a range of different TFMs including TabPFN-2.5, as well as versions which forgo the tuning and use either TabPFN-2.5 or \ourmodel as a fixed TFM.\footnote{The reported results were produced with early checkpoints that did not undergo the full training pipeline and separate binary from multiclass classification. They can be identified on HuggingFace by the \texttt{20260417\_\textless TASK\_TYPE\textgreater} suffix.}

\paragraph{TabPFN-REL sets a new state-of-the-art among RFMs.} We report the aggregate performance of the different RFMs and fully-supervised baselines in~\autoref{fig:relbench} both for entity classification and entity regression on RelBenchV1, as well as per-dataset results in \autoref{sec:app/relbench/per-dataset-results}. \textit{TabPFN-REL achieves state-of-the-art performance among RFMs on both tasks}, with KumoRFMv1/v2 coming second on regression/classification. We attribute KumoRFMv1's strong classification results in part to a potentially different evaluation regime used by the authors, which likely overestimates performance, especially on the \texttt{rel-f1} task suite. We also observe that RDBLearn with the fixed TabPFN-3 backend consistently outperforms the original RDBLearn, which itself tunes over various TFMs including TabPFN-2.5. RDBLearn using \ourmodel hence Pareto-dominates vanilla RDBLearn in terms of runtime and performance, and to the best of our knowledge sets a \textit{new state-of-the-art among open-source RFMs}. \textit{At the time of writing, TabPFN-3 therefore powers both the best overall relational foundation model (TabPFN-REL) and the best open-source alternative (RDBLearn + v3).}

\paragraph{Comparison to fully-supervised baselines.} The fully-supervised RelGNN outperforms TabPFN-REL, with the gap being larger on classification than regression. On regression, the gap between RelGNN and TabPFN-REL is slim, with TabPFN-REL achieving lower mean rank than RelGNN. RelGT and GraphSAGE fall behind TabPFN-REL both in terms of normalized score and rank. We note that training supervised methods is several orders of magnitude more expensive than the in-context learning performed in TabPFN-REL~\cite{rdblearn, kumorfmv1, kumorfmv2}. This is both because training a single supervised model takes significantly longer than the forward pass of TabPFN-REL, and because supervised methods require extensive per-dataset hyperparameter tuning to achieve optimal performance. For example, we identified at least seven axes of variability in RelGNN's per-dataset configs, yielding thousands of possible hyperparameter combinations to search over.

\FloatBarrier

\subsection{Causal Inference}

We follow up on our previous results \cite{TabPFN-2.5}, which showed strong performance of TabPFN-2.5 as a meta (T/X/S) learner \cite{kunzel_meta_learners} on the RealCause benchmark, by providing an evaluation on the \texttt{scikit-uplift} benchmark \cite{user-guide-for-uplift-modeling}. In terms of QINI-score, a real-world evaluation strategy for experimental data, we observe that all TabPFN-3 meta-learners improve over TabPFN-2.5, with the top two spots occupied by T and S-Learners (Figure \ref{fig:causal_inference}). In contrast, we observe slightly worse performance compared to TabPFN-2.5 on RealCause \cite{neal_realcause}.
We provide a more in-depth analysis of the results and description of the QINI evaluation protocol in Appendix \ref{sec:causal_inference}.

\subsection{Embeddings}\label{subsec:results_embed}
\begin{figure}
    \centering
    \includegraphics{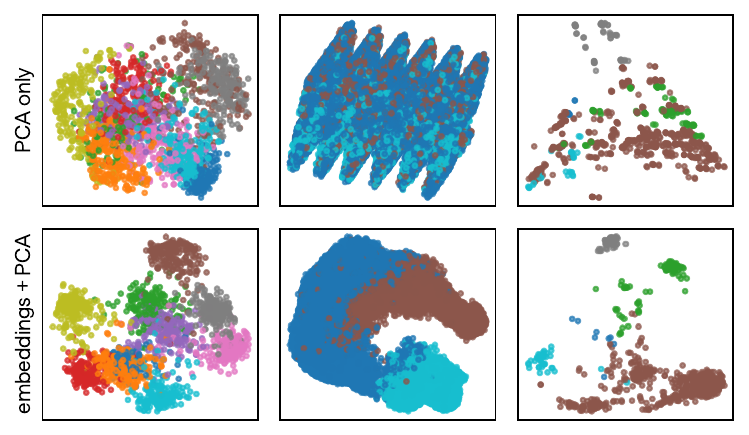}
    \caption{
        \textbf{TabPFN-3 extracts semantically-meaningful row embeddings.}
        The upper plots show 2D PCA applied directly to three classification datasets, where each point is a row, while the lower plots show PCA applied to embeddings of the rows.
        Color indicates the class.
        We observe that the embeddings are clustered by class.
    }
    \label{fig:embeddings-scatter}
\end{figure}

Finally, we demonstrate that \ourmodel generates semantically-meaningful embeddings. We follow the approach developed by \citet{ye2026closer} for TabPFN~v2: we partition the dataset into cross-validation folds, and take the embeddings from the test-portion of the dataset in each fold. The embeddings we capture are the output of the ICL layers at the end of Stage 3 of our model (see Section~/\ref{sec:arch_overview} for more details). Figure~\ref{fig:embeddings-scatter} shows that this approach continues to work well for TabPFN-3, with the generated embeddings capturing the dataset structure.

%% file: tables/time_series/leaderboard_SQL_subset.tex
\begin{tabular}{lrrrrr}
\toprule
\textbf{Model} & \textbf{Win (\%)} & \textbf{Skill (\%)} & \textbf{Runtime (s)} & \textbf{Leak.\ (\%)} & \textbf{\# fails} \\
\midrule
Chronos-2      & 91.7 & 47.3 & 0.8   & 0  & 0 \\
\textcolor{PriorMauve}{TabPFN-TS-3}    & 73.6 & 43.1 & 234.6 & 0  & 0 \\
TiRex          & 83.4 & 42.6 & 0.2   & 1  & 0 \\
TimesFM-2.5    & 78.6 & 42.2 & 1.9   & 10 & 0 \\
Toto-1.0       & 71.6 & 40.7 & 22.1  & 8  & 0 \\
\textcolor{PriorMauve}{TabPFN-v2-TS}   & 64.1 & 39.6 & 88.9  & 0  & 2 \\
Moirai-2.0     & 66.2 & 39.3 & 0.3   & 28 & 0 \\
Chronos-Bolt   & 66.2 & 38.9 & 0.2   & 0  & 0 \\
Sundial-Base   & 47.1 & 33.4 & 8.0   & 1  & 0 \\
TabICL-v2      & 53.8 & 30.8 & 64.7  & 0  & 0 \\
Stat. Ensemble & 43.8 & 20.2 & 148.6 & 0  & 11 \\
Seasonal Naive & 19.1 & 0.0  & 0.5   & 0  & 0 \\
\bottomrule
\end{tabular}

%% file: tables/time_series/leaderboard_MASE_subset.tex
\begin{tabular}{lrrrrr}
\toprule
\textbf{Model} & \textbf{Win (\%)} & \textbf{Skill (\%)} & \textbf{Runtime (s)} & \textbf{Leak.\ (\%)} & \textbf{\# fails} \\
\midrule
Chronos-2      & 86.9 & 35.5 & 0.8   & 0  & 0 \\
\textcolor{PriorMauve}{TabPFN-TS-3}    & 69.8 & 30.6 & 234.6 & 0  & 0 \\
TimesFM-2.5    & 74.9 & 30.2 & 1.9   & 10 & 0 \\
TiRex          & 76.9 & 30.0 & 0.2   & 1  & 0 \\
Toto-1.0       & 66.3 & 28.2 & 22.1  & 8  & 0 \\
\textcolor{PriorMauve}{TabPFN-v2-TS}   & 58.5 & 27.6 & 88.9  & 0  & 2 \\
Moirai-2.0     & 61.4 & 27.3 & 0.3   & 28 & 0 \\
Chronos-Bolt   & 60.7 & 26.5 & 0.2   & 0  & 0 \\
Sundial-Base   & 53.4 & 24.7 & 8.0   & 1  & 0 \\
Stat. Ensemble & 46.7 & 15.7 & 148.6 & 0  & 11 \\
TabICL-v2      & 33.2 & 7.0  & 64.7  & 0  & 0 \\
Seasonal Naive & 20.0 & 0.0  & 0.5   & 0  & 0 \\
\bottomrule
\end{tabular}

%% file: sections/04_adoption.tex
\section{Adoption} \label{sec:usecases_extensions}

TabPFN-3 is shipped into an already sprawling ecosystem. Since the v2 release, TabPFN has been picked up across academic ML research, applied science, and enterprise deployment. A substantial portion of the extension work referenced throughout this report (time-series, causal inference, relational data, interpretability) was driven by that community rather than initiated internally. This section describes the shape of that adoption -- where the model is in production, where it is being evaluated, which platforms make it accessible, and which research areas have published applications -- to give the v3 release its actual operational context.

\subsection{Community and Open-Source Ecosystem}

The open-source \texttt{tabpfn} package has surpassed 3.2 million PyPI downloads, and the original TabPFN Nature paper~\citep{Hollmann2025tabpfnv2} has been cited in over 1{,}000 papers in the sixteen months since publication.\footnote{Google Scholar entry and pepy.tech \texttt{tabpfn} download statistics, both accessed May 8, 2026.} A Discord community of over 2{,}000 users and hundreds of resolved GitHub issues have driven cross-platform stability work, edge-case fixes, and the maturation of the model from research artifact to production-grade library.

A separate \texttt{tabpfn-extensions} repository\footnote{\url{https://github.com/PriorLabs/tabpfn-extensions}} hosts community-driven extensions that compose with the core model: SHAP and SHAP-IQ interpretability, synthetic data generation and missing-value imputation, TabPFN-based feature selection, regression-via-classification, survival analysis and conditional randomization tests. TabPFN-3's reduced KV cache and inference improvements (Section~\ref{sec:methods}) directly accelerate every extension that depends on repeated forward passes -- most notably interpretability and conditional independence testing.

TabPFN also serves as a foundational layer for methods published as independent research, spanning time-series forecasting~\citep{hoo2024tabpfn_ts}, node classification on graphs~\citep{Hayler2025GraphsTablesZeroShot, eremeev2025turningtabularfoundationmodels}, evolving data streams~\citep{Lourenco2025ICLStreams}, causal inference~\citep{robertson_dopfn, balazadeh_causalpfn, feuerriegel_causalfm}, reinforcement learning~\citep{Schiff2025TabPFNRL}, high-dimensional Bayesian optimization~\citep{Yu2025GITBO}, and multimodal encoding~\citep{luo2025timetabpfnintegratedmultimodalengine}. As shown in Section~\ref{sec:results}, many of these extensions move further forward when run with TabPFN-3 as the backend rather than v2.5 or v2.6.

\subsection{Enterprise Engagements}

TabPFN has been deployed and evaluated across a wide range of enterprise settings. Examples include: \textit{Hitachi Rail} deploys TabPFN for predictive maintenance on the Spanish rail network; in initial deployment, TabPFN reduced root-mean-square error by approximately 40\% compared to their existing baseline~\citep{hitachi_case_study}. \textit{Creditplus Bank}, part of the Cr\'{e}dit Agricole group, will use distilled TabPFN models (Section~\ref{sec:distillation}) for assisting CPU-based credit decisioning in motor finance under appropriate credit-risk regulatory constraints~\citep{creditplus_case_study}. \textit{Oxford Cancer Analytics} applies TabPFN to proteomic liquid-biopsy data for early lung-disease detection~\citep{oxcan_case_study}. A longer list of enterprise and commercial engagements is available on the Prior Labs website.

\subsection{Platform Availability}

TabPFN is available through the open-source PyPI distribution for evaluation and non-commercial use, and through a managed API for commercial workloads. The model is currently listed on the \textit{AWS SageMaker Marketplace}\footnote{\url{https://aws.amazon.com/marketplace/pp/prodview-chfhncrdzlb3s}} and the \textit{Azure AI Foundry Model Catalog}\footnote{\url{https://ai.azure.com/catalog/models/TabPFN-2.5}}, with full support for batch and real-time inference on classification and regression tasks; the TabPFN-3 release on both marketplaces follows this report. A reference integration for \textit{Databricks} is available through the Databricks Industry Solutions repository\footnote{\url{https://github.com/databricks-industry-solutions/tabpfn-databricks}}. See Section~\ref{sec:license} for license terms, commercial-use scope, and the contact path for production deployment.

\subsection{Research Adoption Across Domains}

In addition to commercial engagement, we have collected more than 200 published research applications of TabPFN across a broad range of areas; the full list is in Appendix~\ref{app:use_cases}.

Adoption is strongest in \textit{healthcare and life sciences} (98 applications), reflecting TabPFN's relative advantage in data-scarce settings: diagnosis, prognosis, treatment-response prediction, biomarker modeling, survival analysis, drug discovery, pharmacokinetics, radiomics, omics, and multimodal clinical data. \textit{Manufacturing and industrial} applications (41 papers) span concrete and asphalt strength prediction, geotechnical modeling, tunnel construction, steel and semiconductor properties, IIoT intrusion detection, rotating-machinery fault classification, battery and circuit modeling, and materials discovery. \textit{Energy and utilities} (24 papers) cluster around environmental monitoring, renewable-energy and geophysical prediction, water and climate systems, and industrial process optimization. \textit{Financial services} (7 papers) include transaction analytics, churn prediction, return forecasting, actuarial modeling, and credit-risk prediction; the relatively small published count almost certainly underrepresents commercial traction in a domain that publishes little. The remaining 32 applications span uncertainty estimation, hypothesis testing, Shapley value estimation, graph node classification, cybersecurity, geoscience, agriculture, soil and lunar-regolith analysis, fuel-blend prediction, crop-yield forecasting, forensic ancestry prediction, and synthetic tabular data generation.

The distribution of these applications -- weighted toward domains characterized by limited, expensive, or heterogeneous data -- is consistent with the regime TabPFN was designed for, and is the empirical basis for the v3 capability choices described in Section~\ref{sec:methods}.

%% file: sections/05_license.tex
\section{License and Availability}\label{sec:license}We release TabPFN-3 under the \texttt{TABPFN-3.0 License v1.0}, designed to be permissive for academic use, research, and evaluation in commercial settings. The license \emph{explicitly allows} testing, evaluation, and internal benchmarking, so an organization can download the model and run preliminary assessments on its own datasets without a commercial agreement.

The key restriction is that the model, its derivatives, and its outputs cannot be used for commercial or production purposes. This includes, but is not limited to, revenue-generating products, competitive benchmarking for procurement decisions, client deliverables, and using model outputs as inputs to internal commercial decision-making.

For production use, we offer a \emph{Commercial Enterprise License}, available for our managed API, Virtual Private Cloud deployments (at the time of publication: AWS SageMaker \& Azure AI Foundry), and on-prem or other custom deployment modes across other software platforms such as Databricks and SAP. The Commercial Enterprise License provides access to our proprietary high-speed inference engine, dedicated support, integration tooling, additional internal models, and the \ourmodelenhanced variant, which is not available as part of the open-source release. The managed API runs on our optimized GPU infrastructure and is the recommended option for users without dedicated local GPUs; it is accessible via a Python SDK\footnote{The Python client SDK is available on PyPI: \url{https://github.com/PriorLabs/tabpfn-client}.} (\texttt{pip install tabpfn-client}) or a standard REST API.

The full \texttt{TABPFN-3.0 License v1.0} text is available at \url{https://huggingface.co/Prior-Labs/tabpfn_3/blob/main/LICENSE}. For commercial licensing inquiries, please contact \href{mailto:sales@priorlabs.ai}{sales@priorlabs.ai}.

\newpage

\FloatBarrier

\bibliographystyle{unsrtnat}
\bibliography{bib}

\newpage
\appendix

\phantomsection
\addcontentsline{toc}{section}{Appendix}

\addtocontents{toc}{\protect\setcounter{tocdepth}{-10}}

\section*{Appendix Table of Contents}
\startcontents[appendix]
\makeatletter
\protected@write\@auxout{}{\string\ttl@writefile{ptc}{\protect\setcounter{tocdepth}{2}}}
\makeatother
\printcontents[appendix]{}{1}{}
\newpage

%% file: sections/A_contributors.tex
\section{Contributors}\label{app:contributors}

\paragraph{Model Development \& Deployment.}
Noah Hollmann,
Frank Hutter,
L\'eo Grinsztajn,
Klemens Flöge,
Oscar Key,
Felix Birkel,
Philipp Jund,
Brendan Roof,
Mihir Manium,
Shi Bin (Liam) Hoo,
Magnus Bühler,
Anurag Garg,
Dominik Safaric,
Jake Robertson,
Benjamin Jäger,
Simone Alessi,
Adrian Hayler,
Vladyslav Moroshan,
Lennart Purucker,
Philipp Singer,
Alan Arazi,
Julien Siems,
Jan Hendrik Metzen,
Georg Grab,
Nick Erickson,
Siyuan Guo,
Eliott Kalfon,
Simon Bing,
David Salinas
\vspace{-1em}
\paragraph{Distribution \& Product.}

Sauraj Gambhir,
Clara Cornu,
Lilly Charlotte Wehrhahn,
Diana Kriuchkova
\vspace{-2em}
\paragraph{Operations.}

Kursat Kaya,
Lydia Sidhoum,
Marie Salmon,
Jerry Chen
\\

\vspace{-0.5em}
\noindent\textit{Authors are ordered by their date of joining Prior Labs; all authors above affiliated with Prior Labs at the time of contribution; work done at Prior Labs.}

\vspace{0.75em}

\paragraph{Scientific Advisors.}
Samuel Müller, 
Madelon Hulsebos,
Yann LeCun,
Bernhard Schölkopf
\vspace{0.75em}
\\
\noindent\textit{Scientific advisors did not contribute IP.}

\vspace{1em}

%% file: sections/B_acknowledgements.tex
\section{Acknowledgements}
\label{app:acknowledge}
We acknowledge the EuroHPC Joint Undertaking for awarding this project access to the EuroHPC supercomputer LUMI, hosted by CSC (Finland) and the LUMI consortium through a EuroHPC AI Factory Large Scale Access call (project EHPC-AIF-2025LS17-014).

We acknowledge EuroHPC JU for awarding the project ID EHPC-AIF-2025LS14-031 access to MareNostrum 5 ACC hosted by the Barcelona Supercomputing Center (BSC), Spain.

\begin{center}
    \includegraphics[width=0.4\linewidth]{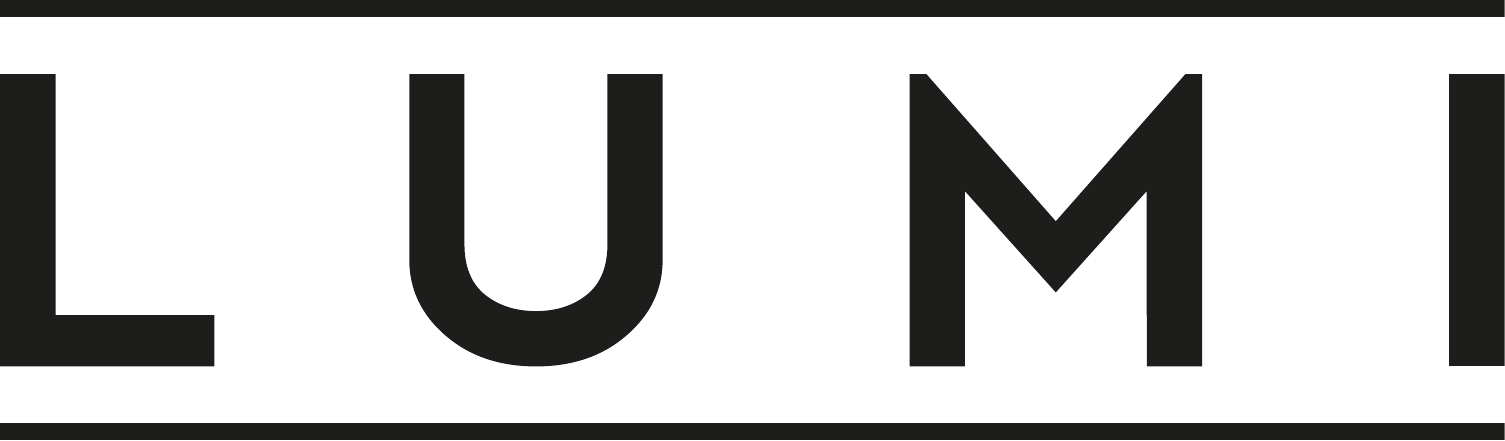}
\end{center}

%% file: sections/C_architectural_hyperparams.tex
\section{Architectural Hyperparameters}\label{app:architecture-hyperparams}

The tables below list the architectural hyperparameters of the released TabPFN-3 classifier and regressor checkpoints.
The two models share all hyperparameters; the only differences are in the output decoder, which is task-specific (noted where applicable).

\begin{table}[H]
\centering
\small
\caption{Stage~1 — Feature embedding.}
\begin{tabular}{lrl}
\toprule
\textbf{Hyperparameter} & \textbf{Value} & \textbf{Description} \\
\midrule
\texttt{embed\_dim}                         & 128 & Base embedding dimension used throughout the model \\
\texttt{feature\_group\_size}               & 3   & Features per circular-shift group \\
\texttt{dist\_embed\_num\_blocks}           & 3   & Induced self-attention blocks \\
\texttt{dist\_embed\_num\_heads}            & 8   & Attention heads per block \\
\texttt{dist\_embed\_num\_inducing\_points} & 128 & Inducing points per column \\
\bottomrule
\end{tabular}
\end{table}

\begin{table}[h]
\centering
\small
\caption{Stage~2 — Feature aggregation.}
\begin{tabular}{lrl}
\toprule
\textbf{Hyperparameter} & \textbf{Value} & \textbf{Description} \\
\midrule
\texttt{feat\_agg\_num\_blocks}     & 3        & Transformer blocks \\
\texttt{feat\_agg\_num\_heads}      & 8        & Attention heads per block \\
\texttt{feat\_agg\_num\_cls\_tokens} & 4       & CLS tokens aggregated per row \\
\texttt{use\_rope}                  & True     & Rotary positional embeddings (RoPE) enabled \\
\texttt{feat\_agg\_rope\_base}      & 100\,000 & RoPE base frequency $\theta$ \\
\bottomrule
\end{tabular}
\end{table}

\begin{table}[H]
\centering
\small
\caption{Stage~3 — ICL transformer.}
\begin{tabular}{lrl}
\toprule
\textbf{Hyperparameter} & \textbf{Value} & \textbf{Description} \\
\midrule
\texttt{icl\_emsize} (derived)     & 512 & $\texttt{embed\_dim} \times \texttt{feat\_agg\_num\_cls\_tokens} = 128 \times 4$ \\
\midrule
\texttt{nlayers}                   & 24  & Transformer blocks \\
\texttt{icl\_num\_heads}           & 8   & Query heads per block \\
\texttt{icl\_num\_kv\_heads}       & 8   & KV heads for train rows (standard MHA) \\
\texttt{icl\_num\_kv\_heads\_test} & 1   & KV heads for test rows \\
\bottomrule
\end{tabular}
\end{table}

\begin{table}[H]
\centering
\small
\caption{Many class output decoder — classifier.}
\begin{tabular}{lrl}
\toprule
\textbf{Hyperparameter} & \textbf{Value} & \textbf{Description} \\
\midrule
\texttt{max\_num\_classes}              & 160  & Maximum supported class count \\
\texttt{decoder\_num\_heads}            & 6    & Attention heads in retrieval decoder \\
\texttt{decoder\_head\_dim}             & 64   & Head dimension in retrieval decoder \\
\bottomrule
\end{tabular}
\end{table}

\begin{table}[h]
\centering
\small
\caption{MLP output decoder — regressor (2-layer MLP).}
\begin{tabular}{lrl}
\toprule
\textbf{Hyperparameter} & \textbf{Value} & \textbf{Description} \\
\midrule
\texttt{architecture} (derived) & $512 \to 1024 \to 5000$ & $\texttt{icl\_emsize} \to \texttt{icl\_emsize} \times \texttt{ff\_factor} \xrightarrow{\text{GELU}} \texttt{num\_buckets}$ \\
\midrule
\texttt{num\_buckets} & 5000 & Output buckets for quantile regression \\
\bottomrule
\end{tabular}
\end{table}

\begin{table}[H]
\centering
\small
\caption{Shared settings (both classifier and regressor).}
\begin{tabular}{lrl}
\toprule
\textbf{Hyperparameter} & \textbf{Value} & \textbf{Description} \\
\midrule
\texttt{ff\_factor}                         & 2    & Feed-forward expansion factor (all stages) \\
\texttt{softmax\_scaling\_mlp\_hidden\_dim} & 64   & Hidden units in query-aware softmax-scaling MLPs \\
\bottomrule
\end{tabular}
\end{table}

%% file: sections/D_prior_visualizations.tex
\section{Prior visualizations}
\begin{figure}[H]
    \centering
    \includegraphics[width=1\linewidth]{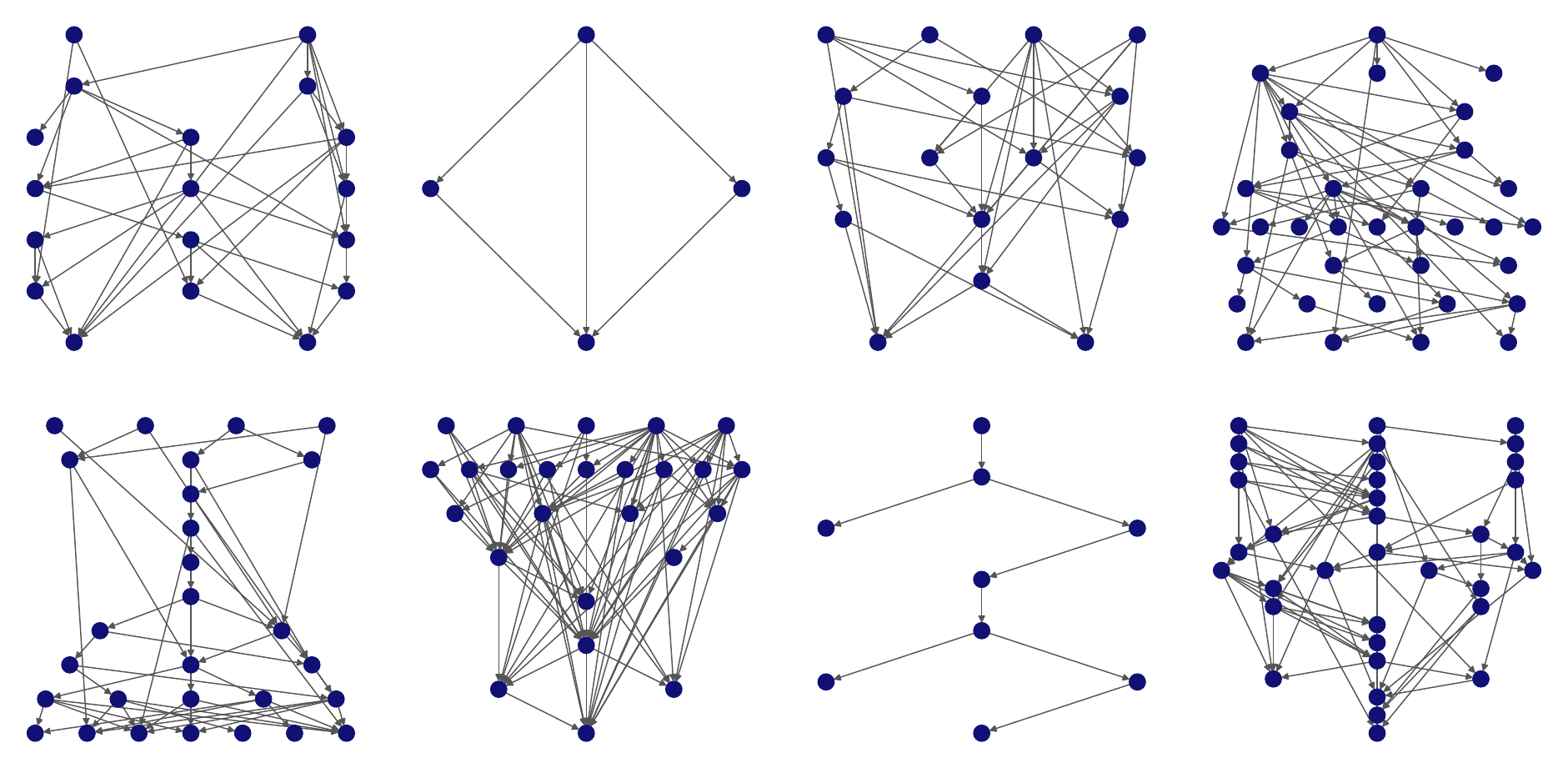}
    \caption{\textbf{Visualization of directed acyclic graphs underlying our SCM prior}, produced by our new graph sampling algorithms.}
    \label{fig:graph_sampling}
\end{figure}

\begin{figure}[H]
    \centering
    \includegraphics[width=1\linewidth]{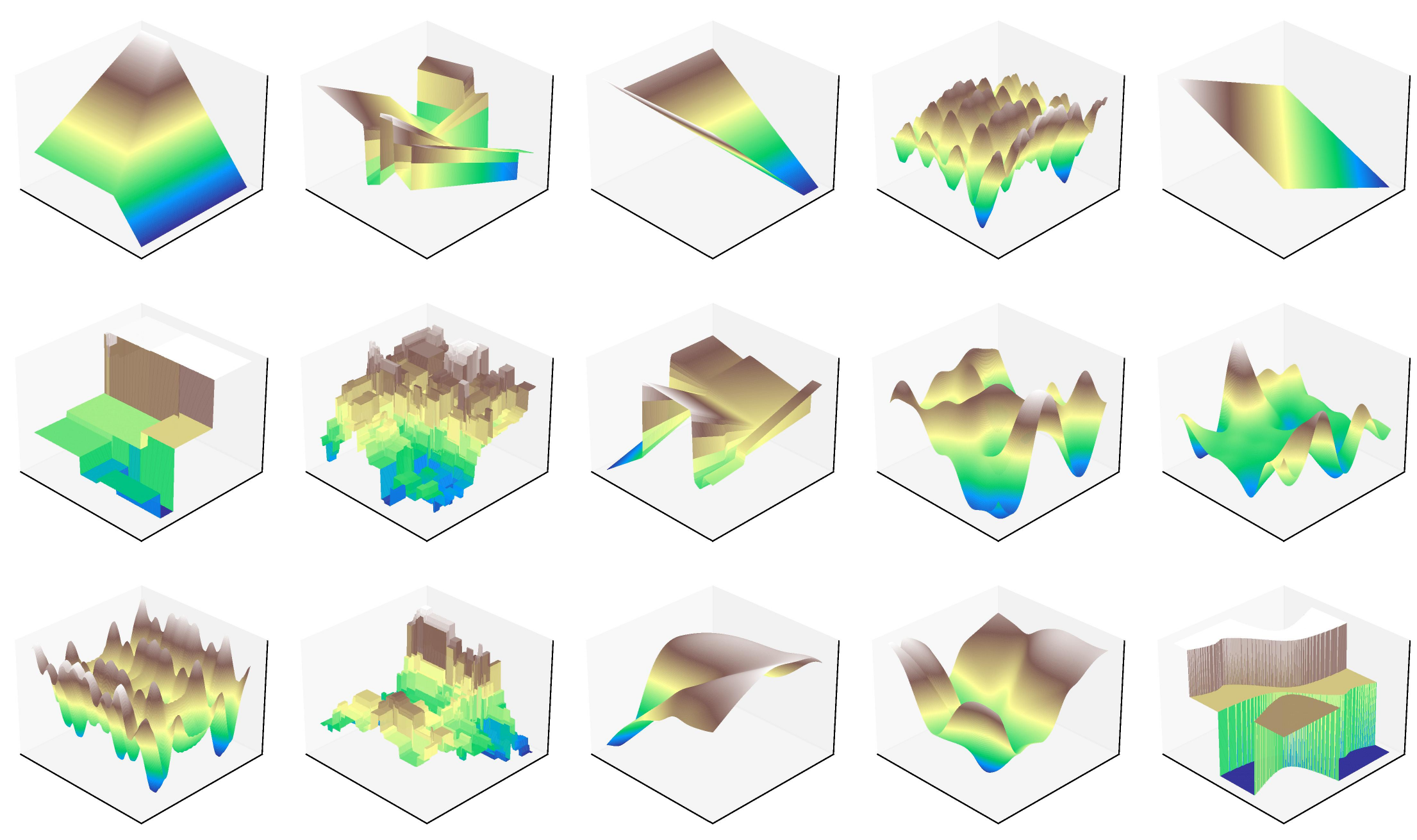}
    \caption{\textbf{Visualization of functional relationships generated by the new combiner mechanisms in our SCM prior.} While mechanisms in the prior have variable dimensionality, for the sake of visualization we plot functions on a two-dimensional grid.}
    \label{fig:mechanisms}
\end{figure}

\begin{figure}[H]
    \centering
    \begin{minipage}[c]{0.8\textwidth}
        \centering
        \includegraphics[width=\linewidth]{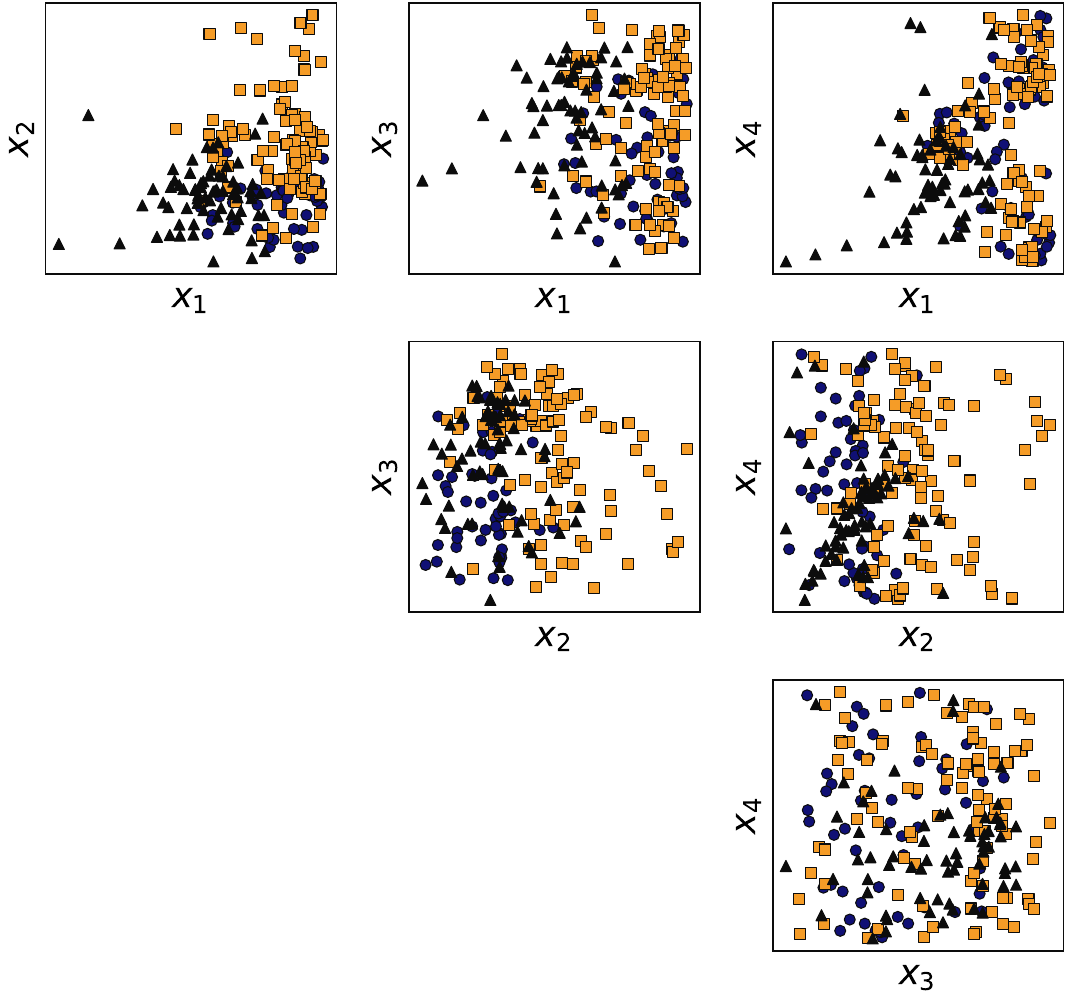}
        \caption{\textbf{Example classification dataset generated from the prior.} There are four covariates and the subplot in row i column j corresponds to a scatter plot of covariates i and j+1 with target class indicated by color.}
        \label{fig:dataset}
    \end{minipage}
\end{figure}

\begin{figure}[H]
    \centering
    \includegraphics[width=1\linewidth]{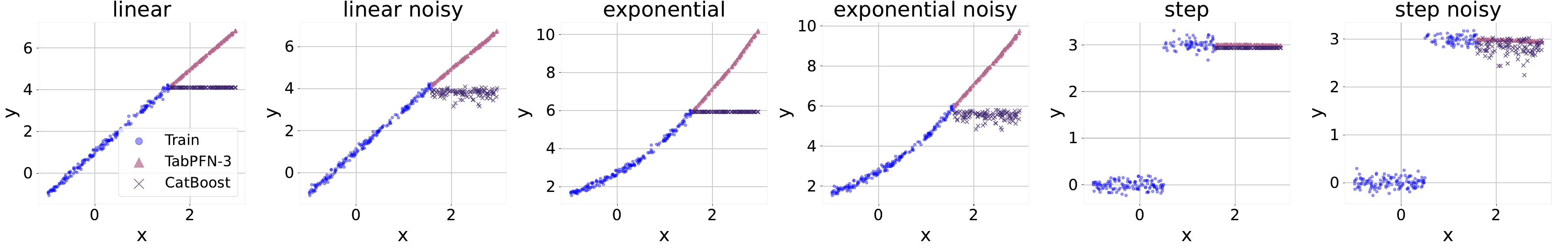}
    \caption{\textbf{Example demonstrating the extrapolation capabilities of TabPFN-3 (using our out-of-distribution compatible preprocessing), comparing to CatBoost.} As can be seen, TabPFN-3 is able to extrapolate successfully, which tree-based algorithms and tabular foundation models often struggle with.}
    \label{fig:ood}
\end{figure}

We provide a number of illustrative visualizations for the improvements to our prior. Figure~\ref{fig:graph_sampling} shows directed acyclic graphs sampled by our new graph-sampling algorithms; Figure~\ref{fig:mechanisms} visualizes the functional relationships generated by the new combiner mechanisms; 
Figure~\ref{fig:dataset} gives an example classification dataset generated from the prior; and Figure~\ref{fig:ood} demonstrates TabPFN-3's extrapolation capabilities, comparing to CatBoost.

%% file: sections/E_experimental_results_details.tex
\section{Experimental results details}

\subsection{Details on Causal Inference Results}
\label{sec:causal_inference}

\paragraph{Causal Inference.} 
Many practical problems are rooted in causal logic, requiring an understanding of how interventions, rather than mere associations, shape outcomes. Estimating Conditional Average Treatment Effects (CATEs) serves as a primary tool for addressing these "what-if" scenarios, quantifying the expected change in an individual’s response when a treatment is applied compared to when it is withheld. Previous results \cite{TabPFN-2.5} have shown that TabPFN-2.5, especially when used as a T-Learner \cite{kunzel_meta_learners}, achieves SOTA performance on the RealCause benchmark \cite{neal_realcause}. While TabPFN-3 does not quite achieve the highest performance on RealCause (still surpassed by TabPFN-2.5), we see substantial improvements on larger datasets with up to 50k samples in the \texttt{scikit-uplift} library \cite{user-guide-for-uplift-modeling}. We describe the details of this evaluation below.

\begin{figure}[t!]
    \centering
    \includegraphics[width=0.49\linewidth]{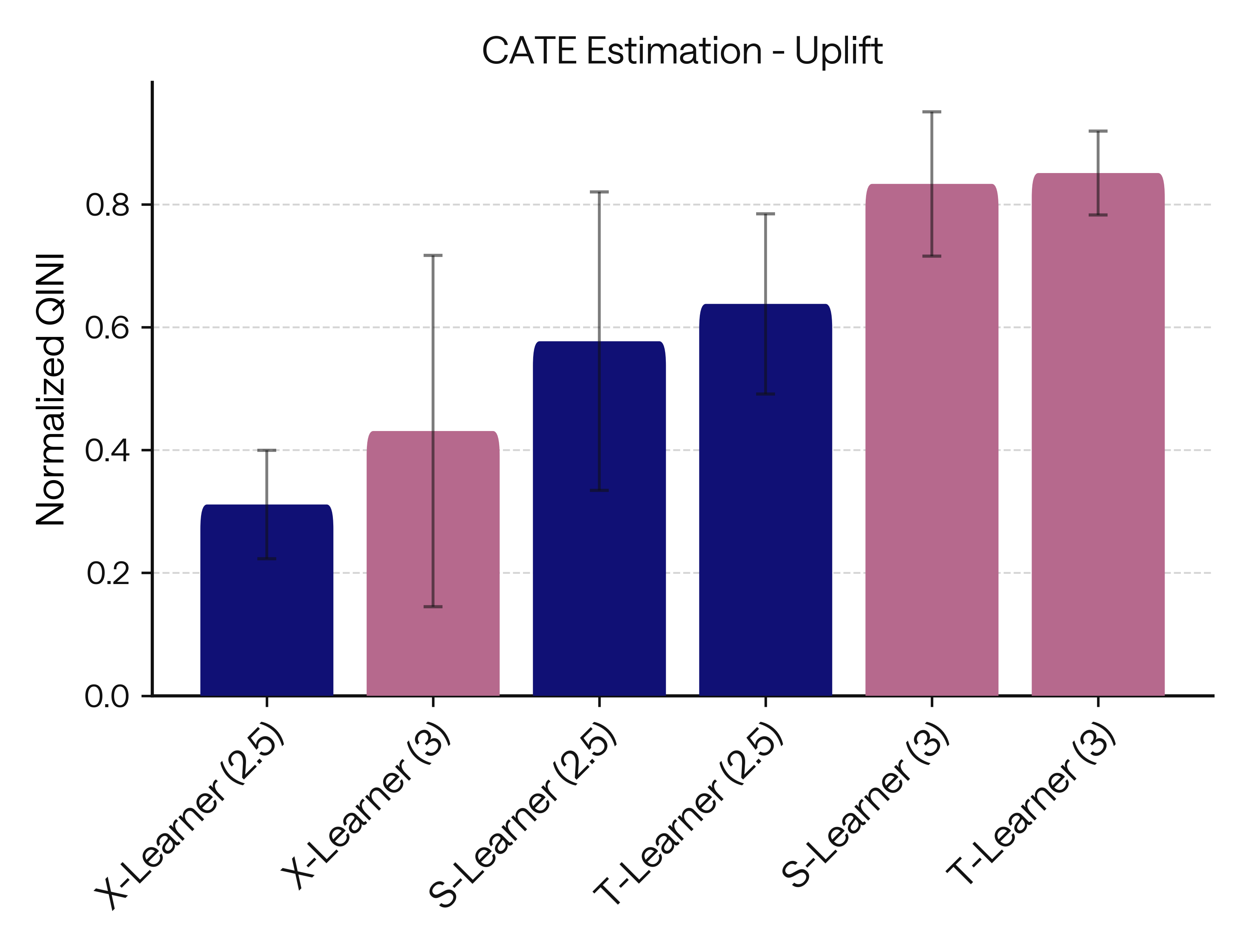}
     \includegraphics[width=0.49\linewidth]{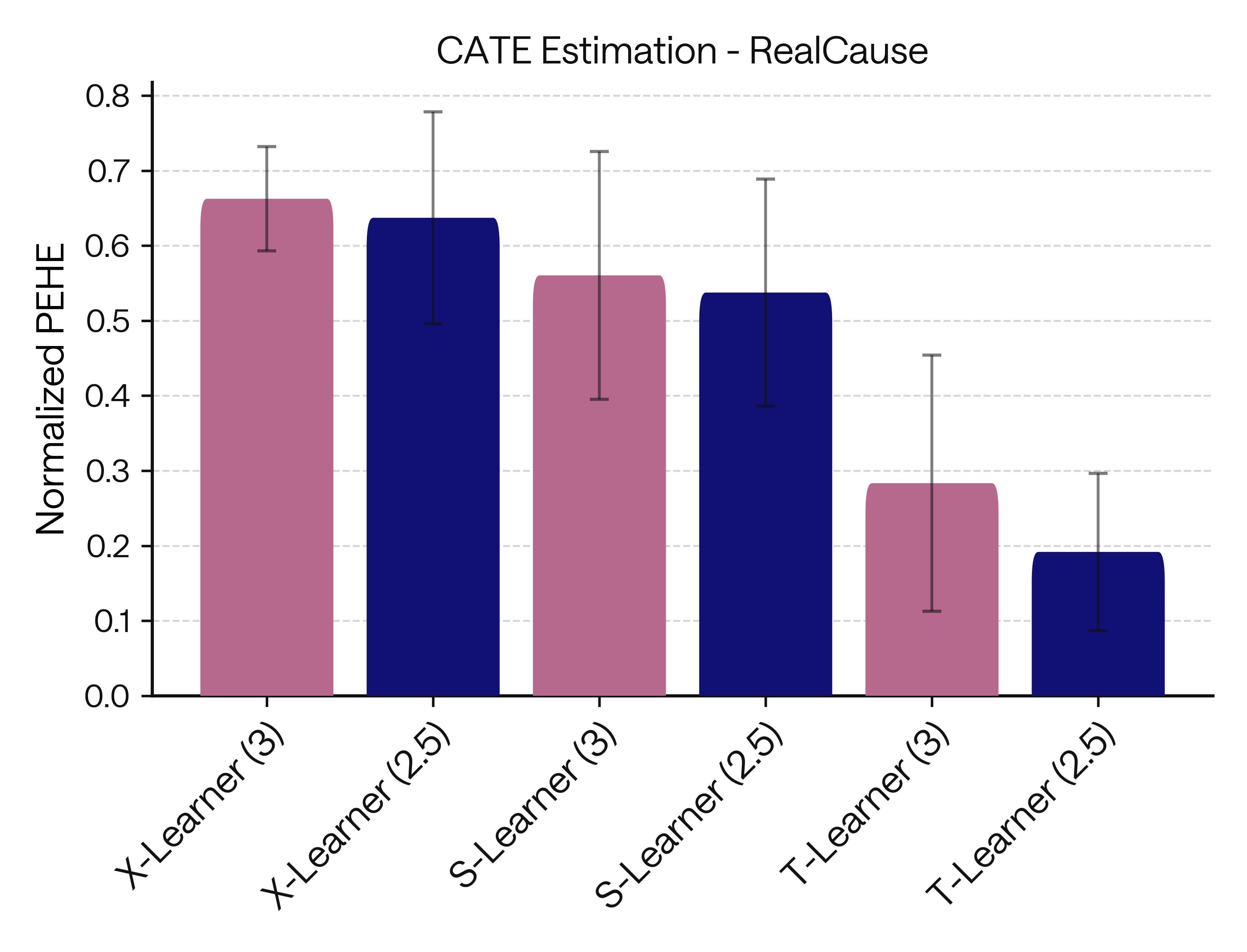}
    \caption{\textbf{TabPFN-3 as a T/X/S-Learner.} TabPFN-3 when used as a T/S-Learner achieves strong performance in terms of QINI-score ($\uparrow$) in Uplift Modeling on the \texttt{scikit-uplift} benchmark. We report worsened performance in terms of PEHE ($\downarrow$) on the RealCause benchmark compared to the previous version.}
    \label{fig:causal_inference}
\end{figure}

\paragraph{Real-World QINI Evaluation.} One of the major drawbacks in evaluating causal inference methods is referred to by \citet{holland1986statistics} as the Fundamental Problem of Causal Inference, which states that individual treatment effects can never actually be observed in the real-world. In simple terms, one cannot experimentally test both potential outcomes without interference. Under the assumption of experimental (RCT) data, Uplift Modeling \cite{user-guide-for-uplift-modeling} allows to evaluate the benefit of using a causal estimator in terms of ranking individuals by treatment effect. Crucially, this evaluation strategy does not require access to ground truth (synthetic) treatment effects, and serves as arguably the most real-world evaluation of CATE estimators, for example, when A/B testing data is available. Using only observed treatment and outcomes, one can compute the Area-Under the QINI Curve (AUC-QINI) to evaluate CATE estimators by their ability to identify individuals for which the treatment has a strong impact. 

\paragraph{Strong Performance in Uplift Modeling.} We report the mean normalized AUC-QINI score for the T/X/S meta-learners using TabPFN-2.5 and 3 (Figure \ref{fig:causal_inference}). TabPFN-3 used as an S and T-Learner achieves stronger performance than other baselines. We observe somewhat worsened performance on the RealCause benchmark \cite{neal_realcause}, which is characterized by smaller sample sizes.

\subsection{Detailed TabArena Results}
\label{app:tabarena-detailed}

\subsubsection{Evaluation Metrics}
\label{app:tabarena-metrics}

We re-use the official TabArena \citep{erickson2025tabarena} evaluation metrics and code for generating TabArena plots and tables.

\textbf{Elo}: Following TabArena, we evaluate models using the Elo rating system~\citep{elo1967proposed}. Elo is a pairwise
comparison-based rating system where each model’s rating predicts its expected win probability against others, with a 400-point Elo gap corresponding to a 10 to 1 (91\%) expected win rate. We calibrate 1000 Elo to the performance of the default TabArena random forest configuration across all figures, and perform 200 rounds of bootstrapping to obtain 95\% confidence intervals, similar to what is done in ChatBot Arena \cite{chiang2024chatbot}. In our TabArena results, Elo scores are computed using ROC AUC for binary classification, log-loss for multiclass classification, and RMSE for regression.

\textbf{Improvability}: The improvability metric introduced in TabArena measures how many percent lower the error of the best method is than the current method on a dataset. This is then averaged over datasets. Formally, for a single dataset,
    \begin{equation*}
        \operatorname{Improvability} := \frac{\err_i - \besterr_i}{\err_i} \cdot 100\%~.
    \end{equation*}
    Improvability is always between $0\%$ and $100\%$.

\subsubsection{Experiment Details}

For all TabArena results, we run experiments using the official TabArena code and evaluation pipeline. We will contribute a reproducible official TabArena submission for \ourmodel shortly after it becomes publicly available. While not strictly necessary to make predictions on test data, we follow TabArena's fit time procedure of fitting an 8-fold bagged ensemble to generate a cross-validation score followed by refitting the model on the full training data at test time, as is done for the other tabular foundation models on TabArena.

All results for non-\ourmodel models in our TabArena experiments were from the official TabArena reported results. All cached results from tabular foundation models (TabPFN-2.5, TabPFN-2.6, TabICLv2 and TabDPT) were run on a single H200 GPU, while all results for \ourmodel and \ourmodelenhanced were run on a single RTX 6000 GPU, a weaker GPU compared to an H200.

For both \ourmodel and \ourmodelenhanced, we ran all splits of TabArena, which includes a total of 816 tasks across 51 datasets. In all cases we report results for all splits of each dataset.

\subsubsection{TabArena Pareto Frontier Explanation}
\label{app:tabarena_pareto_frontier_explanation}

Figure \ref{fig:tabarena_pareto_medium} and Figure \ref{fig:tabarena_pareto} show TabArena Pareto frontiers of models across Improvability and the median combined train + inference time per 1000 samples. The connected points for a given model type indicate tuning + ensembling performance with points from left to right marking ensembles of increasing numbers of random configurations (1, 2, 5, 10, 25, 50, 100, 150, 201). The trajectories are sampled 20 times from all
trials and averaged. The left-most points use the default configuration, and the right-most highlighted points use all configurations.

\subsubsection{TabArena Leaderboard Tables}
\label{sec:tabarena_leaderboard_tables}

We present the leaderboard tables for
\hyperref[tab:tabarena_table]{TabArena},
\hyperref[tab:tabarena_medium_table]{TabArena-medium},
\hyperref[tab:tabarena_small_table]{TabArena-small}, \hyperref[tab:tabarena_classification_table]{TabArena-classification}, and \hyperref[tab:tabarena_regression_table]{TabArena-regression}, below.

For all 5 views, \ourmodel ranks highest among all models on TabArena, while \ourmodelenhanced pushes even futher, strongly outperforming AutoGluon 1.5 extreme and ranking first in Elo, wins, and Improvability in every leaderboard.

\begin{table}[H]
\centering
\caption{\textbf{TabArena leaderboard using all 51 datasets with 816 total tasks.}}
\label{tab:tabarena_table}
\input{figures/tabarena_v3/all/leaderboard.tex}
\end{table}
\begin{table}[H]
\centering
\caption{\textbf{TabArena-medium leaderboard on the 15 largest datasets in TabArena}, with 10k--100k training samples, evaluated on the full 135 tasks with 9 splits per dataset.}
\label{tab:tabarena_medium_table}
\input{figures/tabarena_v3/Medium/leaderboard.tex}
\end{table}

\begin{table}[H]
\centering
\caption{\textbf{TabArena-small leaderboard on the 36 smallest datasets in TabArena}, with 500--10k training samples, evaluated on the full 681 tasks.}
\label{tab:tabarena_small_table}
\input{figures/tabarena_v3/Small/leaderboard.tex}
\end{table}

\begin{table}[H]
\centering
\caption{\textbf{TabArena-classification leaderboard on the 38 classification datasets in TabArena.}}
\label{tab:tabarena_classification_table}
\input{figures/tabarena_v3/classification/leaderboard.tex}
\end{table}

\begin{table}[H]
\centering
\caption{\textbf{TabArena-regression leaderboard on the 13 regression datasets in TabArena.}}
\label{tab:tabarena_regression_table}
\input{figures/tabarena_v3/regression/leaderboard.tex}
\end{table}

\subsection{Details on TALENT benchmark results}
\label{app:TALENT}

\subsubsection{Benchmark description}
TALENT \citep{talent_benchmark_jmlr} base contains 300 datasets (120 binary, 80 multiclass, 100 regression).
Each dataset is split into 64\% training, 16\% validation, and 20\% test sets.

\paragraph{Baselines.} We rely on precomputed baselines provided by the authors of
the TALENT benchmark \citep{talent_benchmark_jmlr} (for the TALENT extensions which we
use for the large-rows slice and the many-class slice) or the TabICLv2 paper
\citep{qu2026tabiclv2} (for the main TALENT slice).

\paragraph{Metrics.} Following the TALENT paper and \cite{qu2026tabiclv2}, we use accuracy for classification and rmse for regression.

\paragraph{Datasets.} Following \cite{qu2026tabiclv2}, we exclude the 26 development datasets used for
TabPFN-2 / TabICLv2 development from the main TALENT benchmark.

\subsubsection{Per-task-type breakdown}
\label{app:TALENT-per-task}

\begin{figure}[H]
  \centering
  \includegraphics[width=\linewidth]{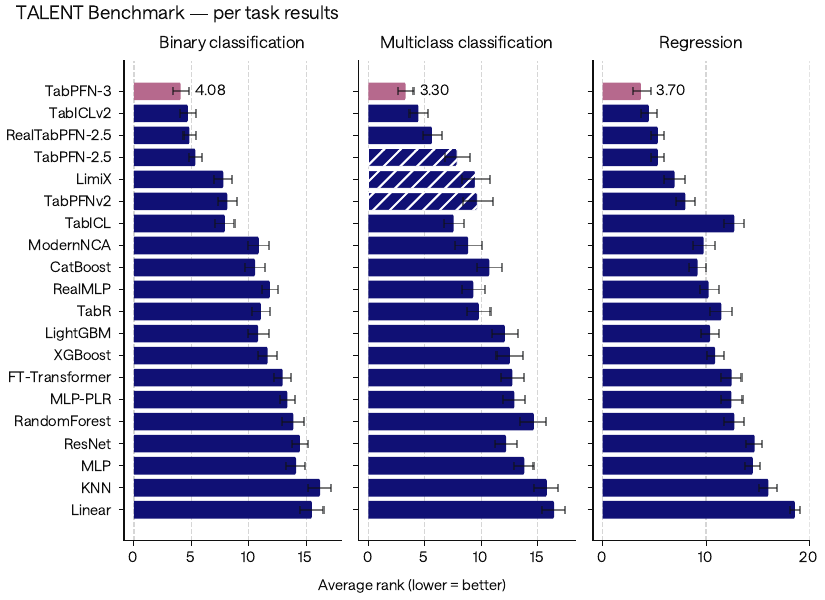}
  \caption{\textbf{Average rank on the TALENT benchmark broken down by task type
    (regression, binary classification, multiclass classification), using the
    TabICLv2 evaluation protocol from \citet{qu2026tabiclv2}.} Bars show mean rank
    (lower is better); error bars are 95\% bootstrap confidence intervals over
    datasets. Hatched bars mark methods with KNN-imputed scores. TabPFN-2.5,
    LimiX, and TabPFNv2 share a 10-class cap, so their scores on the 12
    multiclass datasets with $>$10 classes are KNN-imputed.}
  \label{fig:per-task-TALENT-rank}
\end{figure}

\subsubsection{Many-class TALENT subset}
\label{app:TALENT-many-class}

\begin{figure}[H]
  \centering
  \begin{minipage}[c]{0.52\linewidth}
    \centering
    \footnotesize
    \begin{tabular}{lrrr}
      \toprule
      Dataset                              & Classes & Samples & Feat. \\
      \midrule
      \texttt{one-hundred-plants-margin}   & 100 & 1{,}600  & 64 \\
      \texttt{one-hundred-plants-shape}    & 100 & 1{,}600  & 64 \\
      \texttt{one-hundred-plants-texture}  & 100 & 1{,}599  & 64 \\
      \texttt{helena}                      & 100 & 65{,}196 & 27 \\
      \bottomrule
    \end{tabular}
  \end{minipage}\hfill
  \begin{minipage}[c]{0.45\linewidth}
    \centering
    \includegraphics[width=\linewidth]{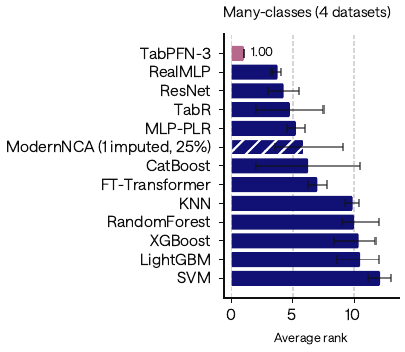}
  \end{minipage}
  \caption{\textbf{Average rank on the many-classes TALENT slice
    (4 datasets, all 100 classes).} Three are the
    \texttt{one-hundred-plants} variants (margin / shape / texture, $\approx$1.6\,k
    samples each) and one is \texttt{helena} (65\,k samples).}
  \label{fig:talent_many_class}
\end{figure}

We report results on the subset of TALENT~\cite{talent_benchmark_jmlr} datasets with more than $50$ classes, which yields 4 datasets with 100 classes, including 3 from the same family. While limited in number, these complement the results on synthetic data from Section \ref{sec:many-class-eval}. Results are shown in Figure~\ref{fig:talent_many_class}.

\subsubsection{Large rows subset}

\begin{figure}[H]
\centering

\begin{minipage}[t]{0.53\linewidth}
\vspace{0pt}
\centering
\scriptsize

\resizebox{\linewidth}{!}{%
\begin{tabular}{lrrl}
\toprule
Dataset & Samples & Feat. & Task \\
\midrule
\texttt{microsoft} & 1{,}200{,}192 & 136 & Reg. \\
\texttt{poker-hand} & 1{,}025{,}009 & 10 & Multi. \\
\texttt{BNG(credit-a)} & 1{,}000{,}000 & 15 & Binary \\
\texttt{Higgs} & 1{,}000{,}000 & 28 & Binary \\
\texttt{Smoking\_and\_Drinking\_Dataset\_with\_body\_signal} & 991{,}346 & 23 & Binary \\
\texttt{yahoo} & 709{,}877 & 699 & Reg. \\
\texttt{Data\_Science\_for\_Good\_Kiva\_Crowdfunding} & 671{,}205 & 11 & Multi. \\
\texttt{covertype} & 581{,}012 & 54 & Multi. \\
\texttt{CDC\_Diabetes\_Health\_Indicators} & 253{,}680 & 21 & Binary \\
\texttt{accelerometer} & 153{,}004 & 4 & Multi. \\
\texttt{walking-activity} & 149{,}332 & 4 & Multi. \\
\texttt{Rain\_in\_Australia} & 145{,}460 & 18 & Multi. \\
\texttt{customer\_satisfaction\_in\_airline} & 129{,}880 & 21 & Binary \\
\texttt{diabetes\_130-us\_hospitals} & 101{,}766 & 20 & Binary \\
\bottomrule
\end{tabular}%
}

\vspace{3.5em}
\captionof{table}{Datasets in the large-rows TALENT slice.}
\label{tab:large_rows_datasets}

\end{minipage}\hfill
\begin{minipage}[t]{0.42\linewidth}
\vspace{0pt}
\centering

\includegraphics[width=0.82\linewidth]
{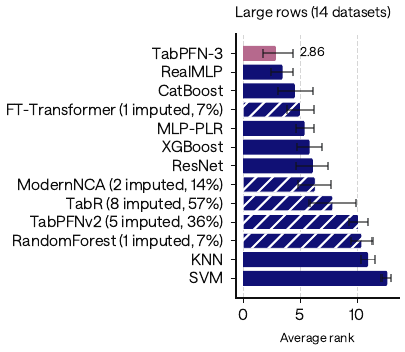}

\captionof{figure}{Average rank on the large-rows (100k-1M rows) TALENT slice.}
\label{fig:large_rows_rank}

\end{minipage}

\end{figure}

We report here the list of datasets in the large-rows subset of TALENT we use in 
Section \ref{sec:large-data}. The datasets are filtered for $>$100k samples and $\leq$1M training samples from the TALENT base and large extension. We report the model ranking in Figure \ref{fig:large_rows_rank}.

\subsubsection{Details}

\paragraph{Per-dataset ranking.} For each (dataset, split) we rank all
methods by their score (best $=1$; ties get average ranks). The reported
\emph{mean rank} of a method is the average of these for ranks across all
(dataset, split) pairs in the slice.

\paragraph{Bootstrap confidence intervals.} 95\% confidence intervals are
non-parametric bootstrap over datasets: for each of $B = 2{,}000$
replicates we resample the (dataset, split) pairs with replacement and
recompute each method's mean rank, then take the empirical $2.5/97.5$
percentiles across replicates.

\subsection{Details on TabSTAR Text-Tabular Benchmark results}
\label{app:TABSTAR}

\begin{figure}[H]
    \centering
    \includegraphics[width=0.8\linewidth]{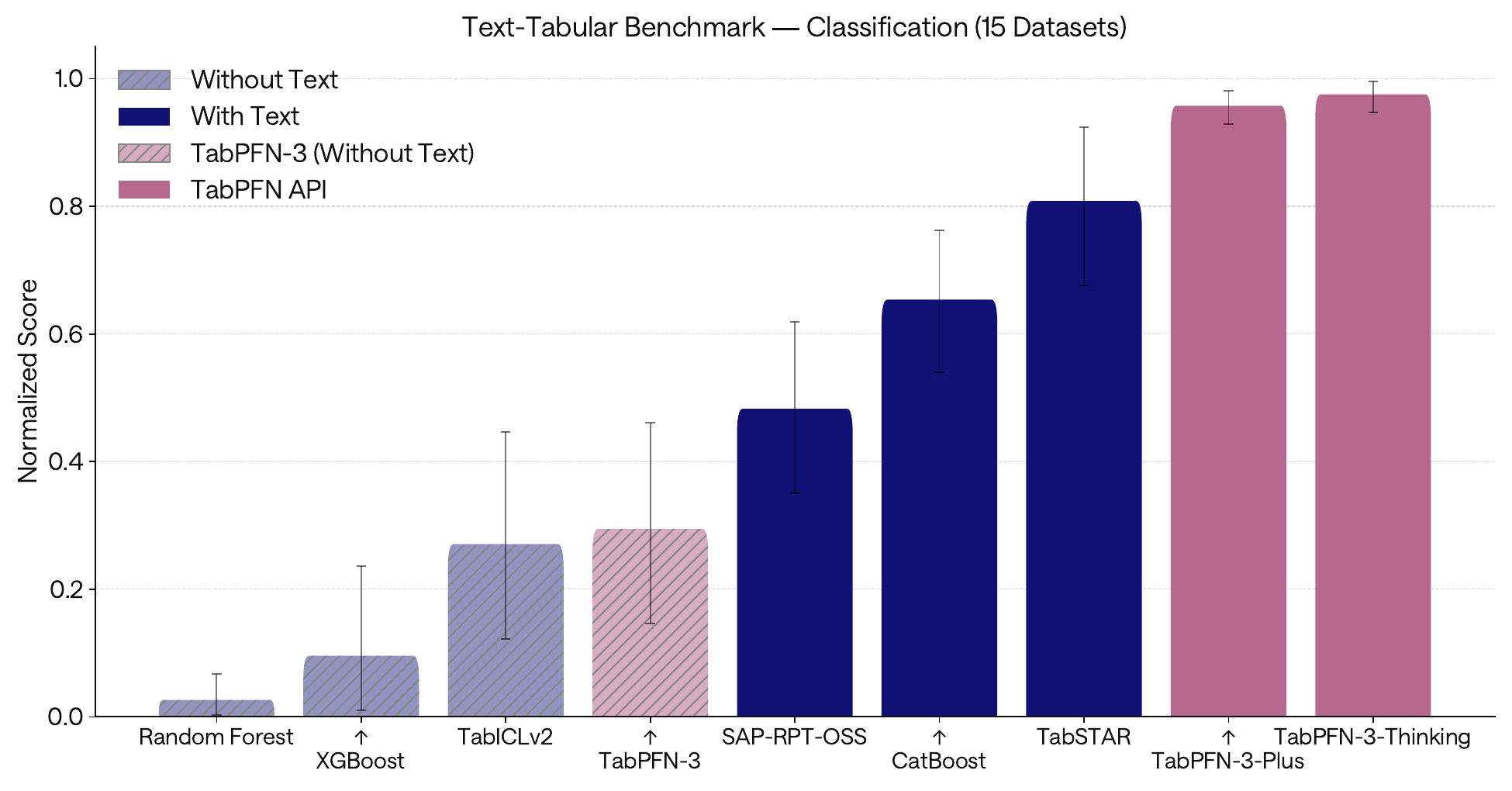}
    \caption{\textbf{Performance on the classification tasks of the TabSTAR text-tabular collection.} \ourmodelenhanced and TabPFN-3-Plus significantly
  outperform the text-aware TabSTAR, which was otherwise the state-of-the-art reference for this task type.}
    \label{fig:text_leaderboard_cls}
  \end{figure}

\begin{figure}[H]
    \centering
    \includegraphics[width=0.8\linewidth]{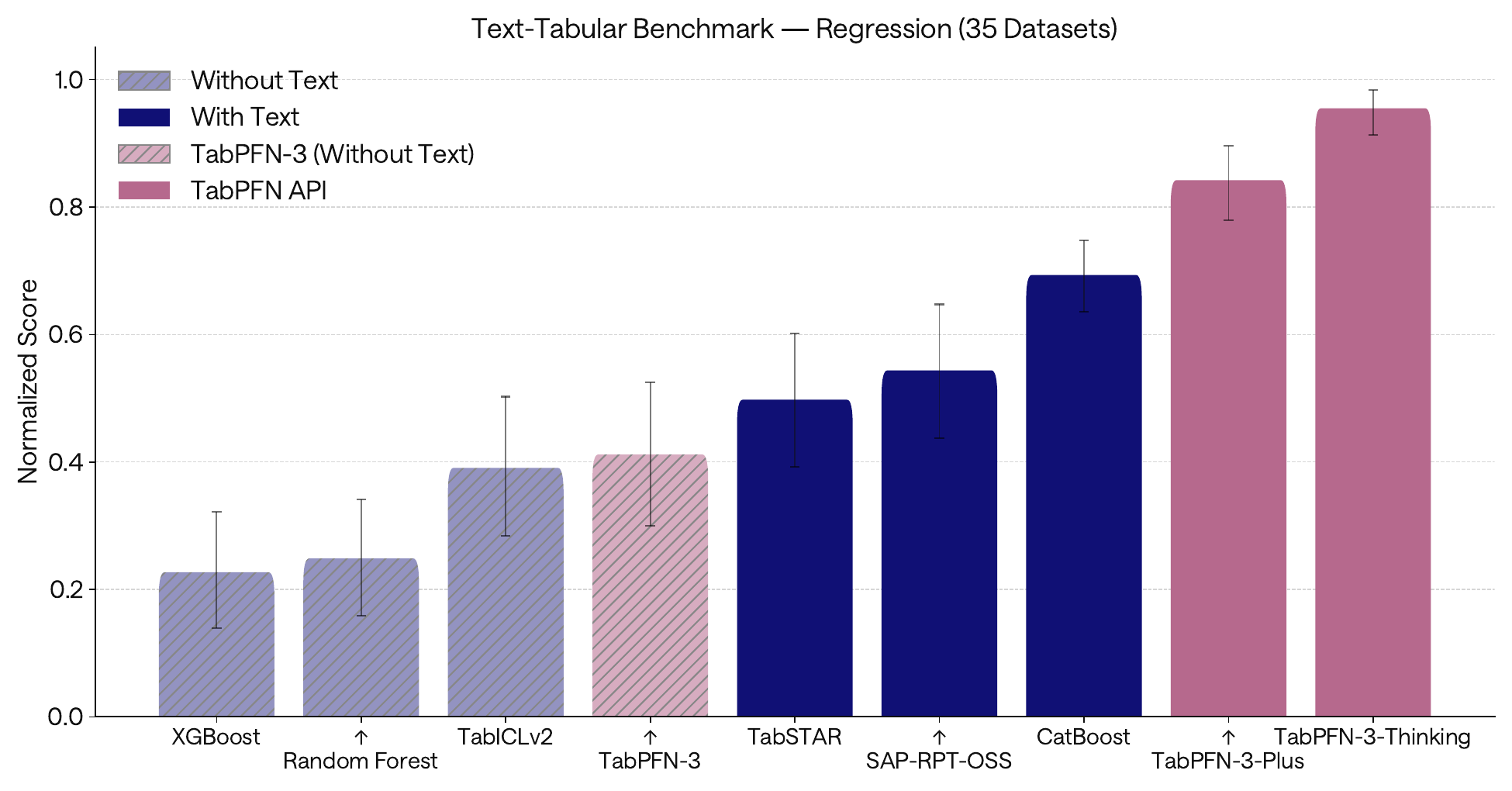}
    \caption{\textbf{Performance on the regression tasks of the TabSTAR text-tabular collection.} \ourmodelenhanced and TabPFN-3-Plus significantly
  outperform all baselines.}
    \label{fig:text_leaderboard_reg}
  \end{figure}

The TabSTAR benchmark is a union of previous text-tabular benchmarks: the Multimodal AutoML Benchmark \cite{shi2021benchmarking},
\citet{grinsztajn2023vectorizing}, and CARTE \cite{kim2024carte}. After deduplication and exclusion of unavailable datasets, the final benchmark contains 50 datasets: 15 classification and 35 regression tasks.\footnote{The TabSTAR paper reports 14 classification tasks, having mistakenly treated \textit{Spotify Genres} as a regression dataset.} Each model is run 5 times, with per-task metrics AUROC (binary classification), log-loss (multiclass), and RMSE (regression); results are normalized with MinMax scaling to the $[0,1]$ range. As in the original paper
\cite{arazi_tabstar_2025}, we limit each run to up to 100,000 examples. Figure~\ref{fig:text_leaderboard_cls} shows the results for
classification, for which the TabSTAR model was reportedly the state of the art; we see that the TabPFN API family significantly outperforms
it. Figure~\ref{fig:text_leaderboard_reg} shows the equivalent regression performance.

\subsection{Per-dataset results on RelBenchV1}
\label{sec:app/relbench/per-dataset-results}

We report per-dataset results for entity regression and entity classification as well as aggregate metrics in \autoref{tab:relbench/classification} and \autoref{tab:relbench/regression}, respectively.

\begin{table}[H]
\centering
\caption{\textbf{RelBenchV1 entity classification: per-task ROC AUC$\times$100.} \textbf{Bold} = best per task / column, \underline{underlined} = second best. Methods marked with $^{*}$ in their name (KumoRFMv1, \rtzero~) indicate methods that are likely following a different evaluation protocol, which overestimates model performance. KumoRFMv2 was reevaluated by us using the evaluation scripts provided by the authors; cells marked $^{*}$ in this row are imputed with the author's results because they were not supported in the evaluation scripts.}
\label{tab:relbench/classification}
\setlength{\tabcolsep}{2pt}
\resizebox{\textwidth}{!}{%
\begin{tabular}{lrrrrrrrrrrrrrr}
\toprule
Method & \multicolumn{2}{c}{\texttt{f1}} & \multicolumn{2}{c}{\texttt{avito}} & \multicolumn{2}{c}{\texttt{event}} & \multicolumn{1}{c}{\texttt{trial}} & \multicolumn{2}{c}{\texttt{amazon}} & \multicolumn{2}{c}{\texttt{stack}} & \multicolumn{1}{c}{\texttt{hm}} & Avg AUROC $\uparrow$ & Rank $\downarrow$ \\
 & \texttt{\scriptsize dnf} & \texttt{\scriptsize top3} & \texttt{\scriptsize click} & \texttt{\scriptsize visit} & \texttt{\scriptsize repeat} & \texttt{\scriptsize ignore} & \texttt{\scriptsize out} & \texttt{\scriptsize user} & \texttt{\scriptsize item} & \texttt{\scriptsize eng} & \texttt{\scriptsize badge} & \texttt{\scriptsize churn} &  &  \\
\midrule
RelGNN & 75.29 & 85.69 & 68.23 & 66.18 & \textbf{79.61} & \underline{86.18} & 71.24 & \textbf{70.99} & \underline{82.64} & \textbf{90.75} & \textbf{88.98} & \textbf{70.93} & \textbf{78.06} & \textbf{2.83} \\
RelGT & 75.87 & 83.52 & 68.30 & \underline{66.78} & 76.09 & 81.57 & 68.61 & 70.39 & 82.55 & 90.53 & 86.32 & 69.27 & 76.65 & 4.50 \\
GraphSAGE & 72.62 & 75.54 & 65.90 & 66.20 & 76.89 & 81.62 & 68.60 & \underline{70.42} & \textbf{82.81} & 90.59 & \underline{88.86} & 69.88 & 75.83 & 5.17 \\
\addlinespace[3pt]
KumoRFMv1$^{*}$ & \textbf{82.41} & \textbf{91.07} & 64.85 & 64.11 & 76.08 & \textbf{89.20} & 70.79 & 67.29 & 79.93 & 87.09 & 80.00 & 67.71 & 76.71 & 6.75 \\
Griffin & 57.70 & 82.50 & 45.90 & 60.70 & 71.88 & 83.27 & 51.00 & 62.30 & 69.00 & 77.50 & 73.50 & 60.20 & 66.29 & 9.92 \\
RT$_{\text{zero}}^{*}$ & \underline{81.20} & \underline{89.30} & 59.50 & 61.80 & 73.22 & 77.47 & 51.80 & 64.00 & 70.90 & 75.70 & 80.10 & 62.80 & 70.65 & 8.67 \\
RDBLearn & 70.87 & 79.69 & \underline{69.04} & 65.49 & 75.04 & 82.52 & 71.58 & 67.57 & 82.07 & 89.39 & 85.26 & 68.05 & 75.55 & 6.83 \\
\addlinespace[3pt]
RDBLearn + v2.5 & 71.72 & 77.60 & 65.72 & 66.47 & 75.55 & 78.65 & \underline{72.90} & 69.74 & 82.18 & 90.23 & 82.81 & 70.11 & 75.31 & 6.46 \\
RDBLearn + v3 & 71.72 & 82.72 & \textbf{69.06} & 66.76 & 76.81 & 73.70 & 72.89 & 69.35 & 82.46 & 90.59 & 85.98 & 70.06 & 76.01 & 4.83 \\
\addlinespace[3pt]
KumoRFMv2 & 72.03 & 82.09 & 67.42$^{*}$ & \textbf{69.41$^{*}$} & \underline{79.34} & 78.86 & 72.03$^{*}$ & 67.71 & 80.18 & 88.69 & 85.40 & 67.81 & 75.91 & 5.75 \\
\addlinespace[3pt]
\textbf{TabPFN-REL} & 70.74 & 79.98 & 67.09 & 66.68 & 77.11 & 85.38 & \textbf{76.43} & 70.27 & \textbf{82.81} & \underline{90.66} & 85.17 & \underline{70.55} & \underline{76.91} & \underline{4.29} \\
\bottomrule
\end{tabular}}
\end{table}

\begin{table}[H]
\centering
\caption{\textbf{RelBenchV1 entity regression: per-task MAE.} \textbf{Bold} = best per task / column (ties bolded together at the displayed precision), \underline{underlined} = second best. Right column: LightGBM-normalised mean $S_{\text{KumoNorm}} = \mathrm{mean}_t \, \mathrm{MAE}_t / \mathrm{MAE}^{\text{LightGBM}}_t$. Methods marked with $^{*}$ in their name (KumoRFMv1, \rtzero~) indicate methods that are likely following a different evaluation protocol, which overestimates model performance. KumoRFMv2 was reevaluated by us using the evaluation scripts provided by the authors; cells marked $^{*}$ in this row are imputed with the author's results because they were not supported in the evaluation scripts. On one task (\texttt{rel-amazon/item}) marked with $^{**}$ we fall back to the default context size of $5000$, because the context exceeded the API's 30MB size limits.}
\label{tab:relbench/regression}
\setlength{\tabcolsep}{2pt}
\resizebox{\textwidth}{!}{%
\begin{tabular}{lrrrrrrrrrrr}
\toprule
Method & \multicolumn{1}{c}{\texttt{f1}} & \multicolumn{1}{c}{\texttt{avito}} & \multicolumn{1}{c}{\texttt{event}} & \multicolumn{2}{c}{\texttt{trial}} & \multicolumn{2}{c}{\texttt{amazon}} & \multicolumn{1}{c}{\texttt{stack}} & \multicolumn{1}{c}{\texttt{hm}} & Avg $S_{\text{KumoNorm}}$ $\downarrow$ & Rank $\downarrow$ \\
 & \texttt{\scriptsize pos} & \texttt{\scriptsize ctr} & \texttt{\scriptsize attend} & \texttt{\scriptsize adverse} & \texttt{\scriptsize succ} & \texttt{\scriptsize user} & \texttt{\scriptsize item} & \texttt{\scriptsize votes} & \texttt{\scriptsize sales} &  &  \\
\midrule
RelGNN & 3.798 & 0.037 & \underline{0.238} & 44.461 & \textbf{0.301} & \textbf{14.230} & 48.767 & \textbf{0.065} & 0.054 & \textbf{0.861} & \underline{3.72} \\
RelGT & 3.917 & 0.035 & 0.250 & 43.992 & \underline{0.326} & \underline{14.267} & 48.922 & \textbf{0.065} & 0.054 & 0.870 & 4.67 \\
GraphSAGE & 4.022 & 0.041 & 0.258 & 44.473 & 0.400 & 14.313 & 50.053 & \textbf{0.065} & 0.056 & 0.918 & 6.39 \\
\addlinespace[3pt]
KumoRFMv1$^{*}$ & \textbf{2.747} & 0.035 & 0.264 & 58.231 & 0.417 & 16.161 & 55.254 & \textbf{0.065} & \textbf{0.040} & 0.908 & 6.06 \\
Griffin & 4.460 & 0.050 & 0.461 & 78.232 & 0.463 & 35.590 & 53.214 & 0.092 & 0.151 & 1.471 & 10.56 \\
RT$_{\text{zero}}^{*}$ & \underline{2.901} & 0.058 & 0.379 & 73.999 & 0.455 & 18.802 & 57.996 & 0.110 & 0.089 & 1.240 & 9.44 \\
RDBLearn & 3.834 & 0.034 & \textbf{0.237} & 43.913 & 0.424 & 14.540 & 48.559 & \underline{0.068} & 0.064 & 0.906 & 5.11 \\
\addlinespace[3pt]
RDBLearn + v2.5 & 3.930 & 0.034 & 0.243 & 43.409 & 0.429 & 14.463 & 49.053 & \underline{0.068} & 0.066 & 0.913 & 6.28 \\
RDBLearn + v3 & 3.835 & 0.034 & 0.245 & 43.290 & 0.375 & 14.720 & 50.097 & \underline{0.068} & 0.064 & 0.898 & 5.89 \\
\addlinespace[3pt]
KumoRFMv2 & 4.022 & \underline{0.033} & 0.241 & \underline{41.974} & 0.433$^{*}$ & 14.627 & \textbf{45.352} & \textbf{0.065$^{**}$} & \underline{0.043} & 0.866 & 4.33 \\
\addlinespace[3pt]
\textbf{TabPFN-REL} & 3.757 & \textbf{0.031} & 0.241 & \textbf{40.202} & 0.385 & 14.359 & \underline{46.199} & \underline{0.068} & 0.059 & \underline{0.864} & \textbf{3.56} \\
\bottomrule
\end{tabular}}
\end{table}

%% file: figures/tabarena_v3/all/leaderboard.tex
\begin{tabular}{llccrr}
\toprule
\textbf{Model} & \textbf{Elo ($\uparrow$)} & \textbf{\#wins ($\uparrow$)} & \textbf{Improva-} & \textbf{Train time} & \textbf{Predict time} \\
 &  &  & \textbf{bility ($\downarrow$)} & \textbf{per 1K [s]} & \textbf{per 1K [s]} \\
\midrule
TabPFN-3-Thinking & \textcolor{gold}{\textbf{1800${}_{-72,+105}$}} & \textcolor{gold}{\textbf{13.2}} & \textcolor{gold}{\textbf{4.7\%}} & 37.69 & 3.26 \\
AutoGluon 1.5 (extreme, 4h) & \textcolor{silver}{\textbf{1695${}_{-68,+83}$}} & \textcolor{bronze}{\textbf{5.8}} & \textcolor{silver}{\textbf{5.7\%}} & 289.07 & 4.03 \\
TabPFN-3 (D) & \textcolor{bronze}{\textbf{1677${}_{-62,+86}$}} & \textcolor{silver}{\textbf{6.3}} & \textcolor{bronze}{\textbf{6.9\%}} & 2.31 & 0.74 \\
TabPFN-2.6 (D) & 1623${}_{-56,+78}$ & 1.3 & 8.7\% & 5.48 & 0.55 \\
RealTabPFN-2.5 (T+E) & 1602${}_{-62,+79}$ & 2.1 & 8.3\% & 2040.22 & 8.92 \\
TabICLv2 (D) & 1599${}_{-64,+77}$ & 5.3 & 7.7\% & 4.02 & 0.38 \\
RealTabPFN-2.5 (T) & 1559${}_{-56,+69}$ & 1.4 & 9.1\% & 2040.22 & 1.22 \\
RealTabPFN-2.5 (D) & 1526${}_{-48,+66}$ & 0.9 & 9.5\% & 5.81 & 0.64 \\
RealMLP (T+E) & 1514${}_{-45,+58}$ & 0.5 & 11.2\% & 2950.72 & 11.99 \\
TabDPT (T+E) & 1461${}_{-54,+63}$ & 2.0 & 11.7\% & 4907.64 & 286.65 \\
TabM (T+E) & 1449${}_{-44,+56}$ & 1.0 & 12.6\% & 3285.87 & 1.47 \\
LightGBM (T+E) & 1438${}_{-31,+36}$ & 0.1 & 13.6\% & 416.98 & 2.64 \\
RealMLP (T) & 1433${}_{-47,+48}$ & 0.4 & 12.5\% & 2950.72 & 0.66 \\
CatBoost (T+E) & 1420${}_{-42,+41}$ & 0.1 & 13.2\% & 1658.41 & 0.65 \\
CatBoost (T) & 1410${}_{-45,+41}$ & 0.5 & 13.4\% & 1658.41 & 0.08 \\
TabDPT (T) & 1405${}_{-56,+60}$ & 0.7 & 12.9\% & 4907.64 & 39.96 \\
TabM (T) & 1392${}_{-43,+54}$ & 0.3 & 13.5\% & 3285.87 & 0.17 \\
LightGBM (T) & 1390${}_{-29,+33}$ & 0.0 & 14.3\% & 416.98 & 0.33 \\
XGBoost (T+E) & 1379${}_{-35,+34}$ & 0.1 & 14.4\% & 693.49 & 1.69 \\
CatBoost (D) & 1371${}_{-44,+40}$ & 0.2 & 14.2\% & 6.83 & 0.08 \\
XGBoost (T) & 1354${}_{-35,+33}$ & 0.0 & 14.7\% & 693.49 & 0.31 \\
TabDPT (D) & 1326${}_{-56,+68}$ & 0.3 & 15.3\% & 47.62 & 43.74 \\
TabM (D) & 1299${}_{-44,+49}$ & 0.2 & 15.7\% & 10.49 & 0.13 \\
RealMLP (D) & 1234${}_{-37,+38}$ & 0.1 & 17.1\% & 10.06 & 1.69 \\
XGBoost (D) & 1215${}_{-38,+39}$ & 0.0 & 17.5\% & 1.94 & 0.12 \\
LightGBM (D) & 1189${}_{-29,+34}$ & 0.0 & 18.0\% & 1.96 & 0.14 \\
\bottomrule
\end{tabular}

%% file: figures/tabarena_v3/Medium/leaderboard.tex
\begin{tabular}{llccrr}
\toprule
\textbf{Model} & \textbf{Elo ($\uparrow$)} & \textbf{\#wins ($\uparrow$)} & \textbf{Improva-} & \textbf{Train time} & \textbf{Predict time} \\
 &  &  & \textbf{bility ($\downarrow$)} & \textbf{per 1K [s]} & \textbf{per 1K [s]} \\
\midrule
TabPFN-3-Thinking & \textcolor{gold}{\textbf{2146${}_{-87,+121}$}} & \textcolor{gold}{\textbf{6.2}} & \textcolor{gold}{\textbf{1.3\%}} & 15.10 & 2.15 \\
AutoGluon 1.5 (extreme, 4h) & \textcolor{silver}{\textbf{1907${}_{-50,+92}$}} & 1.4 & \textcolor{silver}{\textbf{3.4\%}} & 191.18 & 2.21 \\
TabPFN-3 (D) & \textcolor{bronze}{\textbf{1835${}_{-137,+224}$}} & \textcolor{silver}{\textbf{3.3}} & \textcolor{bronze}{\textbf{4.1\%}} & 0.83 & 0.27 \\
TabPFN-2.6 (D) & 1741${}_{-72,+121}$ & 0.0 & 6.4\% & 2.76 & 0.70 \\
TabICLv2 (D) & 1712${}_{-108,+208}$ & \textcolor{bronze}{\textbf{1.9}} & 5.3\% & 0.76 & 0.14 \\
RealTabPFN-2.5 (T+E) & 1663${}_{-111,+149}$ & 0.0 & 7.2\% & 735.58 & 11.74 \\
RealMLP (T+E) & 1645${}_{-94,+91}$ & 0.0 & 7.4\% & 1719.82 & 1.67 \\
CatBoost (T+E) & 1625${}_{-64,+86}$ & 0.0 & 7.4\% & 777.59 & 0.25 \\
CatBoost (T) & 1616${}_{-67,+95}$ & 0.3 & 7.6\% & 777.59 & 0.05 \\
RealTabPFN-2.5 (T) & 1612${}_{-103,+130}$ & 0.1 & 7.9\% & 735.58 & 1.39 \\
LightGBM (T+E) & 1604${}_{-56,+70}$ & 0.0 & 9.2\% & 131.56 & 2.64 \\
CatBoost (D) & 1576${}_{-106,+105}$ & 0.1 & 7.8\% & 3.24 & 0.03 \\
XGBoost (T+E) & 1565${}_{-61,+90}$ & 0.1 & 9.3\% & 282.13 & 0.56 \\
RealMLP (T) & 1554${}_{-85,+106}$ & 0.0 & 8.7\% & 1719.82 & 0.08 \\
TabM (T+E) & 1538${}_{-90,+157}$ & 0.7 & 9.1\% & 1993.14 & 0.62 \\
RealTabPFN-2.5 (D) & 1536${}_{-90,+141}$ & 0.0 & 8.7\% & 1.88 & 0.64 \\
TabDPT (T+E) & 1533${}_{-124,+142}$ & 0.8 & 8.8\% & 4786.55 & 444.54 \\
LightGBM (T) & 1515${}_{-59,+80}$ & 0.0 & 10.3\% & 131.56 & 0.13 \\
XGBoost (T) & 1514${}_{-56,+69}$ & 0.0 & 9.8\% & 282.13 & 0.07 \\
TabM (T) & 1489${}_{-90,+158}$ & 0.0 & 9.9\% & 1993.14 & 0.06 \\
TabDPT (T) & 1411${}_{-125,+121}$ & 0.0 & 11.3\% & 4786.55 & 42.64 \\
XGBoost (D) & 1375${}_{-115,+101}$ & 0.0 & 11.7\% & 0.49 & 0.05 \\
TabDPT (D) & 1336${}_{-144,+131}$ & 0.0 & 14.0\% & 46.62 & 43.74 \\
TabM (D) & 1330${}_{-101,+123}$ & 0.0 & 12.6\% & 5.16 & 0.07 \\
RealMLP (D) & 1280${}_{-71,+79}$ & 0.0 & 13.7\% & 6.75 & 0.23 \\
LightGBM (D) & 1263${}_{-63,+55}$ & 0.0 & 13.5\% & 0.29 & 0.04 \\
\bottomrule
\end{tabular}

%% file: figures/tabarena_v3/Small/leaderboard.tex
\begin{tabular}{llccrr}
\toprule
\textbf{Model} & \textbf{Elo ($\uparrow$)} & \textbf{\#wins ($\uparrow$)} & \textbf{Improva-} & \textbf{Train time} & \textbf{Predict time} \\
 &  &  & \textbf{bility ($\downarrow$)} & \textbf{per 1K [s]} & \textbf{per 1K [s]} \\
\midrule
TabPFN-3-Thinking & \textcolor{gold}{\textbf{1723${}_{-60,+100}$}} & \textcolor{gold}{\textbf{7.0}} & \textcolor{gold}{\textbf{6.1\%}} & 52.78 & 3.40 \\
AutoGluon 1.5 (extreme, 4h) & \textcolor{silver}{\textbf{1641${}_{-57,+79}$}} & \textcolor{silver}{\textbf{4.4}} & \textcolor{silver}{\textbf{6.6\%}} & 346.57 & 6.56 \\
TabPFN-3 (D) & \textcolor{bronze}{\textbf{1638${}_{-58,+85}$}} & 2.9 & \textcolor{bronze}{\textbf{8.1\%}} & 4.84 & 1.54 \\
RealTabPFN-2.5 (T+E) & 1598${}_{-64,+97}$ & 2.1 & 8.7\% & 2289.05 & 8.05 \\
TabPFN-2.6 (D) & 1596${}_{-49,+74}$ & 1.3 & 9.7\% & 7.03 & 0.55 \\
TabICLv2 (D) & 1574${}_{-83,+105}$ & \textcolor{bronze}{\textbf{3.4}} & 8.7\% & 7.06 & 0.67 \\
RealTabPFN-2.5 (T) & 1556${}_{-58,+75}$ & 1.2 & 9.5\% & 2289.05 & 1.14 \\
RealTabPFN-2.5 (D) & 1542${}_{-52,+83}$ & 0.9 & 9.9\% & 6.76 & 0.64 \\
RealMLP (T+E) & 1482${}_{-47,+63}$ & 0.5 & 12.7\% & 3770.75 & 21.90 \\
TabDPT (T+E) & 1448${}_{-59,+76}$ & 1.2 & 12.9\% & 5119.36 & 218.71 \\
TabM (T+E) & 1430${}_{-52,+57}$ & 0.4 & 14.0\% & 3553.12 & 1.74 \\
TabDPT (T) & 1414${}_{-60,+72}$ & 0.7 & 13.6\% & 5119.36 & 28.35 \\
RealMLP (T) & 1402${}_{-42,+55}$ & 0.4 & 14.2\% & 3770.75 & 1.78 \\
LightGBM (T+E) & 1392${}_{-34,+37}$ & 0.1 & 15.5\% & 892.41 & 2.57 \\
TabM (T) & 1368${}_{-53,+56}$ & 0.3 & 15.0\% & 3553.12 & 0.24 \\
CatBoost (T+E) & 1362${}_{-43,+46}$ & 0.1 & 15.6\% & 2476.51 & 0.81 \\
LightGBM (T) & 1357${}_{-30,+36}$ & 0.0 & 15.9\% & 892.41 & 0.35 \\
CatBoost (T) & 1351${}_{-35,+48}$ & 0.1 & 15.8\% & 2476.51 & 0.10 \\
TabDPT (D) & 1331${}_{-67,+74}$ & 0.3 & 15.9\% & 50.32 & 43.71 \\
XGBoost (T+E) & 1326${}_{-37,+34}$ & 0.0 & 16.5\% & 884.18 & 2.37 \\
CatBoost (D) & 1312${}_{-35,+35}$ & 0.1 & 16.9\% & 9.64 & 0.13 \\
XGBoost (T) & 1309${}_{-39,+32}$ & 0.0 & 16.7\% & 884.18 & 0.39 \\
TabM (D) & 1296${}_{-47,+54}$ & 0.2 & 17.0\% & 13.18 & 0.17 \\
RealMLP (D) & 1224${}_{-42,+37}$ & 0.1 & 18.5\% & 15.69 & 4.69 \\
LightGBM (D) & 1169${}_{-40,+42}$ & 0.0 & 19.9\% & 3.61 & 0.17 \\
XGBoost (D) & 1165${}_{-38,+30}$ & 0.0 & 19.9\% & 3.29 & 0.25 \\
\bottomrule
\end{tabular}

%% file: figures/tabarena_v3/classification/leaderboard.tex
\begin{tabular}{llccrr}
\toprule
\textbf{Model} & \textbf{Elo ($\uparrow$)} & \textbf{\#wins ($\uparrow$)} & \textbf{Improva-} & \textbf{Train time} & \textbf{Predict time} \\
 &  &  & \textbf{bility ($\downarrow$)} & \textbf{per 1K [s]} & \textbf{per 1K [s]} \\
\midrule
TabPFN-3-Thinking & \textcolor{gold}{\textbf{1782${}_{-72,+109}$}} & \textcolor{gold}{\textbf{10.0}} & \textcolor{gold}{\textbf{6.0\%}} & 35.70 & 3.00 \\
AutoGluon 1.5 (extreme, 4h) & \textcolor{silver}{\textbf{1689${}_{-82,+96}$}} & \textcolor{silver}{\textbf{4.8}} & \textcolor{silver}{\textbf{6.5\%}} & 267.31 & 3.98 \\
TabPFN-3 (D) & \textcolor{bronze}{\textbf{1660${}_{-75,+91}$}} & 3.7 & \textcolor{bronze}{\textbf{8.7\%}} & 2.43 & 0.75 \\
TabPFN-2.6 (D) & 1604${}_{-69,+69}$ & 0.5 & 10.6\% & 5.17 & 0.54 \\
TabICLv2 (D) & 1593${}_{-75,+94}$ & \textcolor{bronze}{\textbf{4.1}} & 9.3\% & 4.15 & 0.41 \\
RealTabPFN-2.5 (T+E) & 1578${}_{-75,+76}$ & 1.7 & 10.2\% & 2046.25 & 8.98 \\
RealTabPFN-2.5 (T) & 1554${}_{-66,+72}$ & 1.2 & 11.0\% & 2046.25 & 1.33 \\
RealTabPFN-2.5 (D) & 1539${}_{-63,+69}$ & 0.9 & 11.2\% & 5.76 & 0.79 \\
RealMLP (T+E) & 1492${}_{-45,+63}$ & 0.3 & 13.5\% & 2879.46 & 12.49 \\
TabM (T+E) & 1464${}_{-48,+75}$ & 1.0 & 14.8\% & 2466.21 & 1.50 \\
LightGBM (T+E) & 1436${}_{-37,+48}$ & 0.1 & 15.7\% & 382.05 & 1.49 \\
RealMLP (T) & 1413${}_{-47,+55}$ & 0.4 & 15.0\% & 2879.46 & 0.60 \\
CatBoost (T+E) & 1412${}_{-47,+55}$ & 0.1 & 15.2\% & 1372.94 & 0.56 \\
TabM (T) & 1411${}_{-58,+71}$ & 0.3 & 15.6\% & 2466.21 & 0.18 \\
TabDPT (T+E) & 1411${}_{-56,+80}$ & 0.5 & 14.5\% & 4940.61 & 307.75 \\
CatBoost (T) & 1404${}_{-45,+54}$ & 0.4 & 15.4\% & 1372.94 & 0.07 \\
LightGBM (T) & 1392${}_{-33,+43}$ & 0.0 & 16.4\% & 382.05 & 0.25 \\
XGBoost (T+E) & 1382${}_{-48,+50}$ & 0.1 & 16.5\% & 685.87 & 1.45 \\
CatBoost (D) & 1381${}_{-46,+46}$ & 0.2 & 16.0\% & 5.72 & 0.08 \\
XGBoost (T) & 1356${}_{-40,+45}$ & 0.0 & 16.8\% & 685.87 & 0.21 \\
TabDPT (T) & 1351${}_{-58,+66}$ & 0.6 & 16.0\% & 4940.61 & 41.61 \\
TabM (D) & 1315${}_{-48,+56}$ & 0.2 & 18.0\% & 10.21 & 0.14 \\
TabDPT (D) & 1270${}_{-57,+62}$ & 0.3 & 18.9\% & 49.21 & 43.82 \\
RealMLP (D) & 1244${}_{-34,+39}$ & 0.1 & 19.6\% & 10.47 & 1.71 \\
XGBoost (D) & 1231${}_{-50,+47}$ & 0.0 & 19.6\% & 1.77 & 0.12 \\
LightGBM (D) & 1192${}_{-40,+49}$ & 0.0 & 20.6\% & 1.79 & 0.12 \\
\bottomrule
\end{tabular}

%% file: figures/tabarena_v3/regression/leaderboard.tex
\begin{tabular}{llccrr}
\toprule
\textbf{Model} & \textbf{Elo ($\uparrow$)} & \textbf{\#wins ($\uparrow$)} & \textbf{Improva-} & \textbf{Train time} & \textbf{Predict time} \\
 &  &  & \textbf{bility ($\downarrow$)} & \textbf{per 1K [s]} & \textbf{per 1K [s]} \\
\midrule
TabPFN-3-Thinking & \textcolor{gold}{\textbf{1959${}_{-150,+211}$}} & \textcolor{gold}{\textbf{3.2}} & \textcolor{gold}{\textbf{0.9\%}} & 43.00 & 3.26 \\
TabPFN-3 (D) & \textcolor{silver}{\textbf{1827${}_{-142,+255}$}} & \textcolor{silver}{\textbf{2.5}} & \textcolor{silver}{\textbf{1.6\%}} & 1.69 & 0.57 \\
AutoGluon 1.5 (extreme, 4h) & \textcolor{bronze}{\textbf{1804${}_{-97,+133}$}} & 1.1 & 3.2\% & 335.03 & 4.33 \\
TabPFN-2.6 (D) & 1776${}_{-71,+131}$ & 0.8 & 3.3\% & 8.52 & 0.70 \\
RealTabPFN-2.5 (T+E) & 1774${}_{-107,+174}$ & 0.5 & \textcolor{bronze}{\textbf{2.6\%}} & 1709.05 & 8.12 \\
TabDPT (T+E) & 1748${}_{-92,+171}$ & \textcolor{bronze}{\textbf{1.5}} & 3.5\% & 4786.55 & 239.54 \\
TabICLv2 (D) & 1700${}_{-159,+293}$ & 1.2 & 3.2\% & 2.10 & 0.25 \\
TabDPT (T) & 1696${}_{-79,+134}$ & 0.1 & 3.9\% & 4786.55 & 38.50 \\
RealMLP (T+E) & 1677${}_{-68,+126}$ & 0.2 & 4.3\% & 3995.01 & 10.05 \\
RealTabPFN-2.5 (T) & 1654${}_{-113,+165}$ & 0.2 & 3.4\% & 1709.05 & 0.81 \\
TabDPT (D) & 1604${}_{-72,+153}$ & 0.0 & 4.9\% & 46.62 & 39.21 \\
RealMLP (T) & 1574${}_{-84,+114}$ & 0.0 & 5.3\% & 3995.01 & 0.84 \\
RealTabPFN-2.5 (D) & 1558${}_{-110,+159}$ & 0.0 & 4.9\% & 7.04 & 0.51 \\
CatBoost (T+E) & 1513${}_{-73,+113}$ & 0.0 & 7.3\% & 3552.96 & 0.97 \\
LightGBM (T+E) & 1509${}_{-90,+107}$ & 0.0 & 7.7\% & 700.15 & 9.32 \\
CatBoost (T) & 1489${}_{-78,+119}$ & 0.1 & 7.4\% & 3552.96 & 0.10 \\
TabM (T+E) & 1463${}_{-96,+147}$ & 0.0 & 6.2\% & 4158.29 & 1.41 \\
LightGBM (T) & 1440${}_{-77,+119}$ & 0.0 & 8.3\% & 700.15 & 0.97 \\
XGBoost (T+E) & 1424${}_{-52,+72}$ & 0.0 & 8.2\% & 834.93 & 2.61 \\
XGBoost (T) & 1403${}_{-60,+85}$ & 0.0 & 8.4\% & 834.93 & 0.39 \\
CatBoost (D) & 1389${}_{-92,+107}$ & 0.0 & 8.9\% & 10.89 & 0.09 \\
TabM (T) & 1381${}_{-101,+147}$ & 0.0 & 7.1\% & 4158.29 & 0.17 \\
TabM (D) & 1284${}_{-118,+126}$ & 0.0 & 8.8\% & 13.32 & 0.13 \\
RealMLP (D) & 1235${}_{-81,+105}$ & 0.0 & 9.8\% & 8.90 & 1.64 \\
LightGBM (D) & 1210${}_{-35,+40}$ & 0.0 & 10.7\% & 2.11 & 0.27 \\
XGBoost (D) & 1190${}_{-78,+99}$ & 0.0 & 11.3\% & 2.24 & 0.24 \\
\bottomrule
\end{tabular}

%% file: sections/F_additional_internal_benchmarks.tex
\section{Additional Details on Internal Benchmarks}

\subsection{Methodology}
\label{app:methodology}

\paragraph{Metric Normalization.}
\label{sec:metric_norm}

To aggregate heterogeneous metrics across datasets, we apply a per-fold min--max normalization.
For each (dataset, fold) pair and metric $m$, we rescale a model's raw score $s_m^{(b)}$ as
\begin{equation}
    \tilde{s}_m^{(b)} = \frac{s_m^{(b)} - \min_{b' \in \mathcal{B}}\, s_m^{(b')}}{\max_{b' \in \mathcal{B}}\, s_m^{(b')} - \min_{b' \in \mathcal{B}}\, s_m^{(b')}},
\end{equation}
where $\mathcal{B}$ denotes the set of models we evaluate. This allows the model scores to live on a comparable $[0, 1]$ scale for each (dataset, fold) combination. We treat the tuned and default versions of a model as two different models.
For lower-is-better metrics (e.g.\ RMSE, cross-entropy loss), we apply the additional transformation $\tilde{s}_m \mapsto 1 - \tilde{s}_m$, so that all metrics are higher-is-better on a common scale and can be meaningfully averaged or ranked across datasets and metric types.

\paragraph{Statistical significance.}
\label{sec:statistical_significance}

To assess whether performance differences between models are statistically significant, we report critical difference (CD) diagrams using \texttt{scikit-posthocs}~\citep{scikit_posthocs_Terpilowski2019}.
The critical difference diagram from \texttt{scikit-posthocs} summarizes the statistical comparison of methods across multiple datasets. Average ranks are computed per method across all datasets, with lower ranks indicating better performance. 
Methods connected by a horizontal bar are not significantly different from each other.
To assess statistical significance, we use a Friedman test followed by a Conover post hoc analysis at the significance level $\alpha=0.05$.

\subsection{Large Data Benchmark Details}
\label{app:large_data_datasets}

Classification datasets span domains including healthcare (patient
survival, disease diagnosis), customer analytics (satisfaction, credit risk), insurance
(claim prediction), microfinance (loan outcomes), and high-energy physics
(signal/background classification). Regression datasets cover retail sales forecasting,
climate and weather modeling, food delivery logistics, and e-commerce price prediction. All 4 regression datasets use temporal train/test splits reflecting real-world
deployment conditions where the test period strictly follows the training period. For classification, all datasets are IID. These datasets are selected to have between 100K and 1M training rows, and fewer than 200 features, which is the regime TabPFN-3 was designed for.

\begin{figure}[H]
    \centering
    \includegraphics[width=\textwidth]{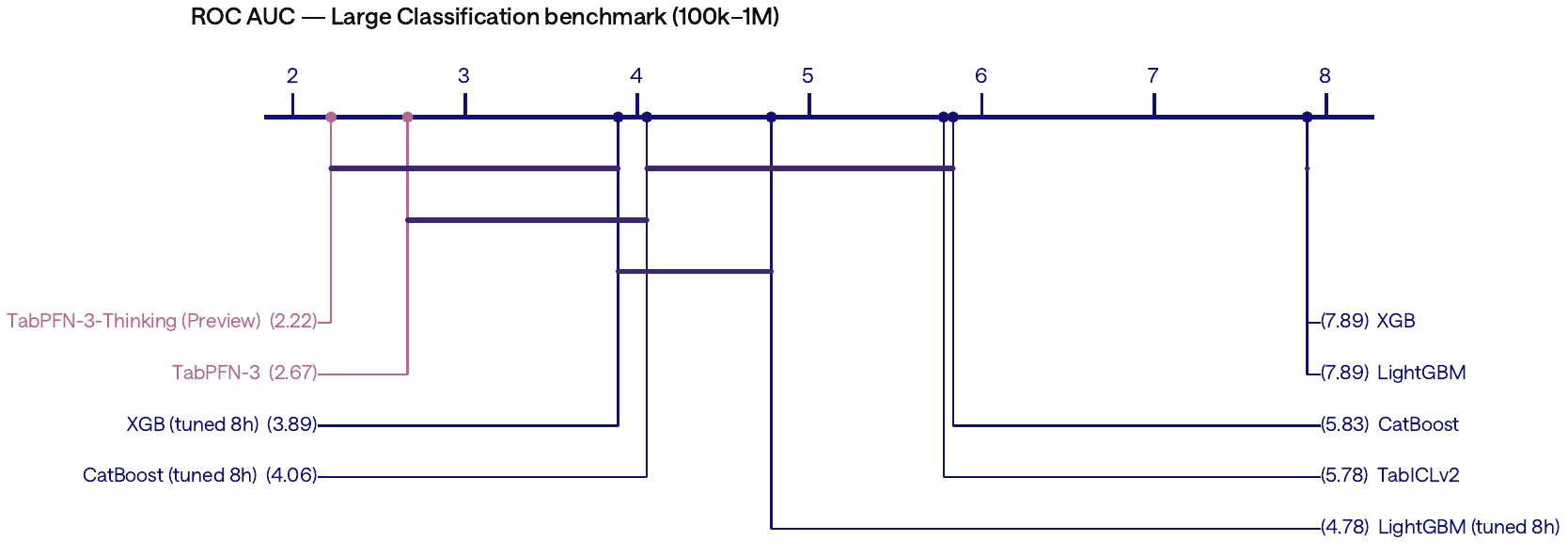}
    \caption{%
        \textbf{Critical difference diagram for ROC-AUC on the large-scale classification
        benchmark (100k--1M training rows).}
        TabPFN-3 ranks first (avg.\ rank 2.11). Its rank differences to the
        8-hour-tuned XGB/CatBoost baselines are not statistically significant,
        while it ranks significantly ahead of tuned LightGBM, all default GBTs,
        and TabICLv2. Bars connect methods whose rank differences are not
        statistically significant at $\alpha = 0.05$ under a Conover-Friedman
        post-hoc test~\citep{scikit_posthocs_Terpilowski2019}..
    }
    \label{fig:large_data_cd_cls}
\end{figure}

\begin{figure}[H]
    \centering
    \includegraphics[width=\textwidth]{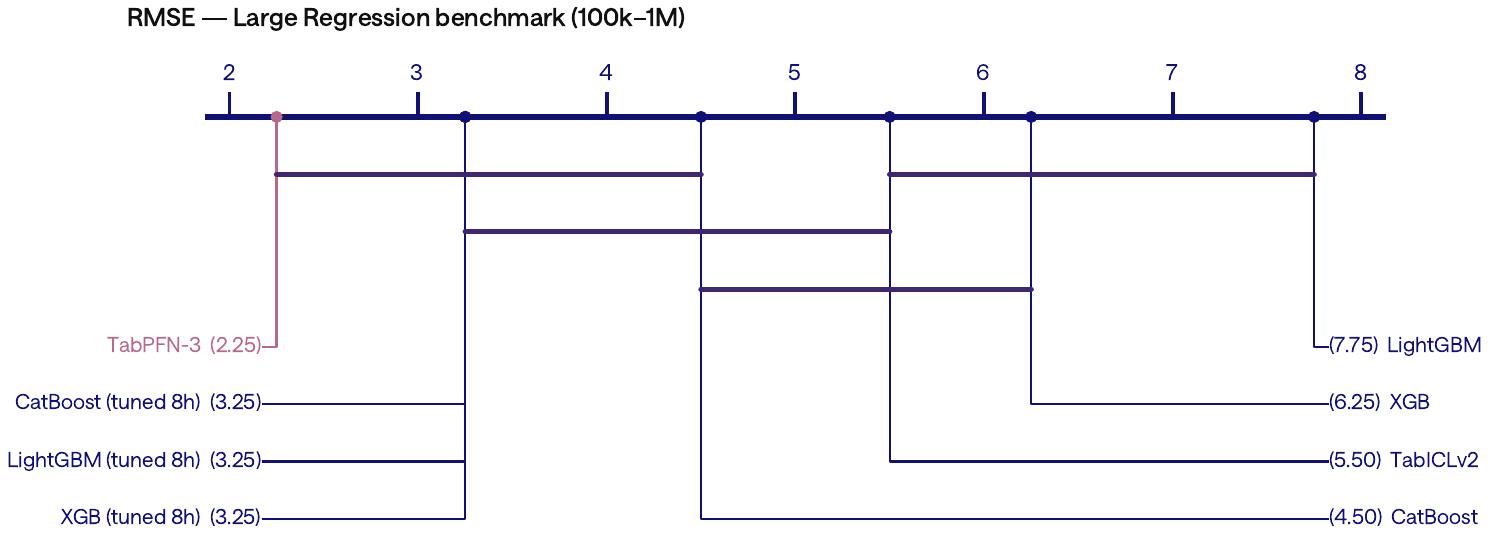}
    \caption{%
         \textbf{Critical difference diagram for RMSE on the large-scale regression
         benchmark (100k--1M rows, 4 datasets, temporal splits).} Methods are
         ranked per (dataset, split); lower rank is better. TabPFN-3 achieves the best average rank ($2.25$). Its rank differences to the three 8h-tuned GBDTs and untuned CatBoost are not statistically significant, while it ranks significantly ahead of the remaining methods. Bars connect methods whose rank differences are not statistically significant at $\alpha = 0.05$ under a Conover-Friedman post-hoc test~\citep{scikit_posthocs_Terpilowski2019}.   
    }
    \label{fig:large_data_cd_reg}
\end{figure}

Figures~\ref{fig:large_data_cd_cls} and~\ref{fig:large_data_cd_reg} show critical difference
diagrams for ROC-AUC and RMSE respectively, based on average ranks across all datasets in
each benchmark.

\subsection{Synthetic Many-Class Benchmark Construction}
\label{app:many_class_construction}

Continuous regression targets are partitioned into $K = 100$ bins using quantile-based
bin edges whose spacings are drawn from a $\mathrm{Dirichlet}(\alpha{=}5.0)$
distribution, producing realistic class imbalance. Bins with fewer than 10 samples are merged
with their nearest neighbour to guarantee sufficient representation for inner cross-validations. 
Class labels are then randomly permuted to remove the implicit ordinal structure inherited from the regression target.

tasets from TabArena whose targets have heavy point masses or too few distinct values to fill 100 quantile bins meaningfully — wine\_quality (7 unique values), Food\_Delivery\_Time (45, discrete times), Fiat-500 (222, discrete prices), and QSAR-TID-11 (concentrated point masses). Dataset statistics are reported in Table~\ref{tab:many_class_datasets}. The resulting benchmark retains a large number of classes for most datasets (median $K=95$), while inducing moderate class imbalance (median IR $=9.9\times$) without collapsing the label distribution onto a few dominant classes (median $H/\log K=0.98$).

\begin{table}[ht]
\centering
\caption{%
    \textbf{Synthetic many-class benchmark datasets derived from continuous regression targets.}
    $N$ is the number of samples before binning. Targets are first partitioned into
    100 quantile-based bins with randomized Dirichlet-spaced bin widths, after which
    bins with fewer than 10 samples are merged into their nearest neighbour. $K$ is
    the resulting number of classes and \textbf{Merged} equals $100-K$. \textbf{Min}
    and \textbf{Max} are the smallest and largest class sizes after merging, and
    \textbf{IR} is their ratio. $H/\log K$ is the Shannon entropy of the class
    distribution normalized by $\log K$, with 1 corresponding to perfectly balanced
    classes.
}
\label{tab:many_class_datasets}
\resizebox{\textwidth}{!}{%
\begin{tabular}{lrrrrrrrrr}
\toprule
\textbf{Dataset} & \textbf{OpenML task} & \textbf{OpenML did} & $N$ & $K$ & \textbf{Merged} & \textbf{Min} & \textbf{Max} & \textbf{IR} & $H/\log K$ \\
\midrule
airfoil\_self\_noise            & 363612 & 46904 & 1{,}503  & 80      & 20 & 10  & 42    & $4.2\times$  & 0.984 \\
concrete\_compressive\_strength & 363625 & 46917 & 1{,}030  & 60      & 40 & 10  & 28    & $2.8\times$  & 0.991 \\
diamonds                        & 363631 & 46923 & 53{,}940 & 100     & 0  & 127 & 1{,}252 & $9.9\times$ & 0.979 \\
healthcare\_insurance\_expenses & 363675 & 46931 & 1{,}338  & 73      & 27 & 10  & 32    & $3.2\times$  & 0.988 \\
houses                          & 363678 & 46934 & 20{,}640 & 97      & 3  & 62  & 965   & $15.6\times$ & 0.974 \\
miami\_housing                  & 363686 & 46942 & 13{,}776 & 95      & 5  & 19  & 339   & $17.8\times$ & 0.970 \\
physiochemical\_protein         & 363693 & 46949 & 45{,}730 & 100     & 0  & 108 & 1{,}214 & $11.2\times$ & 0.982 \\
QSAR\_fish\_toxicity            & 363698 & 46954 & 907      & 51      & 49 & 10  & 37    & $3.7\times$  & 0.980 \\
superconductivity               & 363705 & 46961 & 21{,}263 & 100     & 0  & 28  & 508   & $18.1\times$ & 0.979 \\
\midrule
Aggregate (mean / median)       & ---    & ---   & ---      & 84 / 95 & --- & --- & ---  & $9.6\times$ / $9.9\times$ & 0.98 / 0.98 \\
\bottomrule
\end{tabular}%
}
\end{table}

\subsection{Quantile Regression: Critical Difference Diagram}

\begin{figure}[H]
    \centering
    \includegraphics[width=\textwidth]{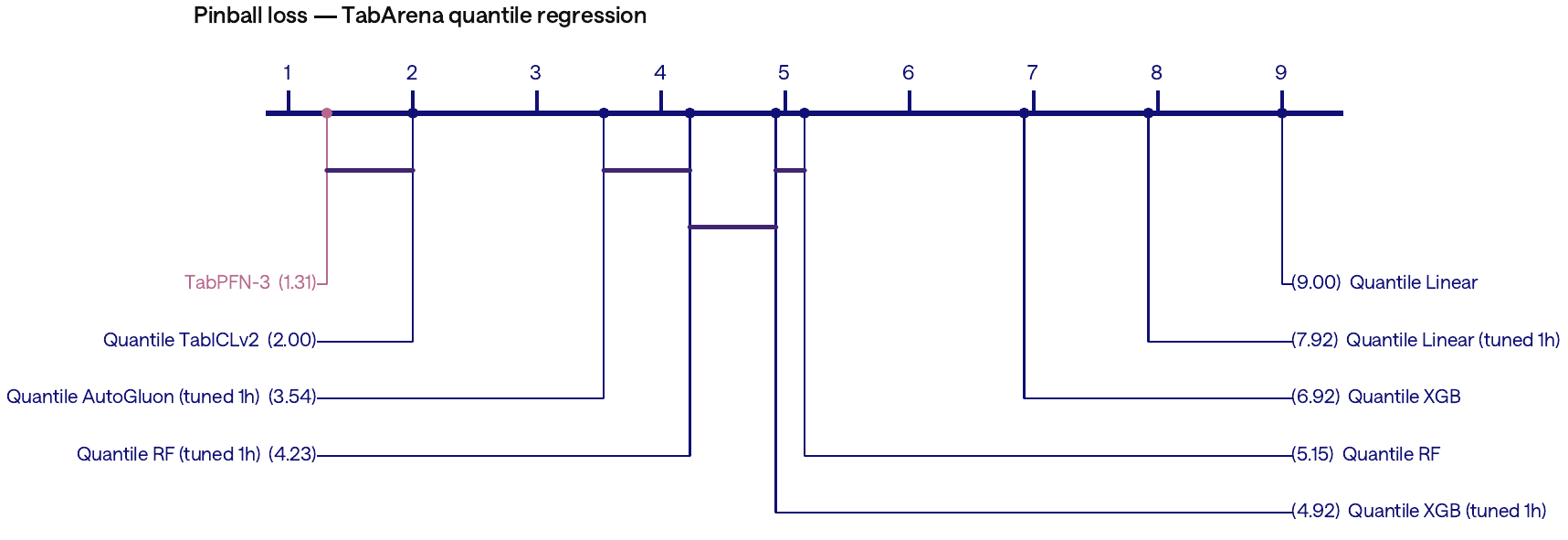}
    \caption{%
        \textbf{Critical difference diagram for pinball loss on our quantile regression benchmark.} The quantile regression benchmark is constructed from TabArena regression datasets and evaluated
        across 10 quantile levels $q \in \{0.1, 0.2, \ldots, 0.9\}$.
        TabPFN-3 ranks first. Its rank difference to Quantile TabICLv2 is not
        statistically significant, while it ranks significantly ahead of all
        remaining baselines. Bars connect methods whose rank differences are not
        statistically significant at $\alpha = 0.05$ under a Conover-Friedman
        post-hoc test~\citep{scikit_posthocs_Terpilowski2019}.
    }
    \label{fig:quantile_cd}
\end{figure}

\subsection{Synthetic Many Class: Critical Difference Diagram}

\begin{figure}[H]
    \centering
    \includegraphics[width=\textwidth]{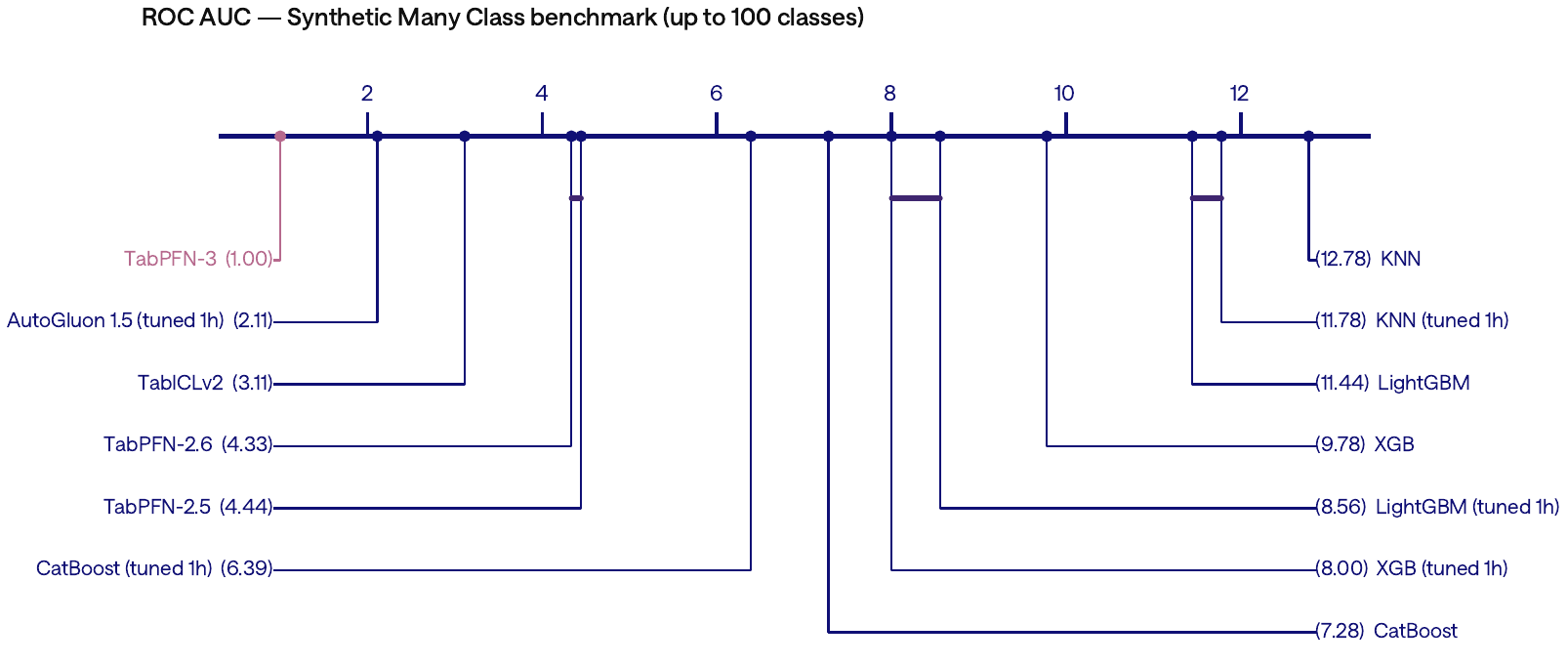}             
    \caption{%
          \textbf{Critical difference diagram for ROC AUC on the synthetic
          many-class benchmark (up to 100 classes).} 
          TabPFN-3 is top-ranked on every (dataset, split) pair and ranks significantly ahead of all baselines. Bars connect methods whose rank differences are not statistically significant at $\alpha = 0.05$ under a Conover-Friedman post-hoc test~\citep{scikit_posthocs_Terpilowski2019}.
      }                                                                                                          
      \label{fig:many_class_roc_auc_cd}                     
  \end{figure}

%% file: sections/G_supplementary_inference_time.tex
\section{Supplementary Inference Time Details}

\subsection{Compilation and FlashAttention-3} \label{app:compile-fa3}

\ourmodel is shipped with two opt-in performance features that target different bottlenecks: \texttt{torch.compile} and FlashAttention-3. At the shapes relevant to large-data inference, the bulk of forward-pass cost is dispatch overhead and attention compute, and the speed-ups of \texttt{torch.compile} and FlashAttention-3 compose cleanly with our chunking strategy without changing the model's behaviour.

\paragraph{torch.compile.} Three hot-path methods are wrapped with \texttt{@torch.compile(dynamic=True)}: feature preprocessing plus embedding grouping, the column-chunk processing block (used in the non-row-chunked path), and the row-chunk processing block (used when chunking is enabled). The \texttt{dynamic=True} mode keeps a single compiled graph across batch and feature-count variation, so the same compiled artefact serves the whole inference grid without re-tracing.

\begin{figure}[!tb]
    \centering

    \begin{subfigure}{\linewidth}
        \centering
        \includegraphics[width=\linewidth]{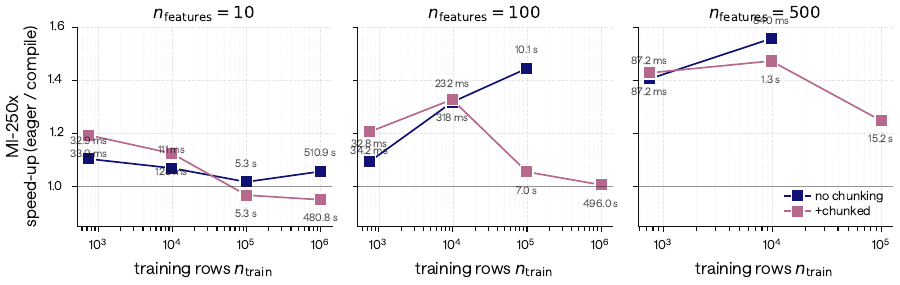}
        \caption{
        MI-250x -- speed-up of \texttt{torch.compile} over eager of
        \ourmodel forward pass, for
        $n_\mathrm{features} \in \{10, 100, 500\}$.
        Values above 1 indicate compile wins.
        }
        \label{fig:compile_vs_eager}
    \end{subfigure}

    \vspace{1em}

    \begin{subfigure}{\linewidth}
        \centering
        \includegraphics[width=\linewidth]{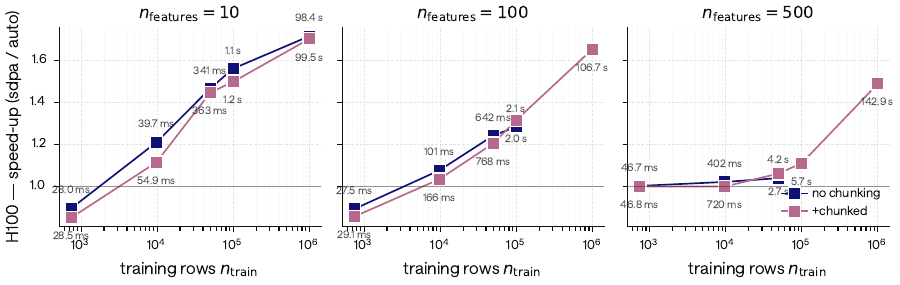}
        \caption{
        H100 -- speed-up of the auto backend
        (Flash Attention 3 where eligible, SDPA fallback elsewhere)
        over the SDPA-only backend on the
        \ourmodel architecture forward pass, for
        $n_\mathrm{features} \in \{10, 100, 500\}$.
        Values above 1 indicate FA3 wins.
        }
        \label{fig:fa3_h100}
    \end{subfigure}

    \caption{
    \textbf{Forward-pass inference speed-ups on the \ourmodel architecture.}
    Top: \texttt{torch.compile} versus eager execution on MI-250x.
    Bottom: auto attention backend (Flash Attention 3 where eligible)
    versus SDPA-only on H100. Marker annotations report absolute execution times.
    }
    \label{fig:inference_speedups}
\end{figure}

Figure~\ref{fig:compile_vs_eager} shows the wall-clock impact on MI-250x. The y-axis is $T_\mathrm{eager} / T_\mathrm{compile}$, so a value above 1 means compile is faster on that shape; each marker is annotated with the absolute time. \texttt{torch.compile} fuses Python-level dispatch into single kernel calls, so it helps most where dispatch is the bottleneck: as $n_\mathrm{features}$ grows, more tensor work becomes compile-able per call. In the non-chunked series the speed-up climbs from $1.04$--$1.15{\times}$ at $n_\mathrm{features}=10$ to $1.10$--$1.46{\times}$ at $n_\mathrm{features}=100$ and $1.40$--$1.58{\times}$ at $n_\mathrm{features}=500$. The chunked series shows the same direction with a different shape: chunking already amortises some dispatch overhead by batching the inner loop, so compile's marginal benefit is largest at small $n_\mathrm{train}$ ($1.21$--$1.43{\times}$ at $n_\mathrm{train}=10^3$) and large $n_\mathrm{features}$ and converges toward parity ($0.95$--$1.06{\times}$) at $n_\mathrm{train} \ge 10^5$ for the smaller feature counts, where the residual cost is dominated by attention itself and compile has no further headroom to claim.

\paragraph{FlashAttention-3.} FlashAttention-3 (FA3) \cite{flash_attention_3} is a Hopper-specific attention kernel that delivers higher throughput and lower memory use than the generic Scaled Dot-Product Attention (SDPA) path. Attention dominates the forward-pass cost of large-$n_\mathrm{train}$ inference, so even a constant-factor improvement in the attention kernel translates into a meaningful end-to-end speed-up. We therefore expose FA3 as an auto-detecting backend: on Hopper-class GPUs with the FA3 library installed, the in-context-learning self-attention -- which carries the bulk of the attention cost at large $n_\mathrm{train}$ -- is routed through FA3, while attention sites whose head dimensions are not FA3-eligible silently fall back to SDPA. On non-Hopper devices (consumer Ada, AMD MI-250x, Blackwell) the same dispatcher selects SDPA.

Figure~\ref{fig:fa3_h100} shows the H100 SDPA-versus-auto comparison in ratio form. The y-axis is $T_\mathrm{sdpa} / T_\mathrm{auto}$, so a value above 1 means FA3 is faster than SDPA on that shape; each marker is annotated with the absolute auto time so the magnitude being sped up is recoverable.
The pattern matches the FA3 design profile. At small training sets ($n_\mathrm{train} \le 1000$) the FA3 dispatch and kernel-launch overhead exceeds the per-call attention work, and SDPA is 10--15\% faster ($T_\mathrm{sdpa}/T_\mathrm{auto} \approx 0.84\text{--}0.91$ across feature counts). The cross-over arrives sooner the smaller $n_\mathrm{features}$: by $n_\mathrm{train}=10^4$ FA3 wins at $n_\mathrm{features}=10$ ($1.21{\times}$), is roughly even at $n_\mathrm{features}=100$ ($1.07{\times}$), and at parity at $n_\mathrm{features}=500$ ($1.02{\times}$). At the inference shapes we care about ($n_\mathrm{train} \ge 10^5$) FA3 is the clear win across all feature counts, with the speed-up climbing to $1.49$--$1.73{\times}$ at $n_\mathrm{train}=10^6$. Chunking does not interact with the FA3-versus-SDPA comparison: the chunked and non-chunked curves overlap to within run-to-run noise, since chunking changes the outer dispatch loop but leaves the underlying attention-kernel selection intact.

\subsection{Interpretability: SHAP-Value Computation}
\label{app:shap-kv-cache}

\begin{figure}[H]
\centering

\begin{subfigure}[t]{0.49\textwidth}
    \centering
    \includegraphics[width=\textwidth]{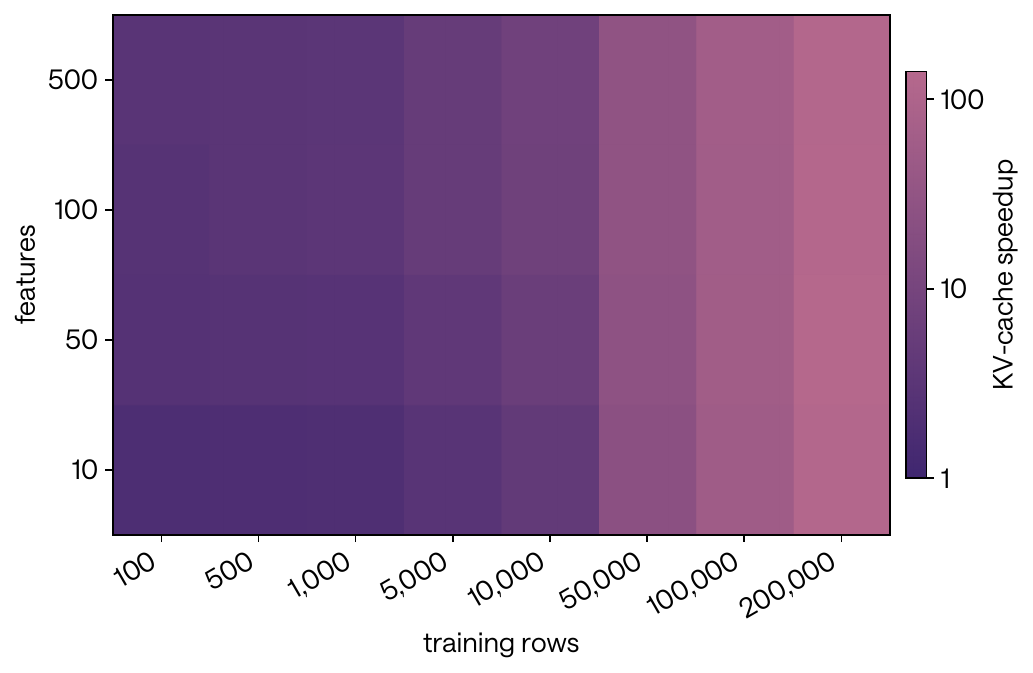}
\end{subfigure}
\hfill
\begin{subfigure}[t]{0.49\textwidth}
    \centering
    \includegraphics[width=\textwidth]{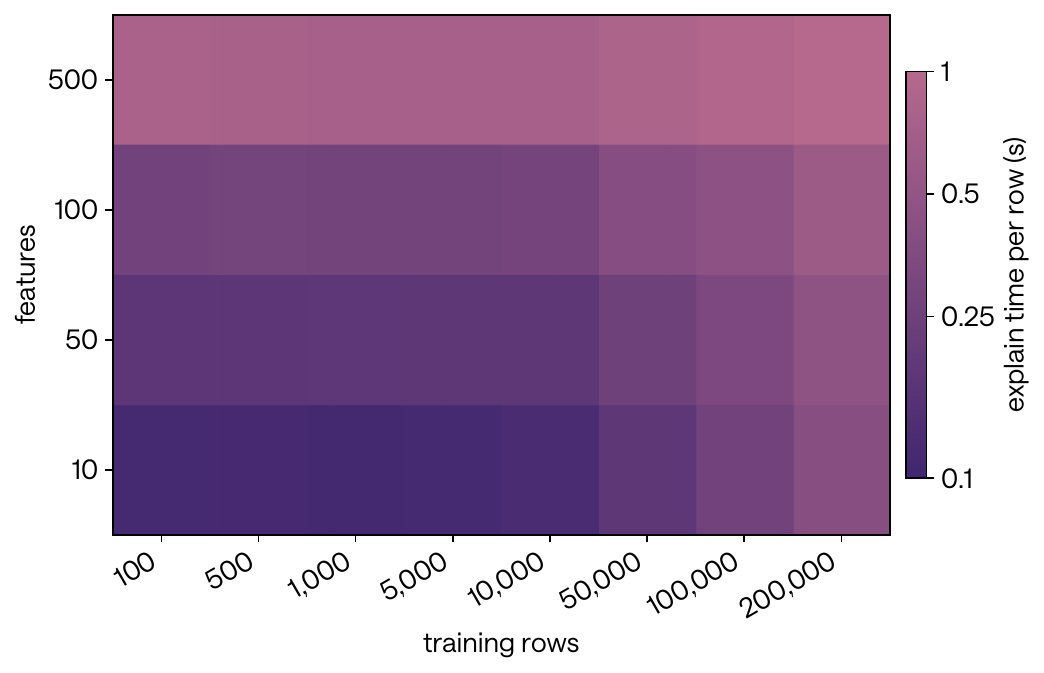}
\end{subfigure}

\caption{\textbf{Efficiency gains for SHAP-value computation with KV-cache across training table dimensions.}
All experiments were conducted on a single RTX Pro 6000 Blackwell with a fixed budget of 1024 coalitions and are averaged over 10 repetitions.
Left: expected speed-up from using KV cache.
Right: expected runtime for computing SHAP values for one test row with KV cache enabled.}
\label{fig:shap-kv-speedup}

\end{figure}

TabPFN-3's improved, smaller KV cache (Section~\ref{sec:kv_cache}) can speed up the computation of SHAP values by multiple order of magnitudes. This is because imputation-based approaches to SHAP-value computation reuse the same \texttt{fit} on many different forward passes. \autoref{fig:shap-kv-speedup} shows the efficiency gains users can expect from enabling the KV cache during SHAP-value-computation.

%% file: sections/H_time_series_results.tex
\section{Detailed Time-Series Forecasting Results on fev-bench}
\label{app:time_series}

\begin{figure}[H]
\centering
\begin{minipage}{0.49\textwidth}
\centering
\includegraphics[width=\linewidth]{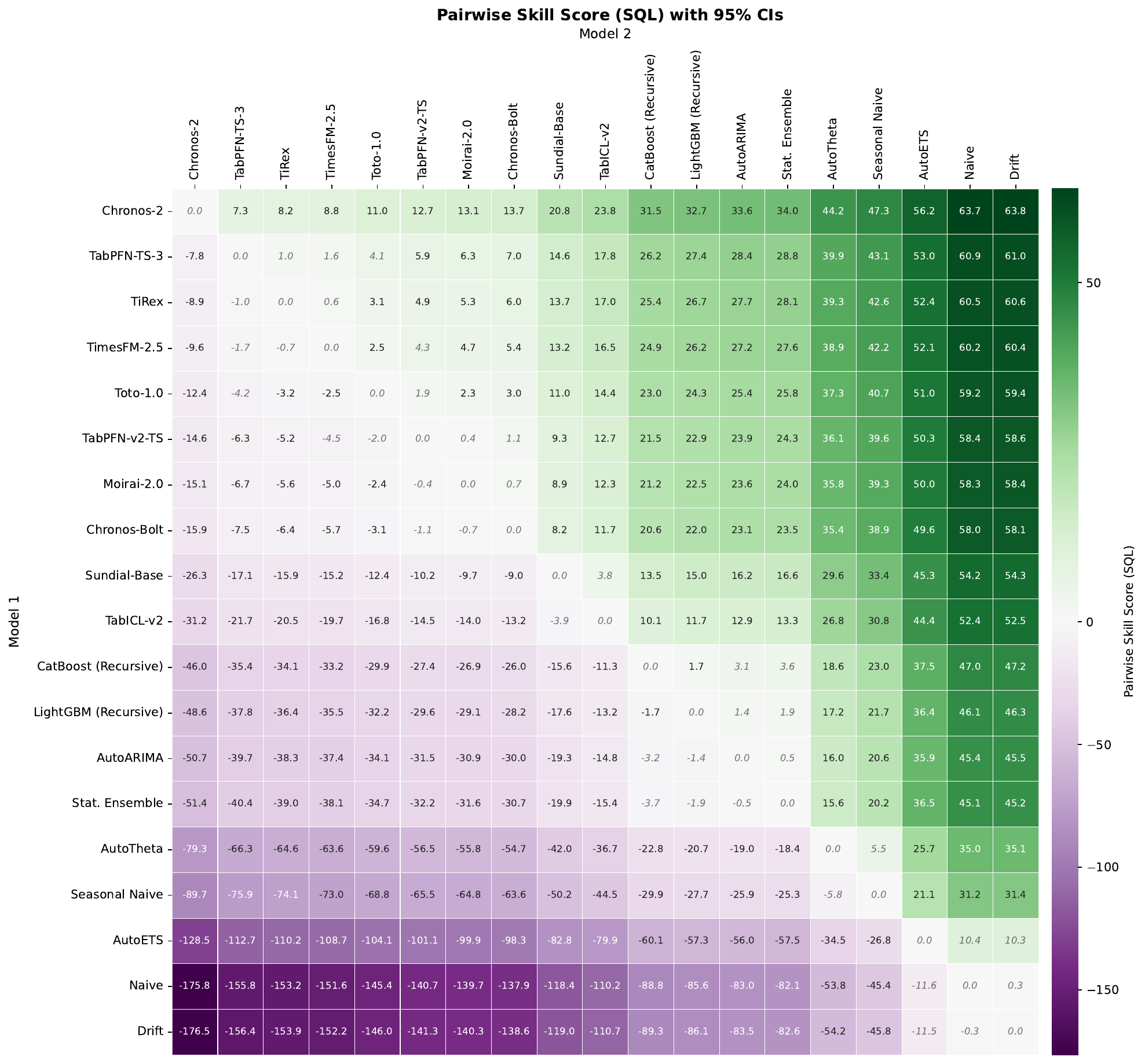}
\end{minipage}
\hfill
\begin{minipage}{0.49\textwidth}
\centering
\includegraphics[width=\linewidth]{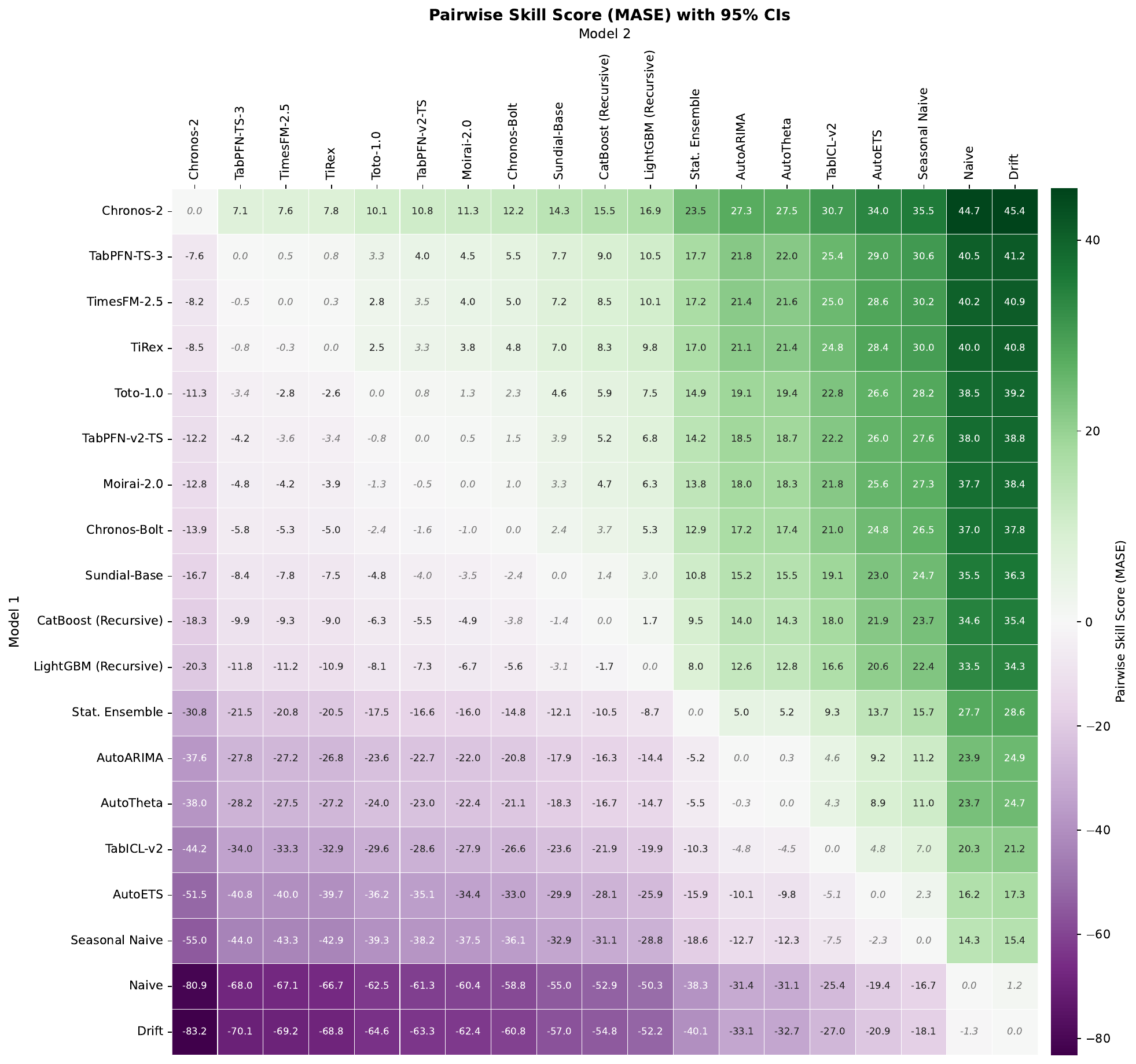}
\end{minipage}
\caption{Pairwise skill-score comparison on fev-bench (100 tasks) under
SQL (left) and MASE (right). Cell $(i, j)$ is the skill score of model
$i$ relative to model $j$, with 95\% confidence intervals from
bootstrapped resampling; cells whose interval overlaps zero are shown
in italics. Rows and columns are ordered by overall skill score.
Best viewed on screen.}
\label{fig:fev-bench-pairwise}
\end{figure}

This appendix complements the body Time-Series subsection
(\Cref{tab:fev-bench}, \Cref{fig:fev-bench-qualitative}) with the full
leaderboards (Table \ref{tab:fev-bench-full}), pairwise comparisons (Figure \ref{fig:fev-bench-pairwise}), additional qualitative forecasts (Section \ref{sec:fev-bench-qualitative} and per-task SQL results (Section \ref{sec:fev-bench-per-task}).

\subsection*{Full leaderboards (SQL and MASE)}

\begin{table}[H]
\centering
\caption{Full marginal forecasting performance on fev-bench (100 tasks),
all 19 baselines, sorted by skill score. The body table
(\Cref{tab:fev-bench}) shows the foundation-model + Stat.\ Ensemble +
Seasonal Naive subset. $^{\dagger}$\,TabICL-v2 results were produced
using the \texttt{tabicl[forecast]} package (v2.0.3) on fev-bench
v0.7.0; results for this model are not currently available in the
official fev-bench results repository.}
\label{tab:fev-bench-full}
\vspace{-3mm}

\begin{minipage}{0.49\textwidth}
\centering
\textbf{(a) SQL (probabilistic)} \\[2pt]
\resizebox{\linewidth}{!}{\input{tables/time_series/leaderboard_SQL}}
\end{minipage}
\hfill
\begin{minipage}{0.49\textwidth}
\centering
\textbf{(b) MASE (point)} \\[2pt]
\resizebox{\linewidth}{!}{\input{tables/time_series/leaderboard_MASE}}
\end{minipage}
\end{table}
 
\subsection*{Qualitative forecast examples}
\label{sec:fev-bench-qualitative}

\begin{figure}[H]
\centering
\includegraphics[width=\textwidth]{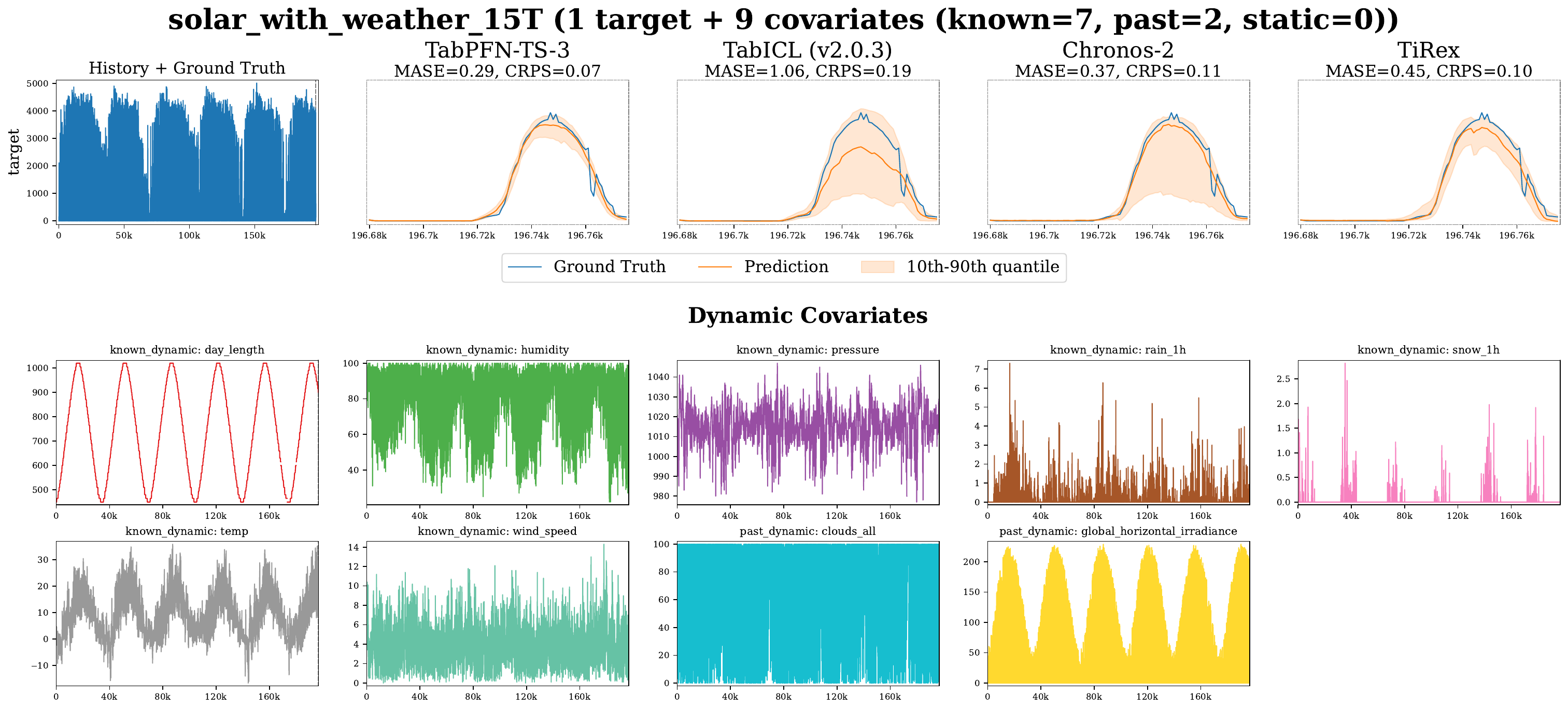}
\caption{\texttt{solar\_with\_weather\_15T} — 15-minute solar
generation with weather covariates.}
\end{figure}

\begin{figure}[H]
\centering
\includegraphics[width=\textwidth]{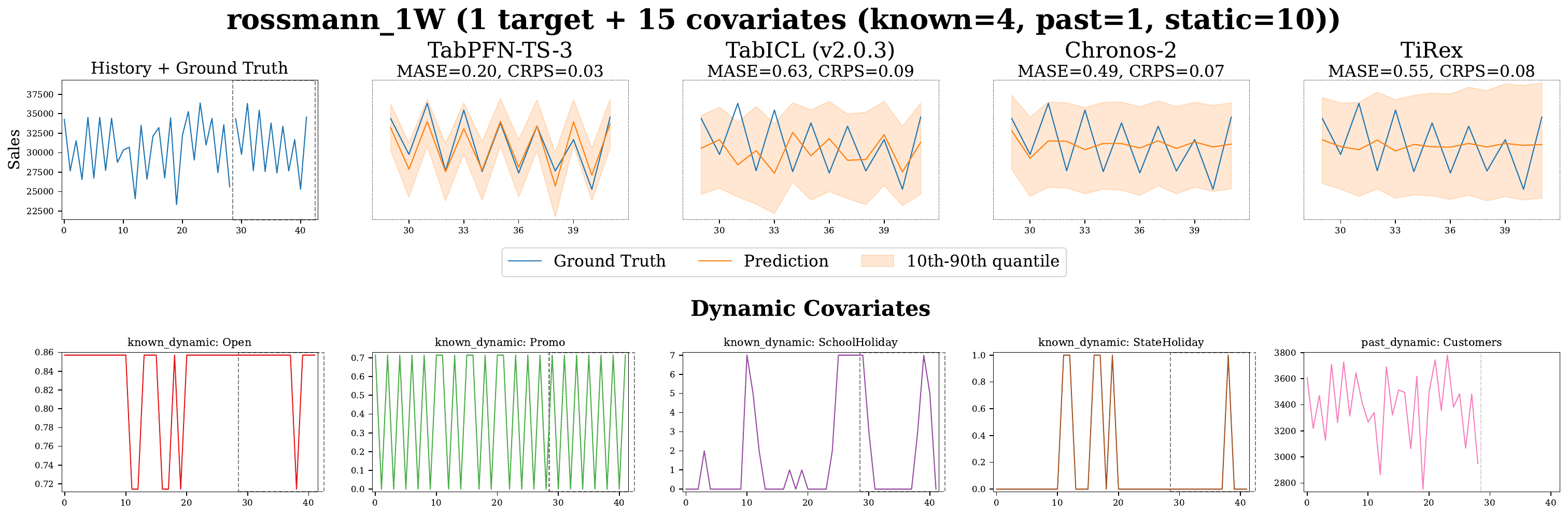}
\caption{\texttt{rossmann\_1W} — weekly Rossmann store sales (series 1).}
\end{figure}

\begin{figure}[H]
\centering
\includegraphics[width=\textwidth]{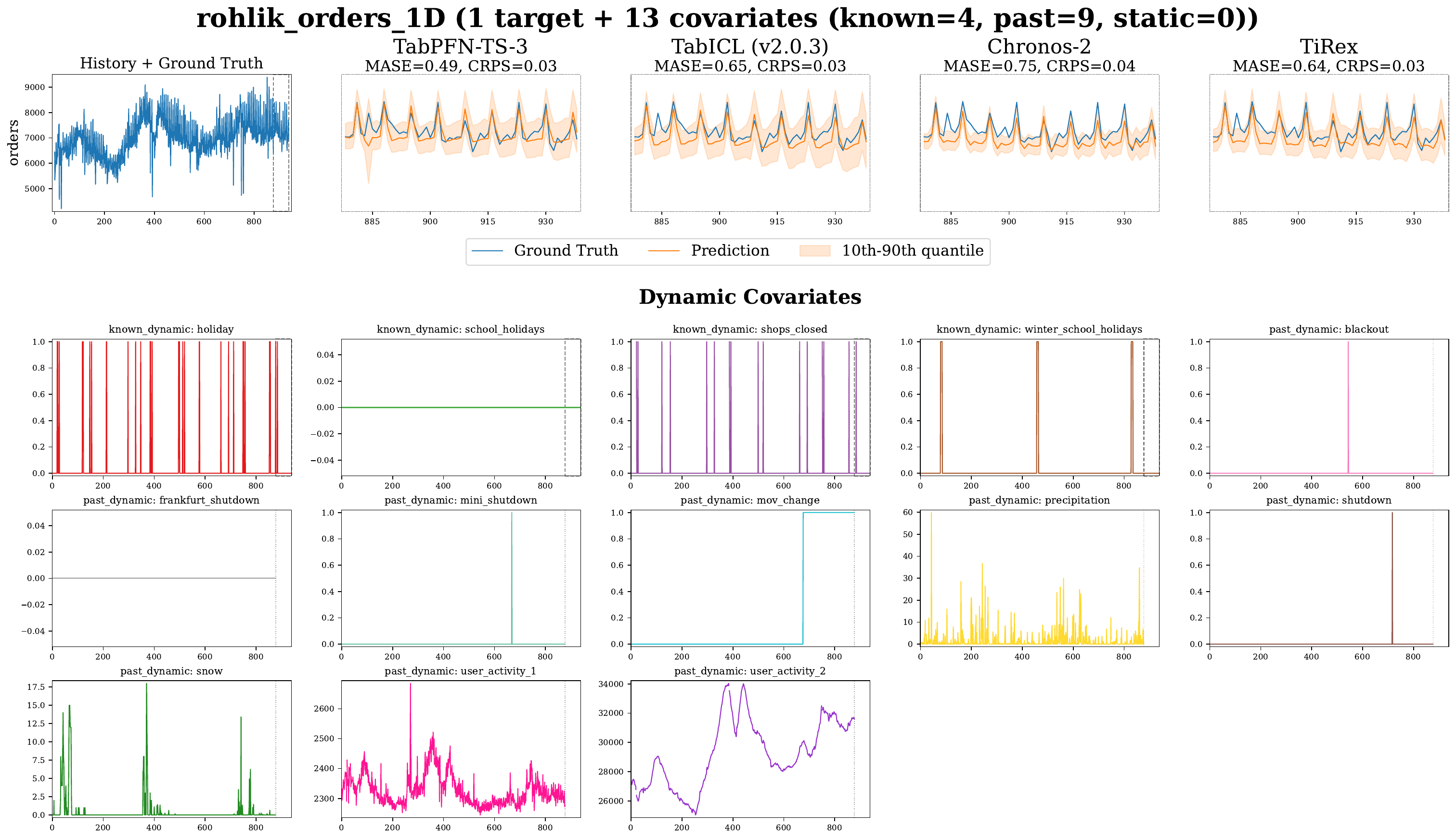}
\caption{\texttt{rohlik\_orders\_1D} — daily online-grocery orders.}
\end{figure}

\begin{figure}[H]
\centering
\includegraphics[width=\textwidth]{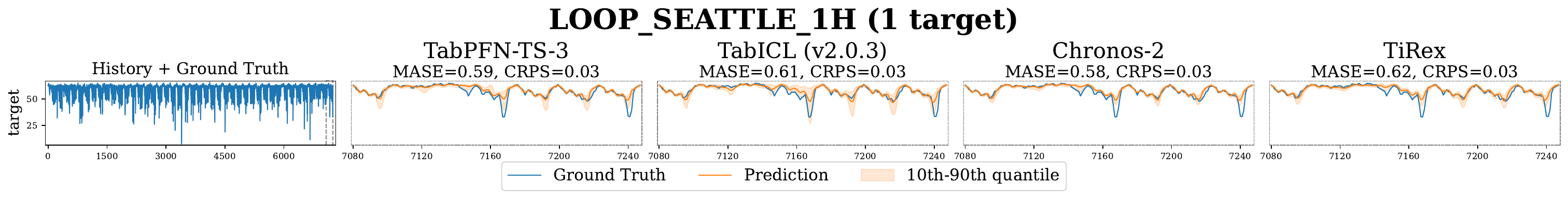}
\caption{\texttt{LOOP\_SEATTLE\_1H} — hourly Seattle freeway loop-detector counts.}
\end{figure}

\begin{figure}[H]
\centering
\includegraphics[width=\textwidth]{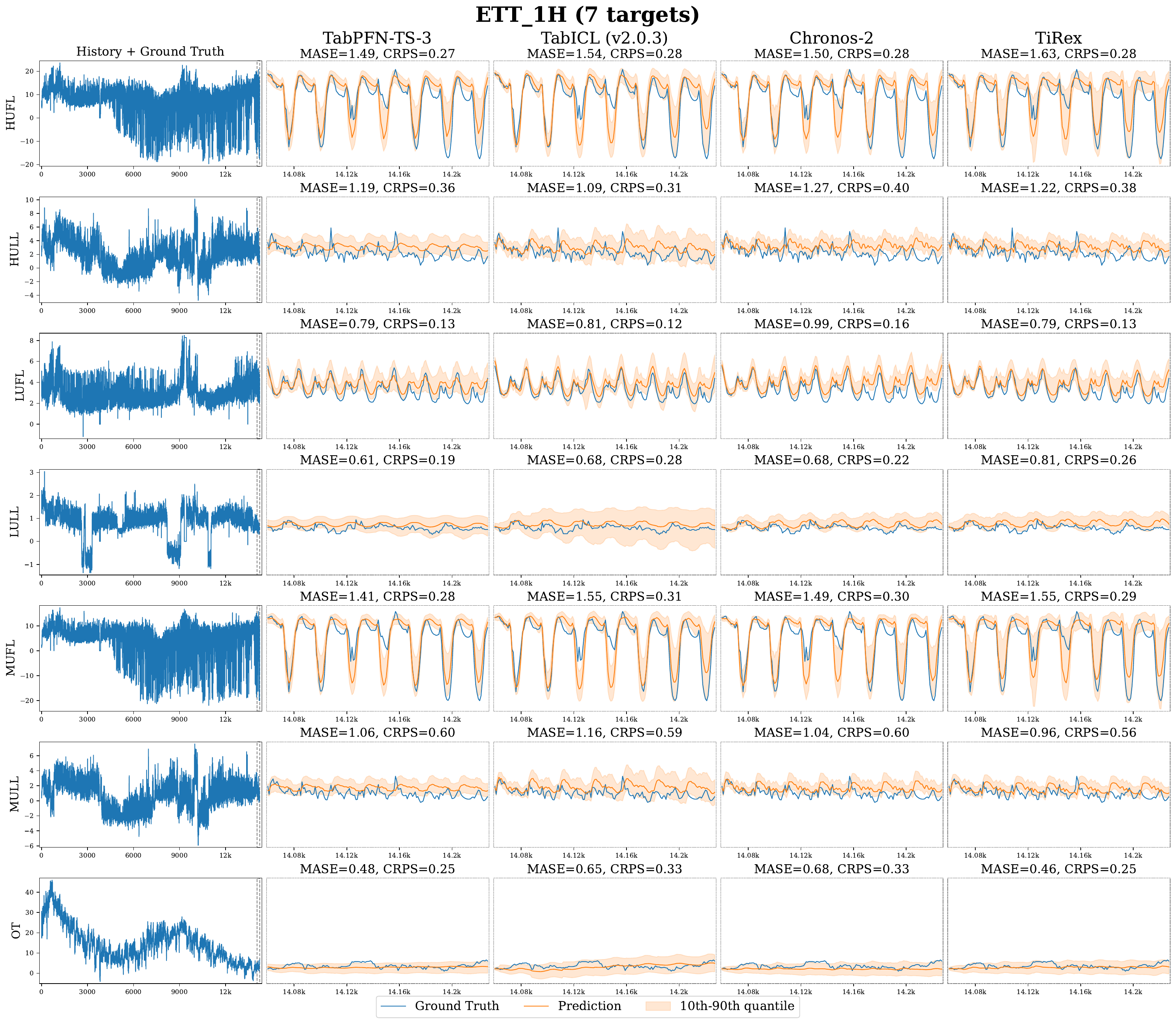}
\caption{\texttt{ETT\_1H} — hourly Electricity Transformer Temperature.}
\end{figure}

\begin{figure}[H]
\centering
\includegraphics[width=\textwidth]{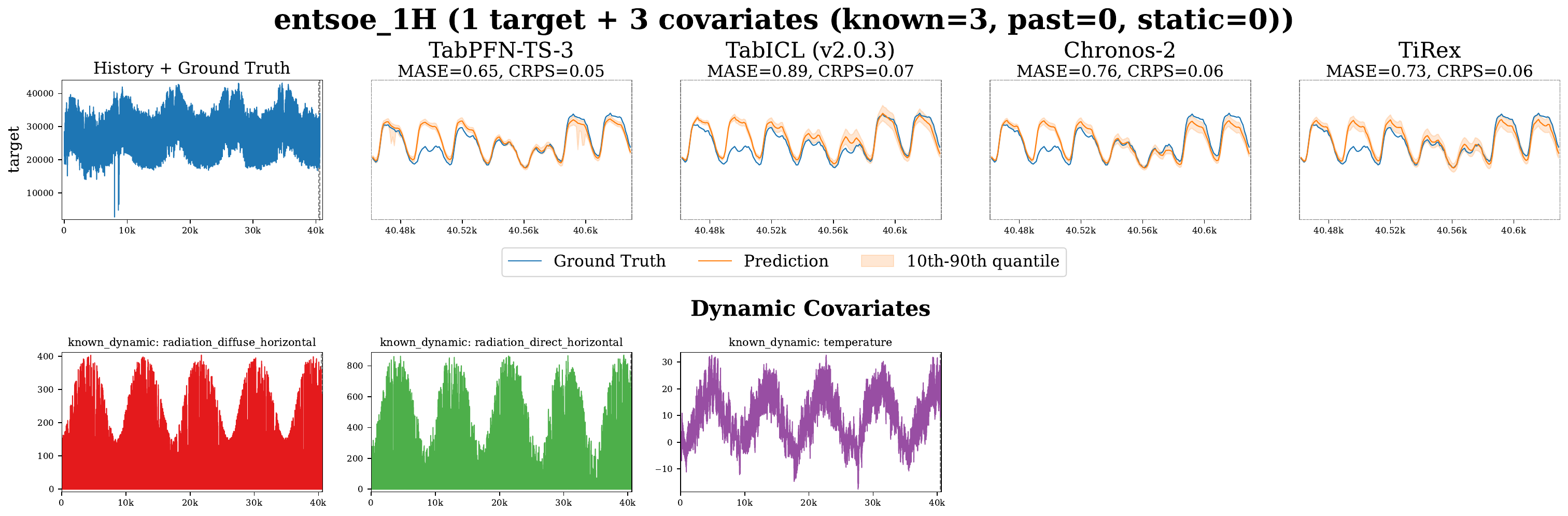}
\caption{\texttt{entsoe\_1H} — hourly ENTSO-E European electricity load.}
\end{figure}

\subsection*{Pairwise skill-score heatmaps}

\subsection*{fev-bench per-task SQL leaderboard}
\label{sec:fev-bench-per-task}

\input{tables/time_series/per_task_SQL}

%% file: tables/time_series/leaderboard_SQL.tex
\begin{tabular}{lrrrrr}
\toprule
\textbf{Model} & \textbf{Win (\%)} & \textbf{Skill (\%)} & \textbf{Runtime (s)} & \textbf{Leak.\ (\%)} & \textbf{\# fails} \\
\midrule
Chronos-2            & 91.7 &  47.3 &   0.8 &  0 &  0 \\
\textcolor{PriorMauve}{TabPFN-TS-3}          & 73.6 &  43.1 & 234.6 &  0 &  0 \\
TiRex                & 83.4 &  42.6 &   0.2 &  1 &  0 \\
TimesFM-2.5          & 78.6 &  42.2 &   1.9 & 10 &  0 \\
Toto-1.0             & 71.6 &  40.7 &  22.1 &  8 &  0 \\
\textcolor{PriorMauve}{TabPFN-v2-TS}         & 64.1 &  39.6 &  88.9 &  0 &  2 \\
Moirai-2.0           & 66.2 &  39.3 &   0.3 & 28 &  0 \\
Chronos-Bolt         & 66.2 &  38.9 &   0.2 &  0 &  0 \\
Sundial-Base         & 47.1 &  33.4 &   8.0 &  1 &  0 \\
TabICL-v2$^{\dagger}$ & 53.8 &  30.8 &  64.7 &  0 &  0 \\
CatBoost (Recursive) & 35.7 &  23.0 &   0.3 &  0 &  0 \\
LightGBM (Recursive) & 33.4 &  21.7 &   0.3 &  0 &  0 \\
AutoARIMA            & 39.6 &  20.6 &  19.5 &  0 & 10 \\
Stat. Ensemble       & 43.8 &  20.2 & 148.6 &  0 & 11 \\
AutoTheta            & 27.1 &   5.5 &   3.3 &  0 &  0 \\
Seasonal Naive       & 19.1 &   0.0 &   0.5 &  0 &  0 \\
AutoETS              & 32.7 & -26.8 &   3.5 &  0 &  3 \\
Naive                & 12.6 & -45.4 &   0.5 &  0 &  0 \\
Drift                &  9.7 & -45.8 &   0.5 &  0 &  0 \\
\bottomrule
\end{tabular}

%% file: tables/time_series/leaderboard_MASE.tex
\begin{tabular}{lrrrrr}
\toprule
\textbf{Model} & \textbf{Win (\%)} & \textbf{Skill (\%)} & \textbf{Runtime (s)} & \textbf{Leak.\ (\%)} & \textbf{\# fails} \\
\midrule
Chronos-2            & 86.9 &  35.5 &   0.8 &  0 &  0 \\
\textcolor{PriorMauve}{TabPFN-TS-3}          & 69.8 &  30.6 & 234.6 &  0 &  0 \\
TimesFM-2.5          & 74.9 &  30.2 &   1.9 & 10 &  0 \\
TiRex                & 76.9 &  30.0 &   0.2 &  1 &  0 \\
Toto-1.0             & 66.3 &  28.2 &  22.1 &  8 &  0 \\
\textcolor{PriorMauve}{TabPFN-v2-TS}         & 58.5 &  27.6 &  88.9 &  0 &  2 \\
Moirai-2.0           & 61.4 &  27.3 &   0.3 & 28 &  0 \\
Chronos-Bolt         & 60.7 &  26.5 &   0.2 &  0 &  0 \\
Sundial-Base         & 53.4 &  24.7 &   8.0 &  1 &  0 \\
CatBoost (Recursive) & 54.0 &  23.7 &   0.3 &  0 &  0 \\
LightGBM (Recursive) & 50.3 &  22.4 &   0.3 &  0 &  0 \\
Stat. Ensemble       & 46.7 &  15.7 & 148.6 &  0 & 11 \\
AutoARIMA            & 36.0 &  11.2 &  19.5 &  0 & 10 \\
AutoTheta            & 34.2 &  11.0 &   3.3 &  0 &  0 \\
TabICL-v2$^{\dagger}$ & 33.2 &   7.0 &  64.7 &  0 &  0 \\
AutoETS              & 33.5 &   2.3 &   3.5 &  0 &  3 \\
Seasonal Naive       & 20.0 &   0.0 &   0.5 &  0 &  0 \\
Naive                & 18.0 & -16.7 &   0.5 &  0 &  0 \\
Drift                & 15.3 & -18.1 &   0.5 &  0 &  0 \\
\bottomrule
\end{tabular}

%% file: tables/time_series/per_task_SQL.tex
\setlength{\tabcolsep}{2pt}
{\scriptsize
\begin{longtable}{@{}p{2.2cm}@{\hspace{6pt}}rrrrrrrrrr@{}}
\caption{\textbf{Per-task SQL on fev-bench (100 tasks).} Lower is better; values
are after leakage and failure imputation. Per-row top-three are
highlighted with gold / silver / bronze backgrounds. Columns are the
ten models with the most medal placements; ordered by overall SQL
skill score. Values exceeding $10^3$ are capped for layout.}
\label{tab:fev-bench-per-task-sql} \\
\toprule
\textbf{Task name} & \rotatebox{60}{\textbf{Chronos-2}} & \rotatebox{60}{\textbf{TabPFN-TS-3}} & \rotatebox{60}{\textbf{TiRex}} & \rotatebox{60}{\textbf{TimesFM-2.5}} & \rotatebox{60}{\textbf{Toto-1.0}} & \rotatebox{60}{\textbf{TabPFN-v2-TS}} & \rotatebox{60}{\textbf{Moirai-2.0}} & \rotatebox{60}{\textbf{Chronos-Bolt}} & \rotatebox{60}{\textbf{StatEns}} & \rotatebox{60}{\textbf{AutoETS}} \\
\midrule
\endfirsthead
\toprule
\textbf{Task name} & \rotatebox{60}{\textbf{Chronos-2}} & \rotatebox{60}{\textbf{TabPFN-TS-3}} & \rotatebox{60}{\textbf{TiRex}} & \rotatebox{60}{\textbf{TimesFM-2.5}} & \rotatebox{60}{\textbf{Toto-1.0}} & \rotatebox{60}{\textbf{TabPFN-v2-TS}} & \rotatebox{60}{\textbf{Moirai-2.0}} & \rotatebox{60}{\textbf{Chronos-Bolt}} & \rotatebox{60}{\textbf{StatEns}} & \rotatebox{60}{\textbf{AutoETS}} \\
\midrule
\endhead
\bottomrule
\endfoot
\ttfamily\scriptsize ETT\_15T & \cellcolor{medalGold}0.546 & 0.626 & \cellcolor{medalSilver}0.568 & 0.577 & 0.593 & 0.602 & \cellcolor{medalBronze}0.574 & 0.574 & 0.762 & 1.263 \\
\ttfamily\scriptsize ETT\_1D & \cellcolor{medalSilver}1.132 & 1.138 & \cellcolor{medalGold}1.101 & 1.144 & 1.143 & 1.230 & \cellcolor{medalBronze}1.132 & 1.132 & 1.271 & 1.356 \\
\ttfamily\scriptsize ETT\_1H & 0.883 & 0.908 & \cellcolor{medalSilver}0.874 & \cellcolor{medalBronze}0.882 & \cellcolor{medalGold}0.873 & 0.933 & 0.944 & 0.944 & 1.272 & 1.765 \\
\ttfamily\scriptsize ETT\_1W & 2.320 & \cellcolor{medalSilver}2.252 & \cellcolor{medalBronze}2.265 & \cellcolor{medalGold}2.249 & 2.281 & 2.411 & 2.280 & 2.280 & 2.407 & 2.394 \\
\ttfamily\scriptsize LOOP\_SEATTLE\_1D & \cellcolor{medalBronze}0.779 & \cellcolor{medalGold}0.769 & 0.792 & \cellcolor{medalSilver}0.774 & 0.831 & 0.780 & 0.805 & 0.805 & 0.820 & 0.825 \\
\ttfamily\scriptsize LOOP\_SEATTLE\_1H & \cellcolor{medalSilver}0.639 & 0.667 & \cellcolor{medalBronze}0.656 & \cellcolor{medalGold}0.621 & 0.698 & 0.679 & 0.765 & 0.765 & 1.501 & 2.639 \\
\ttfamily\scriptsize LOOP\_SEATTLE\_5T & \cellcolor{medalGold}0.533 & 0.710 & \cellcolor{medalSilver}0.549 & 0.595 & \cellcolor{medalBronze}0.561 & 0.641 & 0.710 & 0.710 & 1.044 & 1.155 \\
\ttfamily\scriptsize M\_DENSE\_1D & \cellcolor{medalGold}0.646 & 0.757 & \cellcolor{medalBronze}0.746 & \cellcolor{medalSilver}0.708 & 0.842 & 0.756 & 0.759 & 0.759 & 0.965 & 1.073 \\
\ttfamily\scriptsize M\_DENSE\_1H & \cellcolor{medalBronze}0.585 & \cellcolor{medalSilver}0.585 & 0.587 & \cellcolor{medalGold}0.556 & 0.621 & 0.646 & 0.595 & 0.595 & 1.127 & 59.020 \\
\ttfamily\scriptsize SZ\_TAXI\_15T & \cellcolor{medalGold}0.393 & 0.399 & \cellcolor{medalSilver}0.396 & \cellcolor{medalBronze}0.397 & 0.401 & 0.429 & 0.413 & 0.413 & 0.560 & 2.355 \\
\ttfamily\scriptsize SZ\_TAXI\_1H & \cellcolor{medalGold}0.398 & \cellcolor{medalBronze}0.407 & \cellcolor{medalSilver}0.405 & 0.416 & 0.418 & 0.494 & 0.426 & 0.426 & 0.689 & $>\!10^{3}$ \\
\ttfamily\scriptsize aust...tourism & \cellcolor{medalGold}0.677 & \cellcolor{medalSilver}0.695 & 0.786 & 0.732 & 0.890 & \cellcolor{medalBronze}0.699 & 0.918 & 0.928 & 0.730 & 0.762 \\
\ttfamily\scriptsize bizitobs\_l2c\_1H & \cellcolor{medalGold}0.301 & 0.374 & 0.366 & \cellcolor{medalSilver}0.326 & 0.370 & 0.354 & \cellcolor{medalBronze}0.342 & 0.342 & 0.634 & 0.718 \\
\ttfamily\scriptsize bizitobs\_l2c\_5T & \cellcolor{medalSilver}0.411 & \cellcolor{medalGold}0.370 & 0.679 & \cellcolor{medalBronze}0.461 & 0.595 & 0.485 & 0.757 & 0.757 & 0.720 & 0.731 \\
\ttfamily\scriptsize boomlet\_1062 & \cellcolor{medalSilver}0.552 & \cellcolor{medalBronze}0.554 & 0.555 & 0.573 & \cellcolor{medalGold}0.548 & 0.708 & 0.593 & 0.639 & 0.985 & 1.309 \\
\ttfamily\scriptsize boomlet\_1209 & \cellcolor{medalSilver}0.680 & 0.768 & 0.729 & \cellcolor{medalBronze}0.705 & \cellcolor{medalGold}0.645 & 1.016 & 0.756 & 0.784 & 2.469 & 1.264 \\
\ttfamily\scriptsize boomlet\_1225 & \cellcolor{medalSilver}0.186 & 0.199 & \cellcolor{medalBronze}0.188 & 0.190 & \cellcolor{medalGold}0.183 & 0.215 & 0.195 & 0.203 & 0.280 & 0.318 \\
\ttfamily\scriptsize boomlet\_1230 & 1.201 & 1.292 & \cellcolor{medalSilver}1.186 & \cellcolor{medalBronze}1.187 & \cellcolor{medalGold}1.138 & 1.613 & 1.286 & 1.266 & 3.390 & $>\!10^{3}$ \\
\ttfamily\scriptsize boomlet\_1282 & 0.421 & 0.413 & \cellcolor{medalBronze}0.409 & \cellcolor{medalGold}0.403 & \cellcolor{medalSilver}0.407 & 0.425 & 0.427 & 0.462 & 0.739 & 0.914 \\
\ttfamily\scriptsize boomlet\_1487 & \cellcolor{medalBronze}0.423 & 0.447 & 0.427 & \cellcolor{medalSilver}0.412 & \cellcolor{medalGold}0.400 & 0.745 & 0.456 & 0.482 & 0.681 & 0.724 \\
\ttfamily\scriptsize boomlet\_1631 & \cellcolor{medalGold}0.572 & 0.622 & 0.598 & \cellcolor{medalSilver}0.579 & \cellcolor{medalBronze}0.581 & 0.697 & 0.591 & 0.619 & 0.851 & 0.721 \\
\ttfamily\scriptsize boomlet\_1676 & \cellcolor{medalBronze}0.569 & 0.602 & 0.571 & \cellcolor{medalSilver}0.563 & \cellcolor{medalGold}0.554 & 0.831 & 0.573 & 0.608 & 0.850 & 0.756 \\
\ttfamily\scriptsize boomlet\_1855 & \cellcolor{medalBronze}0.462 & 0.504 & \cellcolor{medalGold}0.450 & 0.473 & \cellcolor{medalSilver}0.452 & 0.623 & 0.465 & 0.470 & 1.123 & 1.185 \\
\ttfamily\scriptsize boomlet\_1975 & \cellcolor{medalSilver}0.133 & 0.251 & 0.192 & \cellcolor{medalBronze}0.167 & \cellcolor{medalGold}0.126 & 0.207 & 0.220 & 0.179 & 0.548 & 0.611 \\
\ttfamily\scriptsize boomlet\_2187 & \cellcolor{medalSilver}0.712 & 0.835 & \cellcolor{medalGold}0.711 & 0.802 & \cellcolor{medalBronze}0.764 & 0.934 & 0.807 & 0.775 & 1.273 & 1.307 \\
\ttfamily\scriptsize boomlet\_285 & \cellcolor{medalGold}0.290 & 0.354 & \cellcolor{medalBronze}0.345 & 0.397 & \cellcolor{medalSilver}0.319 & 0.713 & 0.427 & 0.477 & 1.262 & 1.203 \\
\ttfamily\scriptsize boomlet\_619 & \cellcolor{medalSilver}0.323 & \cellcolor{medalBronze}0.326 & 0.341 & 0.340 & \cellcolor{medalGold}0.310 & 0.331 & 0.329 & 0.471 & 0.777 & 0.894 \\
\ttfamily\scriptsize boomlet\_772 & \cellcolor{medalSilver}0.283 & 0.305 & 0.296 & \cellcolor{medalBronze}0.295 & \cellcolor{medalGold}0.281 & 0.330 & 0.314 & 0.339 & 1.179 & $>\!10^{3}$ \\
\ttfamily\scriptsize boomlet\_963 & \cellcolor{medalGold}0.717 & 0.786 & \cellcolor{medalSilver}0.718 & 0.739 & \cellcolor{medalBronze}0.720 & 0.796 & 0.751 & 0.779 & 1.335 & 1.609 \\
\ttfamily\scriptsize ecdc\_ili & \cellcolor{medalSilver}2.271 & 2.457 & 2.411 & \cellcolor{medalGold}2.215 & 2.554 & \cellcolor{medalBronze}2.382 & 2.454 & 2.653 & 3.837 & 4.079 \\
\ttfamily\scriptsize entsoe\_15T & \cellcolor{medalGold}0.454 & 0.648 & \cellcolor{medalSilver}0.469 & \cellcolor{medalBronze}0.471 & 0.591 & 0.484 & 0.478 & 0.506 & 0.781 & 3.029 \\
\ttfamily\scriptsize entsoe\_1H & \cellcolor{medalSilver}0.429 & \cellcolor{medalGold}0.385 & 0.470 & 0.468 & 0.480 & \cellcolor{medalBronze}0.442 & 0.487 & 0.457 & 0.892 & 1.905 \\
\ttfamily\scriptsize entsoe\_30T & \cellcolor{medalGold}0.434 & 0.579 & 0.523 & 0.566 & \cellcolor{medalBronze}0.496 & 0.512 & \cellcolor{medalSilver}0.488 & 0.529 & 0.847 & 2.493 \\
\ttfamily\scriptsize epf\_be & \cellcolor{medalSilver}0.503 & 0.533 & \cellcolor{medalBronze}0.527 & \cellcolor{medalGold}0.494 & 0.565 & 0.532 & 0.528 & 0.573 & 1.213 & 1.534 \\
\ttfamily\scriptsize epf\_de & \cellcolor{medalBronze}0.491 & \cellcolor{medalGold}0.437 & 1.032 & 1.030 & 1.106 & \cellcolor{medalSilver}0.440 & 1.016 & 1.021 & 1.167 & 1.401 \\
\ttfamily\scriptsize epf\_fr & \cellcolor{medalSilver}0.362 & \cellcolor{medalBronze}0.374 & 0.401 & 0.409 & 0.426 & \cellcolor{medalGold}0.331 & 0.409 & 0.439 & 1.146 & 0.899 \\
\ttfamily\scriptsize epf\_np & \cellcolor{medalSilver}0.658 & \cellcolor{medalGold}0.633 & 0.966 & 1.171 & 1.037 & \cellcolor{medalBronze}0.659 & 0.925 & 0.971 & 1.284 & 1.933 \\
\ttfamily\scriptsize epf\_pjm & \cellcolor{medalGold}0.382 & \cellcolor{medalSilver}0.382 & \cellcolor{medalBronze}0.404 & 0.426 & 0.452 & 0.427 & 0.441 & 0.422 & 0.487 & 0.914 \\
\ttfamily\scriptsize ercot\_1D & 0.869 & \cellcolor{medalBronze}0.845 & \cellcolor{medalGold}0.818 & \cellcolor{medalSilver}0.830 & 0.880 & 0.981 & 0.947 & 0.916 & 1.255 & 1.382 \\
\ttfamily\scriptsize ercot\_1H & \cellcolor{medalGold}1.029 & 1.108 & \cellcolor{medalSilver}1.065 & 1.151 & \cellcolor{medalBronze}1.095 & 1.208 & 1.098 & 1.138 & 1.260 & 2.676 \\
\ttfamily\scriptsize ercot\_1M & \cellcolor{medalGold}0.755 & \cellcolor{medalSilver}0.755 & 0.806 & 0.772 & 1.007 & 0.903 & 0.973 & 0.773 & 0.762 & \cellcolor{medalBronze}0.756 \\
\ttfamily\scriptsize ercot\_1W & 0.966 & 0.996 & \cellcolor{medalSilver}0.955 & \cellcolor{medalGold}0.932 & 1.060 & 1.228 & 1.053 & \cellcolor{medalBronze}0.961 & 2.095 & 2.068 \\
\ttfamily\scriptsize fav...stores\_1D & \cellcolor{medalGold}0.916 & 0.989 & \cellcolor{medalBronze}0.968 & \cellcolor{medalSilver}0.949 & 1.036 & 0.970 & 0.980 & 1.032 & 1.197 & 1.238 \\
\ttfamily\scriptsize fav...stores\_1M & \cellcolor{medalGold}1.794 & \cellcolor{medalBronze}1.923 & \cellcolor{medalSilver}1.856 & 1.998 & 2.009 & 1.934 & 2.091 & 2.087 & 1.943 & 1.942 \\
\ttfamily\scriptsize fav...stores\_1W & \cellcolor{medalSilver}2.024 & 2.054 & \cellcolor{medalBronze}2.046 & \cellcolor{medalGold}1.968 & 2.128 & 2.123 & 2.197 & 2.101 & 2.220 & 2.357 \\
\ttfamily\scriptsize fav...trans\_1D & \cellcolor{medalGold}0.685 & 1.283 & 1.031 & \cellcolor{medalSilver}0.975 & \cellcolor{medalBronze}0.975 & 1.225 & 0.975 & 0.975 & 1.185 & 1.181 \\
\ttfamily\scriptsize fav...trans\_1M & \cellcolor{medalGold}0.943 & 1.214 & \cellcolor{medalSilver}1.089 & \cellcolor{medalBronze}1.133 & 1.397 & 1.244 & 1.390 & 1.358 & 1.152 & 1.179 \\
\ttfamily\scriptsize fav...trans\_1W & \cellcolor{medalGold}1.228 & 1.579 & \cellcolor{medalSilver}1.384 & \cellcolor{medalBronze}1.428 & 1.557 & 1.912 & 1.463 & 1.428 & 1.559 & 1.647 \\
\ttfamily\scriptsize fred\_md\_2025/cee & \cellcolor{medalSilver}3.468 & 4.823 & \cellcolor{medalGold}3.349 & 4.490 & 4.490 & 3.873 & 4.490 & 4.490 & 3.745 & \cellcolor{medalBronze}3.643 \\
\ttfamily\scriptsize fred\_md/macro & \cellcolor{medalSilver}5.680 & 6.623 & \cellcolor{medalGold}5.307 & 5.842 & 5.842 & 6.399 & 5.842 & 5.842 & \cellcolor{medalBronze}5.743 & 5.794 \\
\ttfamily\scriptsize fred\_qd\_2025/cee & 2.192 & 2.455 & \cellcolor{medalBronze}2.046 & 2.181 & \cellcolor{medalGold}1.773 & 2.292 & 2.296 & 2.365 & \cellcolor{medalSilver}1.903 & 2.123 \\
\ttfamily\scriptsize fred\_qd/macro & \cellcolor{medalBronze}3.537 & 4.040 & \cellcolor{medalSilver}3.530 & 3.593 & \cellcolor{medalGold}3.402 & 4.240 & 3.616 & 3.654 & 3.615 & 3.904 \\
\ttfamily\scriptsize gvar & \cellcolor{medalBronze}0.578 & 0.594 & \cellcolor{medalSilver}0.577 & 0.590 & \cellcolor{medalGold}0.576 & 0.674 & 0.593 & 0.596 & 0.590 & 0.593 \\
\ttfamily\scriptsize hermes & \cellcolor{medalGold}0.609 & \cellcolor{medalBronze}0.619 & 0.651 & \cellcolor{medalSilver}0.618 & 0.985 & 0.705 & 0.704 & 0.675 & 1.416 & 1.673 \\
\ttfamily\scriptsize hier...sales\_1D & 0.557 & 0.552 & \cellcolor{medalSilver}0.547 & 0.552 & \cellcolor{medalGold}0.547 & 0.572 & \cellcolor{medalBronze}0.551 & 0.551 & 0.720 & 0.793 \\
\ttfamily\scriptsize hier...sales\_1W & \cellcolor{medalGold}0.616 & 0.625 & \cellcolor{medalBronze}0.621 & \cellcolor{medalSilver}0.618 & 0.637 & 0.637 & 0.637 & 0.637 & 0.746 & 10.477 \\
\ttfamily\scriptsize hospital & \cellcolor{medalBronze}0.686 & \cellcolor{medalGold}0.673 & 0.688 & \cellcolor{medalSilver}0.680 & 0.733 & 0.696 & 0.697 & 0.697 & 0.697 & 0.726 \\
\ttfamily\scriptsize hosp...sions\_1D & \cellcolor{medalSilver}0.554 & \cellcolor{medalGold}0.554 & \cellcolor{medalBronze}0.555 & 0.556 & 0.555 & 0.562 & 0.556 & 0.556 & 0.557 & 0.556 \\
\ttfamily\scriptsize hosp...sions\_1W & \cellcolor{medalGold}0.576 & 0.581 & 0.585 & 0.580 & 0.598 & 0.581 & 0.586 & 0.587 & \cellcolor{medalBronze}0.579 & \cellcolor{medalSilver}0.578 \\
\ttfamily\scriptsize jena\_weather\_10T & \cellcolor{medalGold}0.354 & 0.398 & 0.389 & \cellcolor{medalSilver}0.357 & \cellcolor{medalBronze}0.368 & 0.413 & 0.418 & 0.418 & 0.673 & 0.742 \\
\ttfamily\scriptsize jena\_weather\_1D & 1.111 & 1.143 & \cellcolor{medalGold}1.072 & 1.090 & 1.112 & 1.155 & \cellcolor{medalSilver}1.075 & \cellcolor{medalBronze}1.075 & 1.339 & 1.664 \\
\ttfamily\scriptsize jena\_weather\_1H & \cellcolor{medalGold}0.353 & 0.429 & \cellcolor{medalSilver}0.356 & \cellcolor{medalBronze}0.359 & 0.362 & 0.413 & 0.367 & 0.367 & 0.452 & 0.553 \\
\ttfamily\scriptsize kdd\_cup\_2022\_10T & \cellcolor{medalGold}0.425 & \cellcolor{medalSilver}0.456 & \cellcolor{medalBronze}0.533 & 0.533 & 0.533 & 0.555 & 0.533 & 0.533 & 0.777 & 0.747 \\
\ttfamily\scriptsize kdd\_cup\_2022\_1D & \cellcolor{medalBronze}0.704 & 0.709 & \cellcolor{medalGold}0.697 & \cellcolor{medalSilver}0.698 & 0.704 & 0.715 & 0.708 & 0.709 & 0.730 & 0.751 \\
\ttfamily\scriptsize kdd\_cup\_2022\_30T & 0.439 & 0.459 & \cellcolor{medalBronze}0.432 & 0.505 & \cellcolor{medalSilver}0.429 & 0.543 & \cellcolor{medalGold}0.427 & 0.561 & 0.679 & 0.772 \\
\ttfamily\scriptsize m5\_1D & \cellcolor{medalBronze}0.722 & \cellcolor{medalSilver}0.720 & \cellcolor{medalGold}0.714 & 0.729 & 0.729 & 1.254 & 0.729 & 0.729 & 1.254 & 0.853 \\
\ttfamily\scriptsize m5\_1M & \cellcolor{medalSilver}0.977 & 0.986 & \cellcolor{medalGold}0.974 & \cellcolor{medalBronze}0.980 & 1.044 & 1.002 & 0.996 & 1.000 & 1.022 & 1.108 \\
\ttfamily\scriptsize m5\_1W & \cellcolor{medalGold}0.900 & \cellcolor{medalBronze}0.904 & \cellcolor{medalSilver}0.903 & 0.917 & 0.905 & 0.928 & 0.907 & 0.917 & 0.936 & 0.953 \\
\ttfamily\scriptsize proenfo\_gfc12 & \cellcolor{medalSilver}0.649 & \cellcolor{medalGold}0.614 & 0.908 & 0.917 & 0.917 & \cellcolor{medalBronze}0.834 & 0.917 & 0.917 & 1.305 & 2.431 \\
\ttfamily\scriptsize proenfo\_gfc14 & \cellcolor{medalSilver}0.430 & \cellcolor{medalGold}0.426 & 0.721 & 0.767 & 0.767 & \cellcolor{medalBronze}0.515 & 0.767 & 0.767 & 0.906 & 1.110 \\
\ttfamily\scriptsize proenfo\_gfc17 & \cellcolor{medalGold}0.485 & \cellcolor{medalSilver}0.528 & 0.889 & 0.900 & 0.900 & \cellcolor{medalBronze}0.672 & 0.900 & 0.900 & 1.142 & 2.135 \\
\ttfamily\scriptsize redset\_15T & \cellcolor{medalSilver}0.790 & 1.208 & 0.833 & \cellcolor{medalGold}0.741 & \cellcolor{medalBronze}0.818 & 1.250 & 1.041 & 1.243 & 1.231 & 1.231 \\
\ttfamily\scriptsize redset\_1H & 1.365 & 1.338 & \cellcolor{medalBronze}1.337 & 1.367 & \cellcolor{medalGold}1.306 & \cellcolor{medalSilver}1.321 & 1.410 & 2.279 & 1.859 & 2.377 \\
\ttfamily\scriptsize redset\_5T & \cellcolor{medalGold}0.654 & 0.749 & 0.787 & 0.723 & \cellcolor{medalBronze}0.719 & \cellcolor{medalSilver}0.711 & 0.793 & 1.026 & 2.690 & 1.224 \\
\ttfamily\scriptsize restaurant & \cellcolor{medalBronze}0.685 & 0.686 & \cellcolor{medalSilver}0.682 & \cellcolor{medalGold}0.677 & 0.704 & 0.693 & 0.689 & 0.689 & 0.709 & 1.021 \\
\ttfamily\scriptsize rohlik\_orders\_1D & \cellcolor{medalGold}0.959 & 1.052 & \cellcolor{medalBronze}0.986 & 1.006 & 1.135 & 1.341 & \cellcolor{medalSilver}0.970 & 1.051 & 1.211 & 1.447 \\
\ttfamily\scriptsize rohlik\_orders\_1W & \cellcolor{medalGold}1.300 & 1.415 & \cellcolor{medalSilver}1.300 & \cellcolor{medalBronze}1.328 & 1.493 & 1.524 & 1.532 & 1.428 & 1.398 & 1.419 \\
\ttfamily\scriptsize rohlik\_sales\_1D & \cellcolor{medalGold}0.881 & \cellcolor{medalSilver}0.899 & 1.148 & \cellcolor{medalBronze}1.096 & 1.218 & 1.375 & 1.170 & 1.147 & 1.248 & 1.266 \\
\ttfamily\scriptsize rohlik\_sales\_1W & \cellcolor{medalBronze}1.274 & \cellcolor{medalGold}1.159 & 1.425 & 1.401 & 1.505 & \cellcolor{medalSilver}1.221 & 1.516 & 1.522 & 1.646 & 14.453 \\
\ttfamily\scriptsize rossmann\_1D & \cellcolor{medalBronze}0.283 & \cellcolor{medalSilver}0.245 & 0.539 & 0.502 & 0.568 & \cellcolor{medalGold}0.232 & 0.527 & 0.525 & 0.578 & 0.594 \\
\ttfamily\scriptsize rossmann\_1W & \cellcolor{medalBronze}0.308 & \cellcolor{medalSilver}0.256 & 0.482 & 0.495 & 0.494 & \cellcolor{medalGold}0.254 & 0.497 & 0.487 & 0.501 & 0.518 \\
\ttfamily\scriptsize solar\_1D & \cellcolor{medalGold}0.594 & \cellcolor{medalSilver}0.601 & \cellcolor{medalBronze}0.614 & 0.618 & 0.622 & 0.615 & 0.637 & 0.635 & 0.653 & 0.656 \\
\ttfamily\scriptsize solar\_1W & \cellcolor{medalSilver}0.895 & \cellcolor{medalBronze}0.924 & 1.121 & 1.096 & 1.392 & \cellcolor{medalGold}0.870 & 1.658 & 0.940 & 1.296 & 1.212 \\
\ttfamily\scriptsize s...weather\_15T & \cellcolor{medalSilver}0.677 & \cellcolor{medalGold}0.671 & 0.846 & 0.906 & 0.784 & \cellcolor{medalBronze}0.747 & 0.839 & 0.809 & 1.194 & 2.529 \\
\ttfamily\scriptsize s...weather\_1H & \cellcolor{medalBronze}0.767 & \cellcolor{medalGold}0.660 & 0.900 & 0.815 & 0.876 & \cellcolor{medalSilver}0.701 & 0.907 & 0.816 & 1.458 & 2.182 \\
\ttfamily\scriptsize uci...ality\_1D & \cellcolor{medalGold}1.046 & 1.147 & 1.128 & 1.205 & 1.260 & 1.186 & 1.138 & \cellcolor{medalSilver}1.092 & \cellcolor{medalBronze}1.123 & 1.181 \\
\ttfamily\scriptsize uci...ality\_1H & \cellcolor{medalGold}0.798 & 0.934 & \cellcolor{medalSilver}0.865 & 0.877 & \cellcolor{medalBronze}0.870 & 0.931 & 0.945 & 0.899 & 1.561 & $>\!10^{3}$ \\
\ttfamily\scriptsize uk\_nat\_1D/cum & 7.826 & 10.394 & 7.653 & \cellcolor{medalBronze}7.051 & \cellcolor{medalGold}6.188 & 13.045 & \cellcolor{medalSilver}6.763 & 8.157 & 7.712 & 7.184 \\
\ttfamily\scriptsize uk\_nat\_1D/new & \cellcolor{medalSilver}2.037 & 2.071 & \cellcolor{medalGold}1.992 & 2.135 & \cellcolor{medalBronze}2.039 & 2.076 & 2.135 & 2.122 & 2.799 & 2.741 \\
\ttfamily\scriptsize uk\_nat\_1W/cum & \cellcolor{medalBronze}2.783 & 3.478 & 3.192 & 4.011 & 2.824 & 2.872 & 3.014 & 3.435 & \cellcolor{medalGold}2.238 & \cellcolor{medalSilver}2.399 \\
\ttfamily\scriptsize uk\_nat\_1W/new & 4.968 & 4.784 & 4.532 & \cellcolor{medalGold}3.783 & 5.098 & \cellcolor{medalBronze}4.143 & \cellcolor{medalSilver}3.873 & 4.148 & 5.741 & 5.024 \\
\ttfamily\scriptsize uk\_utla\_1D/new & 3.725 & 3.815 & 3.729 & \cellcolor{medalGold}3.512 & 4.036 & 3.801 & \cellcolor{medalBronze}3.565 & \cellcolor{medalSilver}3.531 & 5.582 & 5.623 \\
\ttfamily\scriptsize uk\_utla\_1W/cum & 17.442 & 18.932 & 19.435 & 18.486 & \cellcolor{medalSilver}16.286 & 16.912 & 19.325 & 17.489 & \cellcolor{medalGold}14.331 & \cellcolor{medalBronze}16.313 \\
\ttfamily\scriptsize us\_cons\_1M & \cellcolor{medalSilver}1.464 & 1.698 & \cellcolor{medalBronze}1.467 & 1.605 & 1.564 & 1.571 & 1.513 & 1.516 & 1.486 & \cellcolor{medalGold}1.445 \\
\ttfamily\scriptsize us\_cons\_1Q & \cellcolor{medalSilver}1.724 & 2.302 & 1.803 & 1.927 & \cellcolor{medalGold}1.707 & 2.673 & 1.796 & \cellcolor{medalBronze}1.764 & 1.908 & 1.886 \\
\ttfamily\scriptsize us\_cons\_1Y & \cellcolor{medalSilver}3.730 & 4.807 & \cellcolor{medalGold}3.634 & 4.007 & 3.898 & 4.180 & 4.807 & 4.108 & \cellcolor{medalBronze}3.786 & 4.081 \\
\ttfamily\scriptsize walmart & \cellcolor{medalGold}0.648 & 0.696 & 0.707 & \cellcolor{medalBronze}0.679 & 0.907 & \cellcolor{medalSilver}0.662 & 0.845 & 0.774 & 1.217 & $>\!10^{3}$ \\
\ttfamily\scriptsize world\_co2\_emis & \cellcolor{medalSilver}2.670 & 2.761 & \cellcolor{medalGold}2.643 & 2.876 & 2.716 & 2.720 & 2.875 & 2.754 & \cellcolor{medalBronze}2.688 & 7.724 \\
\ttfamily\scriptsize world\_life\_exp & \cellcolor{medalBronze}1.187 & 1.190 & \cellcolor{medalGold}1.109 & 1.210 & 1.639 & \cellcolor{medalSilver}1.149 & 1.785 & 1.345 & 1.305 & 1.302 \\
\ttfamily\scriptsize world\_tourism & 3.052 & 3.149 & 3.052 & 3.562 & 3.208 & \cellcolor{medalSilver}2.795 & 3.264 & 3.164 & \cellcolor{medalGold}2.552 & \cellcolor{medalBronze}2.882 \\
\end{longtable}
}

%% file: sections/I_use_cases.tex
\section{TabPFN Use Case Overview}\label{app:use_cases}

Previous TabPFN models have been applied to a broad set of use cases. We now list 201 published use cases across different industries.

\section*{Highlights}

We highlight a selection of representative use cases that demonstrate TabPFN's strengths across domains:
\begin{enumerate}

\item TabPFN enabled non-invasive early detection of pancreatic cancer by integrating NMR metabolomics with clinical and protein biomarkers \cite{wu2026panmetai}.  \href{https://www.nature.com/articles/s41467-026-69426-9}{Link}

\item TabPFN provided highly accurate predictions of donor mobilization success using baseline and post-mobilization variables, facilitating early triage and improved transplantation outcomes \cite{adil2026deep}. \href{https://doi.org/10.1016/j.jtct.2026.02.016}{Link}

\item TabPFN was used for effective differentiation between psychotic and non-psychotic major depression, improving classification accuracy and supporting psychiatric diagnosis \cite{zheng2026differentiation}. \href{https://doi.org/10.1016/j.jad.2026.121454}{Link}

\item TabPFN served as a high-fidelity surrogate model for optimizing geopolymer concrete mix design, achieving superior accuracy, generalization, and low-uncertainty predictions compared to other ML approaches \cite{sichani2025machine}. \href{https://www.nature.com/articles/s41598-025-29088-x}{Link}

\item TabPFN enabled robust prediction of silica nanoparticle cytotoxicity \cite{zhang2025boosting}. \href{https://www.nature.com/articles/s41598-025-33872-0}{Link}

\item TabPFN demonstrated superior performance and translational feasibility for liver fibrosis staging \cite{CHEN2026102726}. \href{https://www.cell.com/cell-reports-medicine/fulltext/S2666-3791(26)00143-6}{Link}

\item TabPFN enables accurate prediction of reaction kinetics, facilitating mechanistic understanding in biochar-catalyzed antibiotic degradation processes \cite{latif2026deep}. \href{https://link.springer.com/article/10.1007/s42773-026-00606-y}{Link}

\item TabPFN serves as the top-performing regression model to estimate degradation kinetics from multi-source experimental data \cite{latif2026deep}. \href{https://doi.org/10.1007/s42773-026-00606-y}{Link}

\item TabPFN was employed as a core modeling component for learning from multimodal tabular data under strict temporal constraints, enabling strong discriminative performance, improved probability calibration, and effective causal forecasting in early rug-pull detection \cite{shoaei2026lroo}. \href{http://arxiv.org/abs/2603.11324v1}{Link}

\item TabPFN was fine-tuned into a domain-specific model (FinPFN) for regime-aware stock return prediction, improving performance in non-stationary financial markets by adapting to evolving feature--return relationships \cite{wang2025metalearning}. \href{https://www.sciencedirect.com/science/article/abs/pii/S1386418125000825}{Link}

\item TabPFN enabled early fault classification in rotating machinery, addressing data scarcity in industrial scenarios \cite{manuf_usecase1_rotating_faults_tabpfn}. \href{https://ieeexplore.ieee.org/abstract/document/10318062}{Link}

\end{enumerate}

\input{use-case-list}

%% file: use-case-list.tex
\section*{Healthcare and Life Sciences}

We collected 98 published TabPFN use cases in this area. Applications span diagnosis, prognosis, treatment response prediction, and biomarker-based modeling under frequent data scarcity.

\begin{enumerate}

\item TabPFN enabled non-invasive early detection of pancreatic cancer by integrating NMR metabolomics with clinical and protein biomarkers. \cite{wu2026panmetai}. \href{https://www.nature.com/articles/s41467-026-69426-9}{Link}

\item TabPFN enables highly accurate and cost-efficient molecular property prediction by pairing in-context learning with frozen molecular embeddings and descriptor \cite{hicham2026tabularfoundationmodelsincontext}. 
\href{https://arxiv.org/abs/2604.16123}{Link}

\item TabPFN enabled robust prediction of silica nanoparticle cytotoxicity \cite{zhang2025boosting}. \href{https://www.nature.com/articles/s41598-025-33872-0}{Link}

\item TabPFN was combined with BulkFormer to improve prediction accuracy of post-transplant kidney function for better assessment of organ viability during machine perfusion or cold storage \cite{tingle2026combining}. \href{https://doi.org/10.21203/rs.3.rs-9242336/v1}{Link}

\item TabPFN enhances survival analysis, leading to superior performance compared to specialized methods \cite{seletkov2026survivalincontextpriorfittedincontext}. \href{https://arxiv.org/pdf/2603.29475}{Link}

\item TabPFN demonstrated superior performance and translational feasibility for liver fibrosis staging \cite{CHEN2026102726}. \href{https://www.cell.com/cell-reports-medicine/fulltext/S2666-3791(26)00143-6}{Link}

\item TabPFN was leveraged in cardiovascular disease diagnosis \cite{hasan2026advancing}. \href{https://doi.org/10.1038/s41598-026-35451-3}{Link}

\item TabPFN enabled accurate prediction of ALM from multimodal clinical data and improved sarcopenia screening by maintaining robust performance despite missing modalities \cite{kita2026transformerbased}. \href{https://doi.org/10.1186/s12967-026-08079-0}{Link}

\item TabPFN was employed in the winning solution for predicting walking function \cite{villines2026asia}. \href{https://doi.org/10.46292/sci25-00137}{Link}

\item TabPFN demonstrated high accuracy and specificity in matching cell line transcriptomes to reference kidney cell types using curated kidney marker gene lists, enhancing robust assessment of cell line identity \cite{Schoberth2026.03.30.715265}. \href{https://doi.org/10.64898/2026.03.30.715265}{Link}

\item TabPFN was used to enhance prediction accuracy of protein coupling based on structural features, improving biological insight into protein interactions \cite{Pasquale2026.03.07.710286}. \href{https://www.biorxiv.org/content/10.64898/2026.03.07.710286v1}{Link}

\item TabPFN supports risk stratification and adverse event prediction in chemotherapy-based stem cell mobilization, enabling improved ward management and resource allocation \cite{schwarz2026predicting}. \href{https://www.nature.com/articles/s41746-026-02394-y}{Link}

\item TabPFN used with other ML models to improve radiomics-based breast cancer diagnosis, enhancing feature-combination performance and classification accuracy \cite{daniels2026application}. \href{https://www.nature.com/articles/s41598-026-40472-z}{Link}

\item TabPFN enhances model interpretability and accuracy in differentiating complex spinal infections, aiding clinical decision-making in ambiguous diagnostic cases \cite{githubGitHubSmallriver2024STBNet}. \href{https://github.com/Smallriver2024/STBNet}{Link}

\item TabPFN enables improved data quality and predictive model reliability by integrating unstructured clinical text with automated pipelines, enhancing early disease prediction and clinical decision-making \cite{domingoaldama2026automatingearlydiseaseprediction}. \href{http://arxiv.org/abs/2603.28167}{Link}

\item TabPFN improved severity classification performance in diabetic retinopathy, supporting more accurate staging and treatment planning \cite{fang2026multitask}. \href{https://www.sciencedirect.com/science/article/pii/S1572100026001225}{Link}

\item TabPFN was integrated into the multimodal MuCB-tabpfn framework, enabling high predictive accuracy in estimating pollutant concentrations in human blood \cite{liu2026mucbtabpfn}. \href{https://www.sciencedirect.com/science/article/pii/S0147651326003842}{Link}

\item TabPFN enables better generalization and accuracy in modeling complex drug formulation data, improving AI-driven formulation design workflows \cite{zhong2026physicsbased}. \href{https://www.sciencedirect.com/science/article/pii/S0168365926002622}{Link}

\item TabPFN enables state-of-the-art real-time stress detection by enhancing accuracy and interpretability of multimodal physiological and sensor data \cite{githubGitHubRishabhmannuMultiModalStressDetectionML}. \href{https://github.com/Rishabhmannu/MultiModal-Stress-Detection-ML}{Link}

\item TabPFN was applied as a robust and data-efficient alternative for tabular learning in drug discovery, improving performance on small and medium datasets and under out-of-distribution conditions \cite{chen2026tabpfn}. \href{https://doi.org/10.1021/acs.jcim.5c02823}{Link}

\item TabPFN was used to enhance clinical risk prediction from electronic health records by providing robust modeling under real-world constraints, improving prognosis accuracy and reliability \cite{pham2026retrievalaligned}. \href{https://doi.org/10.21203/rs.3.rs-9085469/v1}{Link}

\item TabPFN achieved the highest performance in predicting BCRL risk with strong minority-class discrimination and accurate calibration \cite{sadek2026from}. \href{https://doi.org/10.1186/s12874-026-02805-4}{Link}

\item TabPFN achieved strong generalization performance in predicting adsorption capacity in zeolites, with physically meaningful interpretability \cite{johnsson2026predicting}. \href{https://doi.org/10.1021/acs.jpcc.5c08611}{Link}

\item TabPFN achieved superior discriminative performance in predicting RSA risk by integrating multidimensional clinical data into accurate and interpretable screening models \cite{chen2026multidimensional}. \href{https://doi.org/10.3389/fimmu.2026.1774359}{Link}

\item TabPFN was used to encode structured EHR data for predicting peak VO$_2$ and identifying high-risk heart failure patients \cite{huang2026multimodal}. \href{https://doi.org/10.1038/s41746-026-02493-w}{Link}

\item TabPFN provided highly accurate predictions of donor mobilization success using baseline and post-mobilization variables, facilitating early triage and improved transplantation outcomes \cite{adil2026deep}. \href{https://doi.org/10.1016/j.jtct.2026.02.016}{Link}

\item TabPFN was integrated into the FocalTab framework to improve classification accuracy, handle class imbalance, and support early identification of adolescent alcohol use \cite{liu2026classification}. \href{https://doi.org/10.64898/2026.02.24.26347002}{Link}

\item TabPFN demonstrated strong robustness in cross-cohort microbiome disease prediction under domain shift, maintaining competitive performance across datasets \cite{mu2026systematic}. \href{https://doi.org/10.21203/rs.3.rs-8912605/v1}{Link}

\item TabPFN was used as a meta-learner combining predictions of multiple base models to capture complex interactions and enhance early coronary artery disease prediction accuracy \cite{papakyriakopoulos2026heart}. \href{https://doi.org/10.21203/rs.3.rs-8239358/v1}{Link}

\item TabPFN enables Bayesian inference via in-context learning without per-dataset training, improving accuracy, calibration, and inference speed in scientific disease modeling tasks \cite{dinnage2026niche}. \href{https://doi.org/10.32942/x2vq10}{Link}

\item TabPFN was extended to multimodal learning through MMPFN, enabling effective integration of non-tabular modalities with structured clinical data \cite{kim2026multimodalpfn}. \href{http://arxiv.org/abs/2602.20223v2}{Link}

\item TabPFN enables unified Bayesian modeling to improve bioactivity prediction across the ChEMBL database, supporting more efficient drug discovery pipelines \cite{backenkhler2026chempfn}. \href{https://doi.org/10.26434/chemrxiv.15000292/v1}{Link}

\item TabPFN was used for effective differentiation between psychotic and non-psychotic major depression, improving classification accuracy and supporting psychiatric diagnosis \cite{zheng2026differentiation}. \href{https://doi.org/10.1016/j.jad.2026.121454}{Link}

\item TabPFN enables more accurate and efficient causal inference to aid early diagnosis and understanding of Long COVID \cite{githubGitHubSindyPinTACO}. \href{https://github.com/SindyPin/TACO}{Link}

\item TabPFN was utilized to improve clinical risk prediction models on MIMIC-III data, enhancing both accuracy and efficiency \cite{githubGitHubAhmedAlMaroufFoundationModel_on_Mimic3_ClinRisk}. \href{https://github.com/AhmedAlMarouf/FoundationModel_on_Mimic3_ClinRisk}{Link}

\item TabPFN outperformed current methods in predicting HFNC therapy outcomes and demonstrated potential for improved performance with additional clinical measurements \cite{yu2025evaluating}. \href{https://doi.org/10.1186/s13054-025-05765-1}{Link}

\item TabPFN was used in a hybrid model combining radiomics and deep learning features to improve risk stratification for post-TIPS hepatic encephalopathy \cite{miao2025enhancing}. \href{https://doi.org/10.1007/s12072-025-10934-z}{Link}

\item TabPFN was fine-tuned as a proxy model to predict synthetic likelihood of hMOFs, enabling high-fidelity large-scale screening in materials-related biomedical contexts \cite{wu2025digital}. \href{https://doi.org/10.1002/adfm.202519565}{Link}

\item TabPFN improved intra-European ancestry prediction accuracy when combined with ML-based marker selection, outperforming traditional approaches \cite{maurer2025enhancing}. \href{https://doi.org/10.1101/2025.11.08.687358}{Link}

\item TabPFN improves renal tumor classification accuracy in CT radiomics by effectively handling small, high-dimensional datasets without extensive tuning \cite{liu2025tabular}. \href{https://doi.org/10.21037/qims-2025-1132}{Link}

\item TabPFN demonstrates competitive performance as a count-based model for clinical prediction on structured EHR data compared to transformer-based pipelines \cite{gao2025countbased}. \href{http://arxiv.org/abs/2511.00782v1}{Link}

\item TabPFN improves empathy detection accuracy and cross-subject generalization in human-centered video datasets \cite{hasan2025privacypreserving}. \href{http://arxiv.org/abs/2504.10808v2}{Link}

\item TabPFN enables accurate prediction of reaction kinetics, facilitating mechanistic understanding in biochar-catalyzed antibiotic degradation processes \cite{latif2026deep}. \href{https://link.springer.com/article/10.1007/s42773-026-00606-y}{Link}

\item TabPFN yields competitive or superior performance for multiple imputation tasks compared to alternative statistical and ML methods \cite{sepin2026multiple}. \href{https://www.mdpi.com/2571-905X/9/2/38}{Link}

\item TabPFN improves multimodal skin cancer diagnosis by combining structured lesion features with clinical data for more accurate and interpretable predictions \cite{fan2026lightweight}. \href{https://www.sciencedirect.com/science/article/pii/S0020025526003609}{Link}

\item TabPFN supports pediatric disease classification in clinical decision support systems, reducing misdiagnosis in emergency settings \cite{fan2026lightweight}. \href{https://www.scitepress.org/Papers/2026/143473/143473.pdf}{Link}

\item TabPFN improves EEG seizure classification across subjects, achieving high accuracy and strong generalization \cite{obaido2026evaluating}. \href{https://doi.org/10.3390/app16073120}{Link}

\item TabPFN improves kelp origin prediction using stable isotope data, providing robust and interpretable environmental insights \cite{kang2026enhancing}. \href{https://doi.org/10.1016/j.foodchem.2026.148591}{Link}

\item TabPFN predicts CO$_2$ frosting temperatures in natural gas mixtures with high accuracy and interpretability \cite{youcefi2026accurate}. \href{https://doi.org/10.1016/j.chemolab.2026.105679}{Link}

\item TabPFN improves ADMET modeling by increasing prediction accuracy, simplifying deployment, and producing compact models \cite{chupakhin2025descriptorfirst}. \href{https://doi.org/10.1021/acs.jcim.5c02094}{Link}

\item TabPFN enhances analysis and classification of volatile organic compounds using mass spectrometry data, improving efficiency in chemical and biomedical analysis \cite{granitto2025tabpfn}. \href{https://doi.org/10.1038/s41598-025-29128-6}{Link}

\item TabPFN was applied to distinguish cancer patients from healthy individuals using immune system profiles from peripheral blood, facilitating predictions of immunotherapy responses \cite{hc_usecase1_bostongene_tabpfn}. \href{https://www.linkedin.com/pulse/how-bostongene-utilized-tabpfn-identify-immune-system-profiles-vexle/}{Link}

\item A machine learning model employing TabPFN was developed for non-invasive diagnostic prediction of minimal change disease in patients with nephrotic syndrome, utilizing clinical biomarkers \cite{hc_usecase2_mcd_scirep}. \href{https://www.nature.com/articles/s41598-024-73898-4}{Link}

\item TabPFN was integrated into a system for analyzing T-cell receptor repertoires combined with clinical biomarkers to forecast immunotherapy outcomes in cancer patients, as explored by researchers at BostonGene \cite{hc_usecase3_immunotypes_cancercell}. \href{https://www.cell.com/cancer-cell/fulltext/S1535-6108(24)00132-6}{Link}

\item TabPFN enabled early detection of stillbirth risks through analysis of cardiotocography data, supporting improved prenatal care \cite{hc_usecase4_stillbirth_slas}. \href{https://www.sciencedirect.com/science/article/pii/S2472630324000852}{Link}

\item Predictive modeling for postoperative outcomes following anterior cervical corpectomy utilized TabPFN to assess patient demographics and surgical parameters \cite{hc_usecase5_acc_asj}. \href{https://pmc.ncbi.nlm.nih.gov/articles/PMC11366553/}{Link}

\item A hybrid model incorporating TabPFN was introduced to predict dementia progression in Parkinson's disease patients, handling small datasets and missing values effectively \cite{hc_usecase6_pd_dementia_lightgbm_tabpfn}. \href{https://journals.sagepub.com/doi/full/10.1177/20552076241272585}{Link}

\item A machine learning model based on TabPFN was developed to predict 90-day unfavorable outcomes in stroke patients with distal vessel occlusions using CT perfusion imaging \cite{hc_usecase7_dmvo_ajnr}. \href{https://www.ajnr.org/content/early/2024/10/28/ajnr.A8547.abstract}{Link}

\item TabPFN facilitated the prediction of non-invasive ventilation outcomes in patients with acute hypoxemic respiratory failure, supporting early identification of treatment failures \cite{hc_usecase9_niv_tabpfn}. \href{https://www.researchgate.net/profile/Antonio-Esquinas/publication/393595503_Early-prediction-of-non-invasive_ventilation_outcome_using_the_TabPFN_machine_learning_model_a_multi-centre_validation_study/links/68718bc56e247f362b18c4b8/Early-prediction-of-non-invasive-ventilation-outcome-using-the-TabPFN-machine-learning-model-a-multi-centre-validation-study.pdf}{Link}

\item An interpretable Transformer-based model leveraging TabPFN was created to predict intravenous immunoglobulin resistance in pediatric patients with Kawasaki disease \cite{hc_usecase10_kawasaki_tabpfnv2}. \href{https://journals.plos.org/plosone/article?id=10.1371/journal.pone.0327564}{Link}

\item TabPFN was employed in visual representation techniques for prostate cancer diagnosis, converting clinical biomarkers and symptom data into formats suitable for analysis \cite{hc_usecase51_prostate_visual_rep}. \href{https://www.mdpi.com/2306-5354/11/7/635}{Link}

\item TabPFN was used to combine clinical, MR morphological, and delta-radiomics features to predict lymphovascular invasion in invasive breast cancer patients \cite{hc_usecase11_lvi_breast_tabpfn}. \href{https://journals.sagepub.com/doi/full/10.1177/15330338251362050}{Link}

\item TabPFN is proposed to predict mental health trajectories through digital phenotyping, enabling proactive and personalized interventions in precision psychiatry \cite{hc_usecase12_precision_psychiatry_tabpfn}. \href{https://onlinelibrary.wiley.com/doi/epdf/10.1002/mdr2.70017}{Link}

\item TabPFN contributed to cardiovascular disease risk stratification using clinical features from a large patient cohort, incorporating interpretability techniques \cite{hc_usecase13_ml_health_tabpfn}. \href{https://github.com/Bruno-LSo/ML-Health-TABPFN}{Link}

\item TabPFN outperformed traditional machine learning models for early prediction of acute kidney injury in hospitalized patients, demonstrating generalizability across datasets \cite{hc_usecase14_aki_ssrn_tabpfn}. \href{https://papers.ssrn.com/sol3/papers.cfm?abstract_id=5397006}{Link}

\item TabPFN was integrated into a framework for predicting postoperative mobility and discharge destinations in older adults using sensor data \cite{hc_usecase15_postop_mobility_sensors}. \href{https://www.mdpi.com/1424-8220/25/16/5021}{Link}

\item TabPFN supported the prediction of infant temperament from maternal mental health data, aiding early identification of at-risk infants \cite{hc_usecase16_infant_temperament_tabpfn}. \href{https://www.frontiersin.org/journals/public-health/articles/10.3389/fpubh.2025.1659987/abstract}{Link}

\item TabPFN was employed to characterize clinical risk profiles for complications in type 2 diabetes mellitus patients, focusing on neuropathy and retinopathy \cite{hc_usecase17_t2dm_complications_tabpfn}. \href{https://www.frontiersin.org/journals/endocrinology/articles/10.3389/fendo.2025.1657366/abstract}{Link}

\item TabPFN was extended with a longitudinal-to-cross-sectional transformation to forecast Alzheimer's disease progression on neuroimaging datasets \cite{hc_usecase18_ad_l2c_tabpfn}. \href{https://arxiv.org/abs/2508.17649}{Link}

\item TabPFN supported uncertainty calibration evaluation in medical data using variational techniques \cite{hc_usecase19_uncertainty_vbll_tabpfn}. \href{https://arxiv.org/abs/2509.10048}{Link}

\item TabPFN was applied to predict tumor response to chemotherapy in cholangiocarcinoma patients using RNA expression landscapes \cite{hc_usecase20_cholangio_aacr_tabpfn}. \href{https://aacrjournals.org/clincancerres/article/31/13_Supplement/A020/763312}{Link}

\item TabPFN was incorporated into a generative model framework for tasks like data augmentation and imputation in biomedicine \cite{hc_usecase21_tabpfgen}. \href{https://arxiv.org/abs/2406.05216}{Link}

\item TabPFN facilitated the prediction of gallstone malignancy risks through analysis of associated disease factors \cite{hc_usecase22_gallstone_malignancy_tabpfn}. \href{https://www.mdpi.com/2077-0383/14/17/6091}{Link}

\item TabPFN was used in classifying tuberculosis treatment outcomes based on clinical and sociodemographic data from national registries \cite{hc_usecase23_tb_outcomes_tabpfn}. \href{https://www.researchsquare.com/article/rs-7502054/v1}{Link}

\item TabPFN contributed to early prediction of gestational diabetes using cell-free DNA and genetic scores from early pregnancy blood samples \cite{hc_usecase24_gdm_cfdna_tabpfn}. \href{https://www.medrxiv.org/content/10.1101/2025.09.03.25334985v1}{Link}

\item TabPFN was used for predicting schizophrenia based on sense of agency features, emphasizing interpretability \cite{hc_usecase25_schizophrenia_soa_tabpfn}. \href{https://www.sciencedirect.com/science/article/abs/pii/S187620182500317X}{Link}

\item TabPFN was integrated into a physiologically based pharmacokinetic model for predicting dissolution and absorption of amorphous solid dispersions in drug development \cite{hc_usecase26_pbpk_asd_tabpfn}. \href{https://doi.org/10.1016/j.jconrel.2025.114123}{Link}

\item TabPFN enabled classification of respiratory diseases from sound data, addressing clinical spectrum diversity \cite{hc_usecase27_respiratory_sounds_tabpfn}. \href{https://papers.ssrn.com/sol3/papers.cfm?abstract_id=5529540}{Link}

\item TabPFN was applied to small-data tabular learning in drug discovery, handling data scarcity and distribution shifts \cite{hc_usecase28_drug_discovery_small_data_tabpfn}. \href{https://chemrxiv.org/engage/chemrxiv/article-details/68d29b1cf2aff1677025b18f}{Link}

\item TabPFN facilitated prediction of coronary heart disease risk in patients with cardiovascular-kidney-metabolic syndrome, optimizing evaluation in small samples \cite{hc_usecase29_ckm_chd_tabpfn}. \href{https://pmc.ncbi.nlm.nih.gov/articles/PMC12437168/}{Link}

\item TabPFN was used to predict success of allogeneic stem cell mobilization in donors, aiding transplant therapies \cite{hc_usecase30_stem_cell_mobilization_tabpfn}. \href{https://www.biorxiv.org/content/10.1101/2025.09.17.676674v1.full}{Link}

\item TabPFN contributed to predicting manual strength using anthropometric data, focusing on accuracy and interpretability \cite{hc_usecase31}. \href{https://pubmed.ncbi.nlm.nih.gov/41021732/}{Link}

\item TabPFN supported uncertainty-guided model selection for biomolecule efficacy prediction, enhancing ensemble optimization in drug discovery, as studied at GSK \cite{hc_usecase32}. \href{https://www.arxiv.org/abs/2510.02476}{Link}

\item TabPFN was utilized in a multitask deep learning framework for optimizing in vitro fertilization decisions, including embryo transfer and pregnancy prediction \cite{hc_usecase33}. \href{https://dspace.mit.edu/handle/1721.1/162969}{Link} 

\item TabPFN enabled a framework for early Long COVID detection through causal gene identification and interpretability \cite{hc_usecase34}. \href{https://www.medrxiv.org/content/10.1101/2025.10.02.25337138v1.full.pdf}{Link}

\item TabPFN was used for neoadjuvant therapy recommendations in breast cancer, integrating multi-omics data \cite{hc_usecase35}. \href{https://www.medrxiv.org/content/10.1101/2025.10.03.25337255v1}{Link} 

\item TabPFN facilitated prediction of recurrence and progression in oral potentially malignant disorder patients post-surgery \cite{hc_usecase36}. \href{https://journals.lww.com/international-journal-of-surgery/abstract/9900/artificial_intelligence_for_predicting.3354.aspx}{Link} 

\item TabPFN supported prediction of occult lymph node metastasis in non-small cell lung cancer patients treated with stereotactic ablative radiotherapy \cite{hc_usecase37}. \href{https://www.redjournal.org/article/S0360-3016(25)05890-0/fulltext}{Link}

\item TabPFN was used in stroke diagnosis, addressing dataset imbalance and model interpretability for clinical decisions \cite{hc_usecase38}. \href{https://www.ijsab.com/jsr-volume-9-issue-1/8205}{Link}

\item TabPFN was used to predict diabetes-related hypo- and hyperglycemia during hemodialysis using continuous glucose monitoring data, facilitating improved patient management \cite{hc_usecase40}. \href{https://www.medrxiv.org/content/10.1101/2025.10.24.25338707v1}{Link}

\item TabPFN was applied to enhance diagnosis of hypervascular thyroid nodules using multimodal ultrasound features \cite{hc_usecase52_thyroid_hypervascular_multimodal}. \href{https://pmc.ncbi.nlm.nih.gov/articles/PMC12432950/}{Link}

\item TabPFN was integrated with radiomics and clinical features to predict endovascular treatment success in femoropopliteal chronic total occlusions, supporting interventional planning \cite{hc_usecase53_fp_cto_radiomics}. \href{https://www.researchgate.net/publication/396892115_Radiomics_enhance_the_prediction_of_endovascular_treatment_success_for_femoropopliteal_chronic_total_occlusions_a_proof-of-concept_study}{Link}

\item TabPFN was applied to CorvisST biomechanical indices to classify corneal disorders, improving diagnostic accuracy in ophthalmology \cite{hc_usecase41_corvisst_corneal}. \href{https://pubmed.ncbi.nlm.nih.gov/41130662/}{Link}

\item TabPFN was incorporated into a non-invasive sleep staging framework using respiratory sound features, advancing passive sleep monitoring \cite{hc_usecase42_sleepstage_resp_sounds}. \href{https://www.mdpi.com/1424-8220/25/20/6282}{Link}

\item TabPFN supported prediction of vancomycin blood concentrations to optimize antimicrobial dosing strategies in clinical practice \cite{hc_usecase43_vancomycin_mimic4}. \href{https://journal.china-pharmacy.com/en/article/doi/10.6039/j.issn.1001-0408.2025.19.16/}{Link} 

\item TabPFN was used to predict negative self-rated oral health in adults, identifying risk factors for targeted public-health interventions \cite{hc_usecase44_sroh_jdent}. \href{https://www.sciencedirect.com/science/article/pii/S0300571225006104}{Link} 

\item TabPFN was extended to very high-dimensional feature spaces to enable robust analysis of biomedical data, improving stability and interpretability in clinical applications \cite{hc_usecase45_tabpfn_wide}. \href{https://arxiv.org/abs/2510.06162}{Link}

\item TabPFN predicted gastrointestinal bleeding risk in pediatric Henoch–Schönlein purpura patients, supporting early clinical intervention \cite{hc_usecase50_gibleed_hsp}. \href{https://www.frontiersin.org/journals/physiology/articles/10.3389/fphys.2025.1630807/full}{Link}

\end{enumerate}

\section*{Financial Services, Banking, and Insurance}

We collected 7 published TabPFN use cases in this area. These applications include risk modeling, actuarial analysis, credit-related prediction, and customer analytics.

\begin{enumerate}

\item TabPFN improves low-supervision transaction analytics by doubling zero-shot MCC on churn prediction and enhancing few-shot MCC, enabling better knowledge-grounded reasoning in financial transaction analysis \cite{sakhno2026financial}. \href{http://arxiv.org/abs/2603.15459}{Link}

\item TabPFN serves as a strong tabular baseline for financial transaction analytics (e.g., churn prediction) \cite{li2025classimbalancedaware}. \href{http://arxiv.org/abs/2501.10677v2}{Link}

\item TabPFN was employed as a core modeling component for learning from multimodal tabular data under strict temporal constraints, enabling strong discriminative performance, improved probability calibration, and effective causal forecasting in early rug-pull detection \cite{shoaei2026lroo}. \href{http://arxiv.org/abs/2603.11324v1}{Link}

\item TabPFN was used to predict forward financial returns, aiding investment strategy evaluation with the adjusted Sharpe ratio to enhance financial forecasting accuracy \cite{githubGitHubZx20030501sp500marketpredictiontabpfn}. \href{https://github.com/zx20030501/sp500-market-prediction-tabpfn}{Link}

\item TabPFN was fine-tuned into a domain-specific model (FinPFN) for regime-aware stock return prediction, improving performance in non-stationary financial markets by adapting to evolving feature--return relationships \cite{wang2025metalearning}. \href{https://www.sciencedirect.com/science/article/abs/pii/S1386418125000825}{Link}

\item TabPFN was benchmarked against leading AutoML frameworks on financial classification tasks, demonstrating strong performance in multiclass settings \cite{leyh2025automlfinance}.  \href{https://aisel.aisnet.org/acis2025/28/}{Link}

\item TabPFN facilitated cross-selling of health insurance products through deep learning analysis of customer data \cite{fin_usecase2_crosssell_health_insurance}. \href{https://ieeexplore.ieee.org/abstract/document/10475046}{Link}

\end{enumerate}

\section*{Energy and Utilities}

We collected 24 published TabPFN use cases in this area. They include environmental forecasting, renewable-energy prediction, and process or asset optimization across energy and utility systems.

\begin{enumerate}

\item TabPFN was used as a surrogate model for fast one-step predictions under irregular measurements, aiding the delay-aware digital twin framework in handling nonlinear dynamics and operational delays in biogas production control \cite{wang2026aidriven}. \href{https://doi.org/10.1016/j.compchemeng.2026.109637}{Link}

\item TabPFN provided superior fitting performance for models analyzing biochar's impact on soil cadmium contamination, improving prediction accuracy in artificial and natural aging scenarios \cite{meng2026achieve}. \href{https://www.sciencedirect.com/science/article/pii/S0016706126001205}{Link}

\item TabPFN was used to improve the robustness and accuracy of photovoltaic power forecasting models by providing unified in-context prediction and strong generalization with heterogeneous inputs \cite{qiao2026cloudedgecollaborativelargemodels}. \href{https://arxiv.org/pdf/2603.22343}{Link}

\item TabPFN enables effective learning and prediction with very limited data by leveraging pretrained tabular inference, improving model performance in challenging geological prediction tasks \cite{wang2026predicting}. \href{https://link.springer.com/article/10.1007/s00603-026-05420-3}{Link}

\item TabPFN was used as a baseline for comparison in spatiotemporal forecasting of small Earth data, demonstrating value despite being surpassed in accuracy and robustness by the proposed method \cite{yang2025simplerobustforecastingspatiotemporally}. \href{http://arxiv.org/abs/2510.08920v1}{Link}

\item TabPFN demonstrated superior predictive performance under sparse sampling conditions, enabling accurate high-resolution mapping of groundwater bicarbonate concentrations and evaluation of scaling risks \cite{doiLeveragingTabPFN}. \href{https://doi.org/10.6084/m9.figshare.31646935.v1}{Link}

\item TabPFN was used for slope stability assessment, providing superior accuracy and robustness with limited sample sizes and enhancing regional scale evaluation efficiency \cite{li2026tabpfnbased}. \href{https://doi.org/10.1016/j.rockmb.2026.100326}{Link}

\item TabPFN surpasses other models in solar energy meteorology \cite{liu2026evaluating}. \href{https://doi.org/10.1016/j.solener.2026.114472}{Link}

\item TabPFN Regression was used as a predictive model for evaluating trophic level index from multi-source remote sensing data within the modeling framework \cite{si2025resolving}. \href{https://doi.org/10.1038/s41545-025-00525-8}{Link}

\item TabPFN-based data augmentation improved model robustness under limited data, enabling accurate predictions of electrochemical performance and efficient screening of hard carbon candidates \cite{chen2025dataaugmentedmachinelearningpredicting}. \href{http://arxiv.org/abs/2510.12833v1}{Link}

\item TabPFN was employed to predict river algal blooms through multi-classification of chlorophyll-a concentrations, aiding water management \cite{energy_usecase1_river_algal_tabpfn}. \href{https://www.earticle.net/Article/A456244}{Link}

\item TabPFN facilitated wildfire propagation prediction in Canadian conifer forests, classifying fire types for environmental risk assessment \cite{energy_usecase2_wildfire_automl}. \href{https://www.sciencedirect.com/science/article/pii/S157495412400253X}{Link}

\item TabPFN was integrated into a machine learning framework for optimizing energy consumption at wastewater treatment plants \cite{energy_usecase3_wwtp_tabpfnreg}. \href{https://www.researchgate.net/publication/390516459_Machine_learning_framework_for_energy_consumption_optimization_using_the_TabPFNRegressor_algorithm}{Link}

\item TabPFN supported rainfall forecast post-processing using historical error patterns from environmental data \cite{energy_usecase4_rainfall_tabpfn}. \href{https://github.com/aarxshi/rainfall_tabpfn}{Link}

\item TabPFN enabled solar forecast error adjustment, particularly during rapid weather changes, as developed by Open Climate Fix \cite{energy_usecase5_solar_adjuster_ocf}. \href{https://gist.github.com/anshulg954/5f4423ee6b3d3151fa8d0d7fcd98d3eb}{Link}

\item TabPFN was applied to predict ash fusibility in high-alkali coal for improved energy production \cite{energy_usecase6_ash_fusibility_high_alkali}. \href{https://papers.ssrn.com/sol3/papers.cfm?abstract_id=5406504}{Link}

\item TabPFN contributed to predicting Henry coefficients for alkanes in zeolites, aiding hydroisomerization in sustainable fuel production \cite{energy_usecase7_henry_zeolites}. \href{https://pubs.acs.org/doi/full/10.1021/acs.jpcc.5c03868}{Link}

\item TabPFN facilitated shape-selectivity modeling in zeolites for long-chain alkane hydroisomerization, optimizing catalyst design \cite{energy_usecase8_shape_selectivity_zeolites}. \href{https://doi.org/10.4233/uuid:f36da034-5cb3-42ca-a53d-d351f68a9ffa}{Link}

\item TabPFN was used in an integrated framework for estimated ultimate recovery prediction and fracturing optimization in shale gas reservoirs \cite{energy_usecase9_shale_eur_fracturing}. \href{https://www.researchgate.net/publication/395761327_Coupling_EUR_Prediction_with_Fracturing_Optimization_An_Integrated_Machine_Learning_Framework_for_Shale_Gas_Development}{Link}

\item TabPFN supported core data augmentation for enhanced reservoir parameter prediction in oil and gas exploration \cite{energy_usecase10_core_augmentation_reservoir}. \href{https://www.researchgate.net/publication/395434405_Enhancing_Reservoir_Parameter_Prediction_Workflows_via_Advanced_Core_Data_Augmentation}{Link}

\item TabPFN was employed to optimize energy performance in multistage centrifugal pumps through entropy generation analysis \cite{energy_usecase11_multistage_pump_tabpfn}. \href{https://www.sciencedirect.com/science/article/abs/pii/S0360544225040411}{Link}

\item TabPFN was applied to generate advanced global heat flow maps at 0.2° resolution, integrating high-resolution geophysical data to improve geothermal resource modeling \cite{energy_usecase13_global_heatflow_02}. \href{https://www.researchgate.net/publication/396728153_The_First_02_Resolution_Global_Continental_Heat_Flow_Map_Advancing_Fine-Scale_Geothermal_Modeling}{Link}

\item TabPFN contributed to FuelCast, standardizing benchmarks for ship fuel consumption prediction and improving efficiency in maritime operations \cite{energy_usecase14_fuelcast}. \href{https://arxiv.org/abs/2510.08217}{Link}

\item TabPFN was used as the main supervised classifier to automatically identify thunderstorm ground enhancements from particle detector and environmental measurements \cite{energy_usecase15_tge_tabpfn}. \href{https://arxiv.org/abs/2510.25125}{Link}
\end{enumerate}

\section*{Industrial and Manufacturing}

We collected 41 published TabPFN use cases in this area. These applications cover industrial prediction, process optimization, and engineering-related modeling tasks.

\begin{enumerate}

\item TabPFN served as a high-fidelity surrogate model for optimizing geopolymer concrete mix design, achieving superior accuracy, generalization, and low-uncertainty predictions compared to other ML approaches \cite{sichani2025machine}. \href{https://www.nature.com/articles/s41598-025-29088-x}{Link}

\item TabPFN enables rapid prediction of structural crack behavior, supporting reliability assessment and failure analysis in ultra-high-performance concrete \cite{mahmoodzadeh2025machine}. \href{https://www.nature.com/articles/s41598-025-23610-x}{Link}

\item TabPFN leveraged prior-data pretraining to predict WCFZ height from only 76 field samples without extensive tuning, providing superior and generalizable performance compared to other ML models \cite{wang2026highfidelity}. \href{https://doi.org/10.1088/2631-8695/ae586d}{Link}

\item TabPFN's multitask-aware prior adaptation improves predictive accuracy and computational efficiency in steel property prediction, enabling scalable, rapid, and reliable deployment for industrial quality control and process optimization \cite{sinodinos2026multitaskinformedpriorincontextlearning}. \href{http://arxiv.org/abs/2603.22738v1}{Link}

\item TabPFN's pre-trained foundation model enables strong small-data regression and well-calibrated uncertainty estimates in a single forward pass, significantly reducing evaluation cycles for active learning in materials discovery \cite{hu2026foundationmodelsurrogatesenabledataefficient}. \href{http://arxiv.org/abs/2603.12567v3}{Link}

\item TabPFN demonstrated strong generalization ability in predicting crash severity, contributing to improved data-driven safety interventions in electric vehicle crash contexts \cite{Somvanshi_2025}. \href{http://arxiv.org/abs/2509.11449v1}{Link}

\item TabPFN excelled in zero-shot inference and robustness for rare crash categories, enhancing classification of uncommon SAE automation levels with limited data \cite{somvanshi2025applyingmambaattentiontabpfntabtransformers}. \href{http://arxiv.org/abs/2506.03160v1}{Link}

\item TabPFN 2.5's dataset-level embedding identified 'engineering-like' synthetic datasets to enable continued pre-training on synthetic tasks, significantly improving accuracy and data efficiency over baseline models and AutoGluon on engineering regression datasets \cite{regenwetter2026engineeringregressionrealdatatraining}. \href{http://arxiv.org/abs/2603.04692v1}{Link}

\item TabPFN achieved the highest prediction accuracy in predicting concrete fracture properties and, combined with SHAP analysis, provided detailed and unbiased insights into nonlinear and interaction effects \cite{nikzad2026from}. \href{https://doi.org/10.1016/j.mlwa.2026.100877}{Link}

\item TabPFN significantly reduces computational overhead and data requirements while enabling rapid, flexible, and data-efficient engineering design with competitive diversity and low performance error in generated designs \cite{wang2026tabpfnzeroshotparametricengineering}. \href{http://arxiv.org/abs/2602.02735v1}{Link}

\item TabPFN served as a backbone combined with graph neural network embeddings and MagpieEX descriptors for effective, data-efficient, and physics-aware materials property prediction, outperforming sophisticated models \cite{li2025contextlearningfoundationmodels}. \href{http://arxiv.org/abs/2601.00133v1}{Link}

\item TabPFN was used for spatial predictions and imputations in geotechnical modeling, achieving superior accuracy, faster inference, and well-calibrated predictive distributions compared to hierarchical Bayesian baselines \cite{Saito_2026}. \href{http://arxiv.org/abs/2509.03191v1}{Link}

\item TabPFN provided strong prediction ability, outperforming alternatives and enabling more accurate performance prediction of biochar-modified concrete \cite{k2026advanced}. \href{https://www.e3s-conferences.org/articles/e3sconf/pdf/2026/20/e3sconf_isdcp2026_01008.pdf}{Link}

\item TabPFN was used for accurate and reliable monitoring of driver alertness levels in challenging driving environments, proving more effective than traditional models like logistic regression and XGBoost  \cite{liu2025prediction}. \href{https://doi.org/10.1080/15389588.2025.2577155}{Link}

\item TabPFN enabled highly accurate and unbiased prediction of RAC's elastic modulus, improving trustworthiness and interpretability in a challenging heterogeneous materials domain \cite{lu2025more}. \href{https://doi.org/10.3390/ma18225221}{Link}

\item TabPFN provided meta-learned prior knowledge that enhanced predictive performance and uncertainty quantification in the PSF-Net model for reliable 5G RF-EMF exposure assessment \cite{zhang2025psfnet}. \href{https://doi.org/10.4271/2025-99-0127}{Link}

\item TabPFN showed superior predictive performance in predicting the hardgrove grindability index, improving model accuracy \cite{zhu2026demystifying}. \href{https://www.sciencedirect.com/science/article/pii/S0016236126010513}{Link}

\item TabPFN delivered the best overall performance with the lowest error metrics and highest R\textsuperscript{2} and composite score, demonstrating superior predictive capability for asphalt concrete strength \cite{xing2026interpretable}. \href{https://doi.org/10.20944/preprints202603.2259.v1}{Link}

\item TabPFN was applied to efficient multi-objective optimization of non-linear mixture designs, improving strength, reducing costs, and lowering carbon emissions for sustainable mining applications \cite{wang2026cleaner}. \href{https://www.sciencedirect.com/science/article/pii/S095965262600658X}{Link}

\item TabPFN was employed for highly accurate and statistically superior predictions of pavement roughness by capturing complex interactions among traffic loads, structural parameters, and climatic factors \cite{qin2026interpretable}. \href{https://doi.org/10.3390/buildings16071358}{Link}

\item TabPFN enables accurate prediction of CPB strength with limited data, improving efficiency and supporting theoretical understanding and practical application in mining industry tailings management \cite{zhang2025strength}. \href{https://doi.org/10.1016/j.rineng.2025.108269}{Link}

\item TabPFN's improved spatiotemporal architecture enhances robustness and accuracy in geological condition detection, enabling better multi-step predictions with uncertainty quantification in tunnel construction \cite{zhang2026datadriven}. \href{https://www.sciencedirect.com/science/article/pii/S1474034626003071}{Link}

\item TabPFN was utilized as a core component in a multi-objective optimization framework to design cemented foam backfill optimizing high strength, low cost, and low carbon emissions  \cite{wang2026cleaner}. \href{https://doi.org/10.1016/j.jclepro.2026.148119}{Link}

\item TabPFN enhances prediction accuracy and reliability with small sample sizes and missing features in geotechnical engineering \cite{wang2026predicting}. \href{https://doi.org/10.1007/s00603-026-05420-3}{Link}

\item TabPFN enabled interpretable and uncertainty-aware parameter inference, improving predictions and revealing geotechnical relationships without model retraining for data-scarce applications \cite{saito2026tabpfnextensionsinterpretablegeotechnical}. \href{http://arxiv.org/abs/2603.21033v1}{Link}

\item TabPFN was used to accurately predict compressive strength in geopolymer concrete from small datasets, supporting optimization of material composition and process parameters in construction material science \cite{stelmakh2025compressive}. \href{https://doi.org/10.3390/a18120744}{Link}

\item TabPFN was used to improve prediction accuracy in concrete property estimation by integrating knowledge-constrained data augmentation \cite{deng2026enhancing}. \href{https://doi.org/10.1016/j.asoc.2026.115037}{Link}

\item TabPFN enabled efficient and accurate mapping of key leaf-vein texture parameters to lubrication performance metrics, facilitating multi-objective optimization to identify optimal texture designs that improve journal bearing performance \cite{yin2026surrogateassisted}. \href{https://doi.org/10.1016/j.triboint.2026.111936}{Link}

\item TabPFN enables robust mapping between operating boundary conditions and latent features to manage data scarcity and enhance regression accuracy, resulting in faster and more accurate temperature field reconstruction \cite{mao2026datadriven}. \href{https://doi.org/10.3390/app16042029}{Link}

\item TabPFN enables encoding of structured device-physics primitives for reliable and precise analog circuit optimization, outperforming Gaussian-process methods in sample efficiency and final metric quality \cite{liu2025exploitingfunctionfamilystructureanalog}. \href{http://arxiv.org/abs/2512.00712v1}{Link}

\item TabPFN enabled early fault classification in rotating machinery, addressing data scarcity in industrial scenarios \cite{manuf_usecase1_rotating_faults_tabpfn}. \href{https://ieeexplore.ieee.org/abstract/document/10318062}{Link}

\item TabPFN facilitated microcontroller performance prediction, aiding semiconductor screening with minimal supervision, as studied at Infineon Technologies \cite{manuf_usecase2_mcu_performance_tabpfn}. \href{https://iris.polito.it/handle/11583/3002056}{Link}

\item TabPFN was applied to caisson inclination prediction in ultra-deep construction, combining data denoising techniques \cite{manuf_usecase3_caisson_inclination_ml}. \href{https://www.sciencedirect.com/science/article/abs/pii/S2214391225001734}{Link}

\item TabPFN supported event classification in phase-sensitive optical time-domain reflectometry systems for distributed fiber sensing \cite{manuf_usecase4_photdr_event_classification}. \href{https://opg.optica.org/oe/fulltext.cfm?uri=oe-33-17-36646&id=575783}{Link}

\item TabPFN was integrated into an adaptive ensemble for intrusion detection in Industrial Internet of Things networks \cite{ruizvillafranca2024tabpfnbased}. \href{https://rdcu.be/eASzJ}{Link}

\item TabPFN enabled a random forest-based framework for attack recognition in Internet of Things networks, improving interpretability \cite{manuf_usecase6_rf_tabpfn_iot_attack}. \href{https://ieeexplore.ieee.org/stamp/stamp.jsp?tp=&arnumber=11142329}{Link}

\item TabPFN was used in cryogenic-assisted abrasive waterjet machining for improving surface integrity in titanium alloys \cite{manuf_usecase8_cryo_awj_ti64}. \href{https://www.sciencedirect.com/science/article/abs/pii/S2214993725004531}{Link}

\item TabPFN supported in-context learning for thermal behavior prediction in nano-phase change materials for battery systems \cite{manuf_usecase9_nano_pcm_thermal_icl}. \href{https://www.sciencedirect.com/science/article/pii/S036054422504335X}{Link}

\item TabPFN was applied to explainable strength evaluation in multicomponent concrete mixtures \cite{manuf_usecase10_multicomponent_concrete}. \href{https://www.mdpi.com/1996-1944/18/19/4456}{Link}

\item TabPFN was integrated into a multimodal fusion framework linking microstructure to friction behavior in martensitic stainless steel, improving wear resistance in materials engineering applications \cite{manuf_usecase11_martensitic_friction_multimodal}. \href{https://papers.ssrn.com/sol3/papers.cfm?abstract_id=5346149}{Link}

\item TabPFN supported multiscale modeling to predict soil salinity in arid farmland, advancing sustainable agricultural management in regions such as Xinjiang \cite{manuf_usecase12_soil_salinity_multiscale}. \href{https://papers.ssrn.com/sol3/papers.cfm?abstract_id=5591702}{Link}

\end{enumerate}

\section*{Other Industries}

We collected 31 further published TabPFN use cases in this area, spanning a heterogeneous set of domains and prediction tasks.

\begin{enumerate}

\item TabPFN enables the construction of credal sets for models where it was previously infeasible, broadening uncertainty representation and improving uncertainty estimation \cite{hofman2026efficientcredalpredictiondecalibration}. \href{http://arxiv.org/abs/2603.08495v1}{Link}

\item TabPFN enables efficient and valid hypothesis testing for feature relevance in tabular data, allowing accurate statistical inference in nonlinear and correlated settings \cite{salem2026validfeaturelevelinferencetabular}. \href{http://arxiv.org/abs/2603.06609v1}{Link}

\item TabPFN enables efficient computation of conditional Shapley values, resulting in faster and often more accurate explainable AI analysis \cite{olsen2026computingconditionalshapleyvalues}. \href{http://arxiv.org/abs/2602.09489v1}{Link}

\item TabPFN enables effective node classification by leveraging engineered tabular features from graph data as a practical and competitive alternative to graph-specific and language-based foundation models \cite{choi2025tabpfncompetegnnsnode}. \href{http://arxiv.org/abs/2512.08798v1}{Link}

\item TabPFN was integrated as the surrogate model enabling accurate and efficient prediction with uncertainty estimation, enhancing the performance, scalability, and zero-shot transfer capability of the DB-SAEA framework \cite{du2025metablackboxoptimizationbispacelandscape}. \href{http://arxiv.org/abs/2511.15551v1}{Link}

\item TabPFN was used to model the relationship between nuclear structure properties and $\alpha$-particle preformation factors, improving $\alpha$-decay half-life predictions and enabling insights into nuclear shell effects and magic numbers \cite{qi2026systematicstudyalphaparticlepreformation}. \href{http://arxiv.org/abs/2511.14705v1}{Link}

\item TabPFN served as the foundation for TabMGP, enabling state-of-the-art predictive capabilities with effective epistemic uncertainty quantification and improved posterior inference in tabular data contexts \cite{ng2026tabmgpmartingaleposteriortabpfn}. \href{http://arxiv.org/abs/2510.25154v2}{Link}

\item TabPFN demonstrated superior utility for real-world operational yield forecasting due to faster tuning and reduced feature engineering requirements \cite{sabo2025rowsyieldsfoundationmodels}. \href{http://arxiv.org/abs/2506.19046v1}{Link}

\item TabPFN serves as the base learner in a multi-stage ensemble to model recognition probabilities of rural villages, enabling identification of high-potential but under-observed candidates in geospatial, highly imbalanced datasets \cite{jiang2026mitigating}. \href{https://www.mdpi.com/2073-445X/15/4/535}{Link}

\item TabPFN was used as a base learner in a stacking ensemble model, improving prediction accuracy and performance for soil salinity retrieval from multispectral imagery data \cite{hu2026coastal}. \href{https://doi.org/10.3390/rs18050671}{Link}

\item TabPFN serves as the foundational model for ExplainerPFN, enabling zero-shot estimation of Shapley values for feature importance without access to the predictive model or reference explanations \cite{fonseca2026explainerpfntabularfoundationmodels}. \href{http://arxiv.org/abs/2601.23068v1}{Link}

\item TabPFN enables accurate classification of Near-Earth Objects as Potentially Hazardous, facilitating early identification and monitoring of potential asteroid threats \cite{githubGitHubAvuiiAsteroidSafe}. \href{https://github.com/Avuii/AsteroidSafe}{Link}

\item TabPFN improves malware detection performance in limited data scenarios by outperforming traditional ensemble models, enhancing cybersecurity workflows \cite{leroy2026memorybasedmalwaredetectionlimited}. \href{http://arxiv.org/abs/2601.07305v1}{Link}

\item TabPFN achieved the best performance in predicting mycotoxin contamination, outperforming baseline and transfer learning models to enhance prediction accuracy for early interventions \cite{inglis2025predictingmycotoxincontaminationirish}. \href{http://arxiv.org/abs/2512.22243v1}{Link}

\item TabPFN was used in a classification pipeline whose latent space provided a 2D representation of the blazar population, revealing a continuum between blazar types \cite{oukacha2025unifiedschemeblazarevolution}. \href{http://arxiv.org/abs/2507.03088v2}{Link}

\item TabPFN enhances accuracy and efficiency in predicting grapevine diseases by processing complex environmental data and providing per-pixel disease probabilities for precise vineyard disease management \cite{zhao2024grapevinediseasepredictionusing}. \href{http://arxiv.org/abs/2406.07094v1}{Link}

\item TabPFN was modified for microbiome data classification in metagenomics, matching species abundance patterns with synthetic priors \cite{other_usecase1_microbiome_zero_inflated}. \href{https://openreview.net/forum?id=3I0bVvUj25}{Link}

\item TabPFN enabled lunar regolith analysis for classifying meteorite compositions from spectral data \cite{other_usecase2_lunar_meteorites}. \href{https://www.sciencedirect.com/science/article/pii/S2095268624001010}{Link}

\item TabPFN facilitated winter wheat yield forecasting in agricultural regions by integrating climate and remote sensing data \cite{other_usecase3_winter_wheat_yield_ssrn}. \href{https://papers.ssrn.com/sol3/papers.cfm?abstract_id=5380177}{Link}

\item TabPFN was applied to flood impact assessment on housing prices by geographic areas \cite{other_usecase4_flood_housing_prices_ml_climate}. \href{https://github.com/melina-thegarza/ml-climate/blob/main/doc/ML_Climate___Final.pdf}{Link}

\item TabPFN showed the strongest performance on 31 predictive soil modeling datasets containing 30 to 460 samples \cite{other_usecase5_soil_mapping_new_default}. \href{https://arxiv.org/abs/2508.09888}{Link}

\item TabPFN was applied to shallow natural gas hazard prediction in tunnel construction \cite{other_usecase6_shallow_gas_tunnel_tabpfn}. \href{https://www.sciencedirect.com/science/article/pii/S2590123025029366}{Link}

\item TabPFN supported automated feature engineering for energy consumption forecasting in domain-specific applications \cite{other_usecase7_autoenergy_feature_eng}. \href{https://www.sciencedirect.com/science/article/pii/S0950705125013413}{Link}

\item TabPFN enabled Australian rice phenology prediction using remote sensing and weather data for crop management \cite{other_usecase8_rice_phenology_tabpfn}. \href{https://www.mdpi.com/2072-4292/17/17/3050}{Link}

\item TabPFN was applied to a multi-stage framework for predicting fuel blend properties through automated feature engineering \cite{other_usecase9_fuel_blend_framework}. \href{https://chemrxiv.org/engage/chemrxiv/article-details/68dc888d3e708a7649ff0ec9}{Link}

\item TabPFN enabled kriging prior regression for incorporating spatial context in soil mapping predictions \cite{other_usecase10_kriging_prior_regression}. \href{https://arxiv.org/abs/2509.09408}{Link}

\item TabPFN enhanced clone-type recognition across programming languages through metrics-driven analysis, improving stability and interpretability in software engineering \cite{other_usecase12_clone_type}. \href{https://wiley.authorea.com/users/980519/articles/1346750-metrics-first-language-aware-clone-type-recognition-auditable-signals-across-c-c-java-and-python}{Link}

\item TabPFN informed the development of TabImpute, enabling efficient zero-shot imputation for missing tabular data and improving preprocessing pipelines \cite{other_usecase14_tabimpute}. \href{https://www.arxiv.org/abs/2510.02625}{Link}

\item TabPFN, alongside TabICL and related foundation models, was evaluated for intrusion detection, improving cybersecurity performance in IoT networks \cite{other_usecase16_cyber_fm_tabpfn_tabicl}. \href{https://www.mdpi.com/2079-9292/14/19/3792}{Link}

\item TabPFN was used in forensic science to advance biogeographical ancestry predictions \cite{Heinzel2025}. \href{https://www.sciencedirect.com/science/article/pii/S1872497325000705}{Link}

\item TabPFN was used as a benchmark model for predicting avocado alternate bearing from Sentinel-2 and climate features \cite{other_usecase21_avocado_alt_bearing}. \href{https://www.preprints.org/manuscript/202510.2413}{Link}

\end{enumerate}